%% file: arxiv.tex
\pdfoutput=1

\documentclass[10pt,twocolumn,letterpaper]{article}

\usepackage[pagenumbers]{cvpr}  %

\input{preamble}

\definecolor{cvprblue}{rgb}{0.21,0.49,0.74}
\usepackage[pagebackref,breaklinks,colorlinks,allcolors=cvprblue]{hyperref}

\def\confName{CVPR}
\def\confYear{2027}

\def\arxivversion{}

\title{PhasorPT: Continuous Geometry for Convolution-Free Point Transformers}

\author{Christian Schott \qquad Bernd L\"udemann-Ravit\\
Institute for Production and Informatics\\
University of Applied Sciences Kempten, Sonthofen, Germany\\
{\tt\small \{christian.schott, bernd.luedemann-ravit\}@hs-kempten.de}\\[0.3em]
{\small\url{https://phasorpt.github.io/}}
}

\begin{document}
\maketitle
\input{sec/0_abstract}
\input{sec/1_intro}
\input{sec/2_related}
\input{sec/3_method.tex}
\input{sec/4_phasorpt}

\input{sec/5_experiments}

\input{sec/6_conclusion}

\input{sec/7_statements}

{
    \small
    \bibliographystyle{ieeenat_fullname}
    \bibliography{main}
}

\setcounter{figure}{0}
\setcounter{table}{0}
\setcounter{equation}{0}
\setcounter{algorithm}{0}
\renewcommand{\thefigure}{S\arabic{figure}}
\renewcommand{\thetable}{S\arabic{table}}
\renewcommand{\theequation}{S\arabic{equation}}
\renewcommand{\thealgorithm}{S\arabic{algorithm}}
\renewcommand{\theHfigure}{S\arabic{figure}}
\renewcommand{\theHtable}{S\arabic{table}}
\renewcommand{\theHequation}{S\arabic{equation}}
\renewcommand{\theHalgorithm}{S\arabic{algorithm}}
\input{sec/X_suppl}

\end{document}

%% file: preamble.tex
\newcommand{\todo}[1]{{\color{red}#1}}

\newcommand{\best}[1]{\textbf{#1}}
\newcommand{\second}[1]{\underline{#1}}
\newcommand{\ours}{\rowcolor{gray!15}}

\newcommand{\graybg}[1]{{\setlength{\fboxsep}{1pt}\colorbox{gray!15}{#1}}\kern-1pt}
\newcommand{\result}[2]{\ifcsname res@#1@#2\endcsname\csname res@#1@#2\endcsname\else\todo{--}\fi}
\InputIfFileExists{results}{}{}

\usepackage{array}      %
\usepackage{colortbl}   %
\usepackage{algorithm}
\usepackage{algpseudocode}

%% file: results.tex
\expandafter\def\csname res@volt-s-official-scannet@n\endcsname{1}
\expandafter\def\csname res@volt-s-official-scannet@miou\endcsname{76.4}
\expandafter\def\csname res@volt-s-official-scannet@miou-pm\endcsname{76.4}
\expandafter\def\csname res@volt-s-official-scannet@macc\endcsname{85.0}
\expandafter\def\csname res@volt-s-official-scannet@macc-pm\endcsname{85.0}
\expandafter\def\csname res@volt-s-official-scannet@allacc\endcsname{91.3}
\expandafter\def\csname res@volt-s-official-scannet@allacc-pm\endcsname{91.3}
\expandafter\def\csname res@volt-s-official-scannet@iou-wall\endcsname{86.5}
\expandafter\def\csname res@volt-s-official-scannet@iou-wall-pm\endcsname{86.5}
\expandafter\def\csname res@volt-s-official-scannet@iou-floor\endcsname{95.3}
\expandafter\def\csname res@volt-s-official-scannet@iou-floor-pm\endcsname{95.3}
\expandafter\def\csname res@volt-s-official-scannet@iou-cabinet\endcsname{72.3}
\expandafter\def\csname res@volt-s-official-scannet@iou-cabinet-pm\endcsname{72.3}
\expandafter\def\csname res@volt-s-official-scannet@iou-bed\endcsname{83.9}
\expandafter\def\csname res@volt-s-official-scannet@iou-bed-pm\endcsname{83.9}
\expandafter\def\csname res@volt-s-official-scannet@iou-chair\endcsname{91.9}
\expandafter\def\csname res@volt-s-official-scannet@iou-chair-pm\endcsname{91.9}
\expandafter\def\csname res@volt-s-official-scannet@iou-sofa\endcsname{83.4}
\expandafter\def\csname res@volt-s-official-scannet@iou-sofa-pm\endcsname{83.4}
\expandafter\def\csname res@volt-s-official-scannet@iou-table\endcsname{77.6}
\expandafter\def\csname res@volt-s-official-scannet@iou-table-pm\endcsname{77.6}
\expandafter\def\csname res@volt-s-official-scannet@iou-door\endcsname{64.2}
\expandafter\def\csname res@volt-s-official-scannet@iou-door-pm\endcsname{64.2}
\expandafter\def\csname res@volt-s-official-scannet@iou-window\endcsname{68.1}
\expandafter\def\csname res@volt-s-official-scannet@iou-window-pm\endcsname{68.1}
\expandafter\def\csname res@volt-s-official-scannet@iou-bookshelf\endcsname{85.1}
\expandafter\def\csname res@volt-s-official-scannet@iou-bookshelf-pm\endcsname{85.1}
\expandafter\def\csname res@volt-s-official-scannet@iou-picture\endcsname{35.2}
\expandafter\def\csname res@volt-s-official-scannet@iou-picture-pm\endcsname{35.2}
\expandafter\def\csname res@volt-s-official-scannet@iou-counter\endcsname{67.1}
\expandafter\def\csname res@volt-s-official-scannet@iou-counter-pm\endcsname{67.1}
\expandafter\def\csname res@volt-s-official-scannet@iou-desk\endcsname{72.5}
\expandafter\def\csname res@volt-s-official-scannet@iou-desk-pm\endcsname{72.5}
\expandafter\def\csname res@volt-s-official-scannet@iou-curtain\endcsname{80.8}
\expandafter\def\csname res@volt-s-official-scannet@iou-curtain-pm\endcsname{80.8}
\expandafter\def\csname res@volt-s-official-scannet@iou-refridgerator\endcsname{69.0}
\expandafter\def\csname res@volt-s-official-scannet@iou-refridgerator-pm\endcsname{69.0}
\expandafter\def\csname res@volt-s-official-scannet@iou-shower curtain\endcsname{76.6}
\expandafter\def\csname res@volt-s-official-scannet@iou-shower curtain-pm\endcsname{76.6}
\expandafter\def\csname res@volt-s-official-scannet@iou-toilet\endcsname{93.9}
\expandafter\def\csname res@volt-s-official-scannet@iou-toilet-pm\endcsname{93.9}
\expandafter\def\csname res@volt-s-official-scannet@iou-sink\endcsname{71.2}
\expandafter\def\csname res@volt-s-official-scannet@iou-sink-pm\endcsname{71.2}
\expandafter\def\csname res@volt-s-official-scannet@iou-bathtub\endcsname{88.4}
\expandafter\def\csname res@volt-s-official-scannet@iou-bathtub-pm\endcsname{88.4}
\expandafter\def\csname res@volt-s-official-scannet@iou-otherfurniture\endcsname{64.8}
\expandafter\def\csname res@volt-s-official-scannet@iou-otherfurniture-pm\endcsname{64.8}
\expandafter\def\csname res@volt-s-official-scannet200@n\endcsname{1}
\expandafter\def\csname res@volt-s-official-scannet200@miou\endcsname{34.4}
\expandafter\def\csname res@volt-s-official-scannet200@miou-pm\endcsname{34.4}
\expandafter\def\csname res@volt-s-official-scannet200@macc\endcsname{44.4}
\expandafter\def\csname res@volt-s-official-scannet200@macc-pm\endcsname{44.4}
\expandafter\def\csname res@volt-s-official-scannet200@allacc\endcsname{83.2}
\expandafter\def\csname res@volt-s-official-scannet200@allacc-pm\endcsname{83.2}
\expandafter\def\csname res@volt-s-official-scannet200@iou-wall\endcsname{79.8}
\expandafter\def\csname res@volt-s-official-scannet200@iou-wall-pm\endcsname{79.8}
\expandafter\def\csname res@volt-s-official-scannet200@iou-chair\endcsname{78.4}
\expandafter\def\csname res@volt-s-official-scannet200@iou-chair-pm\endcsname{78.4}
\expandafter\def\csname res@volt-s-official-scannet200@iou-floor\endcsname{94.5}
\expandafter\def\csname res@volt-s-official-scannet200@iou-floor-pm\endcsname{94.5}
\expandafter\def\csname res@volt-s-official-scannet200@iou-table\endcsname{67.2}
\expandafter\def\csname res@volt-s-official-scannet200@iou-table-pm\endcsname{67.2}
\expandafter\def\csname res@volt-s-official-scannet200@iou-door\endcsname{61.3}
\expandafter\def\csname res@volt-s-official-scannet200@iou-door-pm\endcsname{61.3}
\expandafter\def\csname res@volt-s-official-scannet200@iou-couch\endcsname{79.9}
\expandafter\def\csname res@volt-s-official-scannet200@iou-couch-pm\endcsname{79.9}
\expandafter\def\csname res@volt-s-official-scannet200@iou-cabinet\endcsname{43.5}
\expandafter\def\csname res@volt-s-official-scannet200@iou-cabinet-pm\endcsname{43.5}
\expandafter\def\csname res@volt-s-official-scannet200@iou-shelf\endcsname{48.3}
\expandafter\def\csname res@volt-s-official-scannet200@iou-shelf-pm\endcsname{48.3}
\expandafter\def\csname res@volt-s-official-scannet200@iou-desk\endcsname{64.4}
\expandafter\def\csname res@volt-s-official-scannet200@iou-desk-pm\endcsname{64.4}
\expandafter\def\csname res@volt-s-official-scannet200@iou-office chair\endcsname{18.1}
\expandafter\def\csname res@volt-s-official-scannet200@iou-office chair-pm\endcsname{18.1}
\expandafter\def\csname res@volt-s-official-scannet200@iou-bed\endcsname{81.4}
\expandafter\def\csname res@volt-s-official-scannet200@iou-bed-pm\endcsname{81.4}
\expandafter\def\csname res@volt-s-official-scannet200@iou-pillow\endcsname{57.0}
\expandafter\def\csname res@volt-s-official-scannet200@iou-pillow-pm\endcsname{57.0}
\expandafter\def\csname res@volt-s-official-scannet200@iou-sink\endcsname{71.8}
\expandafter\def\csname res@volt-s-official-scannet200@iou-sink-pm\endcsname{71.8}
\expandafter\def\csname res@volt-s-official-scannet200@iou-picture\endcsname{35.6}
\expandafter\def\csname res@volt-s-official-scannet200@iou-picture-pm\endcsname{35.6}
\expandafter\def\csname res@volt-s-official-scannet200@iou-window\endcsname{66.4}
\expandafter\def\csname res@volt-s-official-scannet200@iou-window-pm\endcsname{66.4}
\expandafter\def\csname res@volt-s-official-scannet200@iou-toilet\endcsname{92.5}
\expandafter\def\csname res@volt-s-official-scannet200@iou-toilet-pm\endcsname{92.5}
\expandafter\def\csname res@volt-s-official-scannet200@iou-bookshelf\endcsname{51.9}
\expandafter\def\csname res@volt-s-official-scannet200@iou-bookshelf-pm\endcsname{51.9}
\expandafter\def\csname res@volt-s-official-scannet200@iou-monitor\endcsname{89.0}
\expandafter\def\csname res@volt-s-official-scannet200@iou-monitor-pm\endcsname{89.0}
\expandafter\def\csname res@volt-s-official-scannet200@iou-curtain\endcsname{77.3}
\expandafter\def\csname res@volt-s-official-scannet200@iou-curtain-pm\endcsname{77.3}
\expandafter\def\csname res@volt-s-official-scannet200@iou-book\endcsname{32.3}
\expandafter\def\csname res@volt-s-official-scannet200@iou-book-pm\endcsname{32.3}
\expandafter\def\csname res@volt-s-official-scannet200@iou-armchair\endcsname{54.2}
\expandafter\def\csname res@volt-s-official-scannet200@iou-armchair-pm\endcsname{54.2}
\expandafter\def\csname res@volt-s-official-scannet200@iou-coffee table\endcsname{58.3}
\expandafter\def\csname res@volt-s-official-scannet200@iou-coffee table-pm\endcsname{58.3}
\expandafter\def\csname res@volt-s-official-scannet200@iou-box\endcsname{38.7}
\expandafter\def\csname res@volt-s-official-scannet200@iou-box-pm\endcsname{38.7}
\expandafter\def\csname res@volt-s-official-scannet200@iou-refrigerator\endcsname{66.6}
\expandafter\def\csname res@volt-s-official-scannet200@iou-refrigerator-pm\endcsname{66.6}
\expandafter\def\csname res@volt-s-official-scannet200@iou-lamp\endcsname{74.3}
\expandafter\def\csname res@volt-s-official-scannet200@iou-lamp-pm\endcsname{74.3}
\expandafter\def\csname res@volt-s-official-scannet200@iou-kitchen cabinet\endcsname{77.2}
\expandafter\def\csname res@volt-s-official-scannet200@iou-kitchen cabinet-pm\endcsname{77.2}
\expandafter\def\csname res@volt-s-official-scannet200@iou-towel\endcsname{57.4}
\expandafter\def\csname res@volt-s-official-scannet200@iou-towel-pm\endcsname{57.4}
\expandafter\def\csname res@volt-s-official-scannet200@iou-clothes\endcsname{22.2}
\expandafter\def\csname res@volt-s-official-scannet200@iou-clothes-pm\endcsname{22.2}
\expandafter\def\csname res@volt-s-official-scannet200@iou-tv\endcsname{90.9}
\expandafter\def\csname res@volt-s-official-scannet200@iou-tv-pm\endcsname{90.9}
\expandafter\def\csname res@volt-s-official-scannet200@iou-nightstand\endcsname{61.6}
\expandafter\def\csname res@volt-s-official-scannet200@iou-nightstand-pm\endcsname{61.6}
\expandafter\def\csname res@volt-s-official-scannet200@iou-counter\endcsname{33.1}
\expandafter\def\csname res@volt-s-official-scannet200@iou-counter-pm\endcsname{33.1}
\expandafter\def\csname res@volt-s-official-scannet200@iou-dresser\endcsname{42.5}
\expandafter\def\csname res@volt-s-official-scannet200@iou-dresser-pm\endcsname{42.5}
\expandafter\def\csname res@volt-s-official-scannet200@iou-stool\endcsname{49.6}
\expandafter\def\csname res@volt-s-official-scannet200@iou-stool-pm\endcsname{49.6}
\expandafter\def\csname res@volt-s-official-scannet200@iou-cushion\endcsname{0.0}
\expandafter\def\csname res@volt-s-official-scannet200@iou-cushion-pm\endcsname{0.0}
\expandafter\def\csname res@volt-s-official-scannet200@iou-plant\endcsname{78.1}
\expandafter\def\csname res@volt-s-official-scannet200@iou-plant-pm\endcsname{78.1}
\expandafter\def\csname res@volt-s-official-scannet200@iou-ceiling\endcsname{92.2}
\expandafter\def\csname res@volt-s-official-scannet200@iou-ceiling-pm\endcsname{92.2}
\expandafter\def\csname res@volt-s-official-scannet200@iou-bathtub\endcsname{82.2}
\expandafter\def\csname res@volt-s-official-scannet200@iou-bathtub-pm\endcsname{82.2}
\expandafter\def\csname res@volt-s-official-scannet200@iou-end table\endcsname{18.5}
\expandafter\def\csname res@volt-s-official-scannet200@iou-end table-pm\endcsname{18.5}
\expandafter\def\csname res@volt-s-official-scannet200@iou-dining table\endcsname{0.0}
\expandafter\def\csname res@volt-s-official-scannet200@iou-dining table-pm\endcsname{0.0}
\expandafter\def\csname res@volt-s-official-scannet200@iou-keyboard\endcsname{39.7}
\expandafter\def\csname res@volt-s-official-scannet200@iou-keyboard-pm\endcsname{39.7}
\expandafter\def\csname res@volt-s-official-scannet200@iou-bag\endcsname{21.2}
\expandafter\def\csname res@volt-s-official-scannet200@iou-bag-pm\endcsname{21.2}
\expandafter\def\csname res@volt-s-official-scannet200@iou-backpack\endcsname{63.2}
\expandafter\def\csname res@volt-s-official-scannet200@iou-backpack-pm\endcsname{63.2}
\expandafter\def\csname res@volt-s-official-scannet200@iou-toilet paper\endcsname{50.6}
\expandafter\def\csname res@volt-s-official-scannet200@iou-toilet paper-pm\endcsname{50.6}
\expandafter\def\csname res@volt-s-official-scannet200@iou-printer\endcsname{58.3}
\expandafter\def\csname res@volt-s-official-scannet200@iou-printer-pm\endcsname{58.3}
\expandafter\def\csname res@volt-s-official-scannet200@iou-tv stand\endcsname{50.1}
\expandafter\def\csname res@volt-s-official-scannet200@iou-tv stand-pm\endcsname{50.1}
\expandafter\def\csname res@volt-s-official-scannet200@iou-whiteboard\endcsname{45.5}
\expandafter\def\csname res@volt-s-official-scannet200@iou-whiteboard-pm\endcsname{45.5}
\expandafter\def\csname res@volt-s-official-scannet200@iou-blanket\endcsname{3.8}
\expandafter\def\csname res@volt-s-official-scannet200@iou-blanket-pm\endcsname{3.8}
\expandafter\def\csname res@volt-s-official-scannet200@iou-shower curtain\endcsname{69.2}
\expandafter\def\csname res@volt-s-official-scannet200@iou-shower curtain-pm\endcsname{69.2}
\expandafter\def\csname res@volt-s-official-scannet200@iou-trash can\endcsname{67.4}
\expandafter\def\csname res@volt-s-official-scannet200@iou-trash can-pm\endcsname{67.4}
\expandafter\def\csname res@volt-s-official-scannet200@iou-closet\endcsname{10.7}
\expandafter\def\csname res@volt-s-official-scannet200@iou-closet-pm\endcsname{10.7}
\expandafter\def\csname res@volt-s-official-scannet200@iou-stairs\endcsname{39.1}
\expandafter\def\csname res@volt-s-official-scannet200@iou-stairs-pm\endcsname{39.1}
\expandafter\def\csname res@volt-s-official-scannet200@iou-microwave\endcsname{66.8}
\expandafter\def\csname res@volt-s-official-scannet200@iou-microwave-pm\endcsname{66.8}
\expandafter\def\csname res@volt-s-official-scannet200@iou-stove\endcsname{67.3}
\expandafter\def\csname res@volt-s-official-scannet200@iou-stove-pm\endcsname{67.3}
\expandafter\def\csname res@volt-s-official-scannet200@iou-shoe\endcsname{44.5}
\expandafter\def\csname res@volt-s-official-scannet200@iou-shoe-pm\endcsname{44.5}
\expandafter\def\csname res@volt-s-official-scannet200@iou-computer tower\endcsname{59.2}
\expandafter\def\csname res@volt-s-official-scannet200@iou-computer tower-pm\endcsname{59.2}
\expandafter\def\csname res@volt-s-official-scannet200@iou-bottle\endcsname{17.7}
\expandafter\def\csname res@volt-s-official-scannet200@iou-bottle-pm\endcsname{17.7}
\expandafter\def\csname res@volt-s-official-scannet200@iou-bin\endcsname{0.0}
\expandafter\def\csname res@volt-s-official-scannet200@iou-bin-pm\endcsname{0.0}
\expandafter\def\csname res@volt-s-official-scannet200@iou-ottoman\endcsname{52.8}
\expandafter\def\csname res@volt-s-official-scannet200@iou-ottoman-pm\endcsname{52.8}
\expandafter\def\csname res@volt-s-official-scannet200@iou-bench\endcsname{38.7}
\expandafter\def\csname res@volt-s-official-scannet200@iou-bench-pm\endcsname{38.7}
\expandafter\def\csname res@volt-s-official-scannet200@iou-board\endcsname{9.4}
\expandafter\def\csname res@volt-s-official-scannet200@iou-board-pm\endcsname{9.4}
\expandafter\def\csname res@volt-s-official-scannet200@iou-washing machine\endcsname{84.0}
\expandafter\def\csname res@volt-s-official-scannet200@iou-washing machine-pm\endcsname{84.0}
\expandafter\def\csname res@volt-s-official-scannet200@iou-mirror\endcsname{50.2}
\expandafter\def\csname res@volt-s-official-scannet200@iou-mirror-pm\endcsname{50.2}
\expandafter\def\csname res@volt-s-official-scannet200@iou-copier\endcsname{83.8}
\expandafter\def\csname res@volt-s-official-scannet200@iou-copier-pm\endcsname{83.8}
\expandafter\def\csname res@volt-s-official-scannet200@iou-basket\endcsname{4.1}
\expandafter\def\csname res@volt-s-official-scannet200@iou-basket-pm\endcsname{4.1}
\expandafter\def\csname res@volt-s-official-scannet200@iou-sofa chair\endcsname{15.7}
\expandafter\def\csname res@volt-s-official-scannet200@iou-sofa chair-pm\endcsname{15.7}
\expandafter\def\csname res@volt-s-official-scannet200@iou-file cabinet\endcsname{44.0}
\expandafter\def\csname res@volt-s-official-scannet200@iou-file cabinet-pm\endcsname{44.0}
\expandafter\def\csname res@volt-s-official-scannet200@iou-fan\endcsname{56.8}
\expandafter\def\csname res@volt-s-official-scannet200@iou-fan-pm\endcsname{56.8}
\expandafter\def\csname res@volt-s-official-scannet200@iou-laptop\endcsname{55.6}
\expandafter\def\csname res@volt-s-official-scannet200@iou-laptop-pm\endcsname{55.6}
\expandafter\def\csname res@volt-s-official-scannet200@iou-shower\endcsname{56.8}
\expandafter\def\csname res@volt-s-official-scannet200@iou-shower-pm\endcsname{56.8}
\expandafter\def\csname res@volt-s-official-scannet200@iou-paper\endcsname{9.6}
\expandafter\def\csname res@volt-s-official-scannet200@iou-paper-pm\endcsname{9.6}
\expandafter\def\csname res@volt-s-official-scannet200@iou-person\endcsname{18.7}
\expandafter\def\csname res@volt-s-official-scannet200@iou-person-pm\endcsname{18.7}
\expandafter\def\csname res@volt-s-official-scannet200@iou-paper towel dispenser\endcsname{49.3}
\expandafter\def\csname res@volt-s-official-scannet200@iou-paper towel dispenser-pm\endcsname{49.3}
\expandafter\def\csname res@volt-s-official-scannet200@iou-oven\endcsname{15.7}
\expandafter\def\csname res@volt-s-official-scannet200@iou-oven-pm\endcsname{15.7}
\expandafter\def\csname res@volt-s-official-scannet200@iou-blinds\endcsname{0.0}
\expandafter\def\csname res@volt-s-official-scannet200@iou-blinds-pm\endcsname{0.0}
\expandafter\def\csname res@volt-s-official-scannet200@iou-rack\endcsname{0.0}
\expandafter\def\csname res@volt-s-official-scannet200@iou-rack-pm\endcsname{0.0}
\expandafter\def\csname res@volt-s-official-scannet200@iou-plate\endcsname{0.0}
\expandafter\def\csname res@volt-s-official-scannet200@iou-plate-pm\endcsname{0.0}
\expandafter\def\csname res@volt-s-official-scannet200@iou-blackboard\endcsname{66.3}
\expandafter\def\csname res@volt-s-official-scannet200@iou-blackboard-pm\endcsname{66.3}
\expandafter\def\csname res@volt-s-official-scannet200@iou-piano\endcsname{59.3}
\expandafter\def\csname res@volt-s-official-scannet200@iou-piano-pm\endcsname{59.3}
\expandafter\def\csname res@volt-s-official-scannet200@iou-suitcase\endcsname{60.4}
\expandafter\def\csname res@volt-s-official-scannet200@iou-suitcase-pm\endcsname{60.4}
\expandafter\def\csname res@volt-s-official-scannet200@iou-rail\endcsname{7.2}
\expandafter\def\csname res@volt-s-official-scannet200@iou-rail-pm\endcsname{7.2}
\expandafter\def\csname res@volt-s-official-scannet200@iou-radiator\endcsname{70.5}
\expandafter\def\csname res@volt-s-official-scannet200@iou-radiator-pm\endcsname{70.5}
\expandafter\def\csname res@volt-s-official-scannet200@iou-recycling bin\endcsname{59.9}
\expandafter\def\csname res@volt-s-official-scannet200@iou-recycling bin-pm\endcsname{59.9}
\expandafter\def\csname res@volt-s-official-scannet200@iou-container\endcsname{0.0}
\expandafter\def\csname res@volt-s-official-scannet200@iou-container-pm\endcsname{0.0}
\expandafter\def\csname res@volt-s-official-scannet200@iou-wardrobe\endcsname{27.6}
\expandafter\def\csname res@volt-s-official-scannet200@iou-wardrobe-pm\endcsname{27.6}
\expandafter\def\csname res@volt-s-official-scannet200@iou-soap dispenser\endcsname{56.1}
\expandafter\def\csname res@volt-s-official-scannet200@iou-soap dispenser-pm\endcsname{56.1}
\expandafter\def\csname res@volt-s-official-scannet200@iou-telephone\endcsname{48.8}
\expandafter\def\csname res@volt-s-official-scannet200@iou-telephone-pm\endcsname{48.8}
\expandafter\def\csname res@volt-s-official-scannet200@iou-bucket\endcsname{32.9}
\expandafter\def\csname res@volt-s-official-scannet200@iou-bucket-pm\endcsname{32.9}
\expandafter\def\csname res@volt-s-official-scannet200@iou-clock\endcsname{23.1}
\expandafter\def\csname res@volt-s-official-scannet200@iou-clock-pm\endcsname{23.1}
\expandafter\def\csname res@volt-s-official-scannet200@iou-stand\endcsname{0.0}
\expandafter\def\csname res@volt-s-official-scannet200@iou-stand-pm\endcsname{0.0}
\expandafter\def\csname res@volt-s-official-scannet200@iou-light\endcsname{4.1}
\expandafter\def\csname res@volt-s-official-scannet200@iou-light-pm\endcsname{4.1}
\expandafter\def\csname res@volt-s-official-scannet200@iou-laundry basket\endcsname{30.6}
\expandafter\def\csname res@volt-s-official-scannet200@iou-laundry basket-pm\endcsname{30.6}
\expandafter\def\csname res@volt-s-official-scannet200@iou-pipe\endcsname{6.0}
\expandafter\def\csname res@volt-s-official-scannet200@iou-pipe-pm\endcsname{6.0}
\expandafter\def\csname res@volt-s-official-scannet200@iou-clothes dryer\endcsname{72.4}
\expandafter\def\csname res@volt-s-official-scannet200@iou-clothes dryer-pm\endcsname{72.4}
\expandafter\def\csname res@volt-s-official-scannet200@iou-guitar\endcsname{53.7}
\expandafter\def\csname res@volt-s-official-scannet200@iou-guitar-pm\endcsname{53.7}
\expandafter\def\csname res@volt-s-official-scannet200@iou-toilet paper holder\endcsname{0.0}
\expandafter\def\csname res@volt-s-official-scannet200@iou-toilet paper holder-pm\endcsname{0.0}
\expandafter\def\csname res@volt-s-official-scannet200@iou-seat\endcsname{0.0}
\expandafter\def\csname res@volt-s-official-scannet200@iou-seat-pm\endcsname{0.0}
\expandafter\def\csname res@volt-s-official-scannet200@iou-speaker\endcsname{0.0}
\expandafter\def\csname res@volt-s-official-scannet200@iou-speaker-pm\endcsname{0.0}
\expandafter\def\csname res@volt-s-official-scannet200@iou-column\endcsname{0.0}
\expandafter\def\csname res@volt-s-official-scannet200@iou-column-pm\endcsname{0.0}
\expandafter\def\csname res@volt-s-official-scannet200@iou-bicycle\endcsname{0.0}
\expandafter\def\csname res@volt-s-official-scannet200@iou-bicycle-pm\endcsname{0.0}
\expandafter\def\csname res@volt-s-official-scannet200@iou-ladder\endcsname{65.4}
\expandafter\def\csname res@volt-s-official-scannet200@iou-ladder-pm\endcsname{65.4}
\expandafter\def\csname res@volt-s-official-scannet200@iou-bathroom stall\endcsname{56.9}
\expandafter\def\csname res@volt-s-official-scannet200@iou-bathroom stall-pm\endcsname{56.9}
\expandafter\def\csname res@volt-s-official-scannet200@iou-shower wall\endcsname{51.6}
\expandafter\def\csname res@volt-s-official-scannet200@iou-shower wall-pm\endcsname{51.6}
\expandafter\def\csname res@volt-s-official-scannet200@iou-cup\endcsname{17.8}
\expandafter\def\csname res@volt-s-official-scannet200@iou-cup-pm\endcsname{17.8}
\expandafter\def\csname res@volt-s-official-scannet200@iou-jacket\endcsname{24.4}
\expandafter\def\csname res@volt-s-official-scannet200@iou-jacket-pm\endcsname{24.4}
\expandafter\def\csname res@volt-s-official-scannet200@iou-storage bin\endcsname{0.0}
\expandafter\def\csname res@volt-s-official-scannet200@iou-storage bin-pm\endcsname{0.0}
\expandafter\def\csname res@volt-s-official-scannet200@iou-coffee maker\endcsname{87.4}
\expandafter\def\csname res@volt-s-official-scannet200@iou-coffee maker-pm\endcsname{87.4}
\expandafter\def\csname res@volt-s-official-scannet200@iou-dishwasher\endcsname{58.9}
\expandafter\def\csname res@volt-s-official-scannet200@iou-dishwasher-pm\endcsname{58.9}
\expandafter\def\csname res@volt-s-official-scannet200@iou-paper towel roll\endcsname{56.5}
\expandafter\def\csname res@volt-s-official-scannet200@iou-paper towel roll-pm\endcsname{56.5}
\expandafter\def\csname res@volt-s-official-scannet200@iou-machine\endcsname{0.0}
\expandafter\def\csname res@volt-s-official-scannet200@iou-machine-pm\endcsname{0.0}
\expandafter\def\csname res@volt-s-official-scannet200@iou-mat\endcsname{29.9}
\expandafter\def\csname res@volt-s-official-scannet200@iou-mat-pm\endcsname{29.9}
\expandafter\def\csname res@volt-s-official-scannet200@iou-windowsill\endcsname{6.8}
\expandafter\def\csname res@volt-s-official-scannet200@iou-windowsill-pm\endcsname{6.8}
\expandafter\def\csname res@volt-s-official-scannet200@iou-bar\endcsname{14.4}
\expandafter\def\csname res@volt-s-official-scannet200@iou-bar-pm\endcsname{14.4}
\expandafter\def\csname res@volt-s-official-scannet200@iou-toaster\endcsname{68.7}
\expandafter\def\csname res@volt-s-official-scannet200@iou-toaster-pm\endcsname{68.7}
\expandafter\def\csname res@volt-s-official-scannet200@iou-bulletin board\endcsname{58.0}
\expandafter\def\csname res@volt-s-official-scannet200@iou-bulletin board-pm\endcsname{58.0}
\expandafter\def\csname res@volt-s-official-scannet200@iou-ironing board\endcsname{22.0}
\expandafter\def\csname res@volt-s-official-scannet200@iou-ironing board-pm\endcsname{22.0}
\expandafter\def\csname res@volt-s-official-scannet200@iou-fireplace\endcsname{54.7}
\expandafter\def\csname res@volt-s-official-scannet200@iou-fireplace-pm\endcsname{54.7}
\expandafter\def\csname res@volt-s-official-scannet200@iou-soap dish\endcsname{32.6}
\expandafter\def\csname res@volt-s-official-scannet200@iou-soap dish-pm\endcsname{32.6}
\expandafter\def\csname res@volt-s-official-scannet200@iou-kitchen counter\endcsname{54.5}
\expandafter\def\csname res@volt-s-official-scannet200@iou-kitchen counter-pm\endcsname{54.5}
\expandafter\def\csname res@volt-s-official-scannet200@iou-doorframe\endcsname{44.2}
\expandafter\def\csname res@volt-s-official-scannet200@iou-doorframe-pm\endcsname{44.2}
\expandafter\def\csname res@volt-s-official-scannet200@iou-toilet paper dispenser\endcsname{35.2}
\expandafter\def\csname res@volt-s-official-scannet200@iou-toilet paper dispenser-pm\endcsname{35.2}
\expandafter\def\csname res@volt-s-official-scannet200@iou-mini fridge\endcsname{57.8}
\expandafter\def\csname res@volt-s-official-scannet200@iou-mini fridge-pm\endcsname{57.8}
\expandafter\def\csname res@volt-s-official-scannet200@iou-fire extinguisher\endcsname{33.1}
\expandafter\def\csname res@volt-s-official-scannet200@iou-fire extinguisher-pm\endcsname{33.1}
\expandafter\def\csname res@volt-s-official-scannet200@iou-ball\endcsname{51.6}
\expandafter\def\csname res@volt-s-official-scannet200@iou-ball-pm\endcsname{51.6}
\expandafter\def\csname res@volt-s-official-scannet200@iou-hat\endcsname{0.0}
\expandafter\def\csname res@volt-s-official-scannet200@iou-hat-pm\endcsname{0.0}
\expandafter\def\csname res@volt-s-official-scannet200@iou-shower curtain rod\endcsname{40.2}
\expandafter\def\csname res@volt-s-official-scannet200@iou-shower curtain rod-pm\endcsname{40.2}
\expandafter\def\csname res@volt-s-official-scannet200@iou-water cooler\endcsname{86.3}
\expandafter\def\csname res@volt-s-official-scannet200@iou-water cooler-pm\endcsname{86.3}
\expandafter\def\csname res@volt-s-official-scannet200@iou-paper cutter\endcsname{57.2}
\expandafter\def\csname res@volt-s-official-scannet200@iou-paper cutter-pm\endcsname{57.2}
\expandafter\def\csname res@volt-s-official-scannet200@iou-tray\endcsname{16.4}
\expandafter\def\csname res@volt-s-official-scannet200@iou-tray-pm\endcsname{16.4}
\expandafter\def\csname res@volt-s-official-scannet200@iou-shower door\endcsname{36.1}
\expandafter\def\csname res@volt-s-official-scannet200@iou-shower door-pm\endcsname{36.1}
\expandafter\def\csname res@volt-s-official-scannet200@iou-pillar\endcsname{1.5}
\expandafter\def\csname res@volt-s-official-scannet200@iou-pillar-pm\endcsname{1.5}
\expandafter\def\csname res@volt-s-official-scannet200@iou-ledge\endcsname{10.7}
\expandafter\def\csname res@volt-s-official-scannet200@iou-ledge-pm\endcsname{10.7}
\expandafter\def\csname res@volt-s-official-scannet200@iou-toaster oven\endcsname{0.0}
\expandafter\def\csname res@volt-s-official-scannet200@iou-toaster oven-pm\endcsname{0.0}
\expandafter\def\csname res@volt-s-official-scannet200@iou-mouse\endcsname{8.7}
\expandafter\def\csname res@volt-s-official-scannet200@iou-mouse-pm\endcsname{8.7}
\expandafter\def\csname res@volt-s-official-scannet200@iou-toilet seat cover dispenser\endcsname{77.6}
\expandafter\def\csname res@volt-s-official-scannet200@iou-toilet seat cover dispenser-pm\endcsname{77.6}
\expandafter\def\csname res@volt-s-official-scannet200@iou-furniture\endcsname{0.0}
\expandafter\def\csname res@volt-s-official-scannet200@iou-furniture-pm\endcsname{0.0}
\expandafter\def\csname res@volt-s-official-scannet200@iou-cart\endcsname{17.0}
\expandafter\def\csname res@volt-s-official-scannet200@iou-cart-pm\endcsname{17.0}
\expandafter\def\csname res@volt-s-official-scannet200@iou-storage container\endcsname{0.0}
\expandafter\def\csname res@volt-s-official-scannet200@iou-storage container-pm\endcsname{0.0}
\expandafter\def\csname res@volt-s-official-scannet200@iou-scale\endcsname{60.7}
\expandafter\def\csname res@volt-s-official-scannet200@iou-scale-pm\endcsname{60.7}
\expandafter\def\csname res@volt-s-official-scannet200@iou-tissue box\endcsname{17.5}
\expandafter\def\csname res@volt-s-official-scannet200@iou-tissue box-pm\endcsname{17.5}
\expandafter\def\csname res@volt-s-official-scannet200@iou-light switch\endcsname{0.0}
\expandafter\def\csname res@volt-s-official-scannet200@iou-light switch-pm\endcsname{0.0}
\expandafter\def\csname res@volt-s-official-scannet200@iou-crate\endcsname{0.0}
\expandafter\def\csname res@volt-s-official-scannet200@iou-crate-pm\endcsname{0.0}
\expandafter\def\csname res@volt-s-official-scannet200@iou-power outlet\endcsname{0.0}
\expandafter\def\csname res@volt-s-official-scannet200@iou-power outlet-pm\endcsname{0.0}
\expandafter\def\csname res@volt-s-official-scannet200@iou-decoration\endcsname{25.1}
\expandafter\def\csname res@volt-s-official-scannet200@iou-decoration-pm\endcsname{25.1}
\expandafter\def\csname res@volt-s-official-scannet200@iou-sign\endcsname{36.9}
\expandafter\def\csname res@volt-s-official-scannet200@iou-sign-pm\endcsname{36.9}
\expandafter\def\csname res@volt-s-official-scannet200@iou-projector\endcsname{25.4}
\expandafter\def\csname res@volt-s-official-scannet200@iou-projector-pm\endcsname{25.4}
\expandafter\def\csname res@volt-s-official-scannet200@iou-closet door\endcsname{0.4}
\expandafter\def\csname res@volt-s-official-scannet200@iou-closet door-pm\endcsname{0.4}
\expandafter\def\csname res@volt-s-official-scannet200@iou-vacuum cleaner\endcsname{95.0}
\expandafter\def\csname res@volt-s-official-scannet200@iou-vacuum cleaner-pm\endcsname{95.0}
\expandafter\def\csname res@volt-s-official-scannet200@iou-candle\endcsname{0.0}
\expandafter\def\csname res@volt-s-official-scannet200@iou-candle-pm\endcsname{0.0}
\expandafter\def\csname res@volt-s-official-scannet200@iou-plunger\endcsname{36.0}
\expandafter\def\csname res@volt-s-official-scannet200@iou-plunger-pm\endcsname{36.0}
\expandafter\def\csname res@volt-s-official-scannet200@iou-stuffed animal\endcsname{0.0}
\expandafter\def\csname res@volt-s-official-scannet200@iou-stuffed animal-pm\endcsname{0.0}
\expandafter\def\csname res@volt-s-official-scannet200@iou-headphones\endcsname{0.0}
\expandafter\def\csname res@volt-s-official-scannet200@iou-headphones-pm\endcsname{0.0}
\expandafter\def\csname res@volt-s-official-scannet200@iou-dish rack\endcsname{89.5}
\expandafter\def\csname res@volt-s-official-scannet200@iou-dish rack-pm\endcsname{89.5}
\expandafter\def\csname res@volt-s-official-scannet200@iou-broom\endcsname{0.0}
\expandafter\def\csname res@volt-s-official-scannet200@iou-broom-pm\endcsname{0.0}
\expandafter\def\csname res@volt-s-official-scannet200@iou-guitar case\endcsname{0.0}
\expandafter\def\csname res@volt-s-official-scannet200@iou-guitar case-pm\endcsname{0.0}
\expandafter\def\csname res@volt-s-official-scannet200@iou-range hood\endcsname{82.0}
\expandafter\def\csname res@volt-s-official-scannet200@iou-range hood-pm\endcsname{82.0}
\expandafter\def\csname res@volt-s-official-scannet200@iou-dustpan\endcsname{4.6}
\expandafter\def\csname res@volt-s-official-scannet200@iou-dustpan-pm\endcsname{4.6}
\expandafter\def\csname res@volt-s-official-scannet200@iou-hair dryer\endcsname{0.0}
\expandafter\def\csname res@volt-s-official-scannet200@iou-hair dryer-pm\endcsname{0.0}
\expandafter\def\csname res@volt-s-official-scannet200@iou-water bottle\endcsname{0.0}
\expandafter\def\csname res@volt-s-official-scannet200@iou-water bottle-pm\endcsname{0.0}
\expandafter\def\csname res@volt-s-official-scannet200@iou-handicap bar\endcsname{6.0}
\expandafter\def\csname res@volt-s-official-scannet200@iou-handicap bar-pm\endcsname{6.0}
\expandafter\def\csname res@volt-s-official-scannet200@iou-purse\endcsname{0.0}
\expandafter\def\csname res@volt-s-official-scannet200@iou-purse-pm\endcsname{0.0}
\expandafter\def\csname res@volt-s-official-scannet200@iou-vent\endcsname{0.0}
\expandafter\def\csname res@volt-s-official-scannet200@iou-vent-pm\endcsname{0.0}
\expandafter\def\csname res@volt-s-official-scannet200@iou-shower floor\endcsname{65.4}
\expandafter\def\csname res@volt-s-official-scannet200@iou-shower floor-pm\endcsname{65.4}
\expandafter\def\csname res@volt-s-official-scannet200@iou-water pitcher\endcsname{0.0}
\expandafter\def\csname res@volt-s-official-scannet200@iou-water pitcher-pm\endcsname{0.0}
\expandafter\def\csname res@volt-s-official-scannet200@iou-mailbox\endcsname{46.1}
\expandafter\def\csname res@volt-s-official-scannet200@iou-mailbox-pm\endcsname{46.1}
\expandafter\def\csname res@volt-s-official-scannet200@iou-bowl\endcsname{24.1}
\expandafter\def\csname res@volt-s-official-scannet200@iou-bowl-pm\endcsname{24.1}
\expandafter\def\csname res@volt-s-official-scannet200@iou-paper bag\endcsname{2.7}
\expandafter\def\csname res@volt-s-official-scannet200@iou-paper bag-pm\endcsname{2.7}
\expandafter\def\csname res@volt-s-official-scannet200@iou-alarm clock\endcsname{0.0}
\expandafter\def\csname res@volt-s-official-scannet200@iou-alarm clock-pm\endcsname{0.0}
\expandafter\def\csname res@volt-s-official-scannet200@iou-music stand\endcsname{0.0}
\expandafter\def\csname res@volt-s-official-scannet200@iou-music stand-pm\endcsname{0.0}
\expandafter\def\csname res@volt-s-official-scannet200@iou-projector screen\endcsname{2.2}
\expandafter\def\csname res@volt-s-official-scannet200@iou-projector screen-pm\endcsname{2.2}
\expandafter\def\csname res@volt-s-official-scannet200@iou-divider\endcsname{67.0}
\expandafter\def\csname res@volt-s-official-scannet200@iou-divider-pm\endcsname{67.0}
\expandafter\def\csname res@volt-s-official-scannet200@iou-laundry detergent\endcsname{0.0}
\expandafter\def\csname res@volt-s-official-scannet200@iou-laundry detergent-pm\endcsname{0.0}
\expandafter\def\csname res@volt-s-official-scannet200@iou-bathroom counter\endcsname{0.0}
\expandafter\def\csname res@volt-s-official-scannet200@iou-bathroom counter-pm\endcsname{0.0}
\expandafter\def\csname res@volt-s-official-scannet200@iou-object\endcsname{4.4}
\expandafter\def\csname res@volt-s-official-scannet200@iou-object-pm\endcsname{4.4}
\expandafter\def\csname res@volt-s-official-scannet200@iou-bathroom vanity\endcsname{44.3}
\expandafter\def\csname res@volt-s-official-scannet200@iou-bathroom vanity-pm\endcsname{44.3}
\expandafter\def\csname res@volt-s-official-scannet200@iou-closet wall\endcsname{15.3}
\expandafter\def\csname res@volt-s-official-scannet200@iou-closet wall-pm\endcsname{15.3}
\expandafter\def\csname res@volt-s-official-scannet200@iou-laundry hamper\endcsname{35.1}
\expandafter\def\csname res@volt-s-official-scannet200@iou-laundry hamper-pm\endcsname{35.1}
\expandafter\def\csname res@volt-s-official-scannet200@iou-bathroom stall door\endcsname{55.0}
\expandafter\def\csname res@volt-s-official-scannet200@iou-bathroom stall door-pm\endcsname{55.0}
\expandafter\def\csname res@volt-s-official-scannet200@iou-ceiling light\endcsname{31.3}
\expandafter\def\csname res@volt-s-official-scannet200@iou-ceiling light-pm\endcsname{31.3}
\expandafter\def\csname res@volt-s-official-scannet200@iou-trash bin\endcsname{37.4}
\expandafter\def\csname res@volt-s-official-scannet200@iou-trash bin-pm\endcsname{37.4}
\expandafter\def\csname res@volt-s-official-scannet200@iou-dumbbell\endcsname{52.2}
\expandafter\def\csname res@volt-s-official-scannet200@iou-dumbbell-pm\endcsname{52.2}
\expandafter\def\csname res@volt-s-official-scannet200@iou-stair rail\endcsname{36.1}
\expandafter\def\csname res@volt-s-official-scannet200@iou-stair rail-pm\endcsname{36.1}
\expandafter\def\csname res@volt-s-official-scannet200@iou-tube\endcsname{18.9}
\expandafter\def\csname res@volt-s-official-scannet200@iou-tube-pm\endcsname{18.9}
\expandafter\def\csname res@volt-s-official-scannet200@iou-bathroom cabinet\endcsname{20.9}
\expandafter\def\csname res@volt-s-official-scannet200@iou-bathroom cabinet-pm\endcsname{20.9}
\expandafter\def\csname res@volt-s-official-scannet200@iou-cd case\endcsname{0.0}
\expandafter\def\csname res@volt-s-official-scannet200@iou-cd case-pm\endcsname{0.0}
\expandafter\def\csname res@volt-s-official-scannet200@iou-closet rod\endcsname{40.3}
\expandafter\def\csname res@volt-s-official-scannet200@iou-closet rod-pm\endcsname{40.3}
\expandafter\def\csname res@volt-s-official-scannet200@iou-coffee kettle\endcsname{38.4}
\expandafter\def\csname res@volt-s-official-scannet200@iou-coffee kettle-pm\endcsname{38.4}
\expandafter\def\csname res@volt-s-official-scannet200@iou-structure\endcsname{0.0}
\expandafter\def\csname res@volt-s-official-scannet200@iou-structure-pm\endcsname{0.0}
\expandafter\def\csname res@volt-s-official-scannet200@iou-shower head\endcsname{31.1}
\expandafter\def\csname res@volt-s-official-scannet200@iou-shower head-pm\endcsname{31.1}
\expandafter\def\csname res@volt-s-official-scannet200@iou-keyboard piano\endcsname{0.6}
\expandafter\def\csname res@volt-s-official-scannet200@iou-keyboard piano-pm\endcsname{0.6}
\expandafter\def\csname res@volt-s-official-scannet200@iou-case of water bottles\endcsname{0.9}
\expandafter\def\csname res@volt-s-official-scannet200@iou-case of water bottles-pm\endcsname{0.9}
\expandafter\def\csname res@volt-s-official-scannet200@iou-coat rack\endcsname{0.0}
\expandafter\def\csname res@volt-s-official-scannet200@iou-coat rack-pm\endcsname{0.0}
\expandafter\def\csname res@volt-s-official-scannet200@iou-storage organizer\endcsname{0.0}
\expandafter\def\csname res@volt-s-official-scannet200@iou-storage organizer-pm\endcsname{0.0}
\expandafter\def\csname res@volt-s-official-scannet200@iou-folded chair\endcsname{0.0}
\expandafter\def\csname res@volt-s-official-scannet200@iou-folded chair-pm\endcsname{0.0}
\expandafter\def\csname res@volt-s-official-scannet200@iou-fire alarm\endcsname{0.0}
\expandafter\def\csname res@volt-s-official-scannet200@iou-fire alarm-pm\endcsname{0.0}
\expandafter\def\csname res@volt-s-official-scannet200@iou-power strip\endcsname{0.0}
\expandafter\def\csname res@volt-s-official-scannet200@iou-power strip-pm\endcsname{0.0}
\expandafter\def\csname res@volt-s-official-scannet200@iou-calendar\endcsname{0.0}
\expandafter\def\csname res@volt-s-official-scannet200@iou-calendar-pm\endcsname{0.0}
\expandafter\def\csname res@volt-s-official-scannet200@iou-poster\endcsname{0.0}
\expandafter\def\csname res@volt-s-official-scannet200@iou-poster-pm\endcsname{0.0}
\expandafter\def\csname res@volt-s-official-scannet200@iou-potted plant\endcsname{0.0}
\expandafter\def\csname res@volt-s-official-scannet200@iou-potted plant-pm\endcsname{0.0}
\expandafter\def\csname res@volt-s-official-scannet200@iou-luggage\endcsname{0.0}
\expandafter\def\csname res@volt-s-official-scannet200@iou-luggage-pm\endcsname{0.0}
\expandafter\def\csname res@volt-s-official-scannet200@iou-mattress\endcsname{0.0}
\expandafter\def\csname res@volt-s-official-scannet200@iou-mattress-pm\endcsname{0.0}
\expandafter\def\csname res@volt-s-official-nuscenes@n\endcsname{1}
\expandafter\def\csname res@volt-s-official-nuscenes@miou\endcsname{80.5}
\expandafter\def\csname res@volt-s-official-nuscenes@miou-pm\endcsname{80.5}
\expandafter\def\csname res@volt-s-official-nuscenes@macc\endcsname{87.2}
\expandafter\def\csname res@volt-s-official-nuscenes@macc-pm\endcsname{87.2}
\expandafter\def\csname res@volt-s-official-nuscenes@allacc\endcsname{94.5}
\expandafter\def\csname res@volt-s-official-nuscenes@allacc-pm\endcsname{94.5}
\expandafter\def\csname res@volt-s-official-nuscenes@iou-barrier\endcsname{80.5}
\expandafter\def\csname res@volt-s-official-nuscenes@iou-barrier-pm\endcsname{80.5}
\expandafter\def\csname res@volt-s-official-nuscenes@iou-bicycle\endcsname{56.5}
\expandafter\def\csname res@volt-s-official-nuscenes@iou-bicycle-pm\endcsname{56.5}
\expandafter\def\csname res@volt-s-official-nuscenes@iou-bus\endcsname{94.3}
\expandafter\def\csname res@volt-s-official-nuscenes@iou-bus-pm\endcsname{94.3}
\expandafter\def\csname res@volt-s-official-nuscenes@iou-car\endcsname{91.3}
\expandafter\def\csname res@volt-s-official-nuscenes@iou-car-pm\endcsname{91.3}
\expandafter\def\csname res@volt-s-official-nuscenes@iou-construction_vehicle\endcsname{60.5}
\expandafter\def\csname res@volt-s-official-nuscenes@iou-construction_vehicle-pm\endcsname{60.5}
\expandafter\def\csname res@volt-s-official-nuscenes@iou-motorcycle\endcsname{87.3}
\expandafter\def\csname res@volt-s-official-nuscenes@iou-motorcycle-pm\endcsname{87.3}
\expandafter\def\csname res@volt-s-official-nuscenes@iou-pedestrian\endcsname{84.5}
\expandafter\def\csname res@volt-s-official-nuscenes@iou-pedestrian-pm\endcsname{84.5}
\expandafter\def\csname res@volt-s-official-nuscenes@iou-traffic_cone\endcsname{71.3}
\expandafter\def\csname res@volt-s-official-nuscenes@iou-traffic_cone-pm\endcsname{71.3}
\expandafter\def\csname res@volt-s-official-nuscenes@iou-trailer\endcsname{71.7}
\expandafter\def\csname res@volt-s-official-nuscenes@iou-trailer-pm\endcsname{71.7}
\expandafter\def\csname res@volt-s-official-nuscenes@iou-truck\endcsname{83.5}
\expandafter\def\csname res@volt-s-official-nuscenes@iou-truck-pm\endcsname{83.5}
\expandafter\def\csname res@volt-s-official-nuscenes@iou-driveable_surface\endcsname{97.1}
\expandafter\def\csname res@volt-s-official-nuscenes@iou-driveable_surface-pm\endcsname{97.1}
\expandafter\def\csname res@volt-s-official-nuscenes@iou-other_flat\endcsname{75.8}
\expandafter\def\csname res@volt-s-official-nuscenes@iou-other_flat-pm\endcsname{75.8}
\expandafter\def\csname res@volt-s-official-nuscenes@iou-sidewalk\endcsname{76.8}
\expandafter\def\csname res@volt-s-official-nuscenes@iou-sidewalk-pm\endcsname{76.8}
\expandafter\def\csname res@volt-s-official-nuscenes@iou-terrain\endcsname{76.5}
\expandafter\def\csname res@volt-s-official-nuscenes@iou-terrain-pm\endcsname{76.5}
\expandafter\def\csname res@volt-s-official-nuscenes@iou-manmade\endcsname{90.8}
\expandafter\def\csname res@volt-s-official-nuscenes@iou-manmade-pm\endcsname{90.8}
\expandafter\def\csname res@volt-s-official-nuscenes@iou-vegetation\endcsname{89.1}
\expandafter\def\csname res@volt-s-official-nuscenes@iou-vegetation-pm\endcsname{89.1}
\expandafter\def\csname res@volt-s-official-semantickitti@n\endcsname{1}
\expandafter\def\csname res@volt-s-official-semantickitti@miou\endcsname{68.8}
\expandafter\def\csname res@volt-s-official-semantickitti@miou-pm\endcsname{68.8}
\expandafter\def\csname res@volt-s-official-semantickitti@macc\endcsname{76.8}
\expandafter\def\csname res@volt-s-official-semantickitti@macc-pm\endcsname{76.8}
\expandafter\def\csname res@volt-s-official-semantickitti@allacc\endcsname{92.7}
\expandafter\def\csname res@volt-s-official-semantickitti@allacc-pm\endcsname{92.7}
\expandafter\def\csname res@volt-s-official-semantickitti@iou-car\endcsname{96.5}
\expandafter\def\csname res@volt-s-official-semantickitti@iou-car-pm\endcsname{96.5}
\expandafter\def\csname res@volt-s-official-semantickitti@iou-bicycle\endcsname{59.0}
\expandafter\def\csname res@volt-s-official-semantickitti@iou-bicycle-pm\endcsname{59.0}
\expandafter\def\csname res@volt-s-official-semantickitti@iou-motorcycle\endcsname{84.4}
\expandafter\def\csname res@volt-s-official-semantickitti@iou-motorcycle-pm\endcsname{84.4}
\expandafter\def\csname res@volt-s-official-semantickitti@iou-truck\endcsname{85.2}
\expandafter\def\csname res@volt-s-official-semantickitti@iou-truck-pm\endcsname{85.2}
\expandafter\def\csname res@volt-s-official-semantickitti@iou-other-vehicle\endcsname{62.1}
\expandafter\def\csname res@volt-s-official-semantickitti@iou-other-vehicle-pm\endcsname{62.1}
\expandafter\def\csname res@volt-s-official-semantickitti@iou-person\endcsname{77.5}
\expandafter\def\csname res@volt-s-official-semantickitti@iou-person-pm\endcsname{77.5}
\expandafter\def\csname res@volt-s-official-semantickitti@iou-bicyclist\endcsname{89.5}
\expandafter\def\csname res@volt-s-official-semantickitti@iou-bicyclist-pm\endcsname{89.5}
\expandafter\def\csname res@volt-s-official-semantickitti@iou-motorcyclist\endcsname{9.7}
\expandafter\def\csname res@volt-s-official-semantickitti@iou-motorcyclist-pm\endcsname{9.7}
\expandafter\def\csname res@volt-s-official-semantickitti@iou-road\endcsname{95.1}
\expandafter\def\csname res@volt-s-official-semantickitti@iou-road-pm\endcsname{95.1}
\expandafter\def\csname res@volt-s-official-semantickitti@iou-parking\endcsname{49.9}
\expandafter\def\csname res@volt-s-official-semantickitti@iou-parking-pm\endcsname{49.9}
\expandafter\def\csname res@volt-s-official-semantickitti@iou-sidewalk\endcsname{83.4}
\expandafter\def\csname res@volt-s-official-semantickitti@iou-sidewalk-pm\endcsname{83.4}
\expandafter\def\csname res@volt-s-official-semantickitti@iou-other-ground\endcsname{0.3}
\expandafter\def\csname res@volt-s-official-semantickitti@iou-other-ground-pm\endcsname{0.3}
\expandafter\def\csname res@volt-s-official-semantickitti@iou-building\endcsname{92.5}
\expandafter\def\csname res@volt-s-official-semantickitti@iou-building-pm\endcsname{92.5}
\expandafter\def\csname res@volt-s-official-semantickitti@iou-fence\endcsname{66.3}
\expandafter\def\csname res@volt-s-official-semantickitti@iou-fence-pm\endcsname{66.3}
\expandafter\def\csname res@volt-s-official-semantickitti@iou-vegetation\endcsname{88.3}
\expandafter\def\csname res@volt-s-official-semantickitti@iou-vegetation-pm\endcsname{88.3}
\expandafter\def\csname res@volt-s-official-semantickitti@iou-trunk\endcsname{70.6}
\expandafter\def\csname res@volt-s-official-semantickitti@iou-trunk-pm\endcsname{70.6}
\expandafter\def\csname res@volt-s-official-semantickitti@iou-terrain\endcsname{75.3}
\expandafter\def\csname res@volt-s-official-semantickitti@iou-terrain-pm\endcsname{75.3}
\expandafter\def\csname res@volt-s-official-semantickitti@iou-pole\endcsname{67.3}
\expandafter\def\csname res@volt-s-official-semantickitti@iou-pole-pm\endcsname{67.3}
\expandafter\def\csname res@volt-s-official-semantickitti@iou-traffic-sign\endcsname{54.6}
\expandafter\def\csname res@volt-s-official-semantickitti@iou-traffic-sign-pm\endcsname{54.6}
\expandafter\def\csname res@volt-s-vo@n\endcsname{1}
\expandafter\def\csname res@volt-s-vo@miou\endcsname{77.0}
\expandafter\def\csname res@volt-s-vo@miou-pm\endcsname{77.0}
\expandafter\def\csname res@volt-s-vo@macc\endcsname{84.4}
\expandafter\def\csname res@volt-s-vo@macc-pm\endcsname{84.4}
\expandafter\def\csname res@volt-s-vo@allacc\endcsname{91.6}
\expandafter\def\csname res@volt-s-vo@allacc-pm\endcsname{91.6}
\expandafter\def\csname res@volt-s-vo@iou-wall\endcsname{86.7}
\expandafter\def\csname res@volt-s-vo@iou-wall-pm\endcsname{86.7}
\expandafter\def\csname res@volt-s-vo@iou-floor\endcsname{95.3}
\expandafter\def\csname res@volt-s-vo@iou-floor-pm\endcsname{95.3}
\expandafter\def\csname res@volt-s-vo@iou-cabinet\endcsname{73.8}
\expandafter\def\csname res@volt-s-vo@iou-cabinet-pm\endcsname{73.8}
\expandafter\def\csname res@volt-s-vo@iou-bed\endcsname{84.6}
\expandafter\def\csname res@volt-s-vo@iou-bed-pm\endcsname{84.6}
\expandafter\def\csname res@volt-s-vo@iou-chair\endcsname{92.0}
\expandafter\def\csname res@volt-s-vo@iou-chair-pm\endcsname{92.0}
\expandafter\def\csname res@volt-s-vo@iou-sofa\endcsname{83.9}
\expandafter\def\csname res@volt-s-vo@iou-sofa-pm\endcsname{83.9}
\expandafter\def\csname res@volt-s-vo@iou-table\endcsname{77.5}
\expandafter\def\csname res@volt-s-vo@iou-table-pm\endcsname{77.5}
\expandafter\def\csname res@volt-s-vo@iou-door\endcsname{64.7}
\expandafter\def\csname res@volt-s-vo@iou-door-pm\endcsname{64.7}
\expandafter\def\csname res@volt-s-vo@iou-window\endcsname{71.7}
\expandafter\def\csname res@volt-s-vo@iou-window-pm\endcsname{71.7}
\expandafter\def\csname res@volt-s-vo@iou-bookshelf\endcsname{85.9}
\expandafter\def\csname res@volt-s-vo@iou-bookshelf-pm\endcsname{85.9}
\expandafter\def\csname res@volt-s-vo@iou-picture\endcsname{33.6}
\expandafter\def\csname res@volt-s-vo@iou-picture-pm\endcsname{33.6}
\expandafter\def\csname res@volt-s-vo@iou-counter\endcsname{68.8}
\expandafter\def\csname res@volt-s-vo@iou-counter-pm\endcsname{68.8}
\expandafter\def\csname res@volt-s-vo@iou-desk\endcsname{73.7}
\expandafter\def\csname res@volt-s-vo@iou-desk-pm\endcsname{73.7}
\expandafter\def\csname res@volt-s-vo@iou-curtain\endcsname{82.2}
\expandafter\def\csname res@volt-s-vo@iou-curtain-pm\endcsname{82.2}
\expandafter\def\csname res@volt-s-vo@iou-refridgerator\endcsname{67.4}
\expandafter\def\csname res@volt-s-vo@iou-refridgerator-pm\endcsname{67.4}
\expandafter\def\csname res@volt-s-vo@iou-shower curtain\endcsname{77.9}
\expandafter\def\csname res@volt-s-vo@iou-shower curtain-pm\endcsname{77.9}
\expandafter\def\csname res@volt-s-vo@iou-toilet\endcsname{95.2}
\expandafter\def\csname res@volt-s-vo@iou-toilet-pm\endcsname{95.2}
\expandafter\def\csname res@volt-s-vo@iou-sink\endcsname{72.3}
\expandafter\def\csname res@volt-s-vo@iou-sink-pm\endcsname{72.3}
\expandafter\def\csname res@volt-s-vo@iou-bathtub\endcsname{88.2}
\expandafter\def\csname res@volt-s-vo@iou-bathtub-pm\endcsname{88.2}
\expandafter\def\csname res@volt-s-vo@iou-otherfurniture\endcsname{64.9}
\expandafter\def\csname res@volt-s-vo@iou-otherfurniture-pm\endcsname{64.9}
\expandafter\def\csname res@volt-s-vo@miou-delta\endcsname{+0.6}
\expandafter\def\csname res@volt-s-vo@macc-delta\endcsname{\ensuremath{-}0.6}
\expandafter\def\csname res@volt-s-vo@allacc-delta\endcsname{+0.3}
\expandafter\def\csname res@volt-s-vo@iou-wall-delta\endcsname{+0.2}
\expandafter\def\csname res@volt-s-vo@iou-floor-delta\endcsname{+0.0}
\expandafter\def\csname res@volt-s-vo@iou-cabinet-delta\endcsname{+1.5}
\expandafter\def\csname res@volt-s-vo@iou-bed-delta\endcsname{+0.7}
\expandafter\def\csname res@volt-s-vo@iou-chair-delta\endcsname{+0.1}
\expandafter\def\csname res@volt-s-vo@iou-sofa-delta\endcsname{+0.5}
\expandafter\def\csname res@volt-s-vo@iou-table-delta\endcsname{\ensuremath{-}0.1}
\expandafter\def\csname res@volt-s-vo@iou-door-delta\endcsname{+0.5}
\expandafter\def\csname res@volt-s-vo@iou-window-delta\endcsname{+3.6}
\expandafter\def\csname res@volt-s-vo@iou-bookshelf-delta\endcsname{+0.8}
\expandafter\def\csname res@volt-s-vo@iou-picture-delta\endcsname{\ensuremath{-}1.6}
\expandafter\def\csname res@volt-s-vo@iou-counter-delta\endcsname{+1.7}
\expandafter\def\csname res@volt-s-vo@iou-desk-delta\endcsname{+1.2}
\expandafter\def\csname res@volt-s-vo@iou-curtain-delta\endcsname{+1.4}
\expandafter\def\csname res@volt-s-vo@iou-refridgerator-delta\endcsname{\ensuremath{-}1.6}
\expandafter\def\csname res@volt-s-vo@iou-shower curtain-delta\endcsname{+1.3}
\expandafter\def\csname res@volt-s-vo@iou-toilet-delta\endcsname{+1.3}
\expandafter\def\csname res@volt-s-vo@iou-sink-delta\endcsname{+1.1}
\expandafter\def\csname res@volt-s-vo@iou-bathtub-delta\endcsname{\ensuremath{-}0.2}
\expandafter\def\csname res@volt-s-vo@iou-otherfurniture-delta\endcsname{+0.1}
\expandafter\def\csname res@volt-s-distill@n\endcsname{1}
\expandafter\def\csname res@volt-s-distill@miou\endcsname{77.3}
\expandafter\def\csname res@volt-s-distill@miou-pm\endcsname{77.3}
\expandafter\def\csname res@volt-s-distill@macc\endcsname{85.5}
\expandafter\def\csname res@volt-s-distill@macc-pm\endcsname{85.5}
\expandafter\def\csname res@volt-s-distill@allacc\endcsname{91.8}
\expandafter\def\csname res@volt-s-distill@allacc-pm\endcsname{91.8}
\expandafter\def\csname res@volt-s-distill@iou-wall\endcsname{87.4}
\expandafter\def\csname res@volt-s-distill@iou-wall-pm\endcsname{87.4}
\expandafter\def\csname res@volt-s-distill@iou-floor\endcsname{95.5}
\expandafter\def\csname res@volt-s-distill@iou-floor-pm\endcsname{95.5}
\expandafter\def\csname res@volt-s-distill@iou-cabinet\endcsname{74.6}
\expandafter\def\csname res@volt-s-distill@iou-cabinet-pm\endcsname{74.6}
\expandafter\def\csname res@volt-s-distill@iou-bed\endcsname{85.0}
\expandafter\def\csname res@volt-s-distill@iou-bed-pm\endcsname{85.0}
\expandafter\def\csname res@volt-s-distill@iou-chair\endcsname{92.4}
\expandafter\def\csname res@volt-s-distill@iou-chair-pm\endcsname{92.4}
\expandafter\def\csname res@volt-s-distill@iou-sofa\endcsname{83.1}
\expandafter\def\csname res@volt-s-distill@iou-sofa-pm\endcsname{83.1}
\expandafter\def\csname res@volt-s-distill@iou-table\endcsname{76.4}
\expandafter\def\csname res@volt-s-distill@iou-table-pm\endcsname{76.4}
\expandafter\def\csname res@volt-s-distill@iou-door\endcsname{67.8}
\expandafter\def\csname res@volt-s-distill@iou-door-pm\endcsname{67.8}
\expandafter\def\csname res@volt-s-distill@iou-window\endcsname{71.4}
\expandafter\def\csname res@volt-s-distill@iou-window-pm\endcsname{71.4}
\expandafter\def\csname res@volt-s-distill@iou-bookshelf\endcsname{84.4}
\expandafter\def\csname res@volt-s-distill@iou-bookshelf-pm\endcsname{84.4}
\expandafter\def\csname res@volt-s-distill@iou-picture\endcsname{34.8}
\expandafter\def\csname res@volt-s-distill@iou-picture-pm\endcsname{34.8}
\expandafter\def\csname res@volt-s-distill@iou-counter\endcsname{69.6}
\expandafter\def\csname res@volt-s-distill@iou-counter-pm\endcsname{69.6}
\expandafter\def\csname res@volt-s-distill@iou-desk\endcsname{72.9}
\expandafter\def\csname res@volt-s-distill@iou-desk-pm\endcsname{72.9}
\expandafter\def\csname res@volt-s-distill@iou-curtain\endcsname{82.8}
\expandafter\def\csname res@volt-s-distill@iou-curtain-pm\endcsname{82.8}
\expandafter\def\csname res@volt-s-distill@iou-refridgerator\endcsname{71.4}
\expandafter\def\csname res@volt-s-distill@iou-refridgerator-pm\endcsname{71.4}
\expandafter\def\csname res@volt-s-distill@iou-shower curtain\endcsname{78.0}
\expandafter\def\csname res@volt-s-distill@iou-shower curtain-pm\endcsname{78.0}
\expandafter\def\csname res@volt-s-distill@iou-toilet\endcsname{95.6}
\expandafter\def\csname res@volt-s-distill@iou-toilet-pm\endcsname{95.6}
\expandafter\def\csname res@volt-s-distill@iou-sink\endcsname{71.7}
\expandafter\def\csname res@volt-s-distill@iou-sink-pm\endcsname{71.7}
\expandafter\def\csname res@volt-s-distill@iou-bathtub\endcsname{87.9}
\expandafter\def\csname res@volt-s-distill@iou-bathtub-pm\endcsname{87.9}
\expandafter\def\csname res@volt-s-distill@iou-otherfurniture\endcsname{62.8}
\expandafter\def\csname res@volt-s-distill@iou-otherfurniture-pm\endcsname{62.8}
\expandafter\def\csname res@volt-s-distill@miou-delta\endcsname{+0.9}
\expandafter\def\csname res@volt-s-distill@macc-delta\endcsname{+0.5}
\expandafter\def\csname res@volt-s-distill@allacc-delta\endcsname{+0.5}
\expandafter\def\csname res@volt-s-distill@iou-wall-delta\endcsname{+0.9}
\expandafter\def\csname res@volt-s-distill@iou-floor-delta\endcsname{+0.2}
\expandafter\def\csname res@volt-s-distill@iou-cabinet-delta\endcsname{+2.3}
\expandafter\def\csname res@volt-s-distill@iou-bed-delta\endcsname{+1.1}
\expandafter\def\csname res@volt-s-distill@iou-chair-delta\endcsname{+0.5}
\expandafter\def\csname res@volt-s-distill@iou-sofa-delta\endcsname{\ensuremath{-}0.3}
\expandafter\def\csname res@volt-s-distill@iou-table-delta\endcsname{\ensuremath{-}1.2}
\expandafter\def\csname res@volt-s-distill@iou-door-delta\endcsname{+3.6}
\expandafter\def\csname res@volt-s-distill@iou-window-delta\endcsname{+3.3}
\expandafter\def\csname res@volt-s-distill@iou-bookshelf-delta\endcsname{\ensuremath{-}0.7}
\expandafter\def\csname res@volt-s-distill@iou-picture-delta\endcsname{\ensuremath{-}0.4}
\expandafter\def\csname res@volt-s-distill@iou-counter-delta\endcsname{+2.5}
\expandafter\def\csname res@volt-s-distill@iou-desk-delta\endcsname{+0.4}
\expandafter\def\csname res@volt-s-distill@iou-curtain-delta\endcsname{+2.0}
\expandafter\def\csname res@volt-s-distill@iou-refridgerator-delta\endcsname{+2.4}
\expandafter\def\csname res@volt-s-distill@iou-shower curtain-delta\endcsname{+1.4}
\expandafter\def\csname res@volt-s-distill@iou-toilet-delta\endcsname{+1.7}
\expandafter\def\csname res@volt-s-distill@iou-sink-delta\endcsname{+0.5}
\expandafter\def\csname res@volt-s-distill@iou-bathtub-delta\endcsname{\ensuremath{-}0.5}
\expandafter\def\csname res@volt-s-distill@iou-otherfurniture-delta\endcsname{\ensuremath{-}2.0}
\expandafter\def\csname res@volt-s-distill-vo@n\endcsname{1}
\expandafter\def\csname res@volt-s-distill-vo@miou\endcsname{78.0}
\expandafter\def\csname res@volt-s-distill-vo@miou-pm\endcsname{78.0}
\expandafter\def\csname res@volt-s-distill-vo@macc\endcsname{85.3}
\expandafter\def\csname res@volt-s-distill-vo@macc-pm\endcsname{85.3}
\expandafter\def\csname res@volt-s-distill-vo@allacc\endcsname{92.0}
\expandafter\def\csname res@volt-s-distill-vo@allacc-pm\endcsname{92.0}
\expandafter\def\csname res@volt-s-distill-vo@iou-wall\endcsname{87.0}
\expandafter\def\csname res@volt-s-distill-vo@iou-wall-pm\endcsname{87.0}
\expandafter\def\csname res@volt-s-distill-vo@iou-floor\endcsname{95.5}
\expandafter\def\csname res@volt-s-distill-vo@iou-floor-pm\endcsname{95.5}
\expandafter\def\csname res@volt-s-distill-vo@iou-cabinet\endcsname{73.8}
\expandafter\def\csname res@volt-s-distill-vo@iou-cabinet-pm\endcsname{73.8}
\expandafter\def\csname res@volt-s-distill-vo@iou-bed\endcsname{85.2}
\expandafter\def\csname res@volt-s-distill-vo@iou-bed-pm\endcsname{85.2}
\expandafter\def\csname res@volt-s-distill-vo@iou-chair\endcsname{93.0}
\expandafter\def\csname res@volt-s-distill-vo@iou-chair-pm\endcsname{93.0}
\expandafter\def\csname res@volt-s-distill-vo@iou-sofa\endcsname{84.2}
\expandafter\def\csname res@volt-s-distill-vo@iou-sofa-pm\endcsname{84.2}
\expandafter\def\csname res@volt-s-distill-vo@iou-table\endcsname{81.2}
\expandafter\def\csname res@volt-s-distill-vo@iou-table-pm\endcsname{81.2}
\expandafter\def\csname res@volt-s-distill-vo@iou-door\endcsname{66.8}
\expandafter\def\csname res@volt-s-distill-vo@iou-door-pm\endcsname{66.8}
\expandafter\def\csname res@volt-s-distill-vo@iou-window\endcsname{71.1}
\expandafter\def\csname res@volt-s-distill-vo@iou-window-pm\endcsname{71.1}
\expandafter\def\csname res@volt-s-distill-vo@iou-bookshelf\endcsname{85.5}
\expandafter\def\csname res@volt-s-distill-vo@iou-bookshelf-pm\endcsname{85.5}
\expandafter\def\csname res@volt-s-distill-vo@iou-picture\endcsname{33.8}
\expandafter\def\csname res@volt-s-distill-vo@iou-picture-pm\endcsname{33.8}
\expandafter\def\csname res@volt-s-distill-vo@iou-counter\endcsname{71.1}
\expandafter\def\csname res@volt-s-distill-vo@iou-counter-pm\endcsname{71.1}
\expandafter\def\csname res@volt-s-distill-vo@iou-desk\endcsname{77.7}
\expandafter\def\csname res@volt-s-distill-vo@iou-desk-pm\endcsname{77.7}
\expandafter\def\csname res@volt-s-distill-vo@iou-curtain\endcsname{81.2}
\expandafter\def\csname res@volt-s-distill-vo@iou-curtain-pm\endcsname{81.2}
\expandafter\def\csname res@volt-s-distill-vo@iou-refridgerator\endcsname{78.7}
\expandafter\def\csname res@volt-s-distill-vo@iou-refridgerator-pm\endcsname{78.7}
\expandafter\def\csname res@volt-s-distill-vo@iou-shower curtain\endcsname{71.7}
\expandafter\def\csname res@volt-s-distill-vo@iou-shower curtain-pm\endcsname{71.7}
\expandafter\def\csname res@volt-s-distill-vo@iou-toilet\endcsname{94.8}
\expandafter\def\csname res@volt-s-distill-vo@iou-toilet-pm\endcsname{94.8}
\expandafter\def\csname res@volt-s-distill-vo@iou-sink\endcsname{72.7}
\expandafter\def\csname res@volt-s-distill-vo@iou-sink-pm\endcsname{72.7}
\expandafter\def\csname res@volt-s-distill-vo@iou-bathtub\endcsname{88.7}
\expandafter\def\csname res@volt-s-distill-vo@iou-bathtub-pm\endcsname{88.7}
\expandafter\def\csname res@volt-s-distill-vo@iou-otherfurniture\endcsname{66.6}
\expandafter\def\csname res@volt-s-distill-vo@iou-otherfurniture-pm\endcsname{66.6}
\expandafter\def\csname res@volt-s-distill-vo@miou-delta\endcsname{+1.6}
\expandafter\def\csname res@volt-s-distill-vo@macc-delta\endcsname{+0.3}
\expandafter\def\csname res@volt-s-distill-vo@allacc-delta\endcsname{+0.7}
\expandafter\def\csname res@volt-s-distill-vo@iou-wall-delta\endcsname{+0.5}
\expandafter\def\csname res@volt-s-distill-vo@iou-floor-delta\endcsname{+0.2}
\expandafter\def\csname res@volt-s-distill-vo@iou-cabinet-delta\endcsname{+1.5}
\expandafter\def\csname res@volt-s-distill-vo@iou-bed-delta\endcsname{+1.3}
\expandafter\def\csname res@volt-s-distill-vo@iou-chair-delta\endcsname{+1.1}
\expandafter\def\csname res@volt-s-distill-vo@iou-sofa-delta\endcsname{+0.8}
\expandafter\def\csname res@volt-s-distill-vo@iou-table-delta\endcsname{+3.6}
\expandafter\def\csname res@volt-s-distill-vo@iou-door-delta\endcsname{+2.6}
\expandafter\def\csname res@volt-s-distill-vo@iou-window-delta\endcsname{+3.0}
\expandafter\def\csname res@volt-s-distill-vo@iou-bookshelf-delta\endcsname{+0.4}
\expandafter\def\csname res@volt-s-distill-vo@iou-picture-delta\endcsname{\ensuremath{-}1.4}
\expandafter\def\csname res@volt-s-distill-vo@iou-counter-delta\endcsname{+4.0}
\expandafter\def\csname res@volt-s-distill-vo@iou-desk-delta\endcsname{+5.2}
\expandafter\def\csname res@volt-s-distill-vo@iou-curtain-delta\endcsname{+0.4}
\expandafter\def\csname res@volt-s-distill-vo@iou-refridgerator-delta\endcsname{+9.7}
\expandafter\def\csname res@volt-s-distill-vo@iou-shower curtain-delta\endcsname{\ensuremath{-}4.9}
\expandafter\def\csname res@volt-s-distill-vo@iou-toilet-delta\endcsname{+0.9}
\expandafter\def\csname res@volt-s-distill-vo@iou-sink-delta\endcsname{+1.5}
\expandafter\def\csname res@volt-s-distill-vo@iou-bathtub-delta\endcsname{+0.3}
\expandafter\def\csname res@volt-s-distill-vo@iou-otherfurniture-delta\endcsname{+1.8}
\expandafter\def\csname res@volt-s-vo-nuscenes@n\endcsname{1}
\expandafter\def\csname res@volt-s-vo-nuscenes@miou\endcsname{80.3}
\expandafter\def\csname res@volt-s-vo-nuscenes@miou-pm\endcsname{80.3}
\expandafter\def\csname res@volt-s-vo-nuscenes@macc\endcsname{86.1}
\expandafter\def\csname res@volt-s-vo-nuscenes@macc-pm\endcsname{86.1}
\expandafter\def\csname res@volt-s-vo-nuscenes@allacc\endcsname{94.3}
\expandafter\def\csname res@volt-s-vo-nuscenes@allacc-pm\endcsname{94.3}
\expandafter\def\csname res@volt-s-vo-nuscenes@iou-barrier\endcsname{80.0}
\expandafter\def\csname res@volt-s-vo-nuscenes@iou-barrier-pm\endcsname{80.0}
\expandafter\def\csname res@volt-s-vo-nuscenes@iou-bicycle\endcsname{54.7}
\expandafter\def\csname res@volt-s-vo-nuscenes@iou-bicycle-pm\endcsname{54.7}
\expandafter\def\csname res@volt-s-vo-nuscenes@iou-bus\endcsname{96.5}
\expandafter\def\csname res@volt-s-vo-nuscenes@iou-bus-pm\endcsname{96.5}
\expandafter\def\csname res@volt-s-vo-nuscenes@iou-car\endcsname{89.6}
\expandafter\def\csname res@volt-s-vo-nuscenes@iou-car-pm\endcsname{89.6}
\expandafter\def\csname res@volt-s-vo-nuscenes@iou-construction_vehicle\endcsname{62.9}
\expandafter\def\csname res@volt-s-vo-nuscenes@iou-construction_vehicle-pm\endcsname{62.9}
\expandafter\def\csname res@volt-s-vo-nuscenes@iou-motorcycle\endcsname{85.9}
\expandafter\def\csname res@volt-s-vo-nuscenes@iou-motorcycle-pm\endcsname{85.9}
\expandafter\def\csname res@volt-s-vo-nuscenes@iou-pedestrian\endcsname{84.0}
\expandafter\def\csname res@volt-s-vo-nuscenes@iou-pedestrian-pm\endcsname{84.0}
\expandafter\def\csname res@volt-s-vo-nuscenes@iou-traffic_cone\endcsname{72.1}
\expandafter\def\csname res@volt-s-vo-nuscenes@iou-traffic_cone-pm\endcsname{72.1}
\expandafter\def\csname res@volt-s-vo-nuscenes@iou-trailer\endcsname{72.4}
\expandafter\def\csname res@volt-s-vo-nuscenes@iou-trailer-pm\endcsname{72.4}
\expandafter\def\csname res@volt-s-vo-nuscenes@iou-truck\endcsname{84.9}
\expandafter\def\csname res@volt-s-vo-nuscenes@iou-truck-pm\endcsname{84.9}
\expandafter\def\csname res@volt-s-vo-nuscenes@iou-driveable_surface\endcsname{97.0}
\expandafter\def\csname res@volt-s-vo-nuscenes@iou-driveable_surface-pm\endcsname{97.0}
\expandafter\def\csname res@volt-s-vo-nuscenes@iou-other_flat\endcsname{73.3}
\expandafter\def\csname res@volt-s-vo-nuscenes@iou-other_flat-pm\endcsname{73.3}
\expandafter\def\csname res@volt-s-vo-nuscenes@iou-sidewalk\endcsname{76.1}
\expandafter\def\csname res@volt-s-vo-nuscenes@iou-sidewalk-pm\endcsname{76.1}
\expandafter\def\csname res@volt-s-vo-nuscenes@iou-terrain\endcsname{75.6}
\expandafter\def\csname res@volt-s-vo-nuscenes@iou-terrain-pm\endcsname{75.6}
\expandafter\def\csname res@volt-s-vo-nuscenes@iou-manmade\endcsname{90.6}
\expandafter\def\csname res@volt-s-vo-nuscenes@iou-manmade-pm\endcsname{90.6}
\expandafter\def\csname res@volt-s-vo-nuscenes@iou-vegetation\endcsname{88.8}
\expandafter\def\csname res@volt-s-vo-nuscenes@iou-vegetation-pm\endcsname{88.8}
\expandafter\def\csname res@volt-s-vo-nuscenes@miou-delta\endcsname{\ensuremath{-}0.2}
\expandafter\def\csname res@volt-s-vo-nuscenes@macc-delta\endcsname{\ensuremath{-}1.1}
\expandafter\def\csname res@volt-s-vo-nuscenes@allacc-delta\endcsname{\ensuremath{-}0.2}
\expandafter\def\csname res@volt-s-vo-nuscenes@iou-barrier-delta\endcsname{\ensuremath{-}0.5}
\expandafter\def\csname res@volt-s-vo-nuscenes@iou-bicycle-delta\endcsname{\ensuremath{-}1.8}
\expandafter\def\csname res@volt-s-vo-nuscenes@iou-bus-delta\endcsname{+2.2}
\expandafter\def\csname res@volt-s-vo-nuscenes@iou-car-delta\endcsname{\ensuremath{-}1.7}
\expandafter\def\csname res@volt-s-vo-nuscenes@iou-construction_vehicle-delta\endcsname{+2.4}
\expandafter\def\csname res@volt-s-vo-nuscenes@iou-motorcycle-delta\endcsname{\ensuremath{-}1.4}
\expandafter\def\csname res@volt-s-vo-nuscenes@iou-pedestrian-delta\endcsname{\ensuremath{-}0.5}
\expandafter\def\csname res@volt-s-vo-nuscenes@iou-traffic_cone-delta\endcsname{+0.8}
\expandafter\def\csname res@volt-s-vo-nuscenes@iou-trailer-delta\endcsname{+0.7}
\expandafter\def\csname res@volt-s-vo-nuscenes@iou-truck-delta\endcsname{+1.4}
\expandafter\def\csname res@volt-s-vo-nuscenes@iou-driveable_surface-delta\endcsname{\ensuremath{-}0.1}
\expandafter\def\csname res@volt-s-vo-nuscenes@iou-other_flat-delta\endcsname{\ensuremath{-}2.5}
\expandafter\def\csname res@volt-s-vo-nuscenes@iou-sidewalk-delta\endcsname{\ensuremath{-}0.7}
\expandafter\def\csname res@volt-s-vo-nuscenes@iou-terrain-delta\endcsname{\ensuremath{-}0.9}
\expandafter\def\csname res@volt-s-vo-nuscenes@iou-manmade-delta\endcsname{\ensuremath{-}0.2}
\expandafter\def\csname res@volt-s-vo-nuscenes@iou-vegetation-delta\endcsname{\ensuremath{-}0.3}
\expandafter\def\csname res@volt-s-distill-nuscenes@n\endcsname{1}
\expandafter\def\csname res@volt-s-distill-nuscenes@miou\endcsname{81.1}
\expandafter\def\csname res@volt-s-distill-nuscenes@miou-pm\endcsname{81.1}
\expandafter\def\csname res@volt-s-distill-nuscenes@macc\endcsname{87.7}
\expandafter\def\csname res@volt-s-distill-nuscenes@macc-pm\endcsname{87.7}
\expandafter\def\csname res@volt-s-distill-nuscenes@allacc\endcsname{94.7}
\expandafter\def\csname res@volt-s-distill-nuscenes@allacc-pm\endcsname{94.7}
\expandafter\def\csname res@volt-s-distill-nuscenes@iou-barrier\endcsname{80.7}
\expandafter\def\csname res@volt-s-distill-nuscenes@iou-barrier-pm\endcsname{80.7}
\expandafter\def\csname res@volt-s-distill-nuscenes@iou-bicycle\endcsname{56.5}
\expandafter\def\csname res@volt-s-distill-nuscenes@iou-bicycle-pm\endcsname{56.5}
\expandafter\def\csname res@volt-s-distill-nuscenes@iou-bus\endcsname{96.4}
\expandafter\def\csname res@volt-s-distill-nuscenes@iou-bus-pm\endcsname{96.4}
\expandafter\def\csname res@volt-s-distill-nuscenes@iou-car\endcsname{93.1}
\expandafter\def\csname res@volt-s-distill-nuscenes@iou-car-pm\endcsname{93.1}
\expandafter\def\csname res@volt-s-distill-nuscenes@iou-construction_vehicle\endcsname{61.2}
\expandafter\def\csname res@volt-s-distill-nuscenes@iou-construction_vehicle-pm\endcsname{61.2}
\expandafter\def\csname res@volt-s-distill-nuscenes@iou-motorcycle\endcsname{86.9}
\expandafter\def\csname res@volt-s-distill-nuscenes@iou-motorcycle-pm\endcsname{86.9}
\expandafter\def\csname res@volt-s-distill-nuscenes@iou-pedestrian\endcsname{85.3}
\expandafter\def\csname res@volt-s-distill-nuscenes@iou-pedestrian-pm\endcsname{85.3}
\expandafter\def\csname res@volt-s-distill-nuscenes@iou-traffic_cone\endcsname{71.5}
\expandafter\def\csname res@volt-s-distill-nuscenes@iou-traffic_cone-pm\endcsname{71.5}
\expandafter\def\csname res@volt-s-distill-nuscenes@iou-trailer\endcsname{73.6}
\expandafter\def\csname res@volt-s-distill-nuscenes@iou-trailer-pm\endcsname{73.6}
\expandafter\def\csname res@volt-s-distill-nuscenes@iou-truck\endcsname{86.9}
\expandafter\def\csname res@volt-s-distill-nuscenes@iou-truck-pm\endcsname{86.9}
\expandafter\def\csname res@volt-s-distill-nuscenes@iou-driveable_surface\endcsname{97.1}
\expandafter\def\csname res@volt-s-distill-nuscenes@iou-driveable_surface-pm\endcsname{97.1}
\expandafter\def\csname res@volt-s-distill-nuscenes@iou-other_flat\endcsname{75.2}
\expandafter\def\csname res@volt-s-distill-nuscenes@iou-other_flat-pm\endcsname{75.2}
\expandafter\def\csname res@volt-s-distill-nuscenes@iou-sidewalk\endcsname{76.8}
\expandafter\def\csname res@volt-s-distill-nuscenes@iou-sidewalk-pm\endcsname{76.8}
\expandafter\def\csname res@volt-s-distill-nuscenes@iou-terrain\endcsname{76.1}
\expandafter\def\csname res@volt-s-distill-nuscenes@iou-terrain-pm\endcsname{76.1}
\expandafter\def\csname res@volt-s-distill-nuscenes@iou-manmade\endcsname{91.2}
\expandafter\def\csname res@volt-s-distill-nuscenes@iou-manmade-pm\endcsname{91.2}
\expandafter\def\csname res@volt-s-distill-nuscenes@iou-vegetation\endcsname{89.4}
\expandafter\def\csname res@volt-s-distill-nuscenes@iou-vegetation-pm\endcsname{89.4}
\expandafter\def\csname res@volt-s-distill-nuscenes@miou-delta\endcsname{+0.6}
\expandafter\def\csname res@volt-s-distill-nuscenes@macc-delta\endcsname{+0.5}
\expandafter\def\csname res@volt-s-distill-nuscenes@allacc-delta\endcsname{+0.2}
\expandafter\def\csname res@volt-s-distill-nuscenes@iou-barrier-delta\endcsname{+0.2}
\expandafter\def\csname res@volt-s-distill-nuscenes@iou-bicycle-delta\endcsname{+0.0}
\expandafter\def\csname res@volt-s-distill-nuscenes@iou-bus-delta\endcsname{+2.1}
\expandafter\def\csname res@volt-s-distill-nuscenes@iou-car-delta\endcsname{+1.8}
\expandafter\def\csname res@volt-s-distill-nuscenes@iou-construction_vehicle-delta\endcsname{+0.7}
\expandafter\def\csname res@volt-s-distill-nuscenes@iou-motorcycle-delta\endcsname{\ensuremath{-}0.4}
\expandafter\def\csname res@volt-s-distill-nuscenes@iou-pedestrian-delta\endcsname{+0.8}
\expandafter\def\csname res@volt-s-distill-nuscenes@iou-traffic_cone-delta\endcsname{+0.2}
\expandafter\def\csname res@volt-s-distill-nuscenes@iou-trailer-delta\endcsname{+1.9}
\expandafter\def\csname res@volt-s-distill-nuscenes@iou-truck-delta\endcsname{+3.4}
\expandafter\def\csname res@volt-s-distill-nuscenes@iou-driveable_surface-delta\endcsname{+0.0}
\expandafter\def\csname res@volt-s-distill-nuscenes@iou-other_flat-delta\endcsname{\ensuremath{-}0.6}
\expandafter\def\csname res@volt-s-distill-nuscenes@iou-sidewalk-delta\endcsname{+0.0}
\expandafter\def\csname res@volt-s-distill-nuscenes@iou-terrain-delta\endcsname{\ensuremath{-}0.4}
\expandafter\def\csname res@volt-s-distill-nuscenes@iou-manmade-delta\endcsname{+0.4}
\expandafter\def\csname res@volt-s-distill-nuscenes@iou-vegetation-delta\endcsname{+0.3}
\expandafter\def\csname res@volt-s-distill-vo-nuscenes@n\endcsname{1}
\expandafter\def\csname res@volt-s-distill-vo-nuscenes@miou\endcsname{81.6}
\expandafter\def\csname res@volt-s-distill-vo-nuscenes@miou-pm\endcsname{81.6}
\expandafter\def\csname res@volt-s-distill-vo-nuscenes@macc\endcsname{87.4}
\expandafter\def\csname res@volt-s-distill-vo-nuscenes@macc-pm\endcsname{87.4}
\expandafter\def\csname res@volt-s-distill-vo-nuscenes@allacc\endcsname{94.7}
\expandafter\def\csname res@volt-s-distill-vo-nuscenes@allacc-pm\endcsname{94.7}
\expandafter\def\csname res@volt-s-distill-vo-nuscenes@iou-barrier\endcsname{80.3}
\expandafter\def\csname res@volt-s-distill-vo-nuscenes@iou-barrier-pm\endcsname{80.3}
\expandafter\def\csname res@volt-s-distill-vo-nuscenes@iou-bicycle\endcsname{59.1}
\expandafter\def\csname res@volt-s-distill-vo-nuscenes@iou-bicycle-pm\endcsname{59.1}
\expandafter\def\csname res@volt-s-distill-vo-nuscenes@iou-bus\endcsname{95.9}
\expandafter\def\csname res@volt-s-distill-vo-nuscenes@iou-bus-pm\endcsname{95.9}
\expandafter\def\csname res@volt-s-distill-vo-nuscenes@iou-car\endcsname{93.1}
\expandafter\def\csname res@volt-s-distill-vo-nuscenes@iou-car-pm\endcsname{93.1}
\expandafter\def\csname res@volt-s-distill-vo-nuscenes@iou-construction_vehicle\endcsname{64.8}
\expandafter\def\csname res@volt-s-distill-vo-nuscenes@iou-construction_vehicle-pm\endcsname{64.8}
\expandafter\def\csname res@volt-s-distill-vo-nuscenes@iou-motorcycle\endcsname{89.8}
\expandafter\def\csname res@volt-s-distill-vo-nuscenes@iou-motorcycle-pm\endcsname{89.8}
\expandafter\def\csname res@volt-s-distill-vo-nuscenes@iou-pedestrian\endcsname{84.8}
\expandafter\def\csname res@volt-s-distill-vo-nuscenes@iou-pedestrian-pm\endcsname{84.8}
\expandafter\def\csname res@volt-s-distill-vo-nuscenes@iou-traffic_cone\endcsname{72.0}
\expandafter\def\csname res@volt-s-distill-vo-nuscenes@iou-traffic_cone-pm\endcsname{72.0}
\expandafter\def\csname res@volt-s-distill-vo-nuscenes@iou-trailer\endcsname{71.5}
\expandafter\def\csname res@volt-s-distill-vo-nuscenes@iou-trailer-pm\endcsname{71.5}
\expandafter\def\csname res@volt-s-distill-vo-nuscenes@iou-truck\endcsname{87.4}
\expandafter\def\csname res@volt-s-distill-vo-nuscenes@iou-truck-pm\endcsname{87.4}
\expandafter\def\csname res@volt-s-distill-vo-nuscenes@iou-driveable_surface\endcsname{97.2}
\expandafter\def\csname res@volt-s-distill-vo-nuscenes@iou-driveable_surface-pm\endcsname{97.2}
\expandafter\def\csname res@volt-s-distill-vo-nuscenes@iou-other_flat\endcsname{76.2}
\expandafter\def\csname res@volt-s-distill-vo-nuscenes@iou-other_flat-pm\endcsname{76.2}
\expandafter\def\csname res@volt-s-distill-vo-nuscenes@iou-sidewalk\endcsname{77.3}
\expandafter\def\csname res@volt-s-distill-vo-nuscenes@iou-sidewalk-pm\endcsname{77.3}
\expandafter\def\csname res@volt-s-distill-vo-nuscenes@iou-terrain\endcsname{76.5}
\expandafter\def\csname res@volt-s-distill-vo-nuscenes@iou-terrain-pm\endcsname{76.5}
\expandafter\def\csname res@volt-s-distill-vo-nuscenes@iou-manmade\endcsname{90.9}
\expandafter\def\csname res@volt-s-distill-vo-nuscenes@iou-manmade-pm\endcsname{90.9}
\expandafter\def\csname res@volt-s-distill-vo-nuscenes@iou-vegetation\endcsname{89.1}
\expandafter\def\csname res@volt-s-distill-vo-nuscenes@iou-vegetation-pm\endcsname{89.1}
\expandafter\def\csname res@volt-s-distill-vo-nuscenes@miou-delta\endcsname{+1.1}
\expandafter\def\csname res@volt-s-distill-vo-nuscenes@macc-delta\endcsname{+0.2}
\expandafter\def\csname res@volt-s-distill-vo-nuscenes@allacc-delta\endcsname{+0.2}
\expandafter\def\csname res@volt-s-distill-vo-nuscenes@iou-barrier-delta\endcsname{\ensuremath{-}0.2}
\expandafter\def\csname res@volt-s-distill-vo-nuscenes@iou-bicycle-delta\endcsname{+2.6}
\expandafter\def\csname res@volt-s-distill-vo-nuscenes@iou-bus-delta\endcsname{+1.6}
\expandafter\def\csname res@volt-s-distill-vo-nuscenes@iou-car-delta\endcsname{+1.8}
\expandafter\def\csname res@volt-s-distill-vo-nuscenes@iou-construction_vehicle-delta\endcsname{+4.3}
\expandafter\def\csname res@volt-s-distill-vo-nuscenes@iou-motorcycle-delta\endcsname{+2.5}
\expandafter\def\csname res@volt-s-distill-vo-nuscenes@iou-pedestrian-delta\endcsname{+0.3}
\expandafter\def\csname res@volt-s-distill-vo-nuscenes@iou-traffic_cone-delta\endcsname{+0.7}
\expandafter\def\csname res@volt-s-distill-vo-nuscenes@iou-trailer-delta\endcsname{\ensuremath{-}0.2}
\expandafter\def\csname res@volt-s-distill-vo-nuscenes@iou-truck-delta\endcsname{+3.9}
\expandafter\def\csname res@volt-s-distill-vo-nuscenes@iou-driveable_surface-delta\endcsname{+0.1}
\expandafter\def\csname res@volt-s-distill-vo-nuscenes@iou-other_flat-delta\endcsname{+0.4}
\expandafter\def\csname res@volt-s-distill-vo-nuscenes@iou-sidewalk-delta\endcsname{+0.5}
\expandafter\def\csname res@volt-s-distill-vo-nuscenes@iou-terrain-delta\endcsname{+0.0}
\expandafter\def\csname res@volt-s-distill-vo-nuscenes@iou-manmade-delta\endcsname{+0.1}
\expandafter\def\csname res@volt-s-distill-vo-nuscenes@iou-vegetation-delta\endcsname{+0.0}
\expandafter\def\csname res@pvit-d12-scannet@n\endcsname{1}
\expandafter\def\csname res@pvit-d12-scannet@miou\endcsname{77.5}
\expandafter\def\csname res@pvit-d12-scannet@miou-pm\endcsname{77.5}
\expandafter\def\csname res@pvit-d12-scannet@macc\endcsname{85.8}
\expandafter\def\csname res@pvit-d12-scannet@macc-pm\endcsname{85.8}
\expandafter\def\csname res@pvit-d12-scannet@allacc\endcsname{91.9}
\expandafter\def\csname res@pvit-d12-scannet@allacc-pm\endcsname{91.9}
\expandafter\def\csname res@pvit-d12-scannet@iou-wall\endcsname{87.8}
\expandafter\def\csname res@pvit-d12-scannet@iou-wall-pm\endcsname{87.8}
\expandafter\def\csname res@pvit-d12-scannet@iou-floor\endcsname{95.9}
\expandafter\def\csname res@pvit-d12-scannet@iou-floor-pm\endcsname{95.9}
\expandafter\def\csname res@pvit-d12-scannet@iou-cabinet\endcsname{72.4}
\expandafter\def\csname res@pvit-d12-scannet@iou-cabinet-pm\endcsname{72.4}
\expandafter\def\csname res@pvit-d12-scannet@iou-bed\endcsname{83.0}
\expandafter\def\csname res@pvit-d12-scannet@iou-bed-pm\endcsname{83.0}
\expandafter\def\csname res@pvit-d12-scannet@iou-chair\endcsname{92.7}
\expandafter\def\csname res@pvit-d12-scannet@iou-chair-pm\endcsname{92.7}
\expandafter\def\csname res@pvit-d12-scannet@iou-sofa\endcsname{86.8}
\expandafter\def\csname res@pvit-d12-scannet@iou-sofa-pm\endcsname{86.8}
\expandafter\def\csname res@pvit-d12-scannet@iou-table\endcsname{77.1}
\expandafter\def\csname res@pvit-d12-scannet@iou-table-pm\endcsname{77.1}
\expandafter\def\csname res@pvit-d12-scannet@iou-door\endcsname{69.6}
\expandafter\def\csname res@pvit-d12-scannet@iou-door-pm\endcsname{69.6}
\expandafter\def\csname res@pvit-d12-scannet@iou-window\endcsname{71.5}
\expandafter\def\csname res@pvit-d12-scannet@iou-window-pm\endcsname{71.5}
\expandafter\def\csname res@pvit-d12-scannet@iou-bookshelf\endcsname{85.5}
\expandafter\def\csname res@pvit-d12-scannet@iou-bookshelf-pm\endcsname{85.5}
\expandafter\def\csname res@pvit-d12-scannet@iou-picture\endcsname{39.9}
\expandafter\def\csname res@pvit-d12-scannet@iou-picture-pm\endcsname{39.9}
\expandafter\def\csname res@pvit-d12-scannet@iou-counter\endcsname{70.5}
\expandafter\def\csname res@pvit-d12-scannet@iou-counter-pm\endcsname{70.5}
\expandafter\def\csname res@pvit-d12-scannet@iou-desk\endcsname{72.1}
\expandafter\def\csname res@pvit-d12-scannet@iou-desk-pm\endcsname{72.1}
\expandafter\def\csname res@pvit-d12-scannet@iou-curtain\endcsname{80.7}
\expandafter\def\csname res@pvit-d12-scannet@iou-curtain-pm\endcsname{80.7}
\expandafter\def\csname res@pvit-d12-scannet@iou-refridgerator\endcsname{70.9}
\expandafter\def\csname res@pvit-d12-scannet@iou-refridgerator-pm\endcsname{70.9}
\expandafter\def\csname res@pvit-d12-scannet@iou-shower curtain\endcsname{75.4}
\expandafter\def\csname res@pvit-d12-scannet@iou-shower curtain-pm\endcsname{75.4}
\expandafter\def\csname res@pvit-d12-scannet@iou-toilet\endcsname{96.0}
\expandafter\def\csname res@pvit-d12-scannet@iou-toilet-pm\endcsname{96.0}
\expandafter\def\csname res@pvit-d12-scannet@iou-sink\endcsname{71.2}
\expandafter\def\csname res@pvit-d12-scannet@iou-sink-pm\endcsname{71.2}
\expandafter\def\csname res@pvit-d12-scannet@iou-bathtub\endcsname{88.8}
\expandafter\def\csname res@pvit-d12-scannet@iou-bathtub-pm\endcsname{88.8}
\expandafter\def\csname res@pvit-d12-scannet@iou-otherfurniture\endcsname{62.5}
\expandafter\def\csname res@pvit-d12-scannet@iou-otherfurniture-pm\endcsname{62.5}
\expandafter\def\csname res@pvit-d12-scannet@miou-delta\endcsname{\ensuremath{-}0.1}
\expandafter\def\csname res@pvit-d12-scannet@macc-delta\endcsname{\ensuremath{-}0.1}
\expandafter\def\csname res@pvit-d12-scannet@allacc-delta\endcsname{\ensuremath{-}0.1}
\expandafter\def\csname res@pvit-d12-scannet@iou-wall-delta\endcsname{\ensuremath{-}0.1}
\expandafter\def\csname res@pvit-d12-scannet@iou-floor-delta\endcsname{+0.0}
\expandafter\def\csname res@pvit-d12-scannet@iou-cabinet-delta\endcsname{\ensuremath{-}1.0}
\expandafter\def\csname res@pvit-d12-scannet@iou-bed-delta\endcsname{+0.4}
\expandafter\def\csname res@pvit-d12-scannet@iou-chair-delta\endcsname{+0.7}
\expandafter\def\csname res@pvit-d12-scannet@iou-sofa-delta\endcsname{+2.1}
\expandafter\def\csname res@pvit-d12-scannet@iou-table-delta\endcsname{+0.3}
\expandafter\def\csname res@pvit-d12-scannet@iou-door-delta\endcsname{\ensuremath{-}2.9}
\expandafter\def\csname res@pvit-d12-scannet@iou-window-delta\endcsname{\ensuremath{-}2.6}
\expandafter\def\csname res@pvit-d12-scannet@iou-bookshelf-delta\endcsname{\ensuremath{-}0.1}
\expandafter\def\csname res@pvit-d12-scannet@iou-picture-delta\endcsname{\ensuremath{-}1.7}
\expandafter\def\csname res@pvit-d12-scannet@iou-counter-delta\endcsname{\ensuremath{-}0.3}
\expandafter\def\csname res@pvit-d12-scannet@iou-desk-delta\endcsname{+0.3}
\expandafter\def\csname res@pvit-d12-scannet@iou-curtain-delta\endcsname{+1.4}
\expandafter\def\csname res@pvit-d12-scannet@iou-refridgerator-delta\endcsname{\ensuremath{-}1.9}
\expandafter\def\csname res@pvit-d12-scannet@iou-shower curtain-delta\endcsname{+3.5}
\expandafter\def\csname res@pvit-d12-scannet@iou-toilet-delta\endcsname{+0.2}
\expandafter\def\csname res@pvit-d12-scannet@iou-sink-delta\endcsname{+0.7}
\expandafter\def\csname res@pvit-d12-scannet@iou-bathtub-delta\endcsname{+0.2}
\expandafter\def\csname res@pvit-d12-scannet@iou-otherfurniture-delta\endcsname{\ensuremath{-}0.7}
\expandafter\def\csname res@pvit-d12-scannet@miou-dvox\endcsname{+0.9}
\expandafter\def\csname res@pvit-d12-scannet@macc-dvox\endcsname{+0.6}
\expandafter\def\csname res@pvit-d12-scannet@allacc-dvox\endcsname{+0.4}
\expandafter\def\csname res@pvit-d12-sd-scannet@n\endcsname{1}
\expandafter\def\csname res@pvit-d12-sd-scannet@miou\endcsname{78.2}
\expandafter\def\csname res@pvit-d12-sd-scannet@miou-pm\endcsname{78.2}
\expandafter\def\csname res@pvit-d12-sd-scannet@macc\endcsname{86.1}
\expandafter\def\csname res@pvit-d12-sd-scannet@macc-pm\endcsname{86.1}
\expandafter\def\csname res@pvit-d12-sd-scannet@allacc\endcsname{92.1}
\expandafter\def\csname res@pvit-d12-sd-scannet@allacc-pm\endcsname{92.1}
\expandafter\def\csname res@pvit-d12-sd-scannet@iou-wall\endcsname{87.8}
\expandafter\def\csname res@pvit-d12-sd-scannet@iou-wall-pm\endcsname{87.8}
\expandafter\def\csname res@pvit-d12-sd-scannet@iou-floor\endcsname{95.8}
\expandafter\def\csname res@pvit-d12-sd-scannet@iou-floor-pm\endcsname{95.8}
\expandafter\def\csname res@pvit-d12-sd-scannet@iou-cabinet\endcsname{75.4}
\expandafter\def\csname res@pvit-d12-sd-scannet@iou-cabinet-pm\endcsname{75.4}
\expandafter\def\csname res@pvit-d12-sd-scannet@iou-bed\endcsname{82.8}
\expandafter\def\csname res@pvit-d12-sd-scannet@iou-bed-pm\endcsname{82.8}
\expandafter\def\csname res@pvit-d12-sd-scannet@iou-chair\endcsname{92.1}
\expandafter\def\csname res@pvit-d12-sd-scannet@iou-chair-pm\endcsname{92.1}
\expandafter\def\csname res@pvit-d12-sd-scannet@iou-sofa\endcsname{87.2}
\expandafter\def\csname res@pvit-d12-sd-scannet@iou-sofa-pm\endcsname{87.2}
\expandafter\def\csname res@pvit-d12-sd-scannet@iou-table\endcsname{77.8}
\expandafter\def\csname res@pvit-d12-sd-scannet@iou-table-pm\endcsname{77.8}
\expandafter\def\csname res@pvit-d12-sd-scannet@iou-door\endcsname{70.8}
\expandafter\def\csname res@pvit-d12-sd-scannet@iou-door-pm\endcsname{70.8}
\expandafter\def\csname res@pvit-d12-sd-scannet@iou-window\endcsname{74.5}
\expandafter\def\csname res@pvit-d12-sd-scannet@iou-window-pm\endcsname{74.5}
\expandafter\def\csname res@pvit-d12-sd-scannet@iou-bookshelf\endcsname{85.4}
\expandafter\def\csname res@pvit-d12-sd-scannet@iou-bookshelf-pm\endcsname{85.4}
\expandafter\def\csname res@pvit-d12-sd-scannet@iou-picture\endcsname{40.4}
\expandafter\def\csname res@pvit-d12-sd-scannet@iou-picture-pm\endcsname{40.4}
\expandafter\def\csname res@pvit-d12-sd-scannet@iou-counter\endcsname{71.2}
\expandafter\def\csname res@pvit-d12-sd-scannet@iou-counter-pm\endcsname{71.2}
\expandafter\def\csname res@pvit-d12-sd-scannet@iou-desk\endcsname{73.0}
\expandafter\def\csname res@pvit-d12-sd-scannet@iou-desk-pm\endcsname{73.0}
\expandafter\def\csname res@pvit-d12-sd-scannet@iou-curtain\endcsname{81.0}
\expandafter\def\csname res@pvit-d12-sd-scannet@iou-curtain-pm\endcsname{81.0}
\expandafter\def\csname res@pvit-d12-sd-scannet@iou-refridgerator\endcsname{74.2}
\expandafter\def\csname res@pvit-d12-sd-scannet@iou-refridgerator-pm\endcsname{74.2}
\expandafter\def\csname res@pvit-d12-sd-scannet@iou-shower curtain\endcsname{75.9}
\expandafter\def\csname res@pvit-d12-sd-scannet@iou-shower curtain-pm\endcsname{75.9}
\expandafter\def\csname res@pvit-d12-sd-scannet@iou-toilet\endcsname{96.3}
\expandafter\def\csname res@pvit-d12-sd-scannet@iou-toilet-pm\endcsname{96.3}
\expandafter\def\csname res@pvit-d12-sd-scannet@iou-sink\endcsname{71.0}
\expandafter\def\csname res@pvit-d12-sd-scannet@iou-sink-pm\endcsname{71.0}
\expandafter\def\csname res@pvit-d12-sd-scannet@iou-bathtub\endcsname{88.5}
\expandafter\def\csname res@pvit-d12-sd-scannet@iou-bathtub-pm\endcsname{88.5}
\expandafter\def\csname res@pvit-d12-sd-scannet@iou-otherfurniture\endcsname{62.4}
\expandafter\def\csname res@pvit-d12-sd-scannet@iou-otherfurniture-pm\endcsname{62.4}
\expandafter\def\csname res@pvit-d12-sd-scannet@miou-delta\endcsname{+0.7}
\expandafter\def\csname res@pvit-d12-sd-scannet@macc-delta\endcsname{+0.3}
\expandafter\def\csname res@pvit-d12-sd-scannet@allacc-delta\endcsname{+0.2}
\expandafter\def\csname res@pvit-d12-sd-scannet@iou-wall-delta\endcsname{+0.0}
\expandafter\def\csname res@pvit-d12-sd-scannet@iou-floor-delta\endcsname{\ensuremath{-}0.1}
\expandafter\def\csname res@pvit-d12-sd-scannet@iou-cabinet-delta\endcsname{+3.0}
\expandafter\def\csname res@pvit-d12-sd-scannet@iou-bed-delta\endcsname{\ensuremath{-}0.2}
\expandafter\def\csname res@pvit-d12-sd-scannet@iou-chair-delta\endcsname{\ensuremath{-}0.6}
\expandafter\def\csname res@pvit-d12-sd-scannet@iou-sofa-delta\endcsname{+0.4}
\expandafter\def\csname res@pvit-d12-sd-scannet@iou-table-delta\endcsname{+0.7}
\expandafter\def\csname res@pvit-d12-sd-scannet@iou-door-delta\endcsname{+1.2}
\expandafter\def\csname res@pvit-d12-sd-scannet@iou-window-delta\endcsname{+3.0}
\expandafter\def\csname res@pvit-d12-sd-scannet@iou-bookshelf-delta\endcsname{\ensuremath{-}0.1}
\expandafter\def\csname res@pvit-d12-sd-scannet@iou-picture-delta\endcsname{+0.5}
\expandafter\def\csname res@pvit-d12-sd-scannet@iou-counter-delta\endcsname{+0.7}
\expandafter\def\csname res@pvit-d12-sd-scannet@iou-desk-delta\endcsname{+0.9}
\expandafter\def\csname res@pvit-d12-sd-scannet@iou-curtain-delta\endcsname{+0.3}
\expandafter\def\csname res@pvit-d12-sd-scannet@iou-refridgerator-delta\endcsname{+3.3}
\expandafter\def\csname res@pvit-d12-sd-scannet@iou-shower curtain-delta\endcsname{+0.5}
\expandafter\def\csname res@pvit-d12-sd-scannet@iou-toilet-delta\endcsname{+0.3}
\expandafter\def\csname res@pvit-d12-sd-scannet@iou-sink-delta\endcsname{\ensuremath{-}0.2}
\expandafter\def\csname res@pvit-d12-sd-scannet@iou-bathtub-delta\endcsname{\ensuremath{-}0.3}
\expandafter\def\csname res@pvit-d12-sd-scannet@iou-otherfurniture-delta\endcsname{\ensuremath{-}0.1}
\expandafter\def\csname res@pvit-d12-nuscenes@n\endcsname{1}
\expandafter\def\csname res@pvit-d12-nuscenes@miou\endcsname{81.6}
\expandafter\def\csname res@pvit-d12-nuscenes@miou-pm\endcsname{81.6}
\expandafter\def\csname res@pvit-d12-nuscenes@macc\endcsname{88.5}
\expandafter\def\csname res@pvit-d12-nuscenes@macc-pm\endcsname{88.5}
\expandafter\def\csname res@pvit-d12-nuscenes@allacc\endcsname{94.9}
\expandafter\def\csname res@pvit-d12-nuscenes@allacc-pm\endcsname{94.9}
\expandafter\def\csname res@pvit-d12-nuscenes@iou-barrier\endcsname{81.0}
\expandafter\def\csname res@pvit-d12-nuscenes@iou-barrier-pm\endcsname{81.0}
\expandafter\def\csname res@pvit-d12-nuscenes@iou-bicycle\endcsname{52.4}
\expandafter\def\csname res@pvit-d12-nuscenes@iou-bicycle-pm\endcsname{52.4}
\expandafter\def\csname res@pvit-d12-nuscenes@iou-bus\endcsname{97.2}
\expandafter\def\csname res@pvit-d12-nuscenes@iou-bus-pm\endcsname{97.2}
\expandafter\def\csname res@pvit-d12-nuscenes@iou-car\endcsname{93.8}
\expandafter\def\csname res@pvit-d12-nuscenes@iou-car-pm\endcsname{93.8}
\expandafter\def\csname res@pvit-d12-nuscenes@iou-construction_vehicle\endcsname{68.3}
\expandafter\def\csname res@pvit-d12-nuscenes@iou-construction_vehicle-pm\endcsname{68.3}
\expandafter\def\csname res@pvit-d12-nuscenes@iou-motorcycle\endcsname{87.9}
\expandafter\def\csname res@pvit-d12-nuscenes@iou-motorcycle-pm\endcsname{87.9}
\expandafter\def\csname res@pvit-d12-nuscenes@iou-pedestrian\endcsname{84.7}
\expandafter\def\csname res@pvit-d12-nuscenes@iou-pedestrian-pm\endcsname{84.7}
\expandafter\def\csname res@pvit-d12-nuscenes@iou-traffic_cone\endcsname{71.7}
\expandafter\def\csname res@pvit-d12-nuscenes@iou-traffic_cone-pm\endcsname{71.7}
\expandafter\def\csname res@pvit-d12-nuscenes@iou-trailer\endcsname{71.4}
\expandafter\def\csname res@pvit-d12-nuscenes@iou-trailer-pm\endcsname{71.4}
\expandafter\def\csname res@pvit-d12-nuscenes@iou-truck\endcsname{87.9}
\expandafter\def\csname res@pvit-d12-nuscenes@iou-truck-pm\endcsname{87.9}
\expandafter\def\csname res@pvit-d12-nuscenes@iou-driveable_surface\endcsname{97.2}
\expandafter\def\csname res@pvit-d12-nuscenes@iou-driveable_surface-pm\endcsname{97.2}
\expandafter\def\csname res@pvit-d12-nuscenes@iou-other_flat\endcsname{75.8}
\expandafter\def\csname res@pvit-d12-nuscenes@iou-other_flat-pm\endcsname{75.8}
\expandafter\def\csname res@pvit-d12-nuscenes@iou-sidewalk\endcsname{77.7}
\expandafter\def\csname res@pvit-d12-nuscenes@iou-sidewalk-pm\endcsname{77.7}
\expandafter\def\csname res@pvit-d12-nuscenes@iou-terrain\endcsname{76.7}
\expandafter\def\csname res@pvit-d12-nuscenes@iou-terrain-pm\endcsname{76.7}
\expandafter\def\csname res@pvit-d12-nuscenes@iou-manmade\endcsname{91.8}
\expandafter\def\csname res@pvit-d12-nuscenes@iou-manmade-pm\endcsname{91.8}
\expandafter\def\csname res@pvit-d12-nuscenes@iou-vegetation\endcsname{90.3}
\expandafter\def\csname res@pvit-d12-nuscenes@iou-vegetation-pm\endcsname{90.3}
\expandafter\def\csname res@pvit-d12-nuscenes-sd@n\endcsname{1}
\expandafter\def\csname res@pvit-d12-nuscenes-sd@miou\endcsname{82.1}
\expandafter\def\csname res@pvit-d12-nuscenes-sd@miou-pm\endcsname{82.1}
\expandafter\def\csname res@pvit-d12-nuscenes-sd@macc\endcsname{88.4}
\expandafter\def\csname res@pvit-d12-nuscenes-sd@macc-pm\endcsname{88.4}
\expandafter\def\csname res@pvit-d12-nuscenes-sd@allacc\endcsname{94.9}
\expandafter\def\csname res@pvit-d12-nuscenes-sd@allacc-pm\endcsname{94.9}
\expandafter\def\csname res@pvit-d12-nuscenes-sd@iou-barrier\endcsname{81.0}
\expandafter\def\csname res@pvit-d12-nuscenes-sd@iou-barrier-pm\endcsname{81.0}
\expandafter\def\csname res@pvit-d12-nuscenes-sd@iou-bicycle\endcsname{55.4}
\expandafter\def\csname res@pvit-d12-nuscenes-sd@iou-bicycle-pm\endcsname{55.4}
\expandafter\def\csname res@pvit-d12-nuscenes-sd@iou-bus\endcsname{97.2}
\expandafter\def\csname res@pvit-d12-nuscenes-sd@iou-bus-pm\endcsname{97.2}
\expandafter\def\csname res@pvit-d12-nuscenes-sd@iou-car\endcsname{93.8}
\expandafter\def\csname res@pvit-d12-nuscenes-sd@iou-car-pm\endcsname{93.8}
\expandafter\def\csname res@pvit-d12-nuscenes-sd@iou-construction_vehicle\endcsname{68.4}
\expandafter\def\csname res@pvit-d12-nuscenes-sd@iou-construction_vehicle-pm\endcsname{68.4}
\expandafter\def\csname res@pvit-d12-nuscenes-sd@iou-motorcycle\endcsname{90.4}
\expandafter\def\csname res@pvit-d12-nuscenes-sd@iou-motorcycle-pm\endcsname{90.4}
\expandafter\def\csname res@pvit-d12-nuscenes-sd@iou-pedestrian\endcsname{85.2}
\expandafter\def\csname res@pvit-d12-nuscenes-sd@iou-pedestrian-pm\endcsname{85.2}
\expandafter\def\csname res@pvit-d12-nuscenes-sd@iou-traffic_cone\endcsname{72.5}
\expandafter\def\csname res@pvit-d12-nuscenes-sd@iou-traffic_cone-pm\endcsname{72.5}
\expandafter\def\csname res@pvit-d12-nuscenes-sd@iou-trailer\endcsname{71.6}
\expandafter\def\csname res@pvit-d12-nuscenes-sd@iou-trailer-pm\endcsname{71.6}
\expandafter\def\csname res@pvit-d12-nuscenes-sd@iou-truck\endcsname{87.8}
\expandafter\def\csname res@pvit-d12-nuscenes-sd@iou-truck-pm\endcsname{87.8}
\expandafter\def\csname res@pvit-d12-nuscenes-sd@iou-driveable_surface\endcsname{97.3}
\expandafter\def\csname res@pvit-d12-nuscenes-sd@iou-driveable_surface-pm\endcsname{97.3}
\expandafter\def\csname res@pvit-d12-nuscenes-sd@iou-other_flat\endcsname{76.9}
\expandafter\def\csname res@pvit-d12-nuscenes-sd@iou-other_flat-pm\endcsname{76.9}
\expandafter\def\csname res@pvit-d12-nuscenes-sd@iou-sidewalk\endcsname{77.6}
\expandafter\def\csname res@pvit-d12-nuscenes-sd@iou-sidewalk-pm\endcsname{77.6}
\expandafter\def\csname res@pvit-d12-nuscenes-sd@iou-terrain\endcsname{76.6}
\expandafter\def\csname res@pvit-d12-nuscenes-sd@iou-terrain-pm\endcsname{76.6}
\expandafter\def\csname res@pvit-d12-nuscenes-sd@iou-manmade\endcsname{91.7}
\expandafter\def\csname res@pvit-d12-nuscenes-sd@iou-manmade-pm\endcsname{91.7}
\expandafter\def\csname res@pvit-d12-nuscenes-sd@iou-vegetation\endcsname{90.1}
\expandafter\def\csname res@pvit-d12-nuscenes-sd@iou-vegetation-pm\endcsname{90.1}
\expandafter\def\csname res@pvit-d12-nuscenes-sd@miou-delta\endcsname{+0.5}
\expandafter\def\csname res@pvit-d12-nuscenes-sd@macc-delta\endcsname{\ensuremath{-}0.1}
\expandafter\def\csname res@pvit-d12-nuscenes-sd@allacc-delta\endcsname{+0.0}
\expandafter\def\csname res@pvit-d12-nuscenes-sd@iou-barrier-delta\endcsname{+0.0}
\expandafter\def\csname res@pvit-d12-nuscenes-sd@iou-bicycle-delta\endcsname{+3.0}
\expandafter\def\csname res@pvit-d12-nuscenes-sd@iou-bus-delta\endcsname{+0.0}
\expandafter\def\csname res@pvit-d12-nuscenes-sd@iou-car-delta\endcsname{+0.0}
\expandafter\def\csname res@pvit-d12-nuscenes-sd@iou-construction_vehicle-delta\endcsname{+0.1}
\expandafter\def\csname res@pvit-d12-nuscenes-sd@iou-motorcycle-delta\endcsname{+2.5}
\expandafter\def\csname res@pvit-d12-nuscenes-sd@iou-pedestrian-delta\endcsname{+0.5}
\expandafter\def\csname res@pvit-d12-nuscenes-sd@iou-traffic_cone-delta\endcsname{+0.8}
\expandafter\def\csname res@pvit-d12-nuscenes-sd@iou-trailer-delta\endcsname{+0.2}
\expandafter\def\csname res@pvit-d12-nuscenes-sd@iou-truck-delta\endcsname{\ensuremath{-}0.1}
\expandafter\def\csname res@pvit-d12-nuscenes-sd@iou-driveable_surface-delta\endcsname{+0.1}
\expandafter\def\csname res@pvit-d12-nuscenes-sd@iou-other_flat-delta\endcsname{+1.1}
\expandafter\def\csname res@pvit-d12-nuscenes-sd@iou-sidewalk-delta\endcsname{\ensuremath{-}0.1}
\expandafter\def\csname res@pvit-d12-nuscenes-sd@iou-terrain-delta\endcsname{\ensuremath{-}0.1}
\expandafter\def\csname res@pvit-d12-nuscenes-sd@iou-manmade-delta\endcsname{\ensuremath{-}0.1}
\expandafter\def\csname res@pvit-d12-nuscenes-sd@iou-vegetation-delta\endcsname{\ensuremath{-}0.2}
\expandafter\def\csname res@pvit-d12-rope-scannet@n\endcsname{1}
\expandafter\def\csname res@pvit-d12-rope-scannet@miou\endcsname{77.6}
\expandafter\def\csname res@pvit-d12-rope-scannet@miou-pm\endcsname{77.6}
\expandafter\def\csname res@pvit-d12-rope-scannet@macc\endcsname{85.9}
\expandafter\def\csname res@pvit-d12-rope-scannet@macc-pm\endcsname{85.9}
\expandafter\def\csname res@pvit-d12-rope-scannet@allacc\endcsname{92.0}
\expandafter\def\csname res@pvit-d12-rope-scannet@allacc-pm\endcsname{92.0}
\expandafter\def\csname res@pvit-d12-rope-scannet@iou-wall\endcsname{87.9}
\expandafter\def\csname res@pvit-d12-rope-scannet@iou-wall-pm\endcsname{87.9}
\expandafter\def\csname res@pvit-d12-rope-scannet@iou-floor\endcsname{95.9}
\expandafter\def\csname res@pvit-d12-rope-scannet@iou-floor-pm\endcsname{95.9}
\expandafter\def\csname res@pvit-d12-rope-scannet@iou-cabinet\endcsname{73.4}
\expandafter\def\csname res@pvit-d12-rope-scannet@iou-cabinet-pm\endcsname{73.4}
\expandafter\def\csname res@pvit-d12-rope-scannet@iou-bed\endcsname{82.6}
\expandafter\def\csname res@pvit-d12-rope-scannet@iou-bed-pm\endcsname{82.6}
\expandafter\def\csname res@pvit-d12-rope-scannet@iou-chair\endcsname{92.0}
\expandafter\def\csname res@pvit-d12-rope-scannet@iou-chair-pm\endcsname{92.0}
\expandafter\def\csname res@pvit-d12-rope-scannet@iou-sofa\endcsname{84.7}
\expandafter\def\csname res@pvit-d12-rope-scannet@iou-sofa-pm\endcsname{84.7}
\expandafter\def\csname res@pvit-d12-rope-scannet@iou-table\endcsname{76.8}
\expandafter\def\csname res@pvit-d12-rope-scannet@iou-table-pm\endcsname{76.8}
\expandafter\def\csname res@pvit-d12-rope-scannet@iou-door\endcsname{72.5}
\expandafter\def\csname res@pvit-d12-rope-scannet@iou-door-pm\endcsname{72.5}
\expandafter\def\csname res@pvit-d12-rope-scannet@iou-window\endcsname{74.1}
\expandafter\def\csname res@pvit-d12-rope-scannet@iou-window-pm\endcsname{74.1}
\expandafter\def\csname res@pvit-d12-rope-scannet@iou-bookshelf\endcsname{85.6}
\expandafter\def\csname res@pvit-d12-rope-scannet@iou-bookshelf-pm\endcsname{85.6}
\expandafter\def\csname res@pvit-d12-rope-scannet@iou-picture\endcsname{41.6}
\expandafter\def\csname res@pvit-d12-rope-scannet@iou-picture-pm\endcsname{41.6}
\expandafter\def\csname res@pvit-d12-rope-scannet@iou-counter\endcsname{70.8}
\expandafter\def\csname res@pvit-d12-rope-scannet@iou-counter-pm\endcsname{70.8}
\expandafter\def\csname res@pvit-d12-rope-scannet@iou-desk\endcsname{71.8}
\expandafter\def\csname res@pvit-d12-rope-scannet@iou-desk-pm\endcsname{71.8}
\expandafter\def\csname res@pvit-d12-rope-scannet@iou-curtain\endcsname{79.3}
\expandafter\def\csname res@pvit-d12-rope-scannet@iou-curtain-pm\endcsname{79.3}
\expandafter\def\csname res@pvit-d12-rope-scannet@iou-refridgerator\endcsname{72.8}
\expandafter\def\csname res@pvit-d12-rope-scannet@iou-refridgerator-pm\endcsname{72.8}
\expandafter\def\csname res@pvit-d12-rope-scannet@iou-shower curtain\endcsname{71.9}
\expandafter\def\csname res@pvit-d12-rope-scannet@iou-shower curtain-pm\endcsname{71.9}
\expandafter\def\csname res@pvit-d12-rope-scannet@iou-toilet\endcsname{95.8}
\expandafter\def\csname res@pvit-d12-rope-scannet@iou-toilet-pm\endcsname{95.8}
\expandafter\def\csname res@pvit-d12-rope-scannet@iou-sink\endcsname{70.5}
\expandafter\def\csname res@pvit-d12-rope-scannet@iou-sink-pm\endcsname{70.5}
\expandafter\def\csname res@pvit-d12-rope-scannet@iou-bathtub\endcsname{88.6}
\expandafter\def\csname res@pvit-d12-rope-scannet@iou-bathtub-pm\endcsname{88.6}
\expandafter\def\csname res@pvit-d12-rope-scannet@iou-otherfurniture\endcsname{63.2}
\expandafter\def\csname res@pvit-d12-rope-scannet@iou-otherfurniture-pm\endcsname{63.2}
\expandafter\def\csname res@pvit-d12-rope-scannet@miou-dvox\endcsname{+1.0}
\expandafter\def\csname res@pvit-d12-rope-scannet@macc-dvox\endcsname{+0.7}
\expandafter\def\csname res@pvit-d12-rope-scannet@allacc-dvox\endcsname{+0.5}
\expandafter\def\csname res@pvit-d12-voxtok-scannet@n\endcsname{1}
\expandafter\def\csname res@pvit-d12-voxtok-scannet@miou\endcsname{77.4}
\expandafter\def\csname res@pvit-d12-voxtok-scannet@miou-pm\endcsname{77.4}
\expandafter\def\csname res@pvit-d12-voxtok-scannet@macc\endcsname{85.7}
\expandafter\def\csname res@pvit-d12-voxtok-scannet@macc-pm\endcsname{85.7}
\expandafter\def\csname res@pvit-d12-voxtok-scannet@allacc\endcsname{91.7}
\expandafter\def\csname res@pvit-d12-voxtok-scannet@allacc-pm\endcsname{91.7}
\expandafter\def\csname res@pvit-d12-voxtok-scannet@iou-wall\endcsname{87.2}
\expandafter\def\csname res@pvit-d12-voxtok-scannet@iou-wall-pm\endcsname{87.2}
\expandafter\def\csname res@pvit-d12-voxtok-scannet@iou-floor\endcsname{95.8}
\expandafter\def\csname res@pvit-d12-voxtok-scannet@iou-floor-pm\endcsname{95.8}
\expandafter\def\csname res@pvit-d12-voxtok-scannet@iou-cabinet\endcsname{72.1}
\expandafter\def\csname res@pvit-d12-voxtok-scannet@iou-cabinet-pm\endcsname{72.1}
\expandafter\def\csname res@pvit-d12-voxtok-scannet@iou-bed\endcsname{81.7}
\expandafter\def\csname res@pvit-d12-voxtok-scannet@iou-bed-pm\endcsname{81.7}
\expandafter\def\csname res@pvit-d12-voxtok-scannet@iou-chair\endcsname{91.5}
\expandafter\def\csname res@pvit-d12-voxtok-scannet@iou-chair-pm\endcsname{91.5}
\expandafter\def\csname res@pvit-d12-voxtok-scannet@iou-sofa\endcsname{87.4}
\expandafter\def\csname res@pvit-d12-voxtok-scannet@iou-sofa-pm\endcsname{87.4}
\expandafter\def\csname res@pvit-d12-voxtok-scannet@iou-table\endcsname{76.5}
\expandafter\def\csname res@pvit-d12-voxtok-scannet@iou-table-pm\endcsname{76.5}
\expandafter\def\csname res@pvit-d12-voxtok-scannet@iou-door\endcsname{70.5}
\expandafter\def\csname res@pvit-d12-voxtok-scannet@iou-door-pm\endcsname{70.5}
\expandafter\def\csname res@pvit-d12-voxtok-scannet@iou-window\endcsname{73.8}
\expandafter\def\csname res@pvit-d12-voxtok-scannet@iou-window-pm\endcsname{73.8}
\expandafter\def\csname res@pvit-d12-voxtok-scannet@iou-bookshelf\endcsname{84.6}
\expandafter\def\csname res@pvit-d12-voxtok-scannet@iou-bookshelf-pm\endcsname{84.6}
\expandafter\def\csname res@pvit-d12-voxtok-scannet@iou-picture\endcsname{41.5}
\expandafter\def\csname res@pvit-d12-voxtok-scannet@iou-picture-pm\endcsname{41.5}
\expandafter\def\csname res@pvit-d12-voxtok-scannet@iou-counter\endcsname{71.8}
\expandafter\def\csname res@pvit-d12-voxtok-scannet@iou-counter-pm\endcsname{71.8}
\expandafter\def\csname res@pvit-d12-voxtok-scannet@iou-desk\endcsname{72.1}
\expandafter\def\csname res@pvit-d12-voxtok-scannet@iou-desk-pm\endcsname{72.1}
\expandafter\def\csname res@pvit-d12-voxtok-scannet@iou-curtain\endcsname{81.1}
\expandafter\def\csname res@pvit-d12-voxtok-scannet@iou-curtain-pm\endcsname{81.1}
\expandafter\def\csname res@pvit-d12-voxtok-scannet@iou-refridgerator\endcsname{71.1}
\expandafter\def\csname res@pvit-d12-voxtok-scannet@iou-refridgerator-pm\endcsname{71.1}
\expandafter\def\csname res@pvit-d12-voxtok-scannet@iou-shower curtain\endcsname{76.4}
\expandafter\def\csname res@pvit-d12-voxtok-scannet@iou-shower curtain-pm\endcsname{76.4}
\expandafter\def\csname res@pvit-d12-voxtok-scannet@iou-toilet\endcsname{95.2}
\expandafter\def\csname res@pvit-d12-voxtok-scannet@iou-toilet-pm\endcsname{95.2}
\expandafter\def\csname res@pvit-d12-voxtok-scannet@iou-sink\endcsname{71.0}
\expandafter\def\csname res@pvit-d12-voxtok-scannet@iou-sink-pm\endcsname{71.0}
\expandafter\def\csname res@pvit-d12-voxtok-scannet@iou-bathtub\endcsname{88.4}
\expandafter\def\csname res@pvit-d12-voxtok-scannet@iou-bathtub-pm\endcsname{88.4}
\expandafter\def\csname res@pvit-d12-voxtok-scannet@iou-otherfurniture\endcsname{59.2}
\expandafter\def\csname res@pvit-d12-voxtok-scannet@iou-otherfurniture-pm\endcsname{59.2}
\expandafter\def\csname res@pvit-d12-voxtok-scannet@miou-dvox\endcsname{+0.8}
\expandafter\def\csname res@pvit-d12-voxtok-scannet@macc-dvox\endcsname{+0.5}
\expandafter\def\csname res@pvit-d12-voxtok-scannet@allacc-dvox\endcsname{+0.2}
\expandafter\def\csname res@pvit-d12-voxtok-rope-grid-scannet@n\endcsname{1}
\expandafter\def\csname res@pvit-d12-voxtok-rope-grid-scannet@miou\endcsname{76.6}
\expandafter\def\csname res@pvit-d12-voxtok-rope-grid-scannet@miou-pm\endcsname{76.6}
\expandafter\def\csname res@pvit-d12-voxtok-rope-grid-scannet@macc\endcsname{85.2}
\expandafter\def\csname res@pvit-d12-voxtok-rope-grid-scannet@macc-pm\endcsname{85.2}
\expandafter\def\csname res@pvit-d12-voxtok-rope-grid-scannet@allacc\endcsname{91.5}
\expandafter\def\csname res@pvit-d12-voxtok-rope-grid-scannet@allacc-pm\endcsname{91.5}
\expandafter\def\csname res@pvit-d12-voxtok-rope-grid-scannet@iou-wall\endcsname{86.9}
\expandafter\def\csname res@pvit-d12-voxtok-rope-grid-scannet@iou-wall-pm\endcsname{86.9}
\expandafter\def\csname res@pvit-d12-voxtok-rope-grid-scannet@iou-floor\endcsname{95.8}
\expandafter\def\csname res@pvit-d12-voxtok-rope-grid-scannet@iou-floor-pm\endcsname{95.8}
\expandafter\def\csname res@pvit-d12-voxtok-rope-grid-scannet@iou-cabinet\endcsname{71.8}
\expandafter\def\csname res@pvit-d12-voxtok-rope-grid-scannet@iou-cabinet-pm\endcsname{71.8}
\expandafter\def\csname res@pvit-d12-voxtok-rope-grid-scannet@iou-bed\endcsname{82.7}
\expandafter\def\csname res@pvit-d12-voxtok-rope-grid-scannet@iou-bed-pm\endcsname{82.7}
\expandafter\def\csname res@pvit-d12-voxtok-rope-grid-scannet@iou-chair\endcsname{91.4}
\expandafter\def\csname res@pvit-d12-voxtok-rope-grid-scannet@iou-chair-pm\endcsname{91.4}
\expandafter\def\csname res@pvit-d12-voxtok-rope-grid-scannet@iou-sofa\endcsname{83.9}
\expandafter\def\csname res@pvit-d12-voxtok-rope-grid-scannet@iou-sofa-pm\endcsname{83.9}
\expandafter\def\csname res@pvit-d12-voxtok-rope-grid-scannet@iou-table\endcsname{76.6}
\expandafter\def\csname res@pvit-d12-voxtok-rope-grid-scannet@iou-table-pm\endcsname{76.6}
\expandafter\def\csname res@pvit-d12-voxtok-rope-grid-scannet@iou-door\endcsname{69.7}
\expandafter\def\csname res@pvit-d12-voxtok-rope-grid-scannet@iou-door-pm\endcsname{69.7}
\expandafter\def\csname res@pvit-d12-voxtok-rope-grid-scannet@iou-window\endcsname{72.8}
\expandafter\def\csname res@pvit-d12-voxtok-rope-grid-scannet@iou-window-pm\endcsname{72.8}
\expandafter\def\csname res@pvit-d12-voxtok-rope-grid-scannet@iou-bookshelf\endcsname{85.7}
\expandafter\def\csname res@pvit-d12-voxtok-rope-grid-scannet@iou-bookshelf-pm\endcsname{85.7}
\expandafter\def\csname res@pvit-d12-voxtok-rope-grid-scannet@iou-picture\endcsname{40.1}
\expandafter\def\csname res@pvit-d12-voxtok-rope-grid-scannet@iou-picture-pm\endcsname{40.1}
\expandafter\def\csname res@pvit-d12-voxtok-rope-grid-scannet@iou-counter\endcsname{70.9}
\expandafter\def\csname res@pvit-d12-voxtok-rope-grid-scannet@iou-counter-pm\endcsname{70.9}
\expandafter\def\csname res@pvit-d12-voxtok-rope-grid-scannet@iou-desk\endcsname{70.7}
\expandafter\def\csname res@pvit-d12-voxtok-rope-grid-scannet@iou-desk-pm\endcsname{70.7}
\expandafter\def\csname res@pvit-d12-voxtok-rope-grid-scannet@iou-curtain\endcsname{78.7}
\expandafter\def\csname res@pvit-d12-voxtok-rope-grid-scannet@iou-curtain-pm\endcsname{78.7}
\expandafter\def\csname res@pvit-d12-voxtok-rope-grid-scannet@iou-refridgerator\endcsname{70.1}
\expandafter\def\csname res@pvit-d12-voxtok-rope-grid-scannet@iou-refridgerator-pm\endcsname{70.1}
\expandafter\def\csname res@pvit-d12-voxtok-rope-grid-scannet@iou-shower curtain\endcsname{66.6}
\expandafter\def\csname res@pvit-d12-voxtok-rope-grid-scannet@iou-shower curtain-pm\endcsname{66.6}
\expandafter\def\csname res@pvit-d12-voxtok-rope-grid-scannet@iou-toilet\endcsname{95.0}
\expandafter\def\csname res@pvit-d12-voxtok-rope-grid-scannet@iou-toilet-pm\endcsname{95.0}
\expandafter\def\csname res@pvit-d12-voxtok-rope-grid-scannet@iou-sink\endcsname{71.9}
\expandafter\def\csname res@pvit-d12-voxtok-rope-grid-scannet@iou-sink-pm\endcsname{71.9}
\expandafter\def\csname res@pvit-d12-voxtok-rope-grid-scannet@iou-bathtub\endcsname{88.9}
\expandafter\def\csname res@pvit-d12-voxtok-rope-grid-scannet@iou-bathtub-pm\endcsname{88.9}
\expandafter\def\csname res@pvit-d12-voxtok-rope-grid-scannet@iou-otherfurniture\endcsname{61.5}
\expandafter\def\csname res@pvit-d12-voxtok-rope-grid-scannet@iou-otherfurniture-pm\endcsname{61.5}
\expandafter\def\csname res@pvit-d12-voxtok-vo-grid-scannet@n\endcsname{1}
\expandafter\def\csname res@pvit-d12-voxtok-vo-grid-scannet@miou\endcsname{77.1}
\expandafter\def\csname res@pvit-d12-voxtok-vo-grid-scannet@miou-pm\endcsname{77.1}
\expandafter\def\csname res@pvit-d12-voxtok-vo-grid-scannet@macc\endcsname{85.2}
\expandafter\def\csname res@pvit-d12-voxtok-vo-grid-scannet@macc-pm\endcsname{85.2}
\expandafter\def\csname res@pvit-d12-voxtok-vo-grid-scannet@allacc\endcsname{91.5}
\expandafter\def\csname res@pvit-d12-voxtok-vo-grid-scannet@allacc-pm\endcsname{91.5}
\expandafter\def\csname res@pvit-d12-voxtok-vo-grid-scannet@iou-wall\endcsname{86.7}
\expandafter\def\csname res@pvit-d12-voxtok-vo-grid-scannet@iou-wall-pm\endcsname{86.7}
\expandafter\def\csname res@pvit-d12-voxtok-vo-grid-scannet@iou-floor\endcsname{95.7}
\expandafter\def\csname res@pvit-d12-voxtok-vo-grid-scannet@iou-floor-pm\endcsname{95.7}
\expandafter\def\csname res@pvit-d12-voxtok-vo-grid-scannet@iou-cabinet\endcsname{72.1}
\expandafter\def\csname res@pvit-d12-voxtok-vo-grid-scannet@iou-cabinet-pm\endcsname{72.1}
\expandafter\def\csname res@pvit-d12-voxtok-vo-grid-scannet@iou-bed\endcsname{82.3}
\expandafter\def\csname res@pvit-d12-voxtok-vo-grid-scannet@iou-bed-pm\endcsname{82.3}
\expandafter\def\csname res@pvit-d12-voxtok-vo-grid-scannet@iou-chair\endcsname{91.8}
\expandafter\def\csname res@pvit-d12-voxtok-vo-grid-scannet@iou-chair-pm\endcsname{91.8}
\expandafter\def\csname res@pvit-d12-voxtok-vo-grid-scannet@iou-sofa\endcsname{87.5}
\expandafter\def\csname res@pvit-d12-voxtok-vo-grid-scannet@iou-sofa-pm\endcsname{87.5}
\expandafter\def\csname res@pvit-d12-voxtok-vo-grid-scannet@iou-table\endcsname{78.3}
\expandafter\def\csname res@pvit-d12-voxtok-vo-grid-scannet@iou-table-pm\endcsname{78.3}
\expandafter\def\csname res@pvit-d12-voxtok-vo-grid-scannet@iou-door\endcsname{68.1}
\expandafter\def\csname res@pvit-d12-voxtok-vo-grid-scannet@iou-door-pm\endcsname{68.1}
\expandafter\def\csname res@pvit-d12-voxtok-vo-grid-scannet@iou-window\endcsname{72.2}
\expandafter\def\csname res@pvit-d12-voxtok-vo-grid-scannet@iou-window-pm\endcsname{72.2}
\expandafter\def\csname res@pvit-d12-voxtok-vo-grid-scannet@iou-bookshelf\endcsname{80.6}
\expandafter\def\csname res@pvit-d12-voxtok-vo-grid-scannet@iou-bookshelf-pm\endcsname{80.6}
\expandafter\def\csname res@pvit-d12-voxtok-vo-grid-scannet@iou-picture\endcsname{40.6}
\expandafter\def\csname res@pvit-d12-voxtok-vo-grid-scannet@iou-picture-pm\endcsname{40.6}
\expandafter\def\csname res@pvit-d12-voxtok-vo-grid-scannet@iou-counter\endcsname{71.3}
\expandafter\def\csname res@pvit-d12-voxtok-vo-grid-scannet@iou-counter-pm\endcsname{71.3}
\expandafter\def\csname res@pvit-d12-voxtok-vo-grid-scannet@iou-desk\endcsname{72.2}
\expandafter\def\csname res@pvit-d12-voxtok-vo-grid-scannet@iou-desk-pm\endcsname{72.2}
\expandafter\def\csname res@pvit-d12-voxtok-vo-grid-scannet@iou-curtain\endcsname{81.2}
\expandafter\def\csname res@pvit-d12-voxtok-vo-grid-scannet@iou-curtain-pm\endcsname{81.2}
\expandafter\def\csname res@pvit-d12-voxtok-vo-grid-scannet@iou-refridgerator\endcsname{73.0}
\expandafter\def\csname res@pvit-d12-voxtok-vo-grid-scannet@iou-refridgerator-pm\endcsname{73.0}
\expandafter\def\csname res@pvit-d12-voxtok-vo-grid-scannet@iou-shower curtain\endcsname{76.0}
\expandafter\def\csname res@pvit-d12-voxtok-vo-grid-scannet@iou-shower curtain-pm\endcsname{76.0}
\expandafter\def\csname res@pvit-d12-voxtok-vo-grid-scannet@iou-toilet\endcsname{95.6}
\expandafter\def\csname res@pvit-d12-voxtok-vo-grid-scannet@iou-toilet-pm\endcsname{95.6}
\expandafter\def\csname res@pvit-d12-voxtok-vo-grid-scannet@iou-sink\endcsname{70.3}
\expandafter\def\csname res@pvit-d12-voxtok-vo-grid-scannet@iou-sink-pm\endcsname{70.3}
\expandafter\def\csname res@pvit-d12-voxtok-vo-grid-scannet@iou-bathtub\endcsname{88.3}
\expandafter\def\csname res@pvit-d12-voxtok-vo-grid-scannet@iou-bathtub-pm\endcsname{88.3}
\expandafter\def\csname res@pvit-d12-voxtok-vo-grid-scannet@iou-otherfurniture\endcsname{57.8}
\expandafter\def\csname res@pvit-d12-voxtok-vo-grid-scannet@iou-otherfurniture-pm\endcsname{57.8}
\expandafter\def\csname res@pvit-d12-voxtok-vo-grid-scannet@miou-dvox\endcsname{+0.5}
\expandafter\def\csname res@pvit-d12-voxtok-vo-grid-scannet@macc-dvox\endcsname{+0.0}
\expandafter\def\csname res@pvit-d12-voxtok-vo-grid-scannet@allacc-dvox\endcsname{+0.0}
\expandafter\def\csname res@pvit-d12-geom@n\endcsname{1}
\expandafter\def\csname res@pvit-d12-geom@miou\endcsname{75.9}
\expandafter\def\csname res@pvit-d12-geom@miou-pm\endcsname{75.9}
\expandafter\def\csname res@pvit-d12-geom@macc\endcsname{84.4}
\expandafter\def\csname res@pvit-d12-geom@macc-pm\endcsname{84.4}
\expandafter\def\csname res@pvit-d12-geom@allacc\endcsname{90.9}
\expandafter\def\csname res@pvit-d12-geom@allacc-pm\endcsname{90.9}
\expandafter\def\csname res@pvit-d12-geom@iou-wall\endcsname{85.4}
\expandafter\def\csname res@pvit-d12-geom@iou-wall-pm\endcsname{85.4}
\expandafter\def\csname res@pvit-d12-geom@iou-floor\endcsname{95.6}
\expandafter\def\csname res@pvit-d12-geom@iou-floor-pm\endcsname{95.6}
\expandafter\def\csname res@pvit-d12-geom@iou-cabinet\endcsname{71.9}
\expandafter\def\csname res@pvit-d12-geom@iou-cabinet-pm\endcsname{71.9}
\expandafter\def\csname res@pvit-d12-geom@iou-bed\endcsname{80.3}
\expandafter\def\csname res@pvit-d12-geom@iou-bed-pm\endcsname{80.3}
\expandafter\def\csname res@pvit-d12-geom@iou-chair\endcsname{91.3}
\expandafter\def\csname res@pvit-d12-geom@iou-chair-pm\endcsname{91.3}
\expandafter\def\csname res@pvit-d12-geom@iou-sofa\endcsname{87.0}
\expandafter\def\csname res@pvit-d12-geom@iou-sofa-pm\endcsname{87.0}
\expandafter\def\csname res@pvit-d12-geom@iou-table\endcsname{76.6}
\expandafter\def\csname res@pvit-d12-geom@iou-table-pm\endcsname{76.6}
\expandafter\def\csname res@pvit-d12-geom@iou-door\endcsname{66.0}
\expandafter\def\csname res@pvit-d12-geom@iou-door-pm\endcsname{66.0}
\expandafter\def\csname res@pvit-d12-geom@iou-window\endcsname{70.5}
\expandafter\def\csname res@pvit-d12-geom@iou-window-pm\endcsname{70.5}
\expandafter\def\csname res@pvit-d12-geom@iou-bookshelf\endcsname{82.9}
\expandafter\def\csname res@pvit-d12-geom@iou-bookshelf-pm\endcsname{82.9}
\expandafter\def\csname res@pvit-d12-geom@iou-picture\endcsname{30.9}
\expandafter\def\csname res@pvit-d12-geom@iou-picture-pm\endcsname{30.9}
\expandafter\def\csname res@pvit-d12-geom@iou-counter\endcsname{68.5}
\expandafter\def\csname res@pvit-d12-geom@iou-counter-pm\endcsname{68.5}
\expandafter\def\csname res@pvit-d12-geom@iou-desk\endcsname{72.2}
\expandafter\def\csname res@pvit-d12-geom@iou-desk-pm\endcsname{72.2}
\expandafter\def\csname res@pvit-d12-geom@iou-curtain\endcsname{79.7}
\expandafter\def\csname res@pvit-d12-geom@iou-curtain-pm\endcsname{79.7}
\expandafter\def\csname res@pvit-d12-geom@iou-refridgerator\endcsname{68.6}
\expandafter\def\csname res@pvit-d12-geom@iou-refridgerator-pm\endcsname{68.6}
\expandafter\def\csname res@pvit-d12-geom@iou-shower curtain\endcsname{75.3}
\expandafter\def\csname res@pvit-d12-geom@iou-shower curtain-pm\endcsname{75.3}
\expandafter\def\csname res@pvit-d12-geom@iou-toilet\endcsname{95.0}
\expandafter\def\csname res@pvit-d12-geom@iou-toilet-pm\endcsname{95.0}
\expandafter\def\csname res@pvit-d12-geom@iou-sink\endcsname{72.9}
\expandafter\def\csname res@pvit-d12-geom@iou-sink-pm\endcsname{72.9}
\expandafter\def\csname res@pvit-d12-geom@iou-bathtub\endcsname{89.7}
\expandafter\def\csname res@pvit-d12-geom@iou-bathtub-pm\endcsname{89.7}
\expandafter\def\csname res@pvit-d12-geom@iou-otherfurniture\endcsname{57.6}
\expandafter\def\csname res@pvit-d12-geom@iou-otherfurniture-pm\endcsname{57.6}
\expandafter\def\csname res@pvit-d12-geom@miou-dvox\endcsname{+1.9}
\expandafter\def\csname res@pvit-d12-geom@macc-dvox\endcsname{+2.2}
\expandafter\def\csname res@pvit-d12-geom@allacc-dvox\endcsname{+0.7}
\expandafter\def\csname res@pvit-d12-geom-rope@n\endcsname{1}
\expandafter\def\csname res@pvit-d12-geom-rope@miou\endcsname{75.1}
\expandafter\def\csname res@pvit-d12-geom-rope@miou-pm\endcsname{75.1}
\expandafter\def\csname res@pvit-d12-geom-rope@macc\endcsname{83.4}
\expandafter\def\csname res@pvit-d12-geom-rope@macc-pm\endcsname{83.4}
\expandafter\def\csname res@pvit-d12-geom-rope@allacc\endcsname{90.7}
\expandafter\def\csname res@pvit-d12-geom-rope@allacc-pm\endcsname{90.7}
\expandafter\def\csname res@pvit-d12-geom-rope@iou-wall\endcsname{85.0}
\expandafter\def\csname res@pvit-d12-geom-rope@iou-wall-pm\endcsname{85.0}
\expandafter\def\csname res@pvit-d12-geom-rope@iou-floor\endcsname{95.5}
\expandafter\def\csname res@pvit-d12-geom-rope@iou-floor-pm\endcsname{95.5}
\expandafter\def\csname res@pvit-d12-geom-rope@iou-cabinet\endcsname{70.9}
\expandafter\def\csname res@pvit-d12-geom-rope@iou-cabinet-pm\endcsname{70.9}
\expandafter\def\csname res@pvit-d12-geom-rope@iou-bed\endcsname{79.7}
\expandafter\def\csname res@pvit-d12-geom-rope@iou-bed-pm\endcsname{79.7}
\expandafter\def\csname res@pvit-d12-geom-rope@iou-chair\endcsname{91.0}
\expandafter\def\csname res@pvit-d12-geom-rope@iou-chair-pm\endcsname{91.0}
\expandafter\def\csname res@pvit-d12-geom-rope@iou-sofa\endcsname{85.3}
\expandafter\def\csname res@pvit-d12-geom-rope@iou-sofa-pm\endcsname{85.3}
\expandafter\def\csname res@pvit-d12-geom-rope@iou-table\endcsname{75.1}
\expandafter\def\csname res@pvit-d12-geom-rope@iou-table-pm\endcsname{75.1}
\expandafter\def\csname res@pvit-d12-geom-rope@iou-door\endcsname{64.7}
\expandafter\def\csname res@pvit-d12-geom-rope@iou-door-pm\endcsname{64.7}
\expandafter\def\csname res@pvit-d12-geom-rope@iou-window\endcsname{71.2}
\expandafter\def\csname res@pvit-d12-geom-rope@iou-window-pm\endcsname{71.2}
\expandafter\def\csname res@pvit-d12-geom-rope@iou-bookshelf\endcsname{83.5}
\expandafter\def\csname res@pvit-d12-geom-rope@iou-bookshelf-pm\endcsname{83.5}
\expandafter\def\csname res@pvit-d12-geom-rope@iou-picture\endcsname{24.4}
\expandafter\def\csname res@pvit-d12-geom-rope@iou-picture-pm\endcsname{24.4}
\expandafter\def\csname res@pvit-d12-geom-rope@iou-counter\endcsname{67.4}
\expandafter\def\csname res@pvit-d12-geom-rope@iou-counter-pm\endcsname{67.4}
\expandafter\def\csname res@pvit-d12-geom-rope@iou-desk\endcsname{69.0}
\expandafter\def\csname res@pvit-d12-geom-rope@iou-desk-pm\endcsname{69.0}
\expandafter\def\csname res@pvit-d12-geom-rope@iou-curtain\endcsname{81.0}
\expandafter\def\csname res@pvit-d12-geom-rope@iou-curtain-pm\endcsname{81.0}
\expandafter\def\csname res@pvit-d12-geom-rope@iou-refridgerator\endcsname{69.2}
\expandafter\def\csname res@pvit-d12-geom-rope@iou-refridgerator-pm\endcsname{69.2}
\expandafter\def\csname res@pvit-d12-geom-rope@iou-shower curtain\endcsname{74.2}
\expandafter\def\csname res@pvit-d12-geom-rope@iou-shower curtain-pm\endcsname{74.2}
\expandafter\def\csname res@pvit-d12-geom-rope@iou-toilet\endcsname{94.5}
\expandafter\def\csname res@pvit-d12-geom-rope@iou-toilet-pm\endcsname{94.5}
\expandafter\def\csname res@pvit-d12-geom-rope@iou-sink\endcsname{72.5}
\expandafter\def\csname res@pvit-d12-geom-rope@iou-sink-pm\endcsname{72.5}
\expandafter\def\csname res@pvit-d12-geom-rope@iou-bathtub\endcsname{88.9}
\expandafter\def\csname res@pvit-d12-geom-rope@iou-bathtub-pm\endcsname{88.9}
\expandafter\def\csname res@pvit-d12-geom-rope@iou-otherfurniture\endcsname{59.1}
\expandafter\def\csname res@pvit-d12-geom-rope@iou-otherfurniture-pm\endcsname{59.1}
\expandafter\def\csname res@pvit-d12-geom-rope@miou-dvox\endcsname{+1.1}
\expandafter\def\csname res@pvit-d12-geom-rope@macc-dvox\endcsname{+1.2}
\expandafter\def\csname res@pvit-d12-geom-rope@allacc-dvox\endcsname{+0.5}
\expandafter\def\csname res@pvit-d12-voxtok-rope-grid-geom@n\endcsname{1}
\expandafter\def\csname res@pvit-d12-voxtok-rope-grid-geom@miou\endcsname{74.0}
\expandafter\def\csname res@pvit-d12-voxtok-rope-grid-geom@miou-pm\endcsname{74.0}
\expandafter\def\csname res@pvit-d12-voxtok-rope-grid-geom@macc\endcsname{82.2}
\expandafter\def\csname res@pvit-d12-voxtok-rope-grid-geom@macc-pm\endcsname{82.2}
\expandafter\def\csname res@pvit-d12-voxtok-rope-grid-geom@allacc\endcsname{90.2}
\expandafter\def\csname res@pvit-d12-voxtok-rope-grid-geom@allacc-pm\endcsname{90.2}
\expandafter\def\csname res@pvit-d12-voxtok-rope-grid-geom@iou-wall\endcsname{83.9}
\expandafter\def\csname res@pvit-d12-voxtok-rope-grid-geom@iou-wall-pm\endcsname{83.9}
\expandafter\def\csname res@pvit-d12-voxtok-rope-grid-geom@iou-floor\endcsname{95.3}
\expandafter\def\csname res@pvit-d12-voxtok-rope-grid-geom@iou-floor-pm\endcsname{95.3}
\expandafter\def\csname res@pvit-d12-voxtok-rope-grid-geom@iou-cabinet\endcsname{69.3}
\expandafter\def\csname res@pvit-d12-voxtok-rope-grid-geom@iou-cabinet-pm\endcsname{69.3}
\expandafter\def\csname res@pvit-d12-voxtok-rope-grid-geom@iou-bed\endcsname{80.7}
\expandafter\def\csname res@pvit-d12-voxtok-rope-grid-geom@iou-bed-pm\endcsname{80.7}
\expandafter\def\csname res@pvit-d12-voxtok-rope-grid-geom@iou-chair\endcsname{90.2}
\expandafter\def\csname res@pvit-d12-voxtok-rope-grid-geom@iou-chair-pm\endcsname{90.2}
\expandafter\def\csname res@pvit-d12-voxtok-rope-grid-geom@iou-sofa\endcsname{86.9}
\expandafter\def\csname res@pvit-d12-voxtok-rope-grid-geom@iou-sofa-pm\endcsname{86.9}
\expandafter\def\csname res@pvit-d12-voxtok-rope-grid-geom@iou-table\endcsname{75.5}
\expandafter\def\csname res@pvit-d12-voxtok-rope-grid-geom@iou-table-pm\endcsname{75.5}
\expandafter\def\csname res@pvit-d12-voxtok-rope-grid-geom@iou-door\endcsname{60.5}
\expandafter\def\csname res@pvit-d12-voxtok-rope-grid-geom@iou-door-pm\endcsname{60.5}
\expandafter\def\csname res@pvit-d12-voxtok-rope-grid-geom@iou-window\endcsname{67.0}
\expandafter\def\csname res@pvit-d12-voxtok-rope-grid-geom@iou-window-pm\endcsname{67.0}
\expandafter\def\csname res@pvit-d12-voxtok-rope-grid-geom@iou-bookshelf\endcsname{82.1}
\expandafter\def\csname res@pvit-d12-voxtok-rope-grid-geom@iou-bookshelf-pm\endcsname{82.1}
\expandafter\def\csname res@pvit-d12-voxtok-rope-grid-geom@iou-picture\endcsname{20.2}
\expandafter\def\csname res@pvit-d12-voxtok-rope-grid-geom@iou-picture-pm\endcsname{20.2}
\expandafter\def\csname res@pvit-d12-voxtok-rope-grid-geom@iou-counter\endcsname{63.7}
\expandafter\def\csname res@pvit-d12-voxtok-rope-grid-geom@iou-counter-pm\endcsname{63.7}
\expandafter\def\csname res@pvit-d12-voxtok-rope-grid-geom@iou-desk\endcsname{70.7}
\expandafter\def\csname res@pvit-d12-voxtok-rope-grid-geom@iou-desk-pm\endcsname{70.7}
\expandafter\def\csname res@pvit-d12-voxtok-rope-grid-geom@iou-curtain\endcsname{79.9}
\expandafter\def\csname res@pvit-d12-voxtok-rope-grid-geom@iou-curtain-pm\endcsname{79.9}
\expandafter\def\csname res@pvit-d12-voxtok-rope-grid-geom@iou-refridgerator\endcsname{64.2}
\expandafter\def\csname res@pvit-d12-voxtok-rope-grid-geom@iou-refridgerator-pm\endcsname{64.2}
\expandafter\def\csname res@pvit-d12-voxtok-rope-grid-geom@iou-shower curtain\endcsname{74.7}
\expandafter\def\csname res@pvit-d12-voxtok-rope-grid-geom@iou-shower curtain-pm\endcsname{74.7}
\expandafter\def\csname res@pvit-d12-voxtok-rope-grid-geom@iou-toilet\endcsname{94.2}
\expandafter\def\csname res@pvit-d12-voxtok-rope-grid-geom@iou-toilet-pm\endcsname{94.2}
\expandafter\def\csname res@pvit-d12-voxtok-rope-grid-geom@iou-sink\endcsname{72.4}
\expandafter\def\csname res@pvit-d12-voxtok-rope-grid-geom@iou-sink-pm\endcsname{72.4}
\expandafter\def\csname res@pvit-d12-voxtok-rope-grid-geom@iou-bathtub\endcsname{87.7}
\expandafter\def\csname res@pvit-d12-voxtok-rope-grid-geom@iou-bathtub-pm\endcsname{87.7}
\expandafter\def\csname res@pvit-d12-voxtok-rope-grid-geom@iou-otherfurniture\endcsname{61.1}
\expandafter\def\csname res@pvit-d12-voxtok-rope-grid-geom@iou-otherfurniture-pm\endcsname{61.1}
\expandafter\def\csname res@pvit-d12-voxtok-vo-grid-geom@n\endcsname{1}
\expandafter\def\csname res@pvit-d12-voxtok-vo-grid-geom@miou\endcsname{75.1}
\expandafter\def\csname res@pvit-d12-voxtok-vo-grid-geom@miou-pm\endcsname{75.1}
\expandafter\def\csname res@pvit-d12-voxtok-vo-grid-geom@macc\endcsname{82.9}
\expandafter\def\csname res@pvit-d12-voxtok-vo-grid-geom@macc-pm\endcsname{82.9}
\expandafter\def\csname res@pvit-d12-voxtok-vo-grid-geom@allacc\endcsname{90.6}
\expandafter\def\csname res@pvit-d12-voxtok-vo-grid-geom@allacc-pm\endcsname{90.6}
\expandafter\def\csname res@pvit-d12-voxtok-vo-grid-geom@iou-wall\endcsname{84.3}
\expandafter\def\csname res@pvit-d12-voxtok-vo-grid-geom@iou-wall-pm\endcsname{84.3}
\expandafter\def\csname res@pvit-d12-voxtok-vo-grid-geom@iou-floor\endcsname{95.4}
\expandafter\def\csname res@pvit-d12-voxtok-vo-grid-geom@iou-floor-pm\endcsname{95.4}
\expandafter\def\csname res@pvit-d12-voxtok-vo-grid-geom@iou-cabinet\endcsname{70.0}
\expandafter\def\csname res@pvit-d12-voxtok-vo-grid-geom@iou-cabinet-pm\endcsname{70.0}
\expandafter\def\csname res@pvit-d12-voxtok-vo-grid-geom@iou-bed\endcsname{80.1}
\expandafter\def\csname res@pvit-d12-voxtok-vo-grid-geom@iou-bed-pm\endcsname{80.1}
\expandafter\def\csname res@pvit-d12-voxtok-vo-grid-geom@iou-chair\endcsname{90.6}
\expandafter\def\csname res@pvit-d12-voxtok-vo-grid-geom@iou-chair-pm\endcsname{90.6}
\expandafter\def\csname res@pvit-d12-voxtok-vo-grid-geom@iou-sofa\endcsname{86.3}
\expandafter\def\csname res@pvit-d12-voxtok-vo-grid-geom@iou-sofa-pm\endcsname{86.3}
\expandafter\def\csname res@pvit-d12-voxtok-vo-grid-geom@iou-table\endcsname{76.8}
\expandafter\def\csname res@pvit-d12-voxtok-vo-grid-geom@iou-table-pm\endcsname{76.8}
\expandafter\def\csname res@pvit-d12-voxtok-vo-grid-geom@iou-door\endcsname{60.8}
\expandafter\def\csname res@pvit-d12-voxtok-vo-grid-geom@iou-door-pm\endcsname{60.8}
\expandafter\def\csname res@pvit-d12-voxtok-vo-grid-geom@iou-window\endcsname{69.9}
\expandafter\def\csname res@pvit-d12-voxtok-vo-grid-geom@iou-window-pm\endcsname{69.9}
\expandafter\def\csname res@pvit-d12-voxtok-vo-grid-geom@iou-bookshelf\endcsname{82.5}
\expandafter\def\csname res@pvit-d12-voxtok-vo-grid-geom@iou-bookshelf-pm\endcsname{82.5}
\expandafter\def\csname res@pvit-d12-voxtok-vo-grid-geom@iou-picture\endcsname{22.2}
\expandafter\def\csname res@pvit-d12-voxtok-vo-grid-geom@iou-picture-pm\endcsname{22.2}
\expandafter\def\csname res@pvit-d12-voxtok-vo-grid-geom@iou-counter\endcsname{67.4}
\expandafter\def\csname res@pvit-d12-voxtok-vo-grid-geom@iou-counter-pm\endcsname{67.4}
\expandafter\def\csname res@pvit-d12-voxtok-vo-grid-geom@iou-desk\endcsname{72.5}
\expandafter\def\csname res@pvit-d12-voxtok-vo-grid-geom@iou-desk-pm\endcsname{72.5}
\expandafter\def\csname res@pvit-d12-voxtok-vo-grid-geom@iou-curtain\endcsname{80.9}
\expandafter\def\csname res@pvit-d12-voxtok-vo-grid-geom@iou-curtain-pm\endcsname{80.9}
\expandafter\def\csname res@pvit-d12-voxtok-vo-grid-geom@iou-refridgerator\endcsname{65.5}
\expandafter\def\csname res@pvit-d12-voxtok-vo-grid-geom@iou-refridgerator-pm\endcsname{65.5}
\expandafter\def\csname res@pvit-d12-voxtok-vo-grid-geom@iou-shower curtain\endcsname{75.9}
\expandafter\def\csname res@pvit-d12-voxtok-vo-grid-geom@iou-shower curtain-pm\endcsname{75.9}
\expandafter\def\csname res@pvit-d12-voxtok-vo-grid-geom@iou-toilet\endcsname{94.3}
\expandafter\def\csname res@pvit-d12-voxtok-vo-grid-geom@iou-toilet-pm\endcsname{94.3}
\expandafter\def\csname res@pvit-d12-voxtok-vo-grid-geom@iou-sink\endcsname{73.4}
\expandafter\def\csname res@pvit-d12-voxtok-vo-grid-geom@iou-sink-pm\endcsname{73.4}
\expandafter\def\csname res@pvit-d12-voxtok-vo-grid-geom@iou-bathtub\endcsname{89.2}
\expandafter\def\csname res@pvit-d12-voxtok-vo-grid-geom@iou-bathtub-pm\endcsname{89.2}
\expandafter\def\csname res@pvit-d12-voxtok-vo-grid-geom@iou-otherfurniture\endcsname{63.4}
\expandafter\def\csname res@pvit-d12-voxtok-vo-grid-geom@iou-otherfurniture-pm\endcsname{63.4}
\expandafter\def\csname res@pvit-d12-voxtok-vo-grid-geom@miou-dvox\endcsname{+1.1}
\expandafter\def\csname res@pvit-d12-voxtok-vo-grid-geom@macc-dvox\endcsname{+0.7}
\expandafter\def\csname res@pvit-d12-voxtok-vo-grid-geom@allacc-dvox\endcsname{+0.4}
\expandafter\def\csname res@pvit-d12-voxtok-geom@n\endcsname{1}
\expandafter\def\csname res@pvit-d12-voxtok-geom@miou\endcsname{74.6}
\expandafter\def\csname res@pvit-d12-voxtok-geom@miou-pm\endcsname{74.6}
\expandafter\def\csname res@pvit-d12-voxtok-geom@macc\endcsname{82.6}
\expandafter\def\csname res@pvit-d12-voxtok-geom@macc-pm\endcsname{82.6}
\expandafter\def\csname res@pvit-d12-voxtok-geom@allacc\endcsname{90.6}
\expandafter\def\csname res@pvit-d12-voxtok-geom@allacc-pm\endcsname{90.6}
\expandafter\def\csname res@pvit-d12-voxtok-geom@iou-wall\endcsname{84.6}
\expandafter\def\csname res@pvit-d12-voxtok-geom@iou-wall-pm\endcsname{84.6}
\expandafter\def\csname res@pvit-d12-voxtok-geom@iou-floor\endcsname{95.4}
\expandafter\def\csname res@pvit-d12-voxtok-geom@iou-floor-pm\endcsname{95.4}
\expandafter\def\csname res@pvit-d12-voxtok-geom@iou-cabinet\endcsname{69.6}
\expandafter\def\csname res@pvit-d12-voxtok-geom@iou-cabinet-pm\endcsname{69.6}
\expandafter\def\csname res@pvit-d12-voxtok-geom@iou-bed\endcsname{80.0}
\expandafter\def\csname res@pvit-d12-voxtok-geom@iou-bed-pm\endcsname{80.0}
\expandafter\def\csname res@pvit-d12-voxtok-geom@iou-chair\endcsname{90.3}
\expandafter\def\csname res@pvit-d12-voxtok-geom@iou-chair-pm\endcsname{90.3}
\expandafter\def\csname res@pvit-d12-voxtok-geom@iou-sofa\endcsname{85.3}
\expandafter\def\csname res@pvit-d12-voxtok-geom@iou-sofa-pm\endcsname{85.3}
\expandafter\def\csname res@pvit-d12-voxtok-geom@iou-table\endcsname{75.7}
\expandafter\def\csname res@pvit-d12-voxtok-geom@iou-table-pm\endcsname{75.7}
\expandafter\def\csname res@pvit-d12-voxtok-geom@iou-door\endcsname{63.2}
\expandafter\def\csname res@pvit-d12-voxtok-geom@iou-door-pm\endcsname{63.2}
\expandafter\def\csname res@pvit-d12-voxtok-geom@iou-window\endcsname{70.6}
\expandafter\def\csname res@pvit-d12-voxtok-geom@iou-window-pm\endcsname{70.6}
\expandafter\def\csname res@pvit-d12-voxtok-geom@iou-bookshelf\endcsname{81.8}
\expandafter\def\csname res@pvit-d12-voxtok-geom@iou-bookshelf-pm\endcsname{81.8}
\expandafter\def\csname res@pvit-d12-voxtok-geom@iou-picture\endcsname{22.6}
\expandafter\def\csname res@pvit-d12-voxtok-geom@iou-picture-pm\endcsname{22.6}
\expandafter\def\csname res@pvit-d12-voxtok-geom@iou-counter\endcsname{64.8}
\expandafter\def\csname res@pvit-d12-voxtok-geom@iou-counter-pm\endcsname{64.8}
\expandafter\def\csname res@pvit-d12-voxtok-geom@iou-desk\endcsname{69.7}
\expandafter\def\csname res@pvit-d12-voxtok-geom@iou-desk-pm\endcsname{69.7}
\expandafter\def\csname res@pvit-d12-voxtok-geom@iou-curtain\endcsname{80.3}
\expandafter\def\csname res@pvit-d12-voxtok-geom@iou-curtain-pm\endcsname{80.3}
\expandafter\def\csname res@pvit-d12-voxtok-geom@iou-refridgerator\endcsname{64.0}
\expandafter\def\csname res@pvit-d12-voxtok-geom@iou-refridgerator-pm\endcsname{64.0}
\expandafter\def\csname res@pvit-d12-voxtok-geom@iou-shower curtain\endcsname{75.1}
\expandafter\def\csname res@pvit-d12-voxtok-geom@iou-shower curtain-pm\endcsname{75.1}
\expandafter\def\csname res@pvit-d12-voxtok-geom@iou-toilet\endcsname{94.6}
\expandafter\def\csname res@pvit-d12-voxtok-geom@iou-toilet-pm\endcsname{94.6}
\expandafter\def\csname res@pvit-d12-voxtok-geom@iou-sink\endcsname{71.8}
\expandafter\def\csname res@pvit-d12-voxtok-geom@iou-sink-pm\endcsname{71.8}
\expandafter\def\csname res@pvit-d12-voxtok-geom@iou-bathtub\endcsname{89.8}
\expandafter\def\csname res@pvit-d12-voxtok-geom@iou-bathtub-pm\endcsname{89.8}
\expandafter\def\csname res@pvit-d12-voxtok-geom@iou-otherfurniture\endcsname{62.3}
\expandafter\def\csname res@pvit-d12-voxtok-geom@iou-otherfurniture-pm\endcsname{62.3}
\expandafter\def\csname res@pvit-d12-voxtok-geom@miou-dvox\endcsname{+0.6}
\expandafter\def\csname res@pvit-d12-voxtok-geom@macc-dvox\endcsname{+0.4}
\expandafter\def\csname res@pvit-d12-voxtok-geom@allacc-dvox\endcsname{+0.4}
\expandafter\def\csname res@pvit-d12-global@n\endcsname{1}
\expandafter\def\csname res@pvit-d12-global@miou\endcsname{77.6}
\expandafter\def\csname res@pvit-d12-global@miou-pm\endcsname{77.6}
\expandafter\def\csname res@pvit-d12-global@macc\endcsname{85.0}
\expandafter\def\csname res@pvit-d12-global@macc-pm\endcsname{85.0}
\expandafter\def\csname res@pvit-d12-global@allacc\endcsname{91.8}
\expandafter\def\csname res@pvit-d12-global@allacc-pm\endcsname{91.8}
\expandafter\def\csname res@pvit-d12-global@iou-wall\endcsname{87.1}
\expandafter\def\csname res@pvit-d12-global@iou-wall-pm\endcsname{87.1}
\expandafter\def\csname res@pvit-d12-global@iou-floor\endcsname{95.8}
\expandafter\def\csname res@pvit-d12-global@iou-floor-pm\endcsname{95.8}
\expandafter\def\csname res@pvit-d12-global@iou-cabinet\endcsname{72.1}
\expandafter\def\csname res@pvit-d12-global@iou-cabinet-pm\endcsname{72.1}
\expandafter\def\csname res@pvit-d12-global@iou-bed\endcsname{83.1}
\expandafter\def\csname res@pvit-d12-global@iou-bed-pm\endcsname{83.1}
\expandafter\def\csname res@pvit-d12-global@iou-chair\endcsname{91.5}
\expandafter\def\csname res@pvit-d12-global@iou-chair-pm\endcsname{91.5}
\expandafter\def\csname res@pvit-d12-global@iou-sofa\endcsname{85.3}
\expandafter\def\csname res@pvit-d12-global@iou-sofa-pm\endcsname{85.3}
\expandafter\def\csname res@pvit-d12-global@iou-table\endcsname{78.6}
\expandafter\def\csname res@pvit-d12-global@iou-table-pm\endcsname{78.6}
\expandafter\def\csname res@pvit-d12-global@iou-door\endcsname{69.8}
\expandafter\def\csname res@pvit-d12-global@iou-door-pm\endcsname{69.8}
\expandafter\def\csname res@pvit-d12-global@iou-window\endcsname{71.5}
\expandafter\def\csname res@pvit-d12-global@iou-window-pm\endcsname{71.5}
\expandafter\def\csname res@pvit-d12-global@iou-bookshelf\endcsname{87.6}
\expandafter\def\csname res@pvit-d12-global@iou-bookshelf-pm\endcsname{87.6}
\expandafter\def\csname res@pvit-d12-global@iou-picture\endcsname{36.3}
\expandafter\def\csname res@pvit-d12-global@iou-picture-pm\endcsname{36.3}
\expandafter\def\csname res@pvit-d12-global@iou-counter\endcsname{68.4}
\expandafter\def\csname res@pvit-d12-global@iou-counter-pm\endcsname{68.4}
\expandafter\def\csname res@pvit-d12-global@iou-desk\endcsname{74.2}
\expandafter\def\csname res@pvit-d12-global@iou-desk-pm\endcsname{74.2}
\expandafter\def\csname res@pvit-d12-global@iou-curtain\endcsname{80.0}
\expandafter\def\csname res@pvit-d12-global@iou-curtain-pm\endcsname{80.0}
\expandafter\def\csname res@pvit-d12-global@iou-refridgerator\endcsname{77.4}
\expandafter\def\csname res@pvit-d12-global@iou-refridgerator-pm\endcsname{77.4}
\expandafter\def\csname res@pvit-d12-global@iou-shower curtain\endcsname{73.3}
\expandafter\def\csname res@pvit-d12-global@iou-shower curtain-pm\endcsname{73.3}
\expandafter\def\csname res@pvit-d12-global@iou-toilet\endcsname{95.1}
\expandafter\def\csname res@pvit-d12-global@iou-toilet-pm\endcsname{95.1}
\expandafter\def\csname res@pvit-d12-global@iou-sink\endcsname{72.3}
\expandafter\def\csname res@pvit-d12-global@iou-sink-pm\endcsname{72.3}
\expandafter\def\csname res@pvit-d12-global@iou-bathtub\endcsname{90.1}
\expandafter\def\csname res@pvit-d12-global@iou-bathtub-pm\endcsname{90.1}
\expandafter\def\csname res@pvit-d12-global@iou-otherfurniture\endcsname{61.8}
\expandafter\def\csname res@pvit-d12-global@iou-otherfurniture-pm\endcsname{61.8}
\expandafter\def\csname res@pvit-d12-global@miou-delta\endcsname{+0.1}
\expandafter\def\csname res@pvit-d12-global@macc-delta\endcsname{\ensuremath{-}0.8}
\expandafter\def\csname res@pvit-d12-global@allacc-delta\endcsname{\ensuremath{-}0.1}
\expandafter\def\csname res@pvit-d12-global@iou-wall-delta\endcsname{\ensuremath{-}0.7}
\expandafter\def\csname res@pvit-d12-global@iou-floor-delta\endcsname{\ensuremath{-}0.1}
\expandafter\def\csname res@pvit-d12-global@iou-cabinet-delta\endcsname{\ensuremath{-}0.3}
\expandafter\def\csname res@pvit-d12-global@iou-bed-delta\endcsname{+0.1}
\expandafter\def\csname res@pvit-d12-global@iou-chair-delta\endcsname{\ensuremath{-}1.2}
\expandafter\def\csname res@pvit-d12-global@iou-sofa-delta\endcsname{\ensuremath{-}1.5}
\expandafter\def\csname res@pvit-d12-global@iou-table-delta\endcsname{+1.5}
\expandafter\def\csname res@pvit-d12-global@iou-door-delta\endcsname{+0.2}
\expandafter\def\csname res@pvit-d12-global@iou-window-delta\endcsname{+0.0}
\expandafter\def\csname res@pvit-d12-global@iou-bookshelf-delta\endcsname{+2.1}
\expandafter\def\csname res@pvit-d12-global@iou-picture-delta\endcsname{\ensuremath{-}3.6}
\expandafter\def\csname res@pvit-d12-global@iou-counter-delta\endcsname{\ensuremath{-}2.1}
\expandafter\def\csname res@pvit-d12-global@iou-desk-delta\endcsname{+2.1}
\expandafter\def\csname res@pvit-d12-global@iou-curtain-delta\endcsname{\ensuremath{-}0.7}
\expandafter\def\csname res@pvit-d12-global@iou-refridgerator-delta\endcsname{+6.5}
\expandafter\def\csname res@pvit-d12-global@iou-shower curtain-delta\endcsname{\ensuremath{-}2.1}
\expandafter\def\csname res@pvit-d12-global@iou-toilet-delta\endcsname{\ensuremath{-}0.9}
\expandafter\def\csname res@pvit-d12-global@iou-sink-delta\endcsname{+1.1}
\expandafter\def\csname res@pvit-d12-global@iou-bathtub-delta\endcsname{+1.3}
\expandafter\def\csname res@pvit-d12-global@iou-otherfurniture-delta\endcsname{\ensuremath{-}0.7}
\expandafter\def\csname res@pvit-d12-noqknorm@n\endcsname{1}
\expandafter\def\csname res@pvit-d12-noqknorm@miou\endcsname{77.0}
\expandafter\def\csname res@pvit-d12-noqknorm@miou-pm\endcsname{77.0}
\expandafter\def\csname res@pvit-d12-noqknorm@macc\endcsname{85.3}
\expandafter\def\csname res@pvit-d12-noqknorm@macc-pm\endcsname{85.3}
\expandafter\def\csname res@pvit-d12-noqknorm@allacc\endcsname{91.7}
\expandafter\def\csname res@pvit-d12-noqknorm@allacc-pm\endcsname{91.7}
\expandafter\def\csname res@pvit-d12-noqknorm@iou-wall\endcsname{87.5}
\expandafter\def\csname res@pvit-d12-noqknorm@iou-wall-pm\endcsname{87.5}
\expandafter\def\csname res@pvit-d12-noqknorm@iou-floor\endcsname{95.8}
\expandafter\def\csname res@pvit-d12-noqknorm@iou-floor-pm\endcsname{95.8}
\expandafter\def\csname res@pvit-d12-noqknorm@iou-cabinet\endcsname{69.8}
\expandafter\def\csname res@pvit-d12-noqknorm@iou-cabinet-pm\endcsname{69.8}
\expandafter\def\csname res@pvit-d12-noqknorm@iou-bed\endcsname{82.9}
\expandafter\def\csname res@pvit-d12-noqknorm@iou-bed-pm\endcsname{82.9}
\expandafter\def\csname res@pvit-d12-noqknorm@iou-chair\endcsname{92.7}
\expandafter\def\csname res@pvit-d12-noqknorm@iou-chair-pm\endcsname{92.7}
\expandafter\def\csname res@pvit-d12-noqknorm@iou-sofa\endcsname{87.7}
\expandafter\def\csname res@pvit-d12-noqknorm@iou-sofa-pm\endcsname{87.7}
\expandafter\def\csname res@pvit-d12-noqknorm@iou-table\endcsname{77.1}
\expandafter\def\csname res@pvit-d12-noqknorm@iou-table-pm\endcsname{77.1}
\expandafter\def\csname res@pvit-d12-noqknorm@iou-door\endcsname{71.6}
\expandafter\def\csname res@pvit-d12-noqknorm@iou-door-pm\endcsname{71.6}
\expandafter\def\csname res@pvit-d12-noqknorm@iou-window\endcsname{73.2}
\expandafter\def\csname res@pvit-d12-noqknorm@iou-window-pm\endcsname{73.2}
\expandafter\def\csname res@pvit-d12-noqknorm@iou-bookshelf\endcsname{84.2}
\expandafter\def\csname res@pvit-d12-noqknorm@iou-bookshelf-pm\endcsname{84.2}
\expandafter\def\csname res@pvit-d12-noqknorm@iou-picture\endcsname{40.6}
\expandafter\def\csname res@pvit-d12-noqknorm@iou-picture-pm\endcsname{40.6}
\expandafter\def\csname res@pvit-d12-noqknorm@iou-counter\endcsname{71.0}
\expandafter\def\csname res@pvit-d12-noqknorm@iou-counter-pm\endcsname{71.0}
\expandafter\def\csname res@pvit-d12-noqknorm@iou-desk\endcsname{72.9}
\expandafter\def\csname res@pvit-d12-noqknorm@iou-desk-pm\endcsname{72.9}
\expandafter\def\csname res@pvit-d12-noqknorm@iou-curtain\endcsname{80.3}
\expandafter\def\csname res@pvit-d12-noqknorm@iou-curtain-pm\endcsname{80.3}
\expandafter\def\csname res@pvit-d12-noqknorm@iou-refridgerator\endcsname{70.7}
\expandafter\def\csname res@pvit-d12-noqknorm@iou-refridgerator-pm\endcsname{70.7}
\expandafter\def\csname res@pvit-d12-noqknorm@iou-shower curtain\endcsname{68.2}
\expandafter\def\csname res@pvit-d12-noqknorm@iou-shower curtain-pm\endcsname{68.2}
\expandafter\def\csname res@pvit-d12-noqknorm@iou-toilet\endcsname{95.9}
\expandafter\def\csname res@pvit-d12-noqknorm@iou-toilet-pm\endcsname{95.9}
\expandafter\def\csname res@pvit-d12-noqknorm@iou-sink\endcsname{70.6}
\expandafter\def\csname res@pvit-d12-noqknorm@iou-sink-pm\endcsname{70.6}
\expandafter\def\csname res@pvit-d12-noqknorm@iou-bathtub\endcsname{88.1}
\expandafter\def\csname res@pvit-d12-noqknorm@iou-bathtub-pm\endcsname{88.1}
\expandafter\def\csname res@pvit-d12-noqknorm@iou-otherfurniture\endcsname{58.8}
\expandafter\def\csname res@pvit-d12-noqknorm@iou-otherfurniture-pm\endcsname{58.8}
\expandafter\def\csname res@pvit-d12-noqknorm@miou-delta\endcsname{\ensuremath{-}0.5}
\expandafter\def\csname res@pvit-d12-noqknorm@macc-delta\endcsname{\ensuremath{-}0.5}
\expandafter\def\csname res@pvit-d12-noqknorm@allacc-delta\endcsname{\ensuremath{-}0.2}
\expandafter\def\csname res@pvit-d12-noqknorm@iou-wall-delta\endcsname{\ensuremath{-}0.3}
\expandafter\def\csname res@pvit-d12-noqknorm@iou-floor-delta\endcsname{\ensuremath{-}0.1}
\expandafter\def\csname res@pvit-d12-noqknorm@iou-cabinet-delta\endcsname{\ensuremath{-}2.6}
\expandafter\def\csname res@pvit-d12-noqknorm@iou-bed-delta\endcsname{\ensuremath{-}0.1}
\expandafter\def\csname res@pvit-d12-noqknorm@iou-chair-delta\endcsname{+0.0}
\expandafter\def\csname res@pvit-d12-noqknorm@iou-sofa-delta\endcsname{+0.9}
\expandafter\def\csname res@pvit-d12-noqknorm@iou-table-delta\endcsname{+0.0}
\expandafter\def\csname res@pvit-d12-noqknorm@iou-door-delta\endcsname{+2.0}
\expandafter\def\csname res@pvit-d12-noqknorm@iou-window-delta\endcsname{+1.7}
\expandafter\def\csname res@pvit-d12-noqknorm@iou-bookshelf-delta\endcsname{\ensuremath{-}1.3}
\expandafter\def\csname res@pvit-d12-noqknorm@iou-picture-delta\endcsname{+0.7}
\expandafter\def\csname res@pvit-d12-noqknorm@iou-counter-delta\endcsname{+0.5}
\expandafter\def\csname res@pvit-d12-noqknorm@iou-desk-delta\endcsname{+0.8}
\expandafter\def\csname res@pvit-d12-noqknorm@iou-curtain-delta\endcsname{\ensuremath{-}0.4}
\expandafter\def\csname res@pvit-d12-noqknorm@iou-refridgerator-delta\endcsname{\ensuremath{-}0.2}
\expandafter\def\csname res@pvit-d12-noqknorm@iou-shower curtain-delta\endcsname{\ensuremath{-}7.2}
\expandafter\def\csname res@pvit-d12-noqknorm@iou-toilet-delta\endcsname{\ensuremath{-}0.1}
\expandafter\def\csname res@pvit-d12-noqknorm@iou-sink-delta\endcsname{\ensuremath{-}0.6}
\expandafter\def\csname res@pvit-d12-noqknorm@iou-bathtub-delta\endcsname{\ensuremath{-}0.7}
\expandafter\def\csname res@pvit-d12-noqknorm@iou-otherfurniture-delta\endcsname{\ensuremath{-}3.7}
\expandafter\def\csname res@pvit-d12-hd64@n\endcsname{1}
\expandafter\def\csname res@pvit-d12-hd64@miou\endcsname{77.1}
\expandafter\def\csname res@pvit-d12-hd64@miou-pm\endcsname{77.1}
\expandafter\def\csname res@pvit-d12-hd64@macc\endcsname{84.9}
\expandafter\def\csname res@pvit-d12-hd64@macc-pm\endcsname{84.9}
\expandafter\def\csname res@pvit-d12-hd64@allacc\endcsname{91.9}
\expandafter\def\csname res@pvit-d12-hd64@allacc-pm\endcsname{91.9}
\expandafter\def\csname res@pvit-d12-hd64@iou-wall\endcsname{87.3}
\expandafter\def\csname res@pvit-d12-hd64@iou-wall-pm\endcsname{87.3}
\expandafter\def\csname res@pvit-d12-hd64@iou-floor\endcsname{95.8}
\expandafter\def\csname res@pvit-d12-hd64@iou-floor-pm\endcsname{95.8}
\expandafter\def\csname res@pvit-d12-hd64@iou-cabinet\endcsname{72.2}
\expandafter\def\csname res@pvit-d12-hd64@iou-cabinet-pm\endcsname{72.2}
\expandafter\def\csname res@pvit-d12-hd64@iou-bed\endcsname{83.5}
\expandafter\def\csname res@pvit-d12-hd64@iou-bed-pm\endcsname{83.5}
\expandafter\def\csname res@pvit-d12-hd64@iou-chair\endcsname{92.4}
\expandafter\def\csname res@pvit-d12-hd64@iou-chair-pm\endcsname{92.4}
\expandafter\def\csname res@pvit-d12-hd64@iou-sofa\endcsname{85.8}
\expandafter\def\csname res@pvit-d12-hd64@iou-sofa-pm\endcsname{85.8}
\expandafter\def\csname res@pvit-d12-hd64@iou-table\endcsname{77.5}
\expandafter\def\csname res@pvit-d12-hd64@iou-table-pm\endcsname{77.5}
\expandafter\def\csname res@pvit-d12-hd64@iou-door\endcsname{73.5}
\expandafter\def\csname res@pvit-d12-hd64@iou-door-pm\endcsname{73.5}
\expandafter\def\csname res@pvit-d12-hd64@iou-window\endcsname{71.7}
\expandafter\def\csname res@pvit-d12-hd64@iou-window-pm\endcsname{71.7}
\expandafter\def\csname res@pvit-d12-hd64@iou-bookshelf\endcsname{84.7}
\expandafter\def\csname res@pvit-d12-hd64@iou-bookshelf-pm\endcsname{84.7}
\expandafter\def\csname res@pvit-d12-hd64@iou-picture\endcsname{39.1}
\expandafter\def\csname res@pvit-d12-hd64@iou-picture-pm\endcsname{39.1}
\expandafter\def\csname res@pvit-d12-hd64@iou-counter\endcsname{69.1}
\expandafter\def\csname res@pvit-d12-hd64@iou-counter-pm\endcsname{69.1}
\expandafter\def\csname res@pvit-d12-hd64@iou-desk\endcsname{72.9}
\expandafter\def\csname res@pvit-d12-hd64@iou-desk-pm\endcsname{72.9}
\expandafter\def\csname res@pvit-d12-hd64@iou-curtain\endcsname{78.8}
\expandafter\def\csname res@pvit-d12-hd64@iou-curtain-pm\endcsname{78.8}
\expandafter\def\csname res@pvit-d12-hd64@iou-refridgerator\endcsname{71.6}
\expandafter\def\csname res@pvit-d12-hd64@iou-refridgerator-pm\endcsname{71.6}
\expandafter\def\csname res@pvit-d12-hd64@iou-shower curtain\endcsname{64.6}
\expandafter\def\csname res@pvit-d12-hd64@iou-shower curtain-pm\endcsname{64.6}
\expandafter\def\csname res@pvit-d12-hd64@iou-toilet\endcsname{96.0}
\expandafter\def\csname res@pvit-d12-hd64@iou-toilet-pm\endcsname{96.0}
\expandafter\def\csname res@pvit-d12-hd64@iou-sink\endcsname{72.1}
\expandafter\def\csname res@pvit-d12-hd64@iou-sink-pm\endcsname{72.1}
\expandafter\def\csname res@pvit-d12-hd64@iou-bathtub\endcsname{89.4}
\expandafter\def\csname res@pvit-d12-hd64@iou-bathtub-pm\endcsname{89.4}
\expandafter\def\csname res@pvit-d12-hd64@iou-otherfurniture\endcsname{64.2}
\expandafter\def\csname res@pvit-d12-hd64@iou-otherfurniture-pm\endcsname{64.2}
\expandafter\def\csname res@pvit-d12-hd64@miou-delta\endcsname{\ensuremath{-}0.4}
\expandafter\def\csname res@pvit-d12-hd64@macc-delta\endcsname{\ensuremath{-}0.9}
\expandafter\def\csname res@pvit-d12-hd64@allacc-delta\endcsname{+0.0}
\expandafter\def\csname res@pvit-d12-hd64@iou-wall-delta\endcsname{\ensuremath{-}0.5}
\expandafter\def\csname res@pvit-d12-hd64@iou-floor-delta\endcsname{\ensuremath{-}0.1}
\expandafter\def\csname res@pvit-d12-hd64@iou-cabinet-delta\endcsname{\ensuremath{-}0.2}
\expandafter\def\csname res@pvit-d12-hd64@iou-bed-delta\endcsname{+0.5}
\expandafter\def\csname res@pvit-d12-hd64@iou-chair-delta\endcsname{\ensuremath{-}0.3}
\expandafter\def\csname res@pvit-d12-hd64@iou-sofa-delta\endcsname{\ensuremath{-}1.0}
\expandafter\def\csname res@pvit-d12-hd64@iou-table-delta\endcsname{+0.4}
\expandafter\def\csname res@pvit-d12-hd64@iou-door-delta\endcsname{+3.9}
\expandafter\def\csname res@pvit-d12-hd64@iou-window-delta\endcsname{+0.2}
\expandafter\def\csname res@pvit-d12-hd64@iou-bookshelf-delta\endcsname{\ensuremath{-}0.8}
\expandafter\def\csname res@pvit-d12-hd64@iou-picture-delta\endcsname{\ensuremath{-}0.8}
\expandafter\def\csname res@pvit-d12-hd64@iou-counter-delta\endcsname{\ensuremath{-}1.4}
\expandafter\def\csname res@pvit-d12-hd64@iou-desk-delta\endcsname{+0.8}
\expandafter\def\csname res@pvit-d12-hd64@iou-curtain-delta\endcsname{\ensuremath{-}1.9}
\expandafter\def\csname res@pvit-d12-hd64@iou-refridgerator-delta\endcsname{+0.7}
\expandafter\def\csname res@pvit-d12-hd64@iou-shower curtain-delta\endcsname{\ensuremath{-}10.8}
\expandafter\def\csname res@pvit-d12-hd64@iou-toilet-delta\endcsname{+0.0}
\expandafter\def\csname res@pvit-d12-hd64@iou-sink-delta\endcsname{+0.9}
\expandafter\def\csname res@pvit-d12-hd64@iou-bathtub-delta\endcsname{+0.6}
\expandafter\def\csname res@pvit-d12-hd64@iou-otherfurniture-delta\endcsname{+1.7}
\expandafter\def\csname res@pvit-d12-nobranchnorm@n\endcsname{1}
\expandafter\def\csname res@pvit-d12-nobranchnorm@miou\endcsname{76.8}
\expandafter\def\csname res@pvit-d12-nobranchnorm@miou-pm\endcsname{76.8}
\expandafter\def\csname res@pvit-d12-nobranchnorm@macc\endcsname{85.2}
\expandafter\def\csname res@pvit-d12-nobranchnorm@macc-pm\endcsname{85.2}
\expandafter\def\csname res@pvit-d12-nobranchnorm@allacc\endcsname{91.6}
\expandafter\def\csname res@pvit-d12-nobranchnorm@allacc-pm\endcsname{91.6}
\expandafter\def\csname res@pvit-d12-nobranchnorm@iou-wall\endcsname{87.8}
\expandafter\def\csname res@pvit-d12-nobranchnorm@iou-wall-pm\endcsname{87.8}
\expandafter\def\csname res@pvit-d12-nobranchnorm@iou-floor\endcsname{95.8}
\expandafter\def\csname res@pvit-d12-nobranchnorm@iou-floor-pm\endcsname{95.8}
\expandafter\def\csname res@pvit-d12-nobranchnorm@iou-cabinet\endcsname{71.5}
\expandafter\def\csname res@pvit-d12-nobranchnorm@iou-cabinet-pm\endcsname{71.5}
\expandafter\def\csname res@pvit-d12-nobranchnorm@iou-bed\endcsname{82.9}
\expandafter\def\csname res@pvit-d12-nobranchnorm@iou-bed-pm\endcsname{82.9}
\expandafter\def\csname res@pvit-d12-nobranchnorm@iou-chair\endcsname{91.5}
\expandafter\def\csname res@pvit-d12-nobranchnorm@iou-chair-pm\endcsname{91.5}
\expandafter\def\csname res@pvit-d12-nobranchnorm@iou-sofa\endcsname{83.7}
\expandafter\def\csname res@pvit-d12-nobranchnorm@iou-sofa-pm\endcsname{83.7}
\expandafter\def\csname res@pvit-d12-nobranchnorm@iou-table\endcsname{76.3}
\expandafter\def\csname res@pvit-d12-nobranchnorm@iou-table-pm\endcsname{76.3}
\expandafter\def\csname res@pvit-d12-nobranchnorm@iou-door\endcsname{68.2}
\expandafter\def\csname res@pvit-d12-nobranchnorm@iou-door-pm\endcsname{68.2}
\expandafter\def\csname res@pvit-d12-nobranchnorm@iou-window\endcsname{71.7}
\expandafter\def\csname res@pvit-d12-nobranchnorm@iou-window-pm\endcsname{71.7}
\expandafter\def\csname res@pvit-d12-nobranchnorm@iou-bookshelf\endcsname{83.8}
\expandafter\def\csname res@pvit-d12-nobranchnorm@iou-bookshelf-pm\endcsname{83.8}
\expandafter\def\csname res@pvit-d12-nobranchnorm@iou-picture\endcsname{40.7}
\expandafter\def\csname res@pvit-d12-nobranchnorm@iou-picture-pm\endcsname{40.7}
\expandafter\def\csname res@pvit-d12-nobranchnorm@iou-counter\endcsname{70.2}
\expandafter\def\csname res@pvit-d12-nobranchnorm@iou-counter-pm\endcsname{70.2}
\expandafter\def\csname res@pvit-d12-nobranchnorm@iou-desk\endcsname{72.7}
\expandafter\def\csname res@pvit-d12-nobranchnorm@iou-desk-pm\endcsname{72.7}
\expandafter\def\csname res@pvit-d12-nobranchnorm@iou-curtain\endcsname{80.0}
\expandafter\def\csname res@pvit-d12-nobranchnorm@iou-curtain-pm\endcsname{80.0}
\expandafter\def\csname res@pvit-d12-nobranchnorm@iou-refridgerator\endcsname{72.8}
\expandafter\def\csname res@pvit-d12-nobranchnorm@iou-refridgerator-pm\endcsname{72.8}
\expandafter\def\csname res@pvit-d12-nobranchnorm@iou-shower curtain\endcsname{69.5}
\expandafter\def\csname res@pvit-d12-nobranchnorm@iou-shower curtain-pm\endcsname{69.5}
\expandafter\def\csname res@pvit-d12-nobranchnorm@iou-toilet\endcsname{96.4}
\expandafter\def\csname res@pvit-d12-nobranchnorm@iou-toilet-pm\endcsname{96.4}
\expandafter\def\csname res@pvit-d12-nobranchnorm@iou-sink\endcsname{71.0}
\expandafter\def\csname res@pvit-d12-nobranchnorm@iou-sink-pm\endcsname{71.0}
\expandafter\def\csname res@pvit-d12-nobranchnorm@iou-bathtub\endcsname{89.4}
\expandafter\def\csname res@pvit-d12-nobranchnorm@iou-bathtub-pm\endcsname{89.4}
\expandafter\def\csname res@pvit-d12-nobranchnorm@iou-otherfurniture\endcsname{59.4}
\expandafter\def\csname res@pvit-d12-nobranchnorm@iou-otherfurniture-pm\endcsname{59.4}
\expandafter\def\csname res@pvit-d12-nobranchnorm@miou-delta\endcsname{\ensuremath{-}0.7}
\expandafter\def\csname res@pvit-d12-nobranchnorm@macc-delta\endcsname{\ensuremath{-}0.6}
\expandafter\def\csname res@pvit-d12-nobranchnorm@allacc-delta\endcsname{\ensuremath{-}0.3}
\expandafter\def\csname res@pvit-d12-nobranchnorm@iou-wall-delta\endcsname{+0.0}
\expandafter\def\csname res@pvit-d12-nobranchnorm@iou-floor-delta\endcsname{\ensuremath{-}0.1}
\expandafter\def\csname res@pvit-d12-nobranchnorm@iou-cabinet-delta\endcsname{\ensuremath{-}0.9}
\expandafter\def\csname res@pvit-d12-nobranchnorm@iou-bed-delta\endcsname{\ensuremath{-}0.1}
\expandafter\def\csname res@pvit-d12-nobranchnorm@iou-chair-delta\endcsname{\ensuremath{-}1.2}
\expandafter\def\csname res@pvit-d12-nobranchnorm@iou-sofa-delta\endcsname{\ensuremath{-}3.1}
\expandafter\def\csname res@pvit-d12-nobranchnorm@iou-table-delta\endcsname{\ensuremath{-}0.8}
\expandafter\def\csname res@pvit-d12-nobranchnorm@iou-door-delta\endcsname{\ensuremath{-}1.4}
\expandafter\def\csname res@pvit-d12-nobranchnorm@iou-window-delta\endcsname{+0.2}
\expandafter\def\csname res@pvit-d12-nobranchnorm@iou-bookshelf-delta\endcsname{\ensuremath{-}1.7}
\expandafter\def\csname res@pvit-d12-nobranchnorm@iou-picture-delta\endcsname{+0.8}
\expandafter\def\csname res@pvit-d12-nobranchnorm@iou-counter-delta\endcsname{\ensuremath{-}0.3}
\expandafter\def\csname res@pvit-d12-nobranchnorm@iou-desk-delta\endcsname{+0.6}
\expandafter\def\csname res@pvit-d12-nobranchnorm@iou-curtain-delta\endcsname{\ensuremath{-}0.7}
\expandafter\def\csname res@pvit-d12-nobranchnorm@iou-refridgerator-delta\endcsname{+1.9}
\expandafter\def\csname res@pvit-d12-nobranchnorm@iou-shower curtain-delta\endcsname{\ensuremath{-}5.9}
\expandafter\def\csname res@pvit-d12-nobranchnorm@iou-toilet-delta\endcsname{+0.4}
\expandafter\def\csname res@pvit-d12-nobranchnorm@iou-sink-delta\endcsname{\ensuremath{-}0.2}
\expandafter\def\csname res@pvit-d12-nobranchnorm@iou-bathtub-delta\endcsname{+0.6}
\expandafter\def\csname res@pvit-d12-nobranchnorm@iou-otherfurniture-delta\endcsname{\ensuremath{-}3.1}
\expandafter\def\csname res@pvit-d12-unet@n\endcsname{1}
\expandafter\def\csname res@pvit-d12-unet@miou\endcsname{76.8}
\expandafter\def\csname res@pvit-d12-unet@miou-pm\endcsname{76.8}
\expandafter\def\csname res@pvit-d12-unet@macc\endcsname{84.9}
\expandafter\def\csname res@pvit-d12-unet@macc-pm\endcsname{84.9}
\expandafter\def\csname res@pvit-d12-unet@allacc\endcsname{91.8}
\expandafter\def\csname res@pvit-d12-unet@allacc-pm\endcsname{91.8}
\expandafter\def\csname res@pvit-d12-unet@iou-wall\endcsname{87.7}
\expandafter\def\csname res@pvit-d12-unet@iou-wall-pm\endcsname{87.7}
\expandafter\def\csname res@pvit-d12-unet@iou-floor\endcsname{95.8}
\expandafter\def\csname res@pvit-d12-unet@iou-floor-pm\endcsname{95.8}
\expandafter\def\csname res@pvit-d12-unet@iou-cabinet\endcsname{70.7}
\expandafter\def\csname res@pvit-d12-unet@iou-cabinet-pm\endcsname{70.7}
\expandafter\def\csname res@pvit-d12-unet@iou-bed\endcsname{82.9}
\expandafter\def\csname res@pvit-d12-unet@iou-bed-pm\endcsname{82.9}
\expandafter\def\csname res@pvit-d12-unet@iou-chair\endcsname{92.2}
\expandafter\def\csname res@pvit-d12-unet@iou-chair-pm\endcsname{92.2}
\expandafter\def\csname res@pvit-d12-unet@iou-sofa\endcsname{86.1}
\expandafter\def\csname res@pvit-d12-unet@iou-sofa-pm\endcsname{86.1}
\expandafter\def\csname res@pvit-d12-unet@iou-table\endcsname{76.6}
\expandafter\def\csname res@pvit-d12-unet@iou-table-pm\endcsname{76.6}
\expandafter\def\csname res@pvit-d12-unet@iou-door\endcsname{71.7}
\expandafter\def\csname res@pvit-d12-unet@iou-door-pm\endcsname{71.7}
\expandafter\def\csname res@pvit-d12-unet@iou-window\endcsname{73.4}
\expandafter\def\csname res@pvit-d12-unet@iou-window-pm\endcsname{73.4}
\expandafter\def\csname res@pvit-d12-unet@iou-bookshelf\endcsname{85.5}
\expandafter\def\csname res@pvit-d12-unet@iou-bookshelf-pm\endcsname{85.5}
\expandafter\def\csname res@pvit-d12-unet@iou-picture\endcsname{40.8}
\expandafter\def\csname res@pvit-d12-unet@iou-picture-pm\endcsname{40.8}
\expandafter\def\csname res@pvit-d12-unet@iou-counter\endcsname{70.3}
\expandafter\def\csname res@pvit-d12-unet@iou-counter-pm\endcsname{70.3}
\expandafter\def\csname res@pvit-d12-unet@iou-desk\endcsname{70.8}
\expandafter\def\csname res@pvit-d12-unet@iou-desk-pm\endcsname{70.8}
\expandafter\def\csname res@pvit-d12-unet@iou-curtain\endcsname{79.0}
\expandafter\def\csname res@pvit-d12-unet@iou-curtain-pm\endcsname{79.0}
\expandafter\def\csname res@pvit-d12-unet@iou-refridgerator\endcsname{68.6}
\expandafter\def\csname res@pvit-d12-unet@iou-refridgerator-pm\endcsname{68.6}
\expandafter\def\csname res@pvit-d12-unet@iou-shower curtain\endcsname{67.9}
\expandafter\def\csname res@pvit-d12-unet@iou-shower curtain-pm\endcsname{67.9}
\expandafter\def\csname res@pvit-d12-unet@iou-toilet\endcsname{96.1}
\expandafter\def\csname res@pvit-d12-unet@iou-toilet-pm\endcsname{96.1}
\expandafter\def\csname res@pvit-d12-unet@iou-sink\endcsname{70.1}
\expandafter\def\csname res@pvit-d12-unet@iou-sink-pm\endcsname{70.1}
\expandafter\def\csname res@pvit-d12-unet@iou-bathtub\endcsname{88.0}
\expandafter\def\csname res@pvit-d12-unet@iou-bathtub-pm\endcsname{88.0}
\expandafter\def\csname res@pvit-d12-unet@iou-otherfurniture\endcsname{61.0}
\expandafter\def\csname res@pvit-d12-unet@iou-otherfurniture-pm\endcsname{61.0}
\expandafter\def\csname res@pvit-d12-unet@miou-delta\endcsname{\ensuremath{-}0.7}
\expandafter\def\csname res@pvit-d12-unet@macc-delta\endcsname{\ensuremath{-}0.9}
\expandafter\def\csname res@pvit-d12-unet@allacc-delta\endcsname{\ensuremath{-}0.1}
\expandafter\def\csname res@pvit-d12-unet@iou-wall-delta\endcsname{\ensuremath{-}0.1}
\expandafter\def\csname res@pvit-d12-unet@iou-floor-delta\endcsname{\ensuremath{-}0.1}
\expandafter\def\csname res@pvit-d12-unet@iou-cabinet-delta\endcsname{\ensuremath{-}1.7}
\expandafter\def\csname res@pvit-d12-unet@iou-bed-delta\endcsname{\ensuremath{-}0.1}
\expandafter\def\csname res@pvit-d12-unet@iou-chair-delta\endcsname{\ensuremath{-}0.5}
\expandafter\def\csname res@pvit-d12-unet@iou-sofa-delta\endcsname{\ensuremath{-}0.7}
\expandafter\def\csname res@pvit-d12-unet@iou-table-delta\endcsname{\ensuremath{-}0.5}
\expandafter\def\csname res@pvit-d12-unet@iou-door-delta\endcsname{+2.1}
\expandafter\def\csname res@pvit-d12-unet@iou-window-delta\endcsname{+1.9}
\expandafter\def\csname res@pvit-d12-unet@iou-bookshelf-delta\endcsname{+0.0}
\expandafter\def\csname res@pvit-d12-unet@iou-picture-delta\endcsname{+0.9}
\expandafter\def\csname res@pvit-d12-unet@iou-counter-delta\endcsname{\ensuremath{-}0.2}
\expandafter\def\csname res@pvit-d12-unet@iou-desk-delta\endcsname{\ensuremath{-}1.3}
\expandafter\def\csname res@pvit-d12-unet@iou-curtain-delta\endcsname{\ensuremath{-}1.7}
\expandafter\def\csname res@pvit-d12-unet@iou-refridgerator-delta\endcsname{\ensuremath{-}2.3}
\expandafter\def\csname res@pvit-d12-unet@iou-shower curtain-delta\endcsname{\ensuremath{-}7.5}
\expandafter\def\csname res@pvit-d12-unet@iou-toilet-delta\endcsname{+0.1}
\expandafter\def\csname res@pvit-d12-unet@iou-sink-delta\endcsname{\ensuremath{-}1.1}
\expandafter\def\csname res@pvit-d12-unet@iou-bathtub-delta\endcsname{\ensuremath{-}0.8}
\expandafter\def\csname res@pvit-d12-unet@iou-otherfurniture-delta\endcsname{\ensuremath{-}1.5}
\expandafter\def\csname res@pvit-d12-dpscene@n\endcsname{1}
\expandafter\def\csname res@pvit-d12-dpscene@miou\endcsname{77.4}
\expandafter\def\csname res@pvit-d12-dpscene@miou-pm\endcsname{77.4}
\expandafter\def\csname res@pvit-d12-dpscene@macc\endcsname{85.7}
\expandafter\def\csname res@pvit-d12-dpscene@macc-pm\endcsname{85.7}
\expandafter\def\csname res@pvit-d12-dpscene@allacc\endcsname{91.9}
\expandafter\def\csname res@pvit-d12-dpscene@allacc-pm\endcsname{91.9}
\expandafter\def\csname res@pvit-d12-dpscene@iou-wall\endcsname{87.9}
\expandafter\def\csname res@pvit-d12-dpscene@iou-wall-pm\endcsname{87.9}
\expandafter\def\csname res@pvit-d12-dpscene@iou-floor\endcsname{95.9}
\expandafter\def\csname res@pvit-d12-dpscene@iou-floor-pm\endcsname{95.9}
\expandafter\def\csname res@pvit-d12-dpscene@iou-cabinet\endcsname{72.6}
\expandafter\def\csname res@pvit-d12-dpscene@iou-cabinet-pm\endcsname{72.6}
\expandafter\def\csname res@pvit-d12-dpscene@iou-bed\endcsname{82.9}
\expandafter\def\csname res@pvit-d12-dpscene@iou-bed-pm\endcsname{82.9}
\expandafter\def\csname res@pvit-d12-dpscene@iou-chair\endcsname{91.8}
\expandafter\def\csname res@pvit-d12-dpscene@iou-chair-pm\endcsname{91.8}
\expandafter\def\csname res@pvit-d12-dpscene@iou-sofa\endcsname{86.4}
\expandafter\def\csname res@pvit-d12-dpscene@iou-sofa-pm\endcsname{86.4}
\expandafter\def\csname res@pvit-d12-dpscene@iou-table\endcsname{76.5}
\expandafter\def\csname res@pvit-d12-dpscene@iou-table-pm\endcsname{76.5}
\expandafter\def\csname res@pvit-d12-dpscene@iou-door\endcsname{71.6}
\expandafter\def\csname res@pvit-d12-dpscene@iou-door-pm\endcsname{71.6}
\expandafter\def\csname res@pvit-d12-dpscene@iou-window\endcsname{73.4}
\expandafter\def\csname res@pvit-d12-dpscene@iou-window-pm\endcsname{73.4}
\expandafter\def\csname res@pvit-d12-dpscene@iou-bookshelf\endcsname{83.6}
\expandafter\def\csname res@pvit-d12-dpscene@iou-bookshelf-pm\endcsname{83.6}
\expandafter\def\csname res@pvit-d12-dpscene@iou-picture\endcsname{42.4}
\expandafter\def\csname res@pvit-d12-dpscene@iou-picture-pm\endcsname{42.4}
\expandafter\def\csname res@pvit-d12-dpscene@iou-counter\endcsname{69.8}
\expandafter\def\csname res@pvit-d12-dpscene@iou-counter-pm\endcsname{69.8}
\expandafter\def\csname res@pvit-d12-dpscene@iou-desk\endcsname{73.3}
\expandafter\def\csname res@pvit-d12-dpscene@iou-desk-pm\endcsname{73.3}
\expandafter\def\csname res@pvit-d12-dpscene@iou-curtain\endcsname{80.0}
\expandafter\def\csname res@pvit-d12-dpscene@iou-curtain-pm\endcsname{80.0}
\expandafter\def\csname res@pvit-d12-dpscene@iou-refridgerator\endcsname{75.5}
\expandafter\def\csname res@pvit-d12-dpscene@iou-refridgerator-pm\endcsname{75.5}
\expandafter\def\csname res@pvit-d12-dpscene@iou-shower curtain\endcsname{67.4}
\expandafter\def\csname res@pvit-d12-dpscene@iou-shower curtain-pm\endcsname{67.4}
\expandafter\def\csname res@pvit-d12-dpscene@iou-toilet\endcsname{95.9}
\expandafter\def\csname res@pvit-d12-dpscene@iou-toilet-pm\endcsname{95.9}
\expandafter\def\csname res@pvit-d12-dpscene@iou-sink\endcsname{70.5}
\expandafter\def\csname res@pvit-d12-dpscene@iou-sink-pm\endcsname{70.5}
\expandafter\def\csname res@pvit-d12-dpscene@iou-bathtub\endcsname{87.8}
\expandafter\def\csname res@pvit-d12-dpscene@iou-bathtub-pm\endcsname{87.8}
\expandafter\def\csname res@pvit-d12-dpscene@iou-otherfurniture\endcsname{62.7}
\expandafter\def\csname res@pvit-d12-dpscene@iou-otherfurniture-pm\endcsname{62.7}
\expandafter\def\csname res@pvit-d12-dpscene@miou-delta\endcsname{\ensuremath{-}0.1}
\expandafter\def\csname res@pvit-d12-dpscene@macc-delta\endcsname{\ensuremath{-}0.1}
\expandafter\def\csname res@pvit-d12-dpscene@allacc-delta\endcsname{+0.0}
\expandafter\def\csname res@pvit-d12-dpscene@iou-wall-delta\endcsname{+0.1}
\expandafter\def\csname res@pvit-d12-dpscene@iou-floor-delta\endcsname{+0.0}
\expandafter\def\csname res@pvit-d12-dpscene@iou-cabinet-delta\endcsname{+0.2}
\expandafter\def\csname res@pvit-d12-dpscene@iou-bed-delta\endcsname{\ensuremath{-}0.1}
\expandafter\def\csname res@pvit-d12-dpscene@iou-chair-delta\endcsname{\ensuremath{-}0.9}
\expandafter\def\csname res@pvit-d12-dpscene@iou-sofa-delta\endcsname{\ensuremath{-}0.4}
\expandafter\def\csname res@pvit-d12-dpscene@iou-table-delta\endcsname{\ensuremath{-}0.6}
\expandafter\def\csname res@pvit-d12-dpscene@iou-door-delta\endcsname{+2.0}
\expandafter\def\csname res@pvit-d12-dpscene@iou-window-delta\endcsname{+1.9}
\expandafter\def\csname res@pvit-d12-dpscene@iou-bookshelf-delta\endcsname{\ensuremath{-}1.9}
\expandafter\def\csname res@pvit-d12-dpscene@iou-picture-delta\endcsname{+2.5}
\expandafter\def\csname res@pvit-d12-dpscene@iou-counter-delta\endcsname{\ensuremath{-}0.7}
\expandafter\def\csname res@pvit-d12-dpscene@iou-desk-delta\endcsname{+1.2}
\expandafter\def\csname res@pvit-d12-dpscene@iou-curtain-delta\endcsname{\ensuremath{-}0.7}
\expandafter\def\csname res@pvit-d12-dpscene@iou-refridgerator-delta\endcsname{+4.6}
\expandafter\def\csname res@pvit-d12-dpscene@iou-shower curtain-delta\endcsname{\ensuremath{-}8.0}
\expandafter\def\csname res@pvit-d12-dpscene@iou-toilet-delta\endcsname{\ensuremath{-}0.1}
\expandafter\def\csname res@pvit-d12-dpscene@iou-sink-delta\endcsname{\ensuremath{-}0.7}
\expandafter\def\csname res@pvit-d12-dpscene@iou-bathtub-delta\endcsname{\ensuremath{-}1.0}
\expandafter\def\csname res@pvit-d12-dpscene@iou-otherfurniture-delta\endcsname{+0.2}
\expandafter\def\csname res@pvit-d12-rope-sd-scannet@n\endcsname{1}
\expandafter\def\csname res@pvit-d12-rope-sd-scannet@miou\endcsname{77.3}
\expandafter\def\csname res@pvit-d12-rope-sd-scannet@miou-pm\endcsname{77.3}
\expandafter\def\csname res@pvit-d12-rope-sd-scannet@macc\endcsname{85.4}
\expandafter\def\csname res@pvit-d12-rope-sd-scannet@macc-pm\endcsname{85.4}
\expandafter\def\csname res@pvit-d12-rope-sd-scannet@allacc\endcsname{91.8}
\expandafter\def\csname res@pvit-d12-rope-sd-scannet@allacc-pm\endcsname{91.8}
\expandafter\def\csname res@pvit-d12-rope-sd-scannet@iou-wall\endcsname{87.4}
\expandafter\def\csname res@pvit-d12-rope-sd-scannet@iou-wall-pm\endcsname{87.4}
\expandafter\def\csname res@pvit-d12-rope-sd-scannet@iou-floor\endcsname{95.9}
\expandafter\def\csname res@pvit-d12-rope-sd-scannet@iou-floor-pm\endcsname{95.9}
\expandafter\def\csname res@pvit-d12-rope-sd-scannet@iou-cabinet\endcsname{72.8}
\expandafter\def\csname res@pvit-d12-rope-sd-scannet@iou-cabinet-pm\endcsname{72.8}
\expandafter\def\csname res@pvit-d12-rope-sd-scannet@iou-bed\endcsname{83.5}
\expandafter\def\csname res@pvit-d12-rope-sd-scannet@iou-bed-pm\endcsname{83.5}
\expandafter\def\csname res@pvit-d12-rope-sd-scannet@iou-chair\endcsname{92.0}
\expandafter\def\csname res@pvit-d12-rope-sd-scannet@iou-chair-pm\endcsname{92.0}
\expandafter\def\csname res@pvit-d12-rope-sd-scannet@iou-sofa\endcsname{87.6}
\expandafter\def\csname res@pvit-d12-rope-sd-scannet@iou-sofa-pm\endcsname{87.6}
\expandafter\def\csname res@pvit-d12-rope-sd-scannet@iou-table\endcsname{77.1}
\expandafter\def\csname res@pvit-d12-rope-sd-scannet@iou-table-pm\endcsname{77.1}
\expandafter\def\csname res@pvit-d12-rope-sd-scannet@iou-door\endcsname{71.2}
\expandafter\def\csname res@pvit-d12-rope-sd-scannet@iou-door-pm\endcsname{71.2}
\expandafter\def\csname res@pvit-d12-rope-sd-scannet@iou-window\endcsname{74.7}
\expandafter\def\csname res@pvit-d12-rope-sd-scannet@iou-window-pm\endcsname{74.7}
\expandafter\def\csname res@pvit-d12-rope-sd-scannet@iou-bookshelf\endcsname{85.4}
\expandafter\def\csname res@pvit-d12-rope-sd-scannet@iou-bookshelf-pm\endcsname{85.4}
\expandafter\def\csname res@pvit-d12-rope-sd-scannet@iou-picture\endcsname{41.7}
\expandafter\def\csname res@pvit-d12-rope-sd-scannet@iou-picture-pm\endcsname{41.7}
\expandafter\def\csname res@pvit-d12-rope-sd-scannet@iou-counter\endcsname{69.7}
\expandafter\def\csname res@pvit-d12-rope-sd-scannet@iou-counter-pm\endcsname{69.7}
\expandafter\def\csname res@pvit-d12-rope-sd-scannet@iou-desk\endcsname{72.7}
\expandafter\def\csname res@pvit-d12-rope-sd-scannet@iou-desk-pm\endcsname{72.7}
\expandafter\def\csname res@pvit-d12-rope-sd-scannet@iou-curtain\endcsname{78.0}
\expandafter\def\csname res@pvit-d12-rope-sd-scannet@iou-curtain-pm\endcsname{78.0}
\expandafter\def\csname res@pvit-d12-rope-sd-scannet@iou-refridgerator\endcsname{75.3}
\expandafter\def\csname res@pvit-d12-rope-sd-scannet@iou-refridgerator-pm\endcsname{75.3}
\expandafter\def\csname res@pvit-d12-rope-sd-scannet@iou-shower curtain\endcsname{65.6}
\expandafter\def\csname res@pvit-d12-rope-sd-scannet@iou-shower curtain-pm\endcsname{65.6}
\expandafter\def\csname res@pvit-d12-rope-sd-scannet@iou-toilet\endcsname{96.0}
\expandafter\def\csname res@pvit-d12-rope-sd-scannet@iou-toilet-pm\endcsname{96.0}
\expandafter\def\csname res@pvit-d12-rope-sd-scannet@iou-sink\endcsname{71.4}
\expandafter\def\csname res@pvit-d12-rope-sd-scannet@iou-sink-pm\endcsname{71.4}
\expandafter\def\csname res@pvit-d12-rope-sd-scannet@iou-bathtub\endcsname{88.3}
\expandafter\def\csname res@pvit-d12-rope-sd-scannet@iou-bathtub-pm\endcsname{88.3}
\expandafter\def\csname res@pvit-d12-rope-sd-scannet@iou-otherfurniture\endcsname{58.9}
\expandafter\def\csname res@pvit-d12-rope-sd-scannet@iou-otherfurniture-pm\endcsname{58.9}
\expandafter\def\csname res@pvit-d12-rope-sd-scannet@miou-delta\endcsname{\ensuremath{-}0.3}
\expandafter\def\csname res@pvit-d12-rope-sd-scannet@macc-delta\endcsname{\ensuremath{-}0.5}
\expandafter\def\csname res@pvit-d12-rope-sd-scannet@allacc-delta\endcsname{\ensuremath{-}0.2}
\expandafter\def\csname res@pvit-d12-rope-sd-scannet@iou-wall-delta\endcsname{\ensuremath{-}0.5}
\expandafter\def\csname res@pvit-d12-rope-sd-scannet@iou-floor-delta\endcsname{+0.0}
\expandafter\def\csname res@pvit-d12-rope-sd-scannet@iou-cabinet-delta\endcsname{\ensuremath{-}0.6}
\expandafter\def\csname res@pvit-d12-rope-sd-scannet@iou-bed-delta\endcsname{+0.9}
\expandafter\def\csname res@pvit-d12-rope-sd-scannet@iou-chair-delta\endcsname{+0.0}
\expandafter\def\csname res@pvit-d12-rope-sd-scannet@iou-sofa-delta\endcsname{+2.9}
\expandafter\def\csname res@pvit-d12-rope-sd-scannet@iou-table-delta\endcsname{+0.3}
\expandafter\def\csname res@pvit-d12-rope-sd-scannet@iou-door-delta\endcsname{\ensuremath{-}1.3}
\expandafter\def\csname res@pvit-d12-rope-sd-scannet@iou-window-delta\endcsname{+0.6}
\expandafter\def\csname res@pvit-d12-rope-sd-scannet@iou-bookshelf-delta\endcsname{\ensuremath{-}0.2}
\expandafter\def\csname res@pvit-d12-rope-sd-scannet@iou-picture-delta\endcsname{+0.1}
\expandafter\def\csname res@pvit-d12-rope-sd-scannet@iou-counter-delta\endcsname{\ensuremath{-}1.1}
\expandafter\def\csname res@pvit-d12-rope-sd-scannet@iou-desk-delta\endcsname{+0.9}
\expandafter\def\csname res@pvit-d12-rope-sd-scannet@iou-curtain-delta\endcsname{\ensuremath{-}1.3}
\expandafter\def\csname res@pvit-d12-rope-sd-scannet@iou-refridgerator-delta\endcsname{+2.5}
\expandafter\def\csname res@pvit-d12-rope-sd-scannet@iou-shower curtain-delta\endcsname{\ensuremath{-}6.3}
\expandafter\def\csname res@pvit-d12-rope-sd-scannet@iou-toilet-delta\endcsname{+0.2}
\expandafter\def\csname res@pvit-d12-rope-sd-scannet@iou-sink-delta\endcsname{+0.9}
\expandafter\def\csname res@pvit-d12-rope-sd-scannet@iou-bathtub-delta\endcsname{\ensuremath{-}0.3}
\expandafter\def\csname res@pvit-d12-rope-sd-scannet@iou-otherfurniture-delta\endcsname{\ensuremath{-}4.3}
\expandafter\def\csname res@pvit-d12-s3dis@n\endcsname{1}
\expandafter\def\csname res@pvit-d12-s3dis@miou\endcsname{71.0}
\expandafter\def\csname res@pvit-d12-s3dis@miou-pm\endcsname{71.0}
\expandafter\def\csname res@pvit-d12-s3dis@macc\endcsname{76.5}
\expandafter\def\csname res@pvit-d12-s3dis@macc-pm\endcsname{76.5}
\expandafter\def\csname res@pvit-d12-s3dis@allacc\endcsname{91.1}
\expandafter\def\csname res@pvit-d12-s3dis@allacc-pm\endcsname{91.1}
\expandafter\def\csname res@pvit-d12-s3dis@iou-ceiling\endcsname{92.8}
\expandafter\def\csname res@pvit-d12-s3dis@iou-ceiling-pm\endcsname{92.8}
\expandafter\def\csname res@pvit-d12-s3dis@iou-floor\endcsname{98.5}
\expandafter\def\csname res@pvit-d12-s3dis@iou-floor-pm\endcsname{98.5}
\expandafter\def\csname res@pvit-d12-s3dis@iou-wall\endcsname{86.7}
\expandafter\def\csname res@pvit-d12-s3dis@iou-wall-pm\endcsname{86.7}
\expandafter\def\csname res@pvit-d12-s3dis@iou-beam\endcsname{0.0}
\expandafter\def\csname res@pvit-d12-s3dis@iou-beam-pm\endcsname{0.0}
\expandafter\def\csname res@pvit-d12-s3dis@iou-column\endcsname{41.5}
\expandafter\def\csname res@pvit-d12-s3dis@iou-column-pm\endcsname{41.5}
\expandafter\def\csname res@pvit-d12-s3dis@iou-window\endcsname{63.3}
\expandafter\def\csname res@pvit-d12-s3dis@iou-window-pm\endcsname{63.3}
\expandafter\def\csname res@pvit-d12-s3dis@iou-door\endcsname{74.1}
\expandafter\def\csname res@pvit-d12-s3dis@iou-door-pm\endcsname{74.1}
\expandafter\def\csname res@pvit-d12-s3dis@iou-table\endcsname{83.2}
\expandafter\def\csname res@pvit-d12-s3dis@iou-table-pm\endcsname{83.2}
\expandafter\def\csname res@pvit-d12-s3dis@iou-chair\endcsname{91.5}
\expandafter\def\csname res@pvit-d12-s3dis@iou-chair-pm\endcsname{91.5}
\expandafter\def\csname res@pvit-d12-s3dis@iou-sofa\endcsname{66.4}
\expandafter\def\csname res@pvit-d12-s3dis@iou-sofa-pm\endcsname{66.4}
\expandafter\def\csname res@pvit-d12-s3dis@iou-bookcase\endcsname{76.8}
\expandafter\def\csname res@pvit-d12-s3dis@iou-bookcase-pm\endcsname{76.8}
\expandafter\def\csname res@pvit-d12-s3dis@iou-board\endcsname{88.1}
\expandafter\def\csname res@pvit-d12-s3dis@iou-board-pm\endcsname{88.1}
\expandafter\def\csname res@pvit-d12-s3dis@iou-clutter\endcsname{59.9}
\expandafter\def\csname res@pvit-d12-s3dis@iou-clutter-pm\endcsname{59.9}
\expandafter\def\csname res@pvit-d12-scannet200@n\endcsname{1}
\expandafter\def\csname res@pvit-d12-scannet200@miou\endcsname{34.9}
\expandafter\def\csname res@pvit-d12-scannet200@miou-pm\endcsname{34.9}
\expandafter\def\csname res@pvit-d12-scannet200@macc\endcsname{44.9}
\expandafter\def\csname res@pvit-d12-scannet200@macc-pm\endcsname{44.9}
\expandafter\def\csname res@pvit-d12-scannet200@allacc\endcsname{83.4}
\expandafter\def\csname res@pvit-d12-scannet200@allacc-pm\endcsname{83.4}
\expandafter\def\csname res@pvit-d12-scannet200@iou-wall\endcsname{81.3}
\expandafter\def\csname res@pvit-d12-scannet200@iou-wall-pm\endcsname{81.3}
\expandafter\def\csname res@pvit-d12-scannet200@iou-chair\endcsname{74.9}
\expandafter\def\csname res@pvit-d12-scannet200@iou-chair-pm\endcsname{74.9}
\expandafter\def\csname res@pvit-d12-scannet200@iou-floor\endcsname{94.9}
\expandafter\def\csname res@pvit-d12-scannet200@iou-floor-pm\endcsname{94.9}
\expandafter\def\csname res@pvit-d12-scannet200@iou-table\endcsname{67.4}
\expandafter\def\csname res@pvit-d12-scannet200@iou-table-pm\endcsname{67.4}
\expandafter\def\csname res@pvit-d12-scannet200@iou-door\endcsname{63.1}
\expandafter\def\csname res@pvit-d12-scannet200@iou-door-pm\endcsname{63.1}
\expandafter\def\csname res@pvit-d12-scannet200@iou-couch\endcsname{80.2}
\expandafter\def\csname res@pvit-d12-scannet200@iou-couch-pm\endcsname{80.2}
\expandafter\def\csname res@pvit-d12-scannet200@iou-cabinet\endcsname{39.9}
\expandafter\def\csname res@pvit-d12-scannet200@iou-cabinet-pm\endcsname{39.9}
\expandafter\def\csname res@pvit-d12-scannet200@iou-shelf\endcsname{49.1}
\expandafter\def\csname res@pvit-d12-scannet200@iou-shelf-pm\endcsname{49.1}
\expandafter\def\csname res@pvit-d12-scannet200@iou-desk\endcsname{62.7}
\expandafter\def\csname res@pvit-d12-scannet200@iou-desk-pm\endcsname{62.7}
\expandafter\def\csname res@pvit-d12-scannet200@iou-office chair\endcsname{20.5}
\expandafter\def\csname res@pvit-d12-scannet200@iou-office chair-pm\endcsname{20.5}
\expandafter\def\csname res@pvit-d12-scannet200@iou-bed\endcsname{78.8}
\expandafter\def\csname res@pvit-d12-scannet200@iou-bed-pm\endcsname{78.8}
\expandafter\def\csname res@pvit-d12-scannet200@iou-pillow\endcsname{59.8}
\expandafter\def\csname res@pvit-d12-scannet200@iou-pillow-pm\endcsname{59.8}
\expandafter\def\csname res@pvit-d12-scannet200@iou-sink\endcsname{69.9}
\expandafter\def\csname res@pvit-d12-scannet200@iou-sink-pm\endcsname{69.9}
\expandafter\def\csname res@pvit-d12-scannet200@iou-picture\endcsname{35.9}
\expandafter\def\csname res@pvit-d12-scannet200@iou-picture-pm\endcsname{35.9}
\expandafter\def\csname res@pvit-d12-scannet200@iou-window\endcsname{68.6}
\expandafter\def\csname res@pvit-d12-scannet200@iou-window-pm\endcsname{68.6}
\expandafter\def\csname res@pvit-d12-scannet200@iou-toilet\endcsname{93.8}
\expandafter\def\csname res@pvit-d12-scannet200@iou-toilet-pm\endcsname{93.8}
\expandafter\def\csname res@pvit-d12-scannet200@iou-bookshelf\endcsname{51.1}
\expandafter\def\csname res@pvit-d12-scannet200@iou-bookshelf-pm\endcsname{51.1}
\expandafter\def\csname res@pvit-d12-scannet200@iou-monitor\endcsname{90.2}
\expandafter\def\csname res@pvit-d12-scannet200@iou-monitor-pm\endcsname{90.2}
\expandafter\def\csname res@pvit-d12-scannet200@iou-curtain\endcsname{71.8}
\expandafter\def\csname res@pvit-d12-scannet200@iou-curtain-pm\endcsname{71.8}
\expandafter\def\csname res@pvit-d12-scannet200@iou-book\endcsname{44.4}
\expandafter\def\csname res@pvit-d12-scannet200@iou-book-pm\endcsname{44.4}
\expandafter\def\csname res@pvit-d12-scannet200@iou-armchair\endcsname{48.8}
\expandafter\def\csname res@pvit-d12-scannet200@iou-armchair-pm\endcsname{48.8}
\expandafter\def\csname res@pvit-d12-scannet200@iou-coffee table\endcsname{50.7}
\expandafter\def\csname res@pvit-d12-scannet200@iou-coffee table-pm\endcsname{50.7}
\expandafter\def\csname res@pvit-d12-scannet200@iou-box\endcsname{40.6}
\expandafter\def\csname res@pvit-d12-scannet200@iou-box-pm\endcsname{40.6}
\expandafter\def\csname res@pvit-d12-scannet200@iou-refrigerator\endcsname{72.6}
\expandafter\def\csname res@pvit-d12-scannet200@iou-refrigerator-pm\endcsname{72.6}
\expandafter\def\csname res@pvit-d12-scannet200@iou-lamp\endcsname{67.5}
\expandafter\def\csname res@pvit-d12-scannet200@iou-lamp-pm\endcsname{67.5}
\expandafter\def\csname res@pvit-d12-scannet200@iou-kitchen cabinet\endcsname{69.4}
\expandafter\def\csname res@pvit-d12-scannet200@iou-kitchen cabinet-pm\endcsname{69.4}
\expandafter\def\csname res@pvit-d12-scannet200@iou-towel\endcsname{63.7}
\expandafter\def\csname res@pvit-d12-scannet200@iou-towel-pm\endcsname{63.7}
\expandafter\def\csname res@pvit-d12-scannet200@iou-clothes\endcsname{18.9}
\expandafter\def\csname res@pvit-d12-scannet200@iou-clothes-pm\endcsname{18.9}
\expandafter\def\csname res@pvit-d12-scannet200@iou-tv\endcsname{91.3}
\expandafter\def\csname res@pvit-d12-scannet200@iou-tv-pm\endcsname{91.3}
\expandafter\def\csname res@pvit-d12-scannet200@iou-nightstand\endcsname{67.1}
\expandafter\def\csname res@pvit-d12-scannet200@iou-nightstand-pm\endcsname{67.1}
\expandafter\def\csname res@pvit-d12-scannet200@iou-counter\endcsname{39.5}
\expandafter\def\csname res@pvit-d12-scannet200@iou-counter-pm\endcsname{39.5}
\expandafter\def\csname res@pvit-d12-scannet200@iou-dresser\endcsname{48.2}
\expandafter\def\csname res@pvit-d12-scannet200@iou-dresser-pm\endcsname{48.2}
\expandafter\def\csname res@pvit-d12-scannet200@iou-stool\endcsname{62.4}
\expandafter\def\csname res@pvit-d12-scannet200@iou-stool-pm\endcsname{62.4}
\expandafter\def\csname res@pvit-d12-scannet200@iou-cushion\endcsname{0.0}
\expandafter\def\csname res@pvit-d12-scannet200@iou-cushion-pm\endcsname{0.0}
\expandafter\def\csname res@pvit-d12-scannet200@iou-plant\endcsname{80.1}
\expandafter\def\csname res@pvit-d12-scannet200@iou-plant-pm\endcsname{80.1}
\expandafter\def\csname res@pvit-d12-scannet200@iou-ceiling\endcsname{93.0}
\expandafter\def\csname res@pvit-d12-scannet200@iou-ceiling-pm\endcsname{93.0}
\expandafter\def\csname res@pvit-d12-scannet200@iou-bathtub\endcsname{86.3}
\expandafter\def\csname res@pvit-d12-scannet200@iou-bathtub-pm\endcsname{86.3}
\expandafter\def\csname res@pvit-d12-scannet200@iou-end table\endcsname{21.7}
\expandafter\def\csname res@pvit-d12-scannet200@iou-end table-pm\endcsname{21.7}
\expandafter\def\csname res@pvit-d12-scannet200@iou-dining table\endcsname{2.1}
\expandafter\def\csname res@pvit-d12-scannet200@iou-dining table-pm\endcsname{2.1}
\expandafter\def\csname res@pvit-d12-scannet200@iou-keyboard\endcsname{42.1}
\expandafter\def\csname res@pvit-d12-scannet200@iou-keyboard-pm\endcsname{42.1}
\expandafter\def\csname res@pvit-d12-scannet200@iou-bag\endcsname{27.4}
\expandafter\def\csname res@pvit-d12-scannet200@iou-bag-pm\endcsname{27.4}
\expandafter\def\csname res@pvit-d12-scannet200@iou-backpack\endcsname{64.8}
\expandafter\def\csname res@pvit-d12-scannet200@iou-backpack-pm\endcsname{64.8}
\expandafter\def\csname res@pvit-d12-scannet200@iou-toilet paper\endcsname{48.8}
\expandafter\def\csname res@pvit-d12-scannet200@iou-toilet paper-pm\endcsname{48.8}
\expandafter\def\csname res@pvit-d12-scannet200@iou-printer\endcsname{52.8}
\expandafter\def\csname res@pvit-d12-scannet200@iou-printer-pm\endcsname{52.8}
\expandafter\def\csname res@pvit-d12-scannet200@iou-tv stand\endcsname{41.5}
\expandafter\def\csname res@pvit-d12-scannet200@iou-tv stand-pm\endcsname{41.5}
\expandafter\def\csname res@pvit-d12-scannet200@iou-whiteboard\endcsname{56.5}
\expandafter\def\csname res@pvit-d12-scannet200@iou-whiteboard-pm\endcsname{56.5}
\expandafter\def\csname res@pvit-d12-scannet200@iou-blanket\endcsname{3.1}
\expandafter\def\csname res@pvit-d12-scannet200@iou-blanket-pm\endcsname{3.1}
\expandafter\def\csname res@pvit-d12-scannet200@iou-shower curtain\endcsname{62.4}
\expandafter\def\csname res@pvit-d12-scannet200@iou-shower curtain-pm\endcsname{62.4}
\expandafter\def\csname res@pvit-d12-scannet200@iou-trash can\endcsname{65.3}
\expandafter\def\csname res@pvit-d12-scannet200@iou-trash can-pm\endcsname{65.3}
\expandafter\def\csname res@pvit-d12-scannet200@iou-closet\endcsname{12.2}
\expandafter\def\csname res@pvit-d12-scannet200@iou-closet-pm\endcsname{12.2}
\expandafter\def\csname res@pvit-d12-scannet200@iou-stairs\endcsname{49.0}
\expandafter\def\csname res@pvit-d12-scannet200@iou-stairs-pm\endcsname{49.0}
\expandafter\def\csname res@pvit-d12-scannet200@iou-microwave\endcsname{60.0}
\expandafter\def\csname res@pvit-d12-scannet200@iou-microwave-pm\endcsname{60.0}
\expandafter\def\csname res@pvit-d12-scannet200@iou-stove\endcsname{64.7}
\expandafter\def\csname res@pvit-d12-scannet200@iou-stove-pm\endcsname{64.7}
\expandafter\def\csname res@pvit-d12-scannet200@iou-shoe\endcsname{52.7}
\expandafter\def\csname res@pvit-d12-scannet200@iou-shoe-pm\endcsname{52.7}
\expandafter\def\csname res@pvit-d12-scannet200@iou-computer tower\endcsname{68.9}
\expandafter\def\csname res@pvit-d12-scannet200@iou-computer tower-pm\endcsname{68.9}
\expandafter\def\csname res@pvit-d12-scannet200@iou-bottle\endcsname{15.0}
\expandafter\def\csname res@pvit-d12-scannet200@iou-bottle-pm\endcsname{15.0}
\expandafter\def\csname res@pvit-d12-scannet200@iou-bin\endcsname{5.0}
\expandafter\def\csname res@pvit-d12-scannet200@iou-bin-pm\endcsname{5.0}
\expandafter\def\csname res@pvit-d12-scannet200@iou-ottoman\endcsname{36.8}
\expandafter\def\csname res@pvit-d12-scannet200@iou-ottoman-pm\endcsname{36.8}
\expandafter\def\csname res@pvit-d12-scannet200@iou-bench\endcsname{22.8}
\expandafter\def\csname res@pvit-d12-scannet200@iou-bench-pm\endcsname{22.8}
\expandafter\def\csname res@pvit-d12-scannet200@iou-board\endcsname{22.2}
\expandafter\def\csname res@pvit-d12-scannet200@iou-board-pm\endcsname{22.2}
\expandafter\def\csname res@pvit-d12-scannet200@iou-washing machine\endcsname{86.2}
\expandafter\def\csname res@pvit-d12-scannet200@iou-washing machine-pm\endcsname{86.2}
\expandafter\def\csname res@pvit-d12-scannet200@iou-mirror\endcsname{48.6}
\expandafter\def\csname res@pvit-d12-scannet200@iou-mirror-pm\endcsname{48.6}
\expandafter\def\csname res@pvit-d12-scannet200@iou-copier\endcsname{93.6}
\expandafter\def\csname res@pvit-d12-scannet200@iou-copier-pm\endcsname{93.6}
\expandafter\def\csname res@pvit-d12-scannet200@iou-basket\endcsname{9.5}
\expandafter\def\csname res@pvit-d12-scannet200@iou-basket-pm\endcsname{9.5}
\expandafter\def\csname res@pvit-d12-scannet200@iou-sofa chair\endcsname{10.5}
\expandafter\def\csname res@pvit-d12-scannet200@iou-sofa chair-pm\endcsname{10.5}
\expandafter\def\csname res@pvit-d12-scannet200@iou-file cabinet\endcsname{52.7}
\expandafter\def\csname res@pvit-d12-scannet200@iou-file cabinet-pm\endcsname{52.7}
\expandafter\def\csname res@pvit-d12-scannet200@iou-fan\endcsname{41.5}
\expandafter\def\csname res@pvit-d12-scannet200@iou-fan-pm\endcsname{41.5}
\expandafter\def\csname res@pvit-d12-scannet200@iou-laptop\endcsname{58.1}
\expandafter\def\csname res@pvit-d12-scannet200@iou-laptop-pm\endcsname{58.1}
\expandafter\def\csname res@pvit-d12-scannet200@iou-shower\endcsname{47.7}
\expandafter\def\csname res@pvit-d12-scannet200@iou-shower-pm\endcsname{47.7}
\expandafter\def\csname res@pvit-d12-scannet200@iou-paper\endcsname{7.3}
\expandafter\def\csname res@pvit-d12-scannet200@iou-paper-pm\endcsname{7.3}
\expandafter\def\csname res@pvit-d12-scannet200@iou-person\endcsname{36.0}
\expandafter\def\csname res@pvit-d12-scannet200@iou-person-pm\endcsname{36.0}
\expandafter\def\csname res@pvit-d12-scannet200@iou-paper towel dispenser\endcsname{57.3}
\expandafter\def\csname res@pvit-d12-scannet200@iou-paper towel dispenser-pm\endcsname{57.3}
\expandafter\def\csname res@pvit-d12-scannet200@iou-oven\endcsname{12.2}
\expandafter\def\csname res@pvit-d12-scannet200@iou-oven-pm\endcsname{12.2}
\expandafter\def\csname res@pvit-d12-scannet200@iou-blinds\endcsname{1.0}
\expandafter\def\csname res@pvit-d12-scannet200@iou-blinds-pm\endcsname{1.0}
\expandafter\def\csname res@pvit-d12-scannet200@iou-rack\endcsname{0.0}
\expandafter\def\csname res@pvit-d12-scannet200@iou-rack-pm\endcsname{0.0}
\expandafter\def\csname res@pvit-d12-scannet200@iou-plate\endcsname{0.0}
\expandafter\def\csname res@pvit-d12-scannet200@iou-plate-pm\endcsname{0.0}
\expandafter\def\csname res@pvit-d12-scannet200@iou-blackboard\endcsname{78.8}
\expandafter\def\csname res@pvit-d12-scannet200@iou-blackboard-pm\endcsname{78.8}
\expandafter\def\csname res@pvit-d12-scannet200@iou-piano\endcsname{43.1}
\expandafter\def\csname res@pvit-d12-scannet200@iou-piano-pm\endcsname{43.1}
\expandafter\def\csname res@pvit-d12-scannet200@iou-suitcase\endcsname{64.5}
\expandafter\def\csname res@pvit-d12-scannet200@iou-suitcase-pm\endcsname{64.5}
\expandafter\def\csname res@pvit-d12-scannet200@iou-rail\endcsname{3.7}
\expandafter\def\csname res@pvit-d12-scannet200@iou-rail-pm\endcsname{3.7}
\expandafter\def\csname res@pvit-d12-scannet200@iou-radiator\endcsname{72.2}
\expandafter\def\csname res@pvit-d12-scannet200@iou-radiator-pm\endcsname{72.2}
\expandafter\def\csname res@pvit-d12-scannet200@iou-recycling bin\endcsname{53.4}
\expandafter\def\csname res@pvit-d12-scannet200@iou-recycling bin-pm\endcsname{53.4}
\expandafter\def\csname res@pvit-d12-scannet200@iou-container\endcsname{0.0}
\expandafter\def\csname res@pvit-d12-scannet200@iou-container-pm\endcsname{0.0}
\expandafter\def\csname res@pvit-d12-scannet200@iou-wardrobe\endcsname{31.9}
\expandafter\def\csname res@pvit-d12-scannet200@iou-wardrobe-pm\endcsname{31.9}
\expandafter\def\csname res@pvit-d12-scannet200@iou-soap dispenser\endcsname{82.7}
\expandafter\def\csname res@pvit-d12-scannet200@iou-soap dispenser-pm\endcsname{82.7}
\expandafter\def\csname res@pvit-d12-scannet200@iou-telephone\endcsname{37.2}
\expandafter\def\csname res@pvit-d12-scannet200@iou-telephone-pm\endcsname{37.2}
\expandafter\def\csname res@pvit-d12-scannet200@iou-bucket\endcsname{22.0}
\expandafter\def\csname res@pvit-d12-scannet200@iou-bucket-pm\endcsname{22.0}
\expandafter\def\csname res@pvit-d12-scannet200@iou-clock\endcsname{20.1}
\expandafter\def\csname res@pvit-d12-scannet200@iou-clock-pm\endcsname{20.1}
\expandafter\def\csname res@pvit-d12-scannet200@iou-stand\endcsname{0.0}
\expandafter\def\csname res@pvit-d12-scannet200@iou-stand-pm\endcsname{0.0}
\expandafter\def\csname res@pvit-d12-scannet200@iou-light\endcsname{7.5}
\expandafter\def\csname res@pvit-d12-scannet200@iou-light-pm\endcsname{7.5}
\expandafter\def\csname res@pvit-d12-scannet200@iou-laundry basket\endcsname{9.7}
\expandafter\def\csname res@pvit-d12-scannet200@iou-laundry basket-pm\endcsname{9.7}
\expandafter\def\csname res@pvit-d12-scannet200@iou-pipe\endcsname{0.1}
\expandafter\def\csname res@pvit-d12-scannet200@iou-pipe-pm\endcsname{0.1}
\expandafter\def\csname res@pvit-d12-scannet200@iou-clothes dryer\endcsname{78.7}
\expandafter\def\csname res@pvit-d12-scannet200@iou-clothes dryer-pm\endcsname{78.7}
\expandafter\def\csname res@pvit-d12-scannet200@iou-guitar\endcsname{77.5}
\expandafter\def\csname res@pvit-d12-scannet200@iou-guitar-pm\endcsname{77.5}
\expandafter\def\csname res@pvit-d12-scannet200@iou-toilet paper holder\endcsname{0.0}
\expandafter\def\csname res@pvit-d12-scannet200@iou-toilet paper holder-pm\endcsname{0.0}
\expandafter\def\csname res@pvit-d12-scannet200@iou-seat\endcsname{0.0}
\expandafter\def\csname res@pvit-d12-scannet200@iou-seat-pm\endcsname{0.0}
\expandafter\def\csname res@pvit-d12-scannet200@iou-speaker\endcsname{0.0}
\expandafter\def\csname res@pvit-d12-scannet200@iou-speaker-pm\endcsname{0.0}
\expandafter\def\csname res@pvit-d12-scannet200@iou-column\endcsname{0.0}
\expandafter\def\csname res@pvit-d12-scannet200@iou-column-pm\endcsname{0.0}
\expandafter\def\csname res@pvit-d12-scannet200@iou-bicycle\endcsname{0.0}
\expandafter\def\csname res@pvit-d12-scannet200@iou-bicycle-pm\endcsname{0.0}
\expandafter\def\csname res@pvit-d12-scannet200@iou-ladder\endcsname{38.1}
\expandafter\def\csname res@pvit-d12-scannet200@iou-ladder-pm\endcsname{38.1}
\expandafter\def\csname res@pvit-d12-scannet200@iou-bathroom stall\endcsname{62.7}
\expandafter\def\csname res@pvit-d12-scannet200@iou-bathroom stall-pm\endcsname{62.7}
\expandafter\def\csname res@pvit-d12-scannet200@iou-shower wall\endcsname{46.6}
\expandafter\def\csname res@pvit-d12-scannet200@iou-shower wall-pm\endcsname{46.6}
\expandafter\def\csname res@pvit-d12-scannet200@iou-cup\endcsname{19.0}
\expandafter\def\csname res@pvit-d12-scannet200@iou-cup-pm\endcsname{19.0}
\expandafter\def\csname res@pvit-d12-scannet200@iou-jacket\endcsname{12.3}
\expandafter\def\csname res@pvit-d12-scannet200@iou-jacket-pm\endcsname{12.3}
\expandafter\def\csname res@pvit-d12-scannet200@iou-storage bin\endcsname{0.7}
\expandafter\def\csname res@pvit-d12-scannet200@iou-storage bin-pm\endcsname{0.7}
\expandafter\def\csname res@pvit-d12-scannet200@iou-coffee maker\endcsname{69.2}
\expandafter\def\csname res@pvit-d12-scannet200@iou-coffee maker-pm\endcsname{69.2}
\expandafter\def\csname res@pvit-d12-scannet200@iou-dishwasher\endcsname{28.1}
\expandafter\def\csname res@pvit-d12-scannet200@iou-dishwasher-pm\endcsname{28.1}
\expandafter\def\csname res@pvit-d12-scannet200@iou-paper towel roll\endcsname{33.6}
\expandafter\def\csname res@pvit-d12-scannet200@iou-paper towel roll-pm\endcsname{33.6}
\expandafter\def\csname res@pvit-d12-scannet200@iou-machine\endcsname{0.0}
\expandafter\def\csname res@pvit-d12-scannet200@iou-machine-pm\endcsname{0.0}
\expandafter\def\csname res@pvit-d12-scannet200@iou-mat\endcsname{29.0}
\expandafter\def\csname res@pvit-d12-scannet200@iou-mat-pm\endcsname{29.0}
\expandafter\def\csname res@pvit-d12-scannet200@iou-windowsill\endcsname{17.2}
\expandafter\def\csname res@pvit-d12-scannet200@iou-windowsill-pm\endcsname{17.2}
\expandafter\def\csname res@pvit-d12-scannet200@iou-bar\endcsname{12.4}
\expandafter\def\csname res@pvit-d12-scannet200@iou-bar-pm\endcsname{12.4}
\expandafter\def\csname res@pvit-d12-scannet200@iou-toaster\endcsname{84.4}
\expandafter\def\csname res@pvit-d12-scannet200@iou-toaster-pm\endcsname{84.4}
\expandafter\def\csname res@pvit-d12-scannet200@iou-bulletin board\endcsname{37.2}
\expandafter\def\csname res@pvit-d12-scannet200@iou-bulletin board-pm\endcsname{37.2}
\expandafter\def\csname res@pvit-d12-scannet200@iou-ironing board\endcsname{53.4}
\expandafter\def\csname res@pvit-d12-scannet200@iou-ironing board-pm\endcsname{53.4}
\expandafter\def\csname res@pvit-d12-scannet200@iou-fireplace\endcsname{67.2}
\expandafter\def\csname res@pvit-d12-scannet200@iou-fireplace-pm\endcsname{67.2}
\expandafter\def\csname res@pvit-d12-scannet200@iou-soap dish\endcsname{36.3}
\expandafter\def\csname res@pvit-d12-scannet200@iou-soap dish-pm\endcsname{36.3}
\expandafter\def\csname res@pvit-d12-scannet200@iou-kitchen counter\endcsname{58.5}
\expandafter\def\csname res@pvit-d12-scannet200@iou-kitchen counter-pm\endcsname{58.5}
\expandafter\def\csname res@pvit-d12-scannet200@iou-doorframe\endcsname{48.8}
\expandafter\def\csname res@pvit-d12-scannet200@iou-doorframe-pm\endcsname{48.8}
\expandafter\def\csname res@pvit-d12-scannet200@iou-toilet paper dispenser\endcsname{41.1}
\expandafter\def\csname res@pvit-d12-scannet200@iou-toilet paper dispenser-pm\endcsname{41.1}
\expandafter\def\csname res@pvit-d12-scannet200@iou-mini fridge\endcsname{55.9}
\expandafter\def\csname res@pvit-d12-scannet200@iou-mini fridge-pm\endcsname{55.9}
\expandafter\def\csname res@pvit-d12-scannet200@iou-fire extinguisher\endcsname{22.7}
\expandafter\def\csname res@pvit-d12-scannet200@iou-fire extinguisher-pm\endcsname{22.7}
\expandafter\def\csname res@pvit-d12-scannet200@iou-ball\endcsname{45.9}
\expandafter\def\csname res@pvit-d12-scannet200@iou-ball-pm\endcsname{45.9}
\expandafter\def\csname res@pvit-d12-scannet200@iou-hat\endcsname{0.0}
\expandafter\def\csname res@pvit-d12-scannet200@iou-hat-pm\endcsname{0.0}
\expandafter\def\csname res@pvit-d12-scannet200@iou-shower curtain rod\endcsname{53.7}
\expandafter\def\csname res@pvit-d12-scannet200@iou-shower curtain rod-pm\endcsname{53.7}
\expandafter\def\csname res@pvit-d12-scannet200@iou-water cooler\endcsname{88.7}
\expandafter\def\csname res@pvit-d12-scannet200@iou-water cooler-pm\endcsname{88.7}
\expandafter\def\csname res@pvit-d12-scannet200@iou-paper cutter\endcsname{20.6}
\expandafter\def\csname res@pvit-d12-scannet200@iou-paper cutter-pm\endcsname{20.6}
\expandafter\def\csname res@pvit-d12-scannet200@iou-tray\endcsname{6.1}
\expandafter\def\csname res@pvit-d12-scannet200@iou-tray-pm\endcsname{6.1}
\expandafter\def\csname res@pvit-d12-scannet200@iou-shower door\endcsname{41.8}
\expandafter\def\csname res@pvit-d12-scannet200@iou-shower door-pm\endcsname{41.8}
\expandafter\def\csname res@pvit-d12-scannet200@iou-pillar\endcsname{6.7}
\expandafter\def\csname res@pvit-d12-scannet200@iou-pillar-pm\endcsname{6.7}
\expandafter\def\csname res@pvit-d12-scannet200@iou-ledge\endcsname{7.5}
\expandafter\def\csname res@pvit-d12-scannet200@iou-ledge-pm\endcsname{7.5}
\expandafter\def\csname res@pvit-d12-scannet200@iou-toaster oven\endcsname{0.0}
\expandafter\def\csname res@pvit-d12-scannet200@iou-toaster oven-pm\endcsname{0.0}
\expandafter\def\csname res@pvit-d12-scannet200@iou-mouse\endcsname{0.0}
\expandafter\def\csname res@pvit-d12-scannet200@iou-mouse-pm\endcsname{0.0}
\expandafter\def\csname res@pvit-d12-scannet200@iou-toilet seat cover dispenser\endcsname{80.0}
\expandafter\def\csname res@pvit-d12-scannet200@iou-toilet seat cover dispenser-pm\endcsname{80.0}
\expandafter\def\csname res@pvit-d12-scannet200@iou-furniture\endcsname{0.0}
\expandafter\def\csname res@pvit-d12-scannet200@iou-furniture-pm\endcsname{0.0}
\expandafter\def\csname res@pvit-d12-scannet200@iou-cart\endcsname{24.7}
\expandafter\def\csname res@pvit-d12-scannet200@iou-cart-pm\endcsname{24.7}
\expandafter\def\csname res@pvit-d12-scannet200@iou-storage container\endcsname{0.0}
\expandafter\def\csname res@pvit-d12-scannet200@iou-storage container-pm\endcsname{0.0}
\expandafter\def\csname res@pvit-d12-scannet200@iou-scale\endcsname{40.6}
\expandafter\def\csname res@pvit-d12-scannet200@iou-scale-pm\endcsname{40.6}
\expandafter\def\csname res@pvit-d12-scannet200@iou-tissue box\endcsname{12.5}
\expandafter\def\csname res@pvit-d12-scannet200@iou-tissue box-pm\endcsname{12.5}
\expandafter\def\csname res@pvit-d12-scannet200@iou-light switch\endcsname{0.0}
\expandafter\def\csname res@pvit-d12-scannet200@iou-light switch-pm\endcsname{0.0}
\expandafter\def\csname res@pvit-d12-scannet200@iou-crate\endcsname{0.0}
\expandafter\def\csname res@pvit-d12-scannet200@iou-crate-pm\endcsname{0.0}
\expandafter\def\csname res@pvit-d12-scannet200@iou-power outlet\endcsname{0.0}
\expandafter\def\csname res@pvit-d12-scannet200@iou-power outlet-pm\endcsname{0.0}
\expandafter\def\csname res@pvit-d12-scannet200@iou-decoration\endcsname{55.5}
\expandafter\def\csname res@pvit-d12-scannet200@iou-decoration-pm\endcsname{55.5}
\expandafter\def\csname res@pvit-d12-scannet200@iou-sign\endcsname{37.7}
\expandafter\def\csname res@pvit-d12-scannet200@iou-sign-pm\endcsname{37.7}
\expandafter\def\csname res@pvit-d12-scannet200@iou-projector\endcsname{13.7}
\expandafter\def\csname res@pvit-d12-scannet200@iou-projector-pm\endcsname{13.7}
\expandafter\def\csname res@pvit-d12-scannet200@iou-closet door\endcsname{3.6}
\expandafter\def\csname res@pvit-d12-scannet200@iou-closet door-pm\endcsname{3.6}
\expandafter\def\csname res@pvit-d12-scannet200@iou-vacuum cleaner\endcsname{83.7}
\expandafter\def\csname res@pvit-d12-scannet200@iou-vacuum cleaner-pm\endcsname{83.7}
\expandafter\def\csname res@pvit-d12-scannet200@iou-candle\endcsname{0.0}
\expandafter\def\csname res@pvit-d12-scannet200@iou-candle-pm\endcsname{0.0}
\expandafter\def\csname res@pvit-d12-scannet200@iou-plunger\endcsname{35.9}
\expandafter\def\csname res@pvit-d12-scannet200@iou-plunger-pm\endcsname{35.9}
\expandafter\def\csname res@pvit-d12-scannet200@iou-stuffed animal\endcsname{0.0}
\expandafter\def\csname res@pvit-d12-scannet200@iou-stuffed animal-pm\endcsname{0.0}
\expandafter\def\csname res@pvit-d12-scannet200@iou-headphones\endcsname{0.0}
\expandafter\def\csname res@pvit-d12-scannet200@iou-headphones-pm\endcsname{0.0}
\expandafter\def\csname res@pvit-d12-scannet200@iou-dish rack\endcsname{91.9}
\expandafter\def\csname res@pvit-d12-scannet200@iou-dish rack-pm\endcsname{91.9}
\expandafter\def\csname res@pvit-d12-scannet200@iou-broom\endcsname{0.0}
\expandafter\def\csname res@pvit-d12-scannet200@iou-broom-pm\endcsname{0.0}
\expandafter\def\csname res@pvit-d12-scannet200@iou-guitar case\endcsname{0.0}
\expandafter\def\csname res@pvit-d12-scannet200@iou-guitar case-pm\endcsname{0.0}
\expandafter\def\csname res@pvit-d12-scannet200@iou-range hood\endcsname{78.5}
\expandafter\def\csname res@pvit-d12-scannet200@iou-range hood-pm\endcsname{78.5}
\expandafter\def\csname res@pvit-d12-scannet200@iou-dustpan\endcsname{0.7}
\expandafter\def\csname res@pvit-d12-scannet200@iou-dustpan-pm\endcsname{0.7}
\expandafter\def\csname res@pvit-d12-scannet200@iou-hair dryer\endcsname{0.0}
\expandafter\def\csname res@pvit-d12-scannet200@iou-hair dryer-pm\endcsname{0.0}
\expandafter\def\csname res@pvit-d12-scannet200@iou-water bottle\endcsname{0.0}
\expandafter\def\csname res@pvit-d12-scannet200@iou-water bottle-pm\endcsname{0.0}
\expandafter\def\csname res@pvit-d12-scannet200@iou-handicap bar\endcsname{6.2}
\expandafter\def\csname res@pvit-d12-scannet200@iou-handicap bar-pm\endcsname{6.2}
\expandafter\def\csname res@pvit-d12-scannet200@iou-purse\endcsname{0.0}
\expandafter\def\csname res@pvit-d12-scannet200@iou-purse-pm\endcsname{0.0}
\expandafter\def\csname res@pvit-d12-scannet200@iou-vent\endcsname{34.2}
\expandafter\def\csname res@pvit-d12-scannet200@iou-vent-pm\endcsname{34.2}
\expandafter\def\csname res@pvit-d12-scannet200@iou-shower floor\endcsname{64.2}
\expandafter\def\csname res@pvit-d12-scannet200@iou-shower floor-pm\endcsname{64.2}
\expandafter\def\csname res@pvit-d12-scannet200@iou-water pitcher\endcsname{0.0}
\expandafter\def\csname res@pvit-d12-scannet200@iou-water pitcher-pm\endcsname{0.0}
\expandafter\def\csname res@pvit-d12-scannet200@iou-mailbox\endcsname{50.0}
\expandafter\def\csname res@pvit-d12-scannet200@iou-mailbox-pm\endcsname{50.0}
\expandafter\def\csname res@pvit-d12-scannet200@iou-bowl\endcsname{27.4}
\expandafter\def\csname res@pvit-d12-scannet200@iou-bowl-pm\endcsname{27.4}
\expandafter\def\csname res@pvit-d12-scannet200@iou-paper bag\endcsname{5.1}
\expandafter\def\csname res@pvit-d12-scannet200@iou-paper bag-pm\endcsname{5.1}
\expandafter\def\csname res@pvit-d12-scannet200@iou-alarm clock\endcsname{0.0}
\expandafter\def\csname res@pvit-d12-scannet200@iou-alarm clock-pm\endcsname{0.0}
\expandafter\def\csname res@pvit-d12-scannet200@iou-music stand\endcsname{0.0}
\expandafter\def\csname res@pvit-d12-scannet200@iou-music stand-pm\endcsname{0.0}
\expandafter\def\csname res@pvit-d12-scannet200@iou-projector screen\endcsname{0.0}
\expandafter\def\csname res@pvit-d12-scannet200@iou-projector screen-pm\endcsname{0.0}
\expandafter\def\csname res@pvit-d12-scannet200@iou-divider\endcsname{60.9}
\expandafter\def\csname res@pvit-d12-scannet200@iou-divider-pm\endcsname{60.9}
\expandafter\def\csname res@pvit-d12-scannet200@iou-laundry detergent\endcsname{0.0}
\expandafter\def\csname res@pvit-d12-scannet200@iou-laundry detergent-pm\endcsname{0.0}
\expandafter\def\csname res@pvit-d12-scannet200@iou-bathroom counter\endcsname{31.8}
\expandafter\def\csname res@pvit-d12-scannet200@iou-bathroom counter-pm\endcsname{31.8}
\expandafter\def\csname res@pvit-d12-scannet200@iou-object\endcsname{7.6}
\expandafter\def\csname res@pvit-d12-scannet200@iou-object-pm\endcsname{7.6}
\expandafter\def\csname res@pvit-d12-scannet200@iou-bathroom vanity\endcsname{49.7}
\expandafter\def\csname res@pvit-d12-scannet200@iou-bathroom vanity-pm\endcsname{49.7}
\expandafter\def\csname res@pvit-d12-scannet200@iou-closet wall\endcsname{15.0}
\expandafter\def\csname res@pvit-d12-scannet200@iou-closet wall-pm\endcsname{15.0}
\expandafter\def\csname res@pvit-d12-scannet200@iou-laundry hamper\endcsname{43.2}
\expandafter\def\csname res@pvit-d12-scannet200@iou-laundry hamper-pm\endcsname{43.2}
\expandafter\def\csname res@pvit-d12-scannet200@iou-bathroom stall door\endcsname{46.4}
\expandafter\def\csname res@pvit-d12-scannet200@iou-bathroom stall door-pm\endcsname{46.4}
\expandafter\def\csname res@pvit-d12-scannet200@iou-ceiling light\endcsname{48.0}
\expandafter\def\csname res@pvit-d12-scannet200@iou-ceiling light-pm\endcsname{48.0}
\expandafter\def\csname res@pvit-d12-scannet200@iou-trash bin\endcsname{24.9}
\expandafter\def\csname res@pvit-d12-scannet200@iou-trash bin-pm\endcsname{24.9}
\expandafter\def\csname res@pvit-d12-scannet200@iou-dumbbell\endcsname{83.6}
\expandafter\def\csname res@pvit-d12-scannet200@iou-dumbbell-pm\endcsname{83.6}
\expandafter\def\csname res@pvit-d12-scannet200@iou-stair rail\endcsname{14.4}
\expandafter\def\csname res@pvit-d12-scannet200@iou-stair rail-pm\endcsname{14.4}
\expandafter\def\csname res@pvit-d12-scannet200@iou-tube\endcsname{31.4}
\expandafter\def\csname res@pvit-d12-scannet200@iou-tube-pm\endcsname{31.4}
\expandafter\def\csname res@pvit-d12-scannet200@iou-bathroom cabinet\endcsname{28.7}
\expandafter\def\csname res@pvit-d12-scannet200@iou-bathroom cabinet-pm\endcsname{28.7}
\expandafter\def\csname res@pvit-d12-scannet200@iou-cd case\endcsname{0.0}
\expandafter\def\csname res@pvit-d12-scannet200@iou-cd case-pm\endcsname{0.0}
\expandafter\def\csname res@pvit-d12-scannet200@iou-closet rod\endcsname{47.8}
\expandafter\def\csname res@pvit-d12-scannet200@iou-closet rod-pm\endcsname{47.8}
\expandafter\def\csname res@pvit-d12-scannet200@iou-coffee kettle\endcsname{38.8}
\expandafter\def\csname res@pvit-d12-scannet200@iou-coffee kettle-pm\endcsname{38.8}
\expandafter\def\csname res@pvit-d12-scannet200@iou-structure\endcsname{0.0}
\expandafter\def\csname res@pvit-d12-scannet200@iou-structure-pm\endcsname{0.0}
\expandafter\def\csname res@pvit-d12-scannet200@iou-shower head\endcsname{30.4}
\expandafter\def\csname res@pvit-d12-scannet200@iou-shower head-pm\endcsname{30.4}
\expandafter\def\csname res@pvit-d12-scannet200@iou-keyboard piano\endcsname{0.0}
\expandafter\def\csname res@pvit-d12-scannet200@iou-keyboard piano-pm\endcsname{0.0}
\expandafter\def\csname res@pvit-d12-scannet200@iou-case of water bottles\endcsname{0.0}
\expandafter\def\csname res@pvit-d12-scannet200@iou-case of water bottles-pm\endcsname{0.0}
\expandafter\def\csname res@pvit-d12-scannet200@iou-coat rack\endcsname{48.9}
\expandafter\def\csname res@pvit-d12-scannet200@iou-coat rack-pm\endcsname{48.9}
\expandafter\def\csname res@pvit-d12-scannet200@iou-storage organizer\endcsname{0.0}
\expandafter\def\csname res@pvit-d12-scannet200@iou-storage organizer-pm\endcsname{0.0}
\expandafter\def\csname res@pvit-d12-scannet200@iou-folded chair\endcsname{0.0}
\expandafter\def\csname res@pvit-d12-scannet200@iou-folded chair-pm\endcsname{0.0}
\expandafter\def\csname res@pvit-d12-scannet200@iou-fire alarm\endcsname{0.0}
\expandafter\def\csname res@pvit-d12-scannet200@iou-fire alarm-pm\endcsname{0.0}
\expandafter\def\csname res@pvit-d12-scannet200@iou-power strip\endcsname{0.0}
\expandafter\def\csname res@pvit-d12-scannet200@iou-power strip-pm\endcsname{0.0}
\expandafter\def\csname res@pvit-d12-scannet200@iou-calendar\endcsname{0.0}
\expandafter\def\csname res@pvit-d12-scannet200@iou-calendar-pm\endcsname{0.0}
\expandafter\def\csname res@pvit-d12-scannet200@iou-poster\endcsname{0.0}
\expandafter\def\csname res@pvit-d12-scannet200@iou-poster-pm\endcsname{0.0}
\expandafter\def\csname res@pvit-d12-scannet200@iou-potted plant\endcsname{0.0}
\expandafter\def\csname res@pvit-d12-scannet200@iou-potted plant-pm\endcsname{0.0}
\expandafter\def\csname res@pvit-d12-scannet200@iou-luggage\endcsname{0.0}
\expandafter\def\csname res@pvit-d12-scannet200@iou-luggage-pm\endcsname{0.0}
\expandafter\def\csname res@pvit-d12-scannet200@iou-mattress\endcsname{0.0}
\expandafter\def\csname res@pvit-d12-scannet200@iou-mattress-pm\endcsname{0.0}
\expandafter\def\csname res@pvit-d12-scannet200-sd-wide@n\endcsname{1}
\expandafter\def\csname res@pvit-d12-scannet200-sd-wide@miou\endcsname{35.4}
\expandafter\def\csname res@pvit-d12-scannet200-sd-wide@miou-pm\endcsname{35.4}
\expandafter\def\csname res@pvit-d12-scannet200-sd-wide@macc\endcsname{45.8}
\expandafter\def\csname res@pvit-d12-scannet200-sd-wide@macc-pm\endcsname{45.8}
\expandafter\def\csname res@pvit-d12-scannet200-sd-wide@allacc\endcsname{83.8}
\expandafter\def\csname res@pvit-d12-scannet200-sd-wide@allacc-pm\endcsname{83.8}
\expandafter\def\csname res@pvit-d12-scannet200-sd-wide@iou-wall\endcsname{81.5}
\expandafter\def\csname res@pvit-d12-scannet200-sd-wide@iou-wall-pm\endcsname{81.5}
\expandafter\def\csname res@pvit-d12-scannet200-sd-wide@iou-chair\endcsname{75.1}
\expandafter\def\csname res@pvit-d12-scannet200-sd-wide@iou-chair-pm\endcsname{75.1}
\expandafter\def\csname res@pvit-d12-scannet200-sd-wide@iou-floor\endcsname{95.2}
\expandafter\def\csname res@pvit-d12-scannet200-sd-wide@iou-floor-pm\endcsname{95.2}
\expandafter\def\csname res@pvit-d12-scannet200-sd-wide@iou-table\endcsname{69.5}
\expandafter\def\csname res@pvit-d12-scannet200-sd-wide@iou-table-pm\endcsname{69.5}
\expandafter\def\csname res@pvit-d12-scannet200-sd-wide@iou-door\endcsname{66.0}
\expandafter\def\csname res@pvit-d12-scannet200-sd-wide@iou-door-pm\endcsname{66.0}
\expandafter\def\csname res@pvit-d12-scannet200-sd-wide@iou-couch\endcsname{79.0}
\expandafter\def\csname res@pvit-d12-scannet200-sd-wide@iou-couch-pm\endcsname{79.0}
\expandafter\def\csname res@pvit-d12-scannet200-sd-wide@iou-cabinet\endcsname{44.7}
\expandafter\def\csname res@pvit-d12-scannet200-sd-wide@iou-cabinet-pm\endcsname{44.7}
\expandafter\def\csname res@pvit-d12-scannet200-sd-wide@iou-shelf\endcsname{49.8}
\expandafter\def\csname res@pvit-d12-scannet200-sd-wide@iou-shelf-pm\endcsname{49.8}
\expandafter\def\csname res@pvit-d12-scannet200-sd-wide@iou-desk\endcsname{67.8}
\expandafter\def\csname res@pvit-d12-scannet200-sd-wide@iou-desk-pm\endcsname{67.8}
\expandafter\def\csname res@pvit-d12-scannet200-sd-wide@iou-office chair\endcsname{22.8}
\expandafter\def\csname res@pvit-d12-scannet200-sd-wide@iou-office chair-pm\endcsname{22.8}
\expandafter\def\csname res@pvit-d12-scannet200-sd-wide@iou-bed\endcsname{83.1}
\expandafter\def\csname res@pvit-d12-scannet200-sd-wide@iou-bed-pm\endcsname{83.1}
\expandafter\def\csname res@pvit-d12-scannet200-sd-wide@iou-pillow\endcsname{58.5}
\expandafter\def\csname res@pvit-d12-scannet200-sd-wide@iou-pillow-pm\endcsname{58.5}
\expandafter\def\csname res@pvit-d12-scannet200-sd-wide@iou-sink\endcsname{70.3}
\expandafter\def\csname res@pvit-d12-scannet200-sd-wide@iou-sink-pm\endcsname{70.3}
\expandafter\def\csname res@pvit-d12-scannet200-sd-wide@iou-picture\endcsname{34.6}
\expandafter\def\csname res@pvit-d12-scannet200-sd-wide@iou-picture-pm\endcsname{34.6}
\expandafter\def\csname res@pvit-d12-scannet200-sd-wide@iou-window\endcsname{70.4}
\expandafter\def\csname res@pvit-d12-scannet200-sd-wide@iou-window-pm\endcsname{70.4}
\expandafter\def\csname res@pvit-d12-scannet200-sd-wide@iou-toilet\endcsname{94.3}
\expandafter\def\csname res@pvit-d12-scannet200-sd-wide@iou-toilet-pm\endcsname{94.3}
\expandafter\def\csname res@pvit-d12-scannet200-sd-wide@iou-bookshelf\endcsname{51.6}
\expandafter\def\csname res@pvit-d12-scannet200-sd-wide@iou-bookshelf-pm\endcsname{51.6}
\expandafter\def\csname res@pvit-d12-scannet200-sd-wide@iou-monitor\endcsname{90.2}
\expandafter\def\csname res@pvit-d12-scannet200-sd-wide@iou-monitor-pm\endcsname{90.2}
\expandafter\def\csname res@pvit-d12-scannet200-sd-wide@iou-curtain\endcsname{72.6}
\expandafter\def\csname res@pvit-d12-scannet200-sd-wide@iou-curtain-pm\endcsname{72.6}
\expandafter\def\csname res@pvit-d12-scannet200-sd-wide@iou-book\endcsname{42.8}
\expandafter\def\csname res@pvit-d12-scannet200-sd-wide@iou-book-pm\endcsname{42.8}
\expandafter\def\csname res@pvit-d12-scannet200-sd-wide@iou-armchair\endcsname{52.6}
\expandafter\def\csname res@pvit-d12-scannet200-sd-wide@iou-armchair-pm\endcsname{52.6}
\expandafter\def\csname res@pvit-d12-scannet200-sd-wide@iou-coffee table\endcsname{56.0}
\expandafter\def\csname res@pvit-d12-scannet200-sd-wide@iou-coffee table-pm\endcsname{56.0}
\expandafter\def\csname res@pvit-d12-scannet200-sd-wide@iou-box\endcsname{42.6}
\expandafter\def\csname res@pvit-d12-scannet200-sd-wide@iou-box-pm\endcsname{42.6}
\expandafter\def\csname res@pvit-d12-scannet200-sd-wide@iou-refrigerator\endcsname{71.7}
\expandafter\def\csname res@pvit-d12-scannet200-sd-wide@iou-refrigerator-pm\endcsname{71.7}
\expandafter\def\csname res@pvit-d12-scannet200-sd-wide@iou-lamp\endcsname{64.8}
\expandafter\def\csname res@pvit-d12-scannet200-sd-wide@iou-lamp-pm\endcsname{64.8}
\expandafter\def\csname res@pvit-d12-scannet200-sd-wide@iou-kitchen cabinet\endcsname{73.7}
\expandafter\def\csname res@pvit-d12-scannet200-sd-wide@iou-kitchen cabinet-pm\endcsname{73.7}
\expandafter\def\csname res@pvit-d12-scannet200-sd-wide@iou-towel\endcsname{62.4}
\expandafter\def\csname res@pvit-d12-scannet200-sd-wide@iou-towel-pm\endcsname{62.4}
\expandafter\def\csname res@pvit-d12-scannet200-sd-wide@iou-clothes\endcsname{20.8}
\expandafter\def\csname res@pvit-d12-scannet200-sd-wide@iou-clothes-pm\endcsname{20.8}
\expandafter\def\csname res@pvit-d12-scannet200-sd-wide@iou-tv\endcsname{91.9}
\expandafter\def\csname res@pvit-d12-scannet200-sd-wide@iou-tv-pm\endcsname{91.9}
\expandafter\def\csname res@pvit-d12-scannet200-sd-wide@iou-nightstand\endcsname{70.4}
\expandafter\def\csname res@pvit-d12-scannet200-sd-wide@iou-nightstand-pm\endcsname{70.4}
\expandafter\def\csname res@pvit-d12-scannet200-sd-wide@iou-counter\endcsname{35.9}
\expandafter\def\csname res@pvit-d12-scannet200-sd-wide@iou-counter-pm\endcsname{35.9}
\expandafter\def\csname res@pvit-d12-scannet200-sd-wide@iou-dresser\endcsname{49.9}
\expandafter\def\csname res@pvit-d12-scannet200-sd-wide@iou-dresser-pm\endcsname{49.9}
\expandafter\def\csname res@pvit-d12-scannet200-sd-wide@iou-stool\endcsname{57.7}
\expandafter\def\csname res@pvit-d12-scannet200-sd-wide@iou-stool-pm\endcsname{57.7}
\expandafter\def\csname res@pvit-d12-scannet200-sd-wide@iou-cushion\endcsname{0.0}
\expandafter\def\csname res@pvit-d12-scannet200-sd-wide@iou-cushion-pm\endcsname{0.0}
\expandafter\def\csname res@pvit-d12-scannet200-sd-wide@iou-plant\endcsname{82.9}
\expandafter\def\csname res@pvit-d12-scannet200-sd-wide@iou-plant-pm\endcsname{82.9}
\expandafter\def\csname res@pvit-d12-scannet200-sd-wide@iou-ceiling\endcsname{93.0}
\expandafter\def\csname res@pvit-d12-scannet200-sd-wide@iou-ceiling-pm\endcsname{93.0}
\expandafter\def\csname res@pvit-d12-scannet200-sd-wide@iou-bathtub\endcsname{83.5}
\expandafter\def\csname res@pvit-d12-scannet200-sd-wide@iou-bathtub-pm\endcsname{83.5}
\expandafter\def\csname res@pvit-d12-scannet200-sd-wide@iou-end table\endcsname{20.9}
\expandafter\def\csname res@pvit-d12-scannet200-sd-wide@iou-end table-pm\endcsname{20.9}
\expandafter\def\csname res@pvit-d12-scannet200-sd-wide@iou-dining table\endcsname{0.0}
\expandafter\def\csname res@pvit-d12-scannet200-sd-wide@iou-dining table-pm\endcsname{0.0}
\expandafter\def\csname res@pvit-d12-scannet200-sd-wide@iou-keyboard\endcsname{39.5}
\expandafter\def\csname res@pvit-d12-scannet200-sd-wide@iou-keyboard-pm\endcsname{39.5}
\expandafter\def\csname res@pvit-d12-scannet200-sd-wide@iou-bag\endcsname{20.0}
\expandafter\def\csname res@pvit-d12-scannet200-sd-wide@iou-bag-pm\endcsname{20.0}
\expandafter\def\csname res@pvit-d12-scannet200-sd-wide@iou-backpack\endcsname{60.3}
\expandafter\def\csname res@pvit-d12-scannet200-sd-wide@iou-backpack-pm\endcsname{60.3}
\expandafter\def\csname res@pvit-d12-scannet200-sd-wide@iou-toilet paper\endcsname{49.3}
\expandafter\def\csname res@pvit-d12-scannet200-sd-wide@iou-toilet paper-pm\endcsname{49.3}
\expandafter\def\csname res@pvit-d12-scannet200-sd-wide@iou-printer\endcsname{47.3}
\expandafter\def\csname res@pvit-d12-scannet200-sd-wide@iou-printer-pm\endcsname{47.3}
\expandafter\def\csname res@pvit-d12-scannet200-sd-wide@iou-tv stand\endcsname{42.9}
\expandafter\def\csname res@pvit-d12-scannet200-sd-wide@iou-tv stand-pm\endcsname{42.9}
\expandafter\def\csname res@pvit-d12-scannet200-sd-wide@iou-whiteboard\endcsname{54.2}
\expandafter\def\csname res@pvit-d12-scannet200-sd-wide@iou-whiteboard-pm\endcsname{54.2}
\expandafter\def\csname res@pvit-d12-scannet200-sd-wide@iou-blanket\endcsname{5.8}
\expandafter\def\csname res@pvit-d12-scannet200-sd-wide@iou-blanket-pm\endcsname{5.8}
\expandafter\def\csname res@pvit-d12-scannet200-sd-wide@iou-shower curtain\endcsname{61.6}
\expandafter\def\csname res@pvit-d12-scannet200-sd-wide@iou-shower curtain-pm\endcsname{61.6}
\expandafter\def\csname res@pvit-d12-scannet200-sd-wide@iou-trash can\endcsname{65.1}
\expandafter\def\csname res@pvit-d12-scannet200-sd-wide@iou-trash can-pm\endcsname{65.1}
\expandafter\def\csname res@pvit-d12-scannet200-sd-wide@iou-closet\endcsname{15.5}
\expandafter\def\csname res@pvit-d12-scannet200-sd-wide@iou-closet-pm\endcsname{15.5}
\expandafter\def\csname res@pvit-d12-scannet200-sd-wide@iou-stairs\endcsname{51.8}
\expandafter\def\csname res@pvit-d12-scannet200-sd-wide@iou-stairs-pm\endcsname{51.8}
\expandafter\def\csname res@pvit-d12-scannet200-sd-wide@iou-microwave\endcsname{56.3}
\expandafter\def\csname res@pvit-d12-scannet200-sd-wide@iou-microwave-pm\endcsname{56.3}
\expandafter\def\csname res@pvit-d12-scannet200-sd-wide@iou-stove\endcsname{69.9}
\expandafter\def\csname res@pvit-d12-scannet200-sd-wide@iou-stove-pm\endcsname{69.9}
\expandafter\def\csname res@pvit-d12-scannet200-sd-wide@iou-shoe\endcsname{46.7}
\expandafter\def\csname res@pvit-d12-scannet200-sd-wide@iou-shoe-pm\endcsname{46.7}
\expandafter\def\csname res@pvit-d12-scannet200-sd-wide@iou-computer tower\endcsname{55.5}
\expandafter\def\csname res@pvit-d12-scannet200-sd-wide@iou-computer tower-pm\endcsname{55.5}
\expandafter\def\csname res@pvit-d12-scannet200-sd-wide@iou-bottle\endcsname{9.2}
\expandafter\def\csname res@pvit-d12-scannet200-sd-wide@iou-bottle-pm\endcsname{9.2}
\expandafter\def\csname res@pvit-d12-scannet200-sd-wide@iou-bin\endcsname{0.0}
\expandafter\def\csname res@pvit-d12-scannet200-sd-wide@iou-bin-pm\endcsname{0.0}
\expandafter\def\csname res@pvit-d12-scannet200-sd-wide@iou-ottoman\endcsname{43.2}
\expandafter\def\csname res@pvit-d12-scannet200-sd-wide@iou-ottoman-pm\endcsname{43.2}
\expandafter\def\csname res@pvit-d12-scannet200-sd-wide@iou-bench\endcsname{35.9}
\expandafter\def\csname res@pvit-d12-scannet200-sd-wide@iou-bench-pm\endcsname{35.9}
\expandafter\def\csname res@pvit-d12-scannet200-sd-wide@iou-board\endcsname{18.7}
\expandafter\def\csname res@pvit-d12-scannet200-sd-wide@iou-board-pm\endcsname{18.7}
\expandafter\def\csname res@pvit-d12-scannet200-sd-wide@iou-washing machine\endcsname{87.8}
\expandafter\def\csname res@pvit-d12-scannet200-sd-wide@iou-washing machine-pm\endcsname{87.8}
\expandafter\def\csname res@pvit-d12-scannet200-sd-wide@iou-mirror\endcsname{50.9}
\expandafter\def\csname res@pvit-d12-scannet200-sd-wide@iou-mirror-pm\endcsname{50.9}
\expandafter\def\csname res@pvit-d12-scannet200-sd-wide@iou-copier\endcsname{88.5}
\expandafter\def\csname res@pvit-d12-scannet200-sd-wide@iou-copier-pm\endcsname{88.5}
\expandafter\def\csname res@pvit-d12-scannet200-sd-wide@iou-basket\endcsname{5.9}
\expandafter\def\csname res@pvit-d12-scannet200-sd-wide@iou-basket-pm\endcsname{5.9}
\expandafter\def\csname res@pvit-d12-scannet200-sd-wide@iou-sofa chair\endcsname{16.6}
\expandafter\def\csname res@pvit-d12-scannet200-sd-wide@iou-sofa chair-pm\endcsname{16.6}
\expandafter\def\csname res@pvit-d12-scannet200-sd-wide@iou-file cabinet\endcsname{53.1}
\expandafter\def\csname res@pvit-d12-scannet200-sd-wide@iou-file cabinet-pm\endcsname{53.1}
\expandafter\def\csname res@pvit-d12-scannet200-sd-wide@iou-fan\endcsname{35.2}
\expandafter\def\csname res@pvit-d12-scannet200-sd-wide@iou-fan-pm\endcsname{35.2}
\expandafter\def\csname res@pvit-d12-scannet200-sd-wide@iou-laptop\endcsname{55.5}
\expandafter\def\csname res@pvit-d12-scannet200-sd-wide@iou-laptop-pm\endcsname{55.5}
\expandafter\def\csname res@pvit-d12-scannet200-sd-wide@iou-shower\endcsname{28.3}
\expandafter\def\csname res@pvit-d12-scannet200-sd-wide@iou-shower-pm\endcsname{28.3}
\expandafter\def\csname res@pvit-d12-scannet200-sd-wide@iou-paper\endcsname{5.7}
\expandafter\def\csname res@pvit-d12-scannet200-sd-wide@iou-paper-pm\endcsname{5.7}
\expandafter\def\csname res@pvit-d12-scannet200-sd-wide@iou-person\endcsname{25.0}
\expandafter\def\csname res@pvit-d12-scannet200-sd-wide@iou-person-pm\endcsname{25.0}
\expandafter\def\csname res@pvit-d12-scannet200-sd-wide@iou-paper towel dispenser\endcsname{67.7}
\expandafter\def\csname res@pvit-d12-scannet200-sd-wide@iou-paper towel dispenser-pm\endcsname{67.7}
\expandafter\def\csname res@pvit-d12-scannet200-sd-wide@iou-oven\endcsname{7.2}
\expandafter\def\csname res@pvit-d12-scannet200-sd-wide@iou-oven-pm\endcsname{7.2}
\expandafter\def\csname res@pvit-d12-scannet200-sd-wide@iou-blinds\endcsname{0.0}
\expandafter\def\csname res@pvit-d12-scannet200-sd-wide@iou-blinds-pm\endcsname{0.0}
\expandafter\def\csname res@pvit-d12-scannet200-sd-wide@iou-rack\endcsname{0.0}
\expandafter\def\csname res@pvit-d12-scannet200-sd-wide@iou-rack-pm\endcsname{0.0}
\expandafter\def\csname res@pvit-d12-scannet200-sd-wide@iou-plate\endcsname{0.0}
\expandafter\def\csname res@pvit-d12-scannet200-sd-wide@iou-plate-pm\endcsname{0.0}
\expandafter\def\csname res@pvit-d12-scannet200-sd-wide@iou-blackboard\endcsname{82.6}
\expandafter\def\csname res@pvit-d12-scannet200-sd-wide@iou-blackboard-pm\endcsname{82.6}
\expandafter\def\csname res@pvit-d12-scannet200-sd-wide@iou-piano\endcsname{84.3}
\expandafter\def\csname res@pvit-d12-scannet200-sd-wide@iou-piano-pm\endcsname{84.3}
\expandafter\def\csname res@pvit-d12-scannet200-sd-wide@iou-suitcase\endcsname{68.2}
\expandafter\def\csname res@pvit-d12-scannet200-sd-wide@iou-suitcase-pm\endcsname{68.2}
\expandafter\def\csname res@pvit-d12-scannet200-sd-wide@iou-rail\endcsname{5.8}
\expandafter\def\csname res@pvit-d12-scannet200-sd-wide@iou-rail-pm\endcsname{5.8}
\expandafter\def\csname res@pvit-d12-scannet200-sd-wide@iou-radiator\endcsname{69.7}
\expandafter\def\csname res@pvit-d12-scannet200-sd-wide@iou-radiator-pm\endcsname{69.7}
\expandafter\def\csname res@pvit-d12-scannet200-sd-wide@iou-recycling bin\endcsname{53.2}
\expandafter\def\csname res@pvit-d12-scannet200-sd-wide@iou-recycling bin-pm\endcsname{53.2}
\expandafter\def\csname res@pvit-d12-scannet200-sd-wide@iou-container\endcsname{8.6}
\expandafter\def\csname res@pvit-d12-scannet200-sd-wide@iou-container-pm\endcsname{8.6}
\expandafter\def\csname res@pvit-d12-scannet200-sd-wide@iou-wardrobe\endcsname{38.0}
\expandafter\def\csname res@pvit-d12-scannet200-sd-wide@iou-wardrobe-pm\endcsname{38.0}
\expandafter\def\csname res@pvit-d12-scannet200-sd-wide@iou-soap dispenser\endcsname{72.7}
\expandafter\def\csname res@pvit-d12-scannet200-sd-wide@iou-soap dispenser-pm\endcsname{72.7}
\expandafter\def\csname res@pvit-d12-scannet200-sd-wide@iou-telephone\endcsname{37.4}
\expandafter\def\csname res@pvit-d12-scannet200-sd-wide@iou-telephone-pm\endcsname{37.4}
\expandafter\def\csname res@pvit-d12-scannet200-sd-wide@iou-bucket\endcsname{23.4}
\expandafter\def\csname res@pvit-d12-scannet200-sd-wide@iou-bucket-pm\endcsname{23.4}
\expandafter\def\csname res@pvit-d12-scannet200-sd-wide@iou-clock\endcsname{23.8}
\expandafter\def\csname res@pvit-d12-scannet200-sd-wide@iou-clock-pm\endcsname{23.8}
\expandafter\def\csname res@pvit-d12-scannet200-sd-wide@iou-stand\endcsname{0.0}
\expandafter\def\csname res@pvit-d12-scannet200-sd-wide@iou-stand-pm\endcsname{0.0}
\expandafter\def\csname res@pvit-d12-scannet200-sd-wide@iou-light\endcsname{7.5}
\expandafter\def\csname res@pvit-d12-scannet200-sd-wide@iou-light-pm\endcsname{7.5}
\expandafter\def\csname res@pvit-d12-scannet200-sd-wide@iou-laundry basket\endcsname{26.6}
\expandafter\def\csname res@pvit-d12-scannet200-sd-wide@iou-laundry basket-pm\endcsname{26.6}
\expandafter\def\csname res@pvit-d12-scannet200-sd-wide@iou-pipe\endcsname{3.6}
\expandafter\def\csname res@pvit-d12-scannet200-sd-wide@iou-pipe-pm\endcsname{3.6}
\expandafter\def\csname res@pvit-d12-scannet200-sd-wide@iou-clothes dryer\endcsname{66.8}
\expandafter\def\csname res@pvit-d12-scannet200-sd-wide@iou-clothes dryer-pm\endcsname{66.8}
\expandafter\def\csname res@pvit-d12-scannet200-sd-wide@iou-guitar\endcsname{57.3}
\expandafter\def\csname res@pvit-d12-scannet200-sd-wide@iou-guitar-pm\endcsname{57.3}
\expandafter\def\csname res@pvit-d12-scannet200-sd-wide@iou-toilet paper holder\endcsname{0.0}
\expandafter\def\csname res@pvit-d12-scannet200-sd-wide@iou-toilet paper holder-pm\endcsname{0.0}
\expandafter\def\csname res@pvit-d12-scannet200-sd-wide@iou-seat\endcsname{0.0}
\expandafter\def\csname res@pvit-d12-scannet200-sd-wide@iou-seat-pm\endcsname{0.0}
\expandafter\def\csname res@pvit-d12-scannet200-sd-wide@iou-speaker\endcsname{0.0}
\expandafter\def\csname res@pvit-d12-scannet200-sd-wide@iou-speaker-pm\endcsname{0.0}
\expandafter\def\csname res@pvit-d12-scannet200-sd-wide@iou-column\endcsname{0.0}
\expandafter\def\csname res@pvit-d12-scannet200-sd-wide@iou-column-pm\endcsname{0.0}
\expandafter\def\csname res@pvit-d12-scannet200-sd-wide@iou-bicycle\endcsname{0.0}
\expandafter\def\csname res@pvit-d12-scannet200-sd-wide@iou-bicycle-pm\endcsname{0.0}
\expandafter\def\csname res@pvit-d12-scannet200-sd-wide@iou-ladder\endcsname{59.2}
\expandafter\def\csname res@pvit-d12-scannet200-sd-wide@iou-ladder-pm\endcsname{59.2}
\expandafter\def\csname res@pvit-d12-scannet200-sd-wide@iou-bathroom stall\endcsname{55.3}
\expandafter\def\csname res@pvit-d12-scannet200-sd-wide@iou-bathroom stall-pm\endcsname{55.3}
\expandafter\def\csname res@pvit-d12-scannet200-sd-wide@iou-shower wall\endcsname{52.5}
\expandafter\def\csname res@pvit-d12-scannet200-sd-wide@iou-shower wall-pm\endcsname{52.5}
\expandafter\def\csname res@pvit-d12-scannet200-sd-wide@iou-cup\endcsname{31.6}
\expandafter\def\csname res@pvit-d12-scannet200-sd-wide@iou-cup-pm\endcsname{31.6}
\expandafter\def\csname res@pvit-d12-scannet200-sd-wide@iou-jacket\endcsname{11.2}
\expandafter\def\csname res@pvit-d12-scannet200-sd-wide@iou-jacket-pm\endcsname{11.2}
\expandafter\def\csname res@pvit-d12-scannet200-sd-wide@iou-storage bin\endcsname{0.0}
\expandafter\def\csname res@pvit-d12-scannet200-sd-wide@iou-storage bin-pm\endcsname{0.0}
\expandafter\def\csname res@pvit-d12-scannet200-sd-wide@iou-coffee maker\endcsname{89.8}
\expandafter\def\csname res@pvit-d12-scannet200-sd-wide@iou-coffee maker-pm\endcsname{89.8}
\expandafter\def\csname res@pvit-d12-scannet200-sd-wide@iou-dishwasher\endcsname{39.3}
\expandafter\def\csname res@pvit-d12-scannet200-sd-wide@iou-dishwasher-pm\endcsname{39.3}
\expandafter\def\csname res@pvit-d12-scannet200-sd-wide@iou-paper towel roll\endcsname{39.6}
\expandafter\def\csname res@pvit-d12-scannet200-sd-wide@iou-paper towel roll-pm\endcsname{39.6}
\expandafter\def\csname res@pvit-d12-scannet200-sd-wide@iou-machine\endcsname{0.0}
\expandafter\def\csname res@pvit-d12-scannet200-sd-wide@iou-machine-pm\endcsname{0.0}
\expandafter\def\csname res@pvit-d12-scannet200-sd-wide@iou-mat\endcsname{29.8}
\expandafter\def\csname res@pvit-d12-scannet200-sd-wide@iou-mat-pm\endcsname{29.8}
\expandafter\def\csname res@pvit-d12-scannet200-sd-wide@iou-windowsill\endcsname{19.6}
\expandafter\def\csname res@pvit-d12-scannet200-sd-wide@iou-windowsill-pm\endcsname{19.6}
\expandafter\def\csname res@pvit-d12-scannet200-sd-wide@iou-bar\endcsname{18.7}
\expandafter\def\csname res@pvit-d12-scannet200-sd-wide@iou-bar-pm\endcsname{18.7}
\expandafter\def\csname res@pvit-d12-scannet200-sd-wide@iou-toaster\endcsname{74.4}
\expandafter\def\csname res@pvit-d12-scannet200-sd-wide@iou-toaster-pm\endcsname{74.4}
\expandafter\def\csname res@pvit-d12-scannet200-sd-wide@iou-bulletin board\endcsname{43.7}
\expandafter\def\csname res@pvit-d12-scannet200-sd-wide@iou-bulletin board-pm\endcsname{43.7}
\expandafter\def\csname res@pvit-d12-scannet200-sd-wide@iou-ironing board\endcsname{51.2}
\expandafter\def\csname res@pvit-d12-scannet200-sd-wide@iou-ironing board-pm\endcsname{51.2}
\expandafter\def\csname res@pvit-d12-scannet200-sd-wide@iou-fireplace\endcsname{65.1}
\expandafter\def\csname res@pvit-d12-scannet200-sd-wide@iou-fireplace-pm\endcsname{65.1}
\expandafter\def\csname res@pvit-d12-scannet200-sd-wide@iou-soap dish\endcsname{39.2}
\expandafter\def\csname res@pvit-d12-scannet200-sd-wide@iou-soap dish-pm\endcsname{39.2}
\expandafter\def\csname res@pvit-d12-scannet200-sd-wide@iou-kitchen counter\endcsname{56.2}
\expandafter\def\csname res@pvit-d12-scannet200-sd-wide@iou-kitchen counter-pm\endcsname{56.2}
\expandafter\def\csname res@pvit-d12-scannet200-sd-wide@iou-doorframe\endcsname{45.7}
\expandafter\def\csname res@pvit-d12-scannet200-sd-wide@iou-doorframe-pm\endcsname{45.7}
\expandafter\def\csname res@pvit-d12-scannet200-sd-wide@iou-toilet paper dispenser\endcsname{58.4}
\expandafter\def\csname res@pvit-d12-scannet200-sd-wide@iou-toilet paper dispenser-pm\endcsname{58.4}
\expandafter\def\csname res@pvit-d12-scannet200-sd-wide@iou-mini fridge\endcsname{53.4}
\expandafter\def\csname res@pvit-d12-scannet200-sd-wide@iou-mini fridge-pm\endcsname{53.4}
\expandafter\def\csname res@pvit-d12-scannet200-sd-wide@iou-fire extinguisher\endcsname{27.5}
\expandafter\def\csname res@pvit-d12-scannet200-sd-wide@iou-fire extinguisher-pm\endcsname{27.5}
\expandafter\def\csname res@pvit-d12-scannet200-sd-wide@iou-ball\endcsname{85.5}
\expandafter\def\csname res@pvit-d12-scannet200-sd-wide@iou-ball-pm\endcsname{85.5}
\expandafter\def\csname res@pvit-d12-scannet200-sd-wide@iou-hat\endcsname{0.0}
\expandafter\def\csname res@pvit-d12-scannet200-sd-wide@iou-hat-pm\endcsname{0.0}
\expandafter\def\csname res@pvit-d12-scannet200-sd-wide@iou-shower curtain rod\endcsname{49.2}
\expandafter\def\csname res@pvit-d12-scannet200-sd-wide@iou-shower curtain rod-pm\endcsname{49.2}
\expandafter\def\csname res@pvit-d12-scannet200-sd-wide@iou-water cooler\endcsname{84.2}
\expandafter\def\csname res@pvit-d12-scannet200-sd-wide@iou-water cooler-pm\endcsname{84.2}
\expandafter\def\csname res@pvit-d12-scannet200-sd-wide@iou-paper cutter\endcsname{20.8}
\expandafter\def\csname res@pvit-d12-scannet200-sd-wide@iou-paper cutter-pm\endcsname{20.8}
\expandafter\def\csname res@pvit-d12-scannet200-sd-wide@iou-tray\endcsname{0.6}
\expandafter\def\csname res@pvit-d12-scannet200-sd-wide@iou-tray-pm\endcsname{0.6}
\expandafter\def\csname res@pvit-d12-scannet200-sd-wide@iou-shower door\endcsname{32.5}
\expandafter\def\csname res@pvit-d12-scannet200-sd-wide@iou-shower door-pm\endcsname{32.5}
\expandafter\def\csname res@pvit-d12-scannet200-sd-wide@iou-pillar\endcsname{10.1}
\expandafter\def\csname res@pvit-d12-scannet200-sd-wide@iou-pillar-pm\endcsname{10.1}
\expandafter\def\csname res@pvit-d12-scannet200-sd-wide@iou-ledge\endcsname{18.1}
\expandafter\def\csname res@pvit-d12-scannet200-sd-wide@iou-ledge-pm\endcsname{18.1}
\expandafter\def\csname res@pvit-d12-scannet200-sd-wide@iou-toaster oven\endcsname{11.0}
\expandafter\def\csname res@pvit-d12-scannet200-sd-wide@iou-toaster oven-pm\endcsname{11.0}
\expandafter\def\csname res@pvit-d12-scannet200-sd-wide@iou-mouse\endcsname{6.0}
\expandafter\def\csname res@pvit-d12-scannet200-sd-wide@iou-mouse-pm\endcsname{6.0}
\expandafter\def\csname res@pvit-d12-scannet200-sd-wide@iou-toilet seat cover dispenser\endcsname{81.6}
\expandafter\def\csname res@pvit-d12-scannet200-sd-wide@iou-toilet seat cover dispenser-pm\endcsname{81.6}
\expandafter\def\csname res@pvit-d12-scannet200-sd-wide@iou-furniture\endcsname{0.0}
\expandafter\def\csname res@pvit-d12-scannet200-sd-wide@iou-furniture-pm\endcsname{0.0}
\expandafter\def\csname res@pvit-d12-scannet200-sd-wide@iou-cart\endcsname{22.0}
\expandafter\def\csname res@pvit-d12-scannet200-sd-wide@iou-cart-pm\endcsname{22.0}
\expandafter\def\csname res@pvit-d12-scannet200-sd-wide@iou-storage container\endcsname{0.0}
\expandafter\def\csname res@pvit-d12-scannet200-sd-wide@iou-storage container-pm\endcsname{0.0}
\expandafter\def\csname res@pvit-d12-scannet200-sd-wide@iou-scale\endcsname{40.0}
\expandafter\def\csname res@pvit-d12-scannet200-sd-wide@iou-scale-pm\endcsname{40.0}
\expandafter\def\csname res@pvit-d12-scannet200-sd-wide@iou-tissue box\endcsname{15.2}
\expandafter\def\csname res@pvit-d12-scannet200-sd-wide@iou-tissue box-pm\endcsname{15.2}
\expandafter\def\csname res@pvit-d12-scannet200-sd-wide@iou-light switch\endcsname{0.0}
\expandafter\def\csname res@pvit-d12-scannet200-sd-wide@iou-light switch-pm\endcsname{0.0}
\expandafter\def\csname res@pvit-d12-scannet200-sd-wide@iou-crate\endcsname{0.0}
\expandafter\def\csname res@pvit-d12-scannet200-sd-wide@iou-crate-pm\endcsname{0.0}
\expandafter\def\csname res@pvit-d12-scannet200-sd-wide@iou-power outlet\endcsname{0.0}
\expandafter\def\csname res@pvit-d12-scannet200-sd-wide@iou-power outlet-pm\endcsname{0.0}
\expandafter\def\csname res@pvit-d12-scannet200-sd-wide@iou-decoration\endcsname{48.6}
\expandafter\def\csname res@pvit-d12-scannet200-sd-wide@iou-decoration-pm\endcsname{48.6}
\expandafter\def\csname res@pvit-d12-scannet200-sd-wide@iou-sign\endcsname{28.0}
\expandafter\def\csname res@pvit-d12-scannet200-sd-wide@iou-sign-pm\endcsname{28.0}
\expandafter\def\csname res@pvit-d12-scannet200-sd-wide@iou-projector\endcsname{0.0}
\expandafter\def\csname res@pvit-d12-scannet200-sd-wide@iou-projector-pm\endcsname{0.0}
\expandafter\def\csname res@pvit-d12-scannet200-sd-wide@iou-closet door\endcsname{18.5}
\expandafter\def\csname res@pvit-d12-scannet200-sd-wide@iou-closet door-pm\endcsname{18.5}
\expandafter\def\csname res@pvit-d12-scannet200-sd-wide@iou-vacuum cleaner\endcsname{87.5}
\expandafter\def\csname res@pvit-d12-scannet200-sd-wide@iou-vacuum cleaner-pm\endcsname{87.5}
\expandafter\def\csname res@pvit-d12-scannet200-sd-wide@iou-candle\endcsname{0.0}
\expandafter\def\csname res@pvit-d12-scannet200-sd-wide@iou-candle-pm\endcsname{0.0}
\expandafter\def\csname res@pvit-d12-scannet200-sd-wide@iou-plunger\endcsname{33.9}
\expandafter\def\csname res@pvit-d12-scannet200-sd-wide@iou-plunger-pm\endcsname{33.9}
\expandafter\def\csname res@pvit-d12-scannet200-sd-wide@iou-stuffed animal\endcsname{0.0}
\expandafter\def\csname res@pvit-d12-scannet200-sd-wide@iou-stuffed animal-pm\endcsname{0.0}
\expandafter\def\csname res@pvit-d12-scannet200-sd-wide@iou-headphones\endcsname{0.0}
\expandafter\def\csname res@pvit-d12-scannet200-sd-wide@iou-headphones-pm\endcsname{0.0}
\expandafter\def\csname res@pvit-d12-scannet200-sd-wide@iou-dish rack\endcsname{64.7}
\expandafter\def\csname res@pvit-d12-scannet200-sd-wide@iou-dish rack-pm\endcsname{64.7}
\expandafter\def\csname res@pvit-d12-scannet200-sd-wide@iou-broom\endcsname{10.7}
\expandafter\def\csname res@pvit-d12-scannet200-sd-wide@iou-broom-pm\endcsname{10.7}
\expandafter\def\csname res@pvit-d12-scannet200-sd-wide@iou-guitar case\endcsname{0.0}
\expandafter\def\csname res@pvit-d12-scannet200-sd-wide@iou-guitar case-pm\endcsname{0.0}
\expandafter\def\csname res@pvit-d12-scannet200-sd-wide@iou-range hood\endcsname{74.5}
\expandafter\def\csname res@pvit-d12-scannet200-sd-wide@iou-range hood-pm\endcsname{74.5}
\expandafter\def\csname res@pvit-d12-scannet200-sd-wide@iou-dustpan\endcsname{0.3}
\expandafter\def\csname res@pvit-d12-scannet200-sd-wide@iou-dustpan-pm\endcsname{0.3}
\expandafter\def\csname res@pvit-d12-scannet200-sd-wide@iou-hair dryer\endcsname{10.9}
\expandafter\def\csname res@pvit-d12-scannet200-sd-wide@iou-hair dryer-pm\endcsname{10.9}
\expandafter\def\csname res@pvit-d12-scannet200-sd-wide@iou-water bottle\endcsname{0.0}
\expandafter\def\csname res@pvit-d12-scannet200-sd-wide@iou-water bottle-pm\endcsname{0.0}
\expandafter\def\csname res@pvit-d12-scannet200-sd-wide@iou-handicap bar\endcsname{6.3}
\expandafter\def\csname res@pvit-d12-scannet200-sd-wide@iou-handicap bar-pm\endcsname{6.3}
\expandafter\def\csname res@pvit-d12-scannet200-sd-wide@iou-purse\endcsname{0.0}
\expandafter\def\csname res@pvit-d12-scannet200-sd-wide@iou-purse-pm\endcsname{0.0}
\expandafter\def\csname res@pvit-d12-scannet200-sd-wide@iou-vent\endcsname{26.2}
\expandafter\def\csname res@pvit-d12-scannet200-sd-wide@iou-vent-pm\endcsname{26.2}
\expandafter\def\csname res@pvit-d12-scannet200-sd-wide@iou-shower floor\endcsname{79.0}
\expandafter\def\csname res@pvit-d12-scannet200-sd-wide@iou-shower floor-pm\endcsname{79.0}
\expandafter\def\csname res@pvit-d12-scannet200-sd-wide@iou-water pitcher\endcsname{0.0}
\expandafter\def\csname res@pvit-d12-scannet200-sd-wide@iou-water pitcher-pm\endcsname{0.0}
\expandafter\def\csname res@pvit-d12-scannet200-sd-wide@iou-mailbox\endcsname{57.3}
\expandafter\def\csname res@pvit-d12-scannet200-sd-wide@iou-mailbox-pm\endcsname{57.3}
\expandafter\def\csname res@pvit-d12-scannet200-sd-wide@iou-bowl\endcsname{15.4}
\expandafter\def\csname res@pvit-d12-scannet200-sd-wide@iou-bowl-pm\endcsname{15.4}
\expandafter\def\csname res@pvit-d12-scannet200-sd-wide@iou-paper bag\endcsname{8.6}
\expandafter\def\csname res@pvit-d12-scannet200-sd-wide@iou-paper bag-pm\endcsname{8.6}
\expandafter\def\csname res@pvit-d12-scannet200-sd-wide@iou-alarm clock\endcsname{0.0}
\expandafter\def\csname res@pvit-d12-scannet200-sd-wide@iou-alarm clock-pm\endcsname{0.0}
\expandafter\def\csname res@pvit-d12-scannet200-sd-wide@iou-music stand\endcsname{0.0}
\expandafter\def\csname res@pvit-d12-scannet200-sd-wide@iou-music stand-pm\endcsname{0.0}
\expandafter\def\csname res@pvit-d12-scannet200-sd-wide@iou-projector screen\endcsname{0.0}
\expandafter\def\csname res@pvit-d12-scannet200-sd-wide@iou-projector screen-pm\endcsname{0.0}
\expandafter\def\csname res@pvit-d12-scannet200-sd-wide@iou-divider\endcsname{65.1}
\expandafter\def\csname res@pvit-d12-scannet200-sd-wide@iou-divider-pm\endcsname{65.1}
\expandafter\def\csname res@pvit-d12-scannet200-sd-wide@iou-laundry detergent\endcsname{0.0}
\expandafter\def\csname res@pvit-d12-scannet200-sd-wide@iou-laundry detergent-pm\endcsname{0.0}
\expandafter\def\csname res@pvit-d12-scannet200-sd-wide@iou-bathroom counter\endcsname{33.9}
\expandafter\def\csname res@pvit-d12-scannet200-sd-wide@iou-bathroom counter-pm\endcsname{33.9}
\expandafter\def\csname res@pvit-d12-scannet200-sd-wide@iou-object\endcsname{8.2}
\expandafter\def\csname res@pvit-d12-scannet200-sd-wide@iou-object-pm\endcsname{8.2}
\expandafter\def\csname res@pvit-d12-scannet200-sd-wide@iou-bathroom vanity\endcsname{38.7}
\expandafter\def\csname res@pvit-d12-scannet200-sd-wide@iou-bathroom vanity-pm\endcsname{38.7}
\expandafter\def\csname res@pvit-d12-scannet200-sd-wide@iou-closet wall\endcsname{15.6}
\expandafter\def\csname res@pvit-d12-scannet200-sd-wide@iou-closet wall-pm\endcsname{15.6}
\expandafter\def\csname res@pvit-d12-scannet200-sd-wide@iou-laundry hamper\endcsname{40.5}
\expandafter\def\csname res@pvit-d12-scannet200-sd-wide@iou-laundry hamper-pm\endcsname{40.5}
\expandafter\def\csname res@pvit-d12-scannet200-sd-wide@iou-bathroom stall door\endcsname{53.1}
\expandafter\def\csname res@pvit-d12-scannet200-sd-wide@iou-bathroom stall door-pm\endcsname{53.1}
\expandafter\def\csname res@pvit-d12-scannet200-sd-wide@iou-ceiling light\endcsname{49.3}
\expandafter\def\csname res@pvit-d12-scannet200-sd-wide@iou-ceiling light-pm\endcsname{49.3}
\expandafter\def\csname res@pvit-d12-scannet200-sd-wide@iou-trash bin\endcsname{52.6}
\expandafter\def\csname res@pvit-d12-scannet200-sd-wide@iou-trash bin-pm\endcsname{52.6}
\expandafter\def\csname res@pvit-d12-scannet200-sd-wide@iou-dumbbell\endcsname{61.5}
\expandafter\def\csname res@pvit-d12-scannet200-sd-wide@iou-dumbbell-pm\endcsname{61.5}
\expandafter\def\csname res@pvit-d12-scannet200-sd-wide@iou-stair rail\endcsname{23.2}
\expandafter\def\csname res@pvit-d12-scannet200-sd-wide@iou-stair rail-pm\endcsname{23.2}
\expandafter\def\csname res@pvit-d12-scannet200-sd-wide@iou-tube\endcsname{35.6}
\expandafter\def\csname res@pvit-d12-scannet200-sd-wide@iou-tube-pm\endcsname{35.6}
\expandafter\def\csname res@pvit-d12-scannet200-sd-wide@iou-bathroom cabinet\endcsname{7.0}
\expandafter\def\csname res@pvit-d12-scannet200-sd-wide@iou-bathroom cabinet-pm\endcsname{7.0}
\expandafter\def\csname res@pvit-d12-scannet200-sd-wide@iou-cd case\endcsname{0.0}
\expandafter\def\csname res@pvit-d12-scannet200-sd-wide@iou-cd case-pm\endcsname{0.0}
\expandafter\def\csname res@pvit-d12-scannet200-sd-wide@iou-closet rod\endcsname{26.7}
\expandafter\def\csname res@pvit-d12-scannet200-sd-wide@iou-closet rod-pm\endcsname{26.7}
\expandafter\def\csname res@pvit-d12-scannet200-sd-wide@iou-coffee kettle\endcsname{33.1}
\expandafter\def\csname res@pvit-d12-scannet200-sd-wide@iou-coffee kettle-pm\endcsname{33.1}
\expandafter\def\csname res@pvit-d12-scannet200-sd-wide@iou-structure\endcsname{0.0}
\expandafter\def\csname res@pvit-d12-scannet200-sd-wide@iou-structure-pm\endcsname{0.0}
\expandafter\def\csname res@pvit-d12-scannet200-sd-wide@iou-shower head\endcsname{30.5}
\expandafter\def\csname res@pvit-d12-scannet200-sd-wide@iou-shower head-pm\endcsname{30.5}
\expandafter\def\csname res@pvit-d12-scannet200-sd-wide@iou-keyboard piano\endcsname{0.0}
\expandafter\def\csname res@pvit-d12-scannet200-sd-wide@iou-keyboard piano-pm\endcsname{0.0}
\expandafter\def\csname res@pvit-d12-scannet200-sd-wide@iou-case of water bottles\endcsname{0.4}
\expandafter\def\csname res@pvit-d12-scannet200-sd-wide@iou-case of water bottles-pm\endcsname{0.4}
\expandafter\def\csname res@pvit-d12-scannet200-sd-wide@iou-coat rack\endcsname{23.9}
\expandafter\def\csname res@pvit-d12-scannet200-sd-wide@iou-coat rack-pm\endcsname{23.9}
\expandafter\def\csname res@pvit-d12-scannet200-sd-wide@iou-storage organizer\endcsname{0.0}
\expandafter\def\csname res@pvit-d12-scannet200-sd-wide@iou-storage organizer-pm\endcsname{0.0}
\expandafter\def\csname res@pvit-d12-scannet200-sd-wide@iou-folded chair\endcsname{0.0}
\expandafter\def\csname res@pvit-d12-scannet200-sd-wide@iou-folded chair-pm\endcsname{0.0}
\expandafter\def\csname res@pvit-d12-scannet200-sd-wide@iou-fire alarm\endcsname{7.8}
\expandafter\def\csname res@pvit-d12-scannet200-sd-wide@iou-fire alarm-pm\endcsname{7.8}
\expandafter\def\csname res@pvit-d12-scannet200-sd-wide@iou-power strip\endcsname{0.0}
\expandafter\def\csname res@pvit-d12-scannet200-sd-wide@iou-power strip-pm\endcsname{0.0}
\expandafter\def\csname res@pvit-d12-scannet200-sd-wide@iou-calendar\endcsname{0.0}
\expandafter\def\csname res@pvit-d12-scannet200-sd-wide@iou-calendar-pm\endcsname{0.0}
\expandafter\def\csname res@pvit-d12-scannet200-sd-wide@iou-poster\endcsname{0.0}
\expandafter\def\csname res@pvit-d12-scannet200-sd-wide@iou-poster-pm\endcsname{0.0}
\expandafter\def\csname res@pvit-d12-scannet200-sd-wide@iou-potted plant\endcsname{0.0}
\expandafter\def\csname res@pvit-d12-scannet200-sd-wide@iou-potted plant-pm\endcsname{0.0}
\expandafter\def\csname res@pvit-d12-scannet200-sd-wide@iou-luggage\endcsname{0.0}
\expandafter\def\csname res@pvit-d12-scannet200-sd-wide@iou-luggage-pm\endcsname{0.0}
\expandafter\def\csname res@pvit-d12-scannet200-sd-wide@iou-mattress\endcsname{1.8}
\expandafter\def\csname res@pvit-d12-scannet200-sd-wide@iou-mattress-pm\endcsname{1.8}
\expandafter\def\csname res@pvit-d12-scannet200-sd-wide@miou-delta\endcsname{\ensuremath{-}0.7}
\expandafter\def\csname res@pvit-d12-scannet200-sd-wide@macc-delta\endcsname{\ensuremath{-}0.7}
\expandafter\def\csname res@pvit-d12-scannet200-sd-wide@allacc-delta\endcsname{+0.1}
\expandafter\def\csname res@pvit-d12-scannet200-sd-wide@iou-wall-delta\endcsname{\ensuremath{-}0.2}
\expandafter\def\csname res@pvit-d12-scannet200-sd-wide@iou-chair-delta\endcsname{\ensuremath{-}0.7}
\expandafter\def\csname res@pvit-d12-scannet200-sd-wide@iou-floor-delta\endcsname{+0.0}
\expandafter\def\csname res@pvit-d12-scannet200-sd-wide@iou-table-delta\endcsname{\ensuremath{-}0.4}
\expandafter\def\csname res@pvit-d12-scannet200-sd-wide@iou-door-delta\endcsname{\ensuremath{-}0.7}
\expandafter\def\csname res@pvit-d12-scannet200-sd-wide@iou-couch-delta\endcsname{\ensuremath{-}2.8}
\expandafter\def\csname res@pvit-d12-scannet200-sd-wide@iou-cabinet-delta\endcsname{+2.3}
\expandafter\def\csname res@pvit-d12-scannet200-sd-wide@iou-shelf-delta\endcsname{+7.0}
\expandafter\def\csname res@pvit-d12-scannet200-sd-wide@iou-desk-delta\endcsname{+0.9}
\expandafter\def\csname res@pvit-d12-scannet200-sd-wide@iou-office chair-delta\endcsname{\ensuremath{-}7.9}
\expandafter\def\csname res@pvit-d12-scannet200-sd-wide@iou-bed-delta\endcsname{+1.3}
\expandafter\def\csname res@pvit-d12-scannet200-sd-wide@iou-pillow-delta\endcsname{\ensuremath{-}0.8}
\expandafter\def\csname res@pvit-d12-scannet200-sd-wide@iou-sink-delta\endcsname{\ensuremath{-}0.2}
\expandafter\def\csname res@pvit-d12-scannet200-sd-wide@iou-picture-delta\endcsname{+1.1}
\expandafter\def\csname res@pvit-d12-scannet200-sd-wide@iou-window-delta\endcsname{\ensuremath{-}0.6}
\expandafter\def\csname res@pvit-d12-scannet200-sd-wide@iou-toilet-delta\endcsname{+0.3}
\expandafter\def\csname res@pvit-d12-scannet200-sd-wide@iou-bookshelf-delta\endcsname{+3.4}
\expandafter\def\csname res@pvit-d12-scannet200-sd-wide@iou-monitor-delta\endcsname{\ensuremath{-}0.1}
\expandafter\def\csname res@pvit-d12-scannet200-sd-wide@iou-curtain-delta\endcsname{\ensuremath{-}0.9}
\expandafter\def\csname res@pvit-d12-scannet200-sd-wide@iou-book-delta\endcsname{+0.0}
\expandafter\def\csname res@pvit-d12-scannet200-sd-wide@iou-armchair-delta\endcsname{\ensuremath{-}2.5}
\expandafter\def\csname res@pvit-d12-scannet200-sd-wide@iou-coffee table-delta\endcsname{+0.2}
\expandafter\def\csname res@pvit-d12-scannet200-sd-wide@iou-box-delta\endcsname{+0.2}
\expandafter\def\csname res@pvit-d12-scannet200-sd-wide@iou-refrigerator-delta\endcsname{\ensuremath{-}8.7}
\expandafter\def\csname res@pvit-d12-scannet200-sd-wide@iou-lamp-delta\endcsname{\ensuremath{-}8.1}
\expandafter\def\csname res@pvit-d12-scannet200-sd-wide@iou-kitchen cabinet-delta\endcsname{+3.8}
\expandafter\def\csname res@pvit-d12-scannet200-sd-wide@iou-towel-delta\endcsname{+1.0}
\expandafter\def\csname res@pvit-d12-scannet200-sd-wide@iou-clothes-delta\endcsname{\ensuremath{-}1.8}
\expandafter\def\csname res@pvit-d12-scannet200-sd-wide@iou-tv-delta\endcsname{\ensuremath{-}0.9}
\expandafter\def\csname res@pvit-d12-scannet200-sd-wide@iou-nightstand-delta\endcsname{+0.4}
\expandafter\def\csname res@pvit-d12-scannet200-sd-wide@iou-counter-delta\endcsname{\ensuremath{-}2.2}
\expandafter\def\csname res@pvit-d12-scannet200-sd-wide@iou-dresser-delta\endcsname{\ensuremath{-}4.3}
\expandafter\def\csname res@pvit-d12-scannet200-sd-wide@iou-stool-delta\endcsname{\ensuremath{-}13.8}
\expandafter\def\csname res@pvit-d12-scannet200-sd-wide@iou-cushion-delta\endcsname{+0.0}
\expandafter\def\csname res@pvit-d12-scannet200-sd-wide@iou-plant-delta\endcsname{+10.4}
\expandafter\def\csname res@pvit-d12-scannet200-sd-wide@iou-ceiling-delta\endcsname{+0.4}
\expandafter\def\csname res@pvit-d12-scannet200-sd-wide@iou-bathtub-delta\endcsname{+0.7}
\expandafter\def\csname res@pvit-d12-scannet200-sd-wide@iou-end table-delta\endcsname{\ensuremath{-}13.6}
\expandafter\def\csname res@pvit-d12-scannet200-sd-wide@iou-dining table-delta\endcsname{\ensuremath{-}8.7}
\expandafter\def\csname res@pvit-d12-scannet200-sd-wide@iou-keyboard-delta\endcsname{\ensuremath{-}2.2}
\expandafter\def\csname res@pvit-d12-scannet200-sd-wide@iou-bag-delta\endcsname{\ensuremath{-}2.4}
\expandafter\def\csname res@pvit-d12-scannet200-sd-wide@iou-backpack-delta\endcsname{\ensuremath{-}1.9}
\expandafter\def\csname res@pvit-d12-scannet200-sd-wide@iou-toilet paper-delta\endcsname{\ensuremath{-}1.5}
\expandafter\def\csname res@pvit-d12-scannet200-sd-wide@iou-printer-delta\endcsname{\ensuremath{-}6.2}
\expandafter\def\csname res@pvit-d12-scannet200-sd-wide@iou-tv stand-delta\endcsname{\ensuremath{-}17.1}
\expandafter\def\csname res@pvit-d12-scannet200-sd-wide@iou-whiteboard-delta\endcsname{+0.0}
\expandafter\def\csname res@pvit-d12-scannet200-sd-wide@iou-blanket-delta\endcsname{\ensuremath{-}1.9}
\expandafter\def\csname res@pvit-d12-scannet200-sd-wide@iou-shower curtain-delta\endcsname{\ensuremath{-}3.3}
\expandafter\def\csname res@pvit-d12-scannet200-sd-wide@iou-trash can-delta\endcsname{\ensuremath{-}0.7}
\expandafter\def\csname res@pvit-d12-scannet200-sd-wide@iou-closet-delta\endcsname{+4.8}
\expandafter\def\csname res@pvit-d12-scannet200-sd-wide@iou-stairs-delta\endcsname{\ensuremath{-}5.2}
\expandafter\def\csname res@pvit-d12-scannet200-sd-wide@iou-microwave-delta\endcsname{+0.7}
\expandafter\def\csname res@pvit-d12-scannet200-sd-wide@iou-stove-delta\endcsname{+3.3}
\expandafter\def\csname res@pvit-d12-scannet200-sd-wide@iou-shoe-delta\endcsname{\ensuremath{-}9.2}
\expandafter\def\csname res@pvit-d12-scannet200-sd-wide@iou-computer tower-delta\endcsname{\ensuremath{-}3.8}
\expandafter\def\csname res@pvit-d12-scannet200-sd-wide@iou-bottle-delta\endcsname{\ensuremath{-}7.1}
\expandafter\def\csname res@pvit-d12-scannet200-sd-wide@iou-bin-delta\endcsname{+0.0}
\expandafter\def\csname res@pvit-d12-scannet200-sd-wide@iou-ottoman-delta\endcsname{+3.1}
\expandafter\def\csname res@pvit-d12-scannet200-sd-wide@iou-bench-delta\endcsname{+4.7}
\expandafter\def\csname res@pvit-d12-scannet200-sd-wide@iou-board-delta\endcsname{\ensuremath{-}1.3}
\expandafter\def\csname res@pvit-d12-scannet200-sd-wide@iou-washing machine-delta\endcsname{+1.1}
\expandafter\def\csname res@pvit-d12-scannet200-sd-wide@iou-mirror-delta\endcsname{\ensuremath{-}2.4}
\expandafter\def\csname res@pvit-d12-scannet200-sd-wide@iou-copier-delta\endcsname{\ensuremath{-}2.2}
\expandafter\def\csname res@pvit-d12-scannet200-sd-wide@iou-basket-delta\endcsname{+5.9}
\expandafter\def\csname res@pvit-d12-scannet200-sd-wide@iou-sofa chair-delta\endcsname{\ensuremath{-}2.9}
\expandafter\def\csname res@pvit-d12-scannet200-sd-wide@iou-file cabinet-delta\endcsname{\ensuremath{-}12.3}
\expandafter\def\csname res@pvit-d12-scannet200-sd-wide@iou-fan-delta\endcsname{\ensuremath{-}27.2}
\expandafter\def\csname res@pvit-d12-scannet200-sd-wide@iou-laptop-delta\endcsname{+4.5}
\expandafter\def\csname res@pvit-d12-scannet200-sd-wide@iou-shower-delta\endcsname{+1.5}
\expandafter\def\csname res@pvit-d12-scannet200-sd-wide@iou-paper-delta\endcsname{+1.8}
\expandafter\def\csname res@pvit-d12-scannet200-sd-wide@iou-person-delta\endcsname{+2.1}
\expandafter\def\csname res@pvit-d12-scannet200-sd-wide@iou-paper towel dispenser-delta\endcsname{+9.7}
\expandafter\def\csname res@pvit-d12-scannet200-sd-wide@iou-oven-delta\endcsname{\ensuremath{-}2.0}
\expandafter\def\csname res@pvit-d12-scannet200-sd-wide@iou-blinds-delta\endcsname{+0.0}
\expandafter\def\csname res@pvit-d12-scannet200-sd-wide@iou-rack-delta\endcsname{+0.0}
\expandafter\def\csname res@pvit-d12-scannet200-sd-wide@iou-plate-delta\endcsname{+0.0}
\expandafter\def\csname res@pvit-d12-scannet200-sd-wide@iou-blackboard-delta\endcsname{\ensuremath{-}7.0}
\expandafter\def\csname res@pvit-d12-scannet200-sd-wide@iou-piano-delta\endcsname{\ensuremath{-}5.6}
\expandafter\def\csname res@pvit-d12-scannet200-sd-wide@iou-suitcase-delta\endcsname{+8.1}
\expandafter\def\csname res@pvit-d12-scannet200-sd-wide@iou-rail-delta\endcsname{+2.2}
\expandafter\def\csname res@pvit-d12-scannet200-sd-wide@iou-radiator-delta\endcsname{+2.8}
\expandafter\def\csname res@pvit-d12-scannet200-sd-wide@iou-recycling bin-delta\endcsname{\ensuremath{-}1.6}
\expandafter\def\csname res@pvit-d12-scannet200-sd-wide@iou-container-delta\endcsname{+5.6}
\expandafter\def\csname res@pvit-d12-scannet200-sd-wide@iou-wardrobe-delta\endcsname{\ensuremath{-}5.8}
\expandafter\def\csname res@pvit-d12-scannet200-sd-wide@iou-soap dispenser-delta\endcsname{+1.4}
\expandafter\def\csname res@pvit-d12-scannet200-sd-wide@iou-telephone-delta\endcsname{+0.6}
\expandafter\def\csname res@pvit-d12-scannet200-sd-wide@iou-bucket-delta\endcsname{\ensuremath{-}9.7}
\expandafter\def\csname res@pvit-d12-scannet200-sd-wide@iou-clock-delta\endcsname{\ensuremath{-}0.8}
\expandafter\def\csname res@pvit-d12-scannet200-sd-wide@iou-stand-delta\endcsname{+0.0}
\expandafter\def\csname res@pvit-d12-scannet200-sd-wide@iou-light-delta\endcsname{+0.0}
\expandafter\def\csname res@pvit-d12-scannet200-sd-wide@iou-laundry basket-delta\endcsname{+17.6}
\expandafter\def\csname res@pvit-d12-scannet200-sd-wide@iou-pipe-delta\endcsname{\ensuremath{-}4.5}
\expandafter\def\csname res@pvit-d12-scannet200-sd-wide@iou-clothes dryer-delta\endcsname{\ensuremath{-}15.5}
\expandafter\def\csname res@pvit-d12-scannet200-sd-wide@iou-guitar-delta\endcsname{\ensuremath{-}26.1}
\expandafter\def\csname res@pvit-d12-scannet200-sd-wide@iou-toilet paper holder-delta\endcsname{+0.0}
\expandafter\def\csname res@pvit-d12-scannet200-sd-wide@iou-seat-delta\endcsname{+0.0}
\expandafter\def\csname res@pvit-d12-scannet200-sd-wide@iou-speaker-delta\endcsname{+0.0}
\expandafter\def\csname res@pvit-d12-scannet200-sd-wide@iou-column-delta\endcsname{+0.0}
\expandafter\def\csname res@pvit-d12-scannet200-sd-wide@iou-bicycle-delta\endcsname{+0.0}
\expandafter\def\csname res@pvit-d12-scannet200-sd-wide@iou-ladder-delta\endcsname{+9.0}
\expandafter\def\csname res@pvit-d12-scannet200-sd-wide@iou-bathroom stall-delta\endcsname{\ensuremath{-}5.9}
\expandafter\def\csname res@pvit-d12-scannet200-sd-wide@iou-shower wall-delta\endcsname{+4.6}
\expandafter\def\csname res@pvit-d12-scannet200-sd-wide@iou-cup-delta\endcsname{+6.5}
\expandafter\def\csname res@pvit-d12-scannet200-sd-wide@iou-jacket-delta\endcsname{\ensuremath{-}8.2}
\expandafter\def\csname res@pvit-d12-scannet200-sd-wide@iou-storage bin-delta\endcsname{\ensuremath{-}3.3}
\expandafter\def\csname res@pvit-d12-scannet200-sd-wide@iou-coffee maker-delta\endcsname{\ensuremath{-}1.6}
\expandafter\def\csname res@pvit-d12-scannet200-sd-wide@iou-dishwasher-delta\endcsname{\ensuremath{-}8.1}
\expandafter\def\csname res@pvit-d12-scannet200-sd-wide@iou-paper towel roll-delta\endcsname{+3.3}
\expandafter\def\csname res@pvit-d12-scannet200-sd-wide@iou-machine-delta\endcsname{+0.0}
\expandafter\def\csname res@pvit-d12-scannet200-sd-wide@iou-mat-delta\endcsname{\ensuremath{-}7.5}
\expandafter\def\csname res@pvit-d12-scannet200-sd-wide@iou-windowsill-delta\endcsname{+0.7}
\expandafter\def\csname res@pvit-d12-scannet200-sd-wide@iou-bar-delta\endcsname{+3.6}
\expandafter\def\csname res@pvit-d12-scannet200-sd-wide@iou-toaster-delta\endcsname{+10.5}
\expandafter\def\csname res@pvit-d12-scannet200-sd-wide@iou-bulletin board-delta\endcsname{\ensuremath{-}4.2}
\expandafter\def\csname res@pvit-d12-scannet200-sd-wide@iou-ironing board-delta\endcsname{+4.6}
\expandafter\def\csname res@pvit-d12-scannet200-sd-wide@iou-fireplace-delta\endcsname{\ensuremath{-}0.7}
\expandafter\def\csname res@pvit-d12-scannet200-sd-wide@iou-soap dish-delta\endcsname{+6.4}
\expandafter\def\csname res@pvit-d12-scannet200-sd-wide@iou-kitchen counter-delta\endcsname{\ensuremath{-}0.2}
\expandafter\def\csname res@pvit-d12-scannet200-sd-wide@iou-doorframe-delta\endcsname{+0.4}
\expandafter\def\csname res@pvit-d12-scannet200-sd-wide@iou-toilet paper dispenser-delta\endcsname{+1.2}
\expandafter\def\csname res@pvit-d12-scannet200-sd-wide@iou-mini fridge-delta\endcsname{\ensuremath{-}2.8}
\expandafter\def\csname res@pvit-d12-scannet200-sd-wide@iou-fire extinguisher-delta\endcsname{\ensuremath{-}5.1}
\expandafter\def\csname res@pvit-d12-scannet200-sd-wide@iou-ball-delta\endcsname{+5.8}
\expandafter\def\csname res@pvit-d12-scannet200-sd-wide@iou-hat-delta\endcsname{+0.0}
\expandafter\def\csname res@pvit-d12-scannet200-sd-wide@iou-shower curtain rod-delta\endcsname{\ensuremath{-}0.2}
\expandafter\def\csname res@pvit-d12-scannet200-sd-wide@iou-water cooler-delta\endcsname{\ensuremath{-}3.9}
\expandafter\def\csname res@pvit-d12-scannet200-sd-wide@iou-paper cutter-delta\endcsname{\ensuremath{-}1.6}
\expandafter\def\csname res@pvit-d12-scannet200-sd-wide@iou-tray-delta\endcsname{\ensuremath{-}7.0}
\expandafter\def\csname res@pvit-d12-scannet200-sd-wide@iou-shower door-delta\endcsname{+5.9}
\expandafter\def\csname res@pvit-d12-scannet200-sd-wide@iou-pillar-delta\endcsname{+3.7}
\expandafter\def\csname res@pvit-d12-scannet200-sd-wide@iou-ledge-delta\endcsname{+8.8}
\expandafter\def\csname res@pvit-d12-scannet200-sd-wide@iou-toaster oven-delta\endcsname{+10.0}
\expandafter\def\csname res@pvit-d12-scannet200-sd-wide@iou-mouse-delta\endcsname{+0.4}
\expandafter\def\csname res@pvit-d12-scannet200-sd-wide@iou-toilet seat cover dispenser-delta\endcsname{\ensuremath{-}0.8}
\expandafter\def\csname res@pvit-d12-scannet200-sd-wide@iou-furniture-delta\endcsname{+0.0}
\expandafter\def\csname res@pvit-d12-scannet200-sd-wide@iou-cart-delta\endcsname{+3.9}
\expandafter\def\csname res@pvit-d12-scannet200-sd-wide@iou-storage container-delta\endcsname{+0.0}
\expandafter\def\csname res@pvit-d12-scannet200-sd-wide@iou-scale-delta\endcsname{+1.9}
\expandafter\def\csname res@pvit-d12-scannet200-sd-wide@iou-tissue box-delta\endcsname{\ensuremath{-}4.3}
\expandafter\def\csname res@pvit-d12-scannet200-sd-wide@iou-light switch-delta\endcsname{+0.0}
\expandafter\def\csname res@pvit-d12-scannet200-sd-wide@iou-crate-delta\endcsname{+0.0}
\expandafter\def\csname res@pvit-d12-scannet200-sd-wide@iou-power outlet-delta\endcsname{+0.0}
\expandafter\def\csname res@pvit-d12-scannet200-sd-wide@iou-decoration-delta\endcsname{+6.5}
\expandafter\def\csname res@pvit-d12-scannet200-sd-wide@iou-sign-delta\endcsname{\ensuremath{-}9.3}
\expandafter\def\csname res@pvit-d12-scannet200-sd-wide@iou-projector-delta\endcsname{+0.0}
\expandafter\def\csname res@pvit-d12-scannet200-sd-wide@iou-closet door-delta\endcsname{+10.3}
\expandafter\def\csname res@pvit-d12-scannet200-sd-wide@iou-vacuum cleaner-delta\endcsname{+2.7}
\expandafter\def\csname res@pvit-d12-scannet200-sd-wide@iou-candle-delta\endcsname{+0.0}
\expandafter\def\csname res@pvit-d12-scannet200-sd-wide@iou-plunger-delta\endcsname{+5.1}
\expandafter\def\csname res@pvit-d12-scannet200-sd-wide@iou-stuffed animal-delta\endcsname{+0.0}
\expandafter\def\csname res@pvit-d12-scannet200-sd-wide@iou-headphones-delta\endcsname{+0.0}
\expandafter\def\csname res@pvit-d12-scannet200-sd-wide@iou-dish rack-delta\endcsname{\ensuremath{-}26.6}
\expandafter\def\csname res@pvit-d12-scannet200-sd-wide@iou-broom-delta\endcsname{+8.6}
\expandafter\def\csname res@pvit-d12-scannet200-sd-wide@iou-guitar case-delta\endcsname{+0.0}
\expandafter\def\csname res@pvit-d12-scannet200-sd-wide@iou-range hood-delta\endcsname{+0.4}
\expandafter\def\csname res@pvit-d12-scannet200-sd-wide@iou-dustpan-delta\endcsname{\ensuremath{-}1.4}
\expandafter\def\csname res@pvit-d12-scannet200-sd-wide@iou-hair dryer-delta\endcsname{\ensuremath{-}12.9}
\expandafter\def\csname res@pvit-d12-scannet200-sd-wide@iou-water bottle-delta\endcsname{+0.0}
\expandafter\def\csname res@pvit-d12-scannet200-sd-wide@iou-handicap bar-delta\endcsname{+0.2}
\expandafter\def\csname res@pvit-d12-scannet200-sd-wide@iou-purse-delta\endcsname{+0.0}
\expandafter\def\csname res@pvit-d12-scannet200-sd-wide@iou-vent-delta\endcsname{\ensuremath{-}4.3}
\expandafter\def\csname res@pvit-d12-scannet200-sd-wide@iou-shower floor-delta\endcsname{+40.5}
\expandafter\def\csname res@pvit-d12-scannet200-sd-wide@iou-water pitcher-delta\endcsname{+0.0}
\expandafter\def\csname res@pvit-d12-scannet200-sd-wide@iou-mailbox-delta\endcsname{+8.7}
\expandafter\def\csname res@pvit-d12-scannet200-sd-wide@iou-bowl-delta\endcsname{+7.7}
\expandafter\def\csname res@pvit-d12-scannet200-sd-wide@iou-paper bag-delta\endcsname{+2.4}
\expandafter\def\csname res@pvit-d12-scannet200-sd-wide@iou-alarm clock-delta\endcsname{+0.0}
\expandafter\def\csname res@pvit-d12-scannet200-sd-wide@iou-music stand-delta\endcsname{+0.0}
\expandafter\def\csname res@pvit-d12-scannet200-sd-wide@iou-projector screen-delta\endcsname{+0.0}
\expandafter\def\csname res@pvit-d12-scannet200-sd-wide@iou-divider-delta\endcsname{\ensuremath{-}2.3}
\expandafter\def\csname res@pvit-d12-scannet200-sd-wide@iou-laundry detergent-delta\endcsname{+0.0}
\expandafter\def\csname res@pvit-d12-scannet200-sd-wide@iou-bathroom counter-delta\endcsname{+7.0}
\expandafter\def\csname res@pvit-d12-scannet200-sd-wide@iou-object-delta\endcsname{+1.2}
\expandafter\def\csname res@pvit-d12-scannet200-sd-wide@iou-bathroom vanity-delta\endcsname{\ensuremath{-}10.4}
\expandafter\def\csname res@pvit-d12-scannet200-sd-wide@iou-closet wall-delta\endcsname{+4.7}
\expandafter\def\csname res@pvit-d12-scannet200-sd-wide@iou-laundry hamper-delta\endcsname{\ensuremath{-}0.8}
\expandafter\def\csname res@pvit-d12-scannet200-sd-wide@iou-bathroom stall door-delta\endcsname{\ensuremath{-}7.0}
\expandafter\def\csname res@pvit-d12-scannet200-sd-wide@iou-ceiling light-delta\endcsname{+7.5}
\expandafter\def\csname res@pvit-d12-scannet200-sd-wide@iou-trash bin-delta\endcsname{+20.7}
\expandafter\def\csname res@pvit-d12-scannet200-sd-wide@iou-dumbbell-delta\endcsname{\ensuremath{-}25.9}
\expandafter\def\csname res@pvit-d12-scannet200-sd-wide@iou-stair rail-delta\endcsname{\ensuremath{-}9.1}
\expandafter\def\csname res@pvit-d12-scannet200-sd-wide@iou-tube-delta\endcsname{+0.2}
\expandafter\def\csname res@pvit-d12-scannet200-sd-wide@iou-bathroom cabinet-delta\endcsname{\ensuremath{-}23.9}
\expandafter\def\csname res@pvit-d12-scannet200-sd-wide@iou-cd case-delta\endcsname{+0.0}
\expandafter\def\csname res@pvit-d12-scannet200-sd-wide@iou-closet rod-delta\endcsname{+2.3}
\expandafter\def\csname res@pvit-d12-scannet200-sd-wide@iou-coffee kettle-delta\endcsname{\ensuremath{-}6.3}
\expandafter\def\csname res@pvit-d12-scannet200-sd-wide@iou-structure-delta\endcsname{+0.0}
\expandafter\def\csname res@pvit-d12-scannet200-sd-wide@iou-shower head-delta\endcsname{\ensuremath{-}3.7}
\expandafter\def\csname res@pvit-d12-scannet200-sd-wide@iou-keyboard piano-delta\endcsname{+0.0}
\expandafter\def\csname res@pvit-d12-scannet200-sd-wide@iou-case of water bottles-delta\endcsname{\ensuremath{-}34.9}
\expandafter\def\csname res@pvit-d12-scannet200-sd-wide@iou-coat rack-delta\endcsname{+23.9}
\expandafter\def\csname res@pvit-d12-scannet200-sd-wide@iou-storage organizer-delta\endcsname{+0.0}
\expandafter\def\csname res@pvit-d12-scannet200-sd-wide@iou-folded chair-delta\endcsname{+0.0}
\expandafter\def\csname res@pvit-d12-scannet200-sd-wide@iou-fire alarm-delta\endcsname{+7.8}
\expandafter\def\csname res@pvit-d12-scannet200-sd-wide@iou-power strip-delta\endcsname{+0.0}
\expandafter\def\csname res@pvit-d12-scannet200-sd-wide@iou-calendar-delta\endcsname{+0.0}
\expandafter\def\csname res@pvit-d12-scannet200-sd-wide@iou-poster-delta\endcsname{+0.0}
\expandafter\def\csname res@pvit-d12-scannet200-sd-wide@iou-potted plant-delta\endcsname{+0.0}
\expandafter\def\csname res@pvit-d12-scannet200-sd-wide@iou-luggage-delta\endcsname{+0.0}
\expandafter\def\csname res@pvit-d12-scannet200-sd-wide@iou-mattress-delta\endcsname{\ensuremath{-}1.5}
\expandafter\def\csname res@pvit-d12-scannet200-wide@n\endcsname{1}
\expandafter\def\csname res@pvit-d12-scannet200-wide@miou\endcsname{36.1}
\expandafter\def\csname res@pvit-d12-scannet200-wide@miou-pm\endcsname{36.1}
\expandafter\def\csname res@pvit-d12-scannet200-wide@macc\endcsname{46.5}
\expandafter\def\csname res@pvit-d12-scannet200-wide@macc-pm\endcsname{46.5}
\expandafter\def\csname res@pvit-d12-scannet200-wide@allacc\endcsname{83.7}
\expandafter\def\csname res@pvit-d12-scannet200-wide@allacc-pm\endcsname{83.7}
\expandafter\def\csname res@pvit-d12-scannet200-wide@iou-wall\endcsname{81.7}
\expandafter\def\csname res@pvit-d12-scannet200-wide@iou-wall-pm\endcsname{81.7}
\expandafter\def\csname res@pvit-d12-scannet200-wide@iou-chair\endcsname{75.8}
\expandafter\def\csname res@pvit-d12-scannet200-wide@iou-chair-pm\endcsname{75.8}
\expandafter\def\csname res@pvit-d12-scannet200-wide@iou-floor\endcsname{95.2}
\expandafter\def\csname res@pvit-d12-scannet200-wide@iou-floor-pm\endcsname{95.2}
\expandafter\def\csname res@pvit-d12-scannet200-wide@iou-table\endcsname{69.9}
\expandafter\def\csname res@pvit-d12-scannet200-wide@iou-table-pm\endcsname{69.9}
\expandafter\def\csname res@pvit-d12-scannet200-wide@iou-door\endcsname{66.7}
\expandafter\def\csname res@pvit-d12-scannet200-wide@iou-door-pm\endcsname{66.7}
\expandafter\def\csname res@pvit-d12-scannet200-wide@iou-couch\endcsname{81.8}
\expandafter\def\csname res@pvit-d12-scannet200-wide@iou-couch-pm\endcsname{81.8}
\expandafter\def\csname res@pvit-d12-scannet200-wide@iou-cabinet\endcsname{42.4}
\expandafter\def\csname res@pvit-d12-scannet200-wide@iou-cabinet-pm\endcsname{42.4}
\expandafter\def\csname res@pvit-d12-scannet200-wide@iou-shelf\endcsname{42.8}
\expandafter\def\csname res@pvit-d12-scannet200-wide@iou-shelf-pm\endcsname{42.8}
\expandafter\def\csname res@pvit-d12-scannet200-wide@iou-desk\endcsname{66.9}
\expandafter\def\csname res@pvit-d12-scannet200-wide@iou-desk-pm\endcsname{66.9}
\expandafter\def\csname res@pvit-d12-scannet200-wide@iou-office chair\endcsname{30.7}
\expandafter\def\csname res@pvit-d12-scannet200-wide@iou-office chair-pm\endcsname{30.7}
\expandafter\def\csname res@pvit-d12-scannet200-wide@iou-bed\endcsname{81.8}
\expandafter\def\csname res@pvit-d12-scannet200-wide@iou-bed-pm\endcsname{81.8}
\expandafter\def\csname res@pvit-d12-scannet200-wide@iou-pillow\endcsname{59.3}
\expandafter\def\csname res@pvit-d12-scannet200-wide@iou-pillow-pm\endcsname{59.3}
\expandafter\def\csname res@pvit-d12-scannet200-wide@iou-sink\endcsname{70.5}
\expandafter\def\csname res@pvit-d12-scannet200-wide@iou-sink-pm\endcsname{70.5}
\expandafter\def\csname res@pvit-d12-scannet200-wide@iou-picture\endcsname{33.5}
\expandafter\def\csname res@pvit-d12-scannet200-wide@iou-picture-pm\endcsname{33.5}
\expandafter\def\csname res@pvit-d12-scannet200-wide@iou-window\endcsname{71.0}
\expandafter\def\csname res@pvit-d12-scannet200-wide@iou-window-pm\endcsname{71.0}
\expandafter\def\csname res@pvit-d12-scannet200-wide@iou-toilet\endcsname{94.0}
\expandafter\def\csname res@pvit-d12-scannet200-wide@iou-toilet-pm\endcsname{94.0}
\expandafter\def\csname res@pvit-d12-scannet200-wide@iou-bookshelf\endcsname{48.2}
\expandafter\def\csname res@pvit-d12-scannet200-wide@iou-bookshelf-pm\endcsname{48.2}
\expandafter\def\csname res@pvit-d12-scannet200-wide@iou-monitor\endcsname{90.3}
\expandafter\def\csname res@pvit-d12-scannet200-wide@iou-monitor-pm\endcsname{90.3}
\expandafter\def\csname res@pvit-d12-scannet200-wide@iou-curtain\endcsname{73.5}
\expandafter\def\csname res@pvit-d12-scannet200-wide@iou-curtain-pm\endcsname{73.5}
\expandafter\def\csname res@pvit-d12-scannet200-wide@iou-book\endcsname{42.8}
\expandafter\def\csname res@pvit-d12-scannet200-wide@iou-book-pm\endcsname{42.8}
\expandafter\def\csname res@pvit-d12-scannet200-wide@iou-armchair\endcsname{55.1}
\expandafter\def\csname res@pvit-d12-scannet200-wide@iou-armchair-pm\endcsname{55.1}
\expandafter\def\csname res@pvit-d12-scannet200-wide@iou-coffee table\endcsname{55.8}
\expandafter\def\csname res@pvit-d12-scannet200-wide@iou-coffee table-pm\endcsname{55.8}
\expandafter\def\csname res@pvit-d12-scannet200-wide@iou-box\endcsname{42.4}
\expandafter\def\csname res@pvit-d12-scannet200-wide@iou-box-pm\endcsname{42.4}
\expandafter\def\csname res@pvit-d12-scannet200-wide@iou-refrigerator\endcsname{80.4}
\expandafter\def\csname res@pvit-d12-scannet200-wide@iou-refrigerator-pm\endcsname{80.4}
\expandafter\def\csname res@pvit-d12-scannet200-wide@iou-lamp\endcsname{72.9}
\expandafter\def\csname res@pvit-d12-scannet200-wide@iou-lamp-pm\endcsname{72.9}
\expandafter\def\csname res@pvit-d12-scannet200-wide@iou-kitchen cabinet\endcsname{69.9}
\expandafter\def\csname res@pvit-d12-scannet200-wide@iou-kitchen cabinet-pm\endcsname{69.9}
\expandafter\def\csname res@pvit-d12-scannet200-wide@iou-towel\endcsname{61.4}
\expandafter\def\csname res@pvit-d12-scannet200-wide@iou-towel-pm\endcsname{61.4}
\expandafter\def\csname res@pvit-d12-scannet200-wide@iou-clothes\endcsname{22.6}
\expandafter\def\csname res@pvit-d12-scannet200-wide@iou-clothes-pm\endcsname{22.6}
\expandafter\def\csname res@pvit-d12-scannet200-wide@iou-tv\endcsname{92.8}
\expandafter\def\csname res@pvit-d12-scannet200-wide@iou-tv-pm\endcsname{92.8}
\expandafter\def\csname res@pvit-d12-scannet200-wide@iou-nightstand\endcsname{70.0}
\expandafter\def\csname res@pvit-d12-scannet200-wide@iou-nightstand-pm\endcsname{70.0}
\expandafter\def\csname res@pvit-d12-scannet200-wide@iou-counter\endcsname{38.1}
\expandafter\def\csname res@pvit-d12-scannet200-wide@iou-counter-pm\endcsname{38.1}
\expandafter\def\csname res@pvit-d12-scannet200-wide@iou-dresser\endcsname{54.2}
\expandafter\def\csname res@pvit-d12-scannet200-wide@iou-dresser-pm\endcsname{54.2}
\expandafter\def\csname res@pvit-d12-scannet200-wide@iou-stool\endcsname{71.5}
\expandafter\def\csname res@pvit-d12-scannet200-wide@iou-stool-pm\endcsname{71.5}
\expandafter\def\csname res@pvit-d12-scannet200-wide@iou-cushion\endcsname{0.0}
\expandafter\def\csname res@pvit-d12-scannet200-wide@iou-cushion-pm\endcsname{0.0}
\expandafter\def\csname res@pvit-d12-scannet200-wide@iou-plant\endcsname{72.5}
\expandafter\def\csname res@pvit-d12-scannet200-wide@iou-plant-pm\endcsname{72.5}
\expandafter\def\csname res@pvit-d12-scannet200-wide@iou-ceiling\endcsname{92.6}
\expandafter\def\csname res@pvit-d12-scannet200-wide@iou-ceiling-pm\endcsname{92.6}
\expandafter\def\csname res@pvit-d12-scannet200-wide@iou-bathtub\endcsname{82.8}
\expandafter\def\csname res@pvit-d12-scannet200-wide@iou-bathtub-pm\endcsname{82.8}
\expandafter\def\csname res@pvit-d12-scannet200-wide@iou-end table\endcsname{34.5}
\expandafter\def\csname res@pvit-d12-scannet200-wide@iou-end table-pm\endcsname{34.5}
\expandafter\def\csname res@pvit-d12-scannet200-wide@iou-dining table\endcsname{8.7}
\expandafter\def\csname res@pvit-d12-scannet200-wide@iou-dining table-pm\endcsname{8.7}
\expandafter\def\csname res@pvit-d12-scannet200-wide@iou-keyboard\endcsname{41.7}
\expandafter\def\csname res@pvit-d12-scannet200-wide@iou-keyboard-pm\endcsname{41.7}
\expandafter\def\csname res@pvit-d12-scannet200-wide@iou-bag\endcsname{22.4}
\expandafter\def\csname res@pvit-d12-scannet200-wide@iou-bag-pm\endcsname{22.4}
\expandafter\def\csname res@pvit-d12-scannet200-wide@iou-backpack\endcsname{62.2}
\expandafter\def\csname res@pvit-d12-scannet200-wide@iou-backpack-pm\endcsname{62.2}
\expandafter\def\csname res@pvit-d12-scannet200-wide@iou-toilet paper\endcsname{50.8}
\expandafter\def\csname res@pvit-d12-scannet200-wide@iou-toilet paper-pm\endcsname{50.8}
\expandafter\def\csname res@pvit-d12-scannet200-wide@iou-printer\endcsname{53.5}
\expandafter\def\csname res@pvit-d12-scannet200-wide@iou-printer-pm\endcsname{53.5}
\expandafter\def\csname res@pvit-d12-scannet200-wide@iou-tv stand\endcsname{60.0}
\expandafter\def\csname res@pvit-d12-scannet200-wide@iou-tv stand-pm\endcsname{60.0}
\expandafter\def\csname res@pvit-d12-scannet200-wide@iou-whiteboard\endcsname{54.2}
\expandafter\def\csname res@pvit-d12-scannet200-wide@iou-whiteboard-pm\endcsname{54.2}
\expandafter\def\csname res@pvit-d12-scannet200-wide@iou-blanket\endcsname{7.7}
\expandafter\def\csname res@pvit-d12-scannet200-wide@iou-blanket-pm\endcsname{7.7}
\expandafter\def\csname res@pvit-d12-scannet200-wide@iou-shower curtain\endcsname{64.9}
\expandafter\def\csname res@pvit-d12-scannet200-wide@iou-shower curtain-pm\endcsname{64.9}
\expandafter\def\csname res@pvit-d12-scannet200-wide@iou-trash can\endcsname{65.8}
\expandafter\def\csname res@pvit-d12-scannet200-wide@iou-trash can-pm\endcsname{65.8}
\expandafter\def\csname res@pvit-d12-scannet200-wide@iou-closet\endcsname{10.7}
\expandafter\def\csname res@pvit-d12-scannet200-wide@iou-closet-pm\endcsname{10.7}
\expandafter\def\csname res@pvit-d12-scannet200-wide@iou-stairs\endcsname{57.0}
\expandafter\def\csname res@pvit-d12-scannet200-wide@iou-stairs-pm\endcsname{57.0}
\expandafter\def\csname res@pvit-d12-scannet200-wide@iou-microwave\endcsname{55.6}
\expandafter\def\csname res@pvit-d12-scannet200-wide@iou-microwave-pm\endcsname{55.6}
\expandafter\def\csname res@pvit-d12-scannet200-wide@iou-stove\endcsname{66.6}
\expandafter\def\csname res@pvit-d12-scannet200-wide@iou-stove-pm\endcsname{66.6}
\expandafter\def\csname res@pvit-d12-scannet200-wide@iou-shoe\endcsname{55.9}
\expandafter\def\csname res@pvit-d12-scannet200-wide@iou-shoe-pm\endcsname{55.9}
\expandafter\def\csname res@pvit-d12-scannet200-wide@iou-computer tower\endcsname{59.3}
\expandafter\def\csname res@pvit-d12-scannet200-wide@iou-computer tower-pm\endcsname{59.3}
\expandafter\def\csname res@pvit-d12-scannet200-wide@iou-bottle\endcsname{16.3}
\expandafter\def\csname res@pvit-d12-scannet200-wide@iou-bottle-pm\endcsname{16.3}
\expandafter\def\csname res@pvit-d12-scannet200-wide@iou-bin\endcsname{0.0}
\expandafter\def\csname res@pvit-d12-scannet200-wide@iou-bin-pm\endcsname{0.0}
\expandafter\def\csname res@pvit-d12-scannet200-wide@iou-ottoman\endcsname{40.1}
\expandafter\def\csname res@pvit-d12-scannet200-wide@iou-ottoman-pm\endcsname{40.1}
\expandafter\def\csname res@pvit-d12-scannet200-wide@iou-bench\endcsname{31.2}
\expandafter\def\csname res@pvit-d12-scannet200-wide@iou-bench-pm\endcsname{31.2}
\expandafter\def\csname res@pvit-d12-scannet200-wide@iou-board\endcsname{20.0}
\expandafter\def\csname res@pvit-d12-scannet200-wide@iou-board-pm\endcsname{20.0}
\expandafter\def\csname res@pvit-d12-scannet200-wide@iou-washing machine\endcsname{86.7}
\expandafter\def\csname res@pvit-d12-scannet200-wide@iou-washing machine-pm\endcsname{86.7}
\expandafter\def\csname res@pvit-d12-scannet200-wide@iou-mirror\endcsname{53.3}
\expandafter\def\csname res@pvit-d12-scannet200-wide@iou-mirror-pm\endcsname{53.3}
\expandafter\def\csname res@pvit-d12-scannet200-wide@iou-copier\endcsname{90.7}
\expandafter\def\csname res@pvit-d12-scannet200-wide@iou-copier-pm\endcsname{90.7}
\expandafter\def\csname res@pvit-d12-scannet200-wide@iou-basket\endcsname{0.0}
\expandafter\def\csname res@pvit-d12-scannet200-wide@iou-basket-pm\endcsname{0.0}
\expandafter\def\csname res@pvit-d12-scannet200-wide@iou-sofa chair\endcsname{19.5}
\expandafter\def\csname res@pvit-d12-scannet200-wide@iou-sofa chair-pm\endcsname{19.5}
\expandafter\def\csname res@pvit-d12-scannet200-wide@iou-file cabinet\endcsname{65.4}
\expandafter\def\csname res@pvit-d12-scannet200-wide@iou-file cabinet-pm\endcsname{65.4}
\expandafter\def\csname res@pvit-d12-scannet200-wide@iou-fan\endcsname{62.4}
\expandafter\def\csname res@pvit-d12-scannet200-wide@iou-fan-pm\endcsname{62.4}
\expandafter\def\csname res@pvit-d12-scannet200-wide@iou-laptop\endcsname{51.0}
\expandafter\def\csname res@pvit-d12-scannet200-wide@iou-laptop-pm\endcsname{51.0}
\expandafter\def\csname res@pvit-d12-scannet200-wide@iou-shower\endcsname{26.8}
\expandafter\def\csname res@pvit-d12-scannet200-wide@iou-shower-pm\endcsname{26.8}
\expandafter\def\csname res@pvit-d12-scannet200-wide@iou-paper\endcsname{3.9}
\expandafter\def\csname res@pvit-d12-scannet200-wide@iou-paper-pm\endcsname{3.9}
\expandafter\def\csname res@pvit-d12-scannet200-wide@iou-person\endcsname{22.9}
\expandafter\def\csname res@pvit-d12-scannet200-wide@iou-person-pm\endcsname{22.9}
\expandafter\def\csname res@pvit-d12-scannet200-wide@iou-paper towel dispenser\endcsname{58.0}
\expandafter\def\csname res@pvit-d12-scannet200-wide@iou-paper towel dispenser-pm\endcsname{58.0}
\expandafter\def\csname res@pvit-d12-scannet200-wide@iou-oven\endcsname{9.2}
\expandafter\def\csname res@pvit-d12-scannet200-wide@iou-oven-pm\endcsname{9.2}
\expandafter\def\csname res@pvit-d12-scannet200-wide@iou-blinds\endcsname{0.0}
\expandafter\def\csname res@pvit-d12-scannet200-wide@iou-blinds-pm\endcsname{0.0}
\expandafter\def\csname res@pvit-d12-scannet200-wide@iou-rack\endcsname{0.0}
\expandafter\def\csname res@pvit-d12-scannet200-wide@iou-rack-pm\endcsname{0.0}
\expandafter\def\csname res@pvit-d12-scannet200-wide@iou-plate\endcsname{0.0}
\expandafter\def\csname res@pvit-d12-scannet200-wide@iou-plate-pm\endcsname{0.0}
\expandafter\def\csname res@pvit-d12-scannet200-wide@iou-blackboard\endcsname{89.6}
\expandafter\def\csname res@pvit-d12-scannet200-wide@iou-blackboard-pm\endcsname{89.6}
\expandafter\def\csname res@pvit-d12-scannet200-wide@iou-piano\endcsname{89.9}
\expandafter\def\csname res@pvit-d12-scannet200-wide@iou-piano-pm\endcsname{89.9}
\expandafter\def\csname res@pvit-d12-scannet200-wide@iou-suitcase\endcsname{60.1}
\expandafter\def\csname res@pvit-d12-scannet200-wide@iou-suitcase-pm\endcsname{60.1}
\expandafter\def\csname res@pvit-d12-scannet200-wide@iou-rail\endcsname{3.6}
\expandafter\def\csname res@pvit-d12-scannet200-wide@iou-rail-pm\endcsname{3.6}
\expandafter\def\csname res@pvit-d12-scannet200-wide@iou-radiator\endcsname{66.9}
\expandafter\def\csname res@pvit-d12-scannet200-wide@iou-radiator-pm\endcsname{66.9}
\expandafter\def\csname res@pvit-d12-scannet200-wide@iou-recycling bin\endcsname{54.8}
\expandafter\def\csname res@pvit-d12-scannet200-wide@iou-recycling bin-pm\endcsname{54.8}
\expandafter\def\csname res@pvit-d12-scannet200-wide@iou-container\endcsname{3.0}
\expandafter\def\csname res@pvit-d12-scannet200-wide@iou-container-pm\endcsname{3.0}
\expandafter\def\csname res@pvit-d12-scannet200-wide@iou-wardrobe\endcsname{43.8}
\expandafter\def\csname res@pvit-d12-scannet200-wide@iou-wardrobe-pm\endcsname{43.8}
\expandafter\def\csname res@pvit-d12-scannet200-wide@iou-soap dispenser\endcsname{71.3}
\expandafter\def\csname res@pvit-d12-scannet200-wide@iou-soap dispenser-pm\endcsname{71.3}
\expandafter\def\csname res@pvit-d12-scannet200-wide@iou-telephone\endcsname{36.8}
\expandafter\def\csname res@pvit-d12-scannet200-wide@iou-telephone-pm\endcsname{36.8}
\expandafter\def\csname res@pvit-d12-scannet200-wide@iou-bucket\endcsname{33.1}
\expandafter\def\csname res@pvit-d12-scannet200-wide@iou-bucket-pm\endcsname{33.1}
\expandafter\def\csname res@pvit-d12-scannet200-wide@iou-clock\endcsname{24.6}
\expandafter\def\csname res@pvit-d12-scannet200-wide@iou-clock-pm\endcsname{24.6}
\expandafter\def\csname res@pvit-d12-scannet200-wide@iou-stand\endcsname{0.0}
\expandafter\def\csname res@pvit-d12-scannet200-wide@iou-stand-pm\endcsname{0.0}
\expandafter\def\csname res@pvit-d12-scannet200-wide@iou-light\endcsname{7.5}
\expandafter\def\csname res@pvit-d12-scannet200-wide@iou-light-pm\endcsname{7.5}
\expandafter\def\csname res@pvit-d12-scannet200-wide@iou-laundry basket\endcsname{9.0}
\expandafter\def\csname res@pvit-d12-scannet200-wide@iou-laundry basket-pm\endcsname{9.0}
\expandafter\def\csname res@pvit-d12-scannet200-wide@iou-pipe\endcsname{8.1}
\expandafter\def\csname res@pvit-d12-scannet200-wide@iou-pipe-pm\endcsname{8.1}
\expandafter\def\csname res@pvit-d12-scannet200-wide@iou-clothes dryer\endcsname{82.3}
\expandafter\def\csname res@pvit-d12-scannet200-wide@iou-clothes dryer-pm\endcsname{82.3}
\expandafter\def\csname res@pvit-d12-scannet200-wide@iou-guitar\endcsname{83.4}
\expandafter\def\csname res@pvit-d12-scannet200-wide@iou-guitar-pm\endcsname{83.4}
\expandafter\def\csname res@pvit-d12-scannet200-wide@iou-toilet paper holder\endcsname{0.0}
\expandafter\def\csname res@pvit-d12-scannet200-wide@iou-toilet paper holder-pm\endcsname{0.0}
\expandafter\def\csname res@pvit-d12-scannet200-wide@iou-seat\endcsname{0.0}
\expandafter\def\csname res@pvit-d12-scannet200-wide@iou-seat-pm\endcsname{0.0}
\expandafter\def\csname res@pvit-d12-scannet200-wide@iou-speaker\endcsname{0.0}
\expandafter\def\csname res@pvit-d12-scannet200-wide@iou-speaker-pm\endcsname{0.0}
\expandafter\def\csname res@pvit-d12-scannet200-wide@iou-column\endcsname{0.0}
\expandafter\def\csname res@pvit-d12-scannet200-wide@iou-column-pm\endcsname{0.0}
\expandafter\def\csname res@pvit-d12-scannet200-wide@iou-bicycle\endcsname{0.0}
\expandafter\def\csname res@pvit-d12-scannet200-wide@iou-bicycle-pm\endcsname{0.0}
\expandafter\def\csname res@pvit-d12-scannet200-wide@iou-ladder\endcsname{50.2}
\expandafter\def\csname res@pvit-d12-scannet200-wide@iou-ladder-pm\endcsname{50.2}
\expandafter\def\csname res@pvit-d12-scannet200-wide@iou-bathroom stall\endcsname{61.2}
\expandafter\def\csname res@pvit-d12-scannet200-wide@iou-bathroom stall-pm\endcsname{61.2}
\expandafter\def\csname res@pvit-d12-scannet200-wide@iou-shower wall\endcsname{47.9}
\expandafter\def\csname res@pvit-d12-scannet200-wide@iou-shower wall-pm\endcsname{47.9}
\expandafter\def\csname res@pvit-d12-scannet200-wide@iou-cup\endcsname{25.1}
\expandafter\def\csname res@pvit-d12-scannet200-wide@iou-cup-pm\endcsname{25.1}
\expandafter\def\csname res@pvit-d12-scannet200-wide@iou-jacket\endcsname{19.4}
\expandafter\def\csname res@pvit-d12-scannet200-wide@iou-jacket-pm\endcsname{19.4}
\expandafter\def\csname res@pvit-d12-scannet200-wide@iou-storage bin\endcsname{3.3}
\expandafter\def\csname res@pvit-d12-scannet200-wide@iou-storage bin-pm\endcsname{3.3}
\expandafter\def\csname res@pvit-d12-scannet200-wide@iou-coffee maker\endcsname{91.4}
\expandafter\def\csname res@pvit-d12-scannet200-wide@iou-coffee maker-pm\endcsname{91.4}
\expandafter\def\csname res@pvit-d12-scannet200-wide@iou-dishwasher\endcsname{47.4}
\expandafter\def\csname res@pvit-d12-scannet200-wide@iou-dishwasher-pm\endcsname{47.4}
\expandafter\def\csname res@pvit-d12-scannet200-wide@iou-paper towel roll\endcsname{36.3}
\expandafter\def\csname res@pvit-d12-scannet200-wide@iou-paper towel roll-pm\endcsname{36.3}
\expandafter\def\csname res@pvit-d12-scannet200-wide@iou-machine\endcsname{0.0}
\expandafter\def\csname res@pvit-d12-scannet200-wide@iou-machine-pm\endcsname{0.0}
\expandafter\def\csname res@pvit-d12-scannet200-wide@iou-mat\endcsname{37.3}
\expandafter\def\csname res@pvit-d12-scannet200-wide@iou-mat-pm\endcsname{37.3}
\expandafter\def\csname res@pvit-d12-scannet200-wide@iou-windowsill\endcsname{18.9}
\expandafter\def\csname res@pvit-d12-scannet200-wide@iou-windowsill-pm\endcsname{18.9}
\expandafter\def\csname res@pvit-d12-scannet200-wide@iou-bar\endcsname{15.1}
\expandafter\def\csname res@pvit-d12-scannet200-wide@iou-bar-pm\endcsname{15.1}
\expandafter\def\csname res@pvit-d12-scannet200-wide@iou-toaster\endcsname{63.9}
\expandafter\def\csname res@pvit-d12-scannet200-wide@iou-toaster-pm\endcsname{63.9}
\expandafter\def\csname res@pvit-d12-scannet200-wide@iou-bulletin board\endcsname{47.9}
\expandafter\def\csname res@pvit-d12-scannet200-wide@iou-bulletin board-pm\endcsname{47.9}
\expandafter\def\csname res@pvit-d12-scannet200-wide@iou-ironing board\endcsname{46.6}
\expandafter\def\csname res@pvit-d12-scannet200-wide@iou-ironing board-pm\endcsname{46.6}
\expandafter\def\csname res@pvit-d12-scannet200-wide@iou-fireplace\endcsname{65.8}
\expandafter\def\csname res@pvit-d12-scannet200-wide@iou-fireplace-pm\endcsname{65.8}
\expandafter\def\csname res@pvit-d12-scannet200-wide@iou-soap dish\endcsname{32.8}
\expandafter\def\csname res@pvit-d12-scannet200-wide@iou-soap dish-pm\endcsname{32.8}
\expandafter\def\csname res@pvit-d12-scannet200-wide@iou-kitchen counter\endcsname{56.4}
\expandafter\def\csname res@pvit-d12-scannet200-wide@iou-kitchen counter-pm\endcsname{56.4}
\expandafter\def\csname res@pvit-d12-scannet200-wide@iou-doorframe\endcsname{45.3}
\expandafter\def\csname res@pvit-d12-scannet200-wide@iou-doorframe-pm\endcsname{45.3}
\expandafter\def\csname res@pvit-d12-scannet200-wide@iou-toilet paper dispenser\endcsname{57.2}
\expandafter\def\csname res@pvit-d12-scannet200-wide@iou-toilet paper dispenser-pm\endcsname{57.2}
\expandafter\def\csname res@pvit-d12-scannet200-wide@iou-mini fridge\endcsname{56.2}
\expandafter\def\csname res@pvit-d12-scannet200-wide@iou-mini fridge-pm\endcsname{56.2}
\expandafter\def\csname res@pvit-d12-scannet200-wide@iou-fire extinguisher\endcsname{32.6}
\expandafter\def\csname res@pvit-d12-scannet200-wide@iou-fire extinguisher-pm\endcsname{32.6}
\expandafter\def\csname res@pvit-d12-scannet200-wide@iou-ball\endcsname{79.7}
\expandafter\def\csname res@pvit-d12-scannet200-wide@iou-ball-pm\endcsname{79.7}
\expandafter\def\csname res@pvit-d12-scannet200-wide@iou-hat\endcsname{0.0}
\expandafter\def\csname res@pvit-d12-scannet200-wide@iou-hat-pm\endcsname{0.0}
\expandafter\def\csname res@pvit-d12-scannet200-wide@iou-shower curtain rod\endcsname{49.4}
\expandafter\def\csname res@pvit-d12-scannet200-wide@iou-shower curtain rod-pm\endcsname{49.4}
\expandafter\def\csname res@pvit-d12-scannet200-wide@iou-water cooler\endcsname{88.1}
\expandafter\def\csname res@pvit-d12-scannet200-wide@iou-water cooler-pm\endcsname{88.1}
\expandafter\def\csname res@pvit-d12-scannet200-wide@iou-paper cutter\endcsname{22.4}
\expandafter\def\csname res@pvit-d12-scannet200-wide@iou-paper cutter-pm\endcsname{22.4}
\expandafter\def\csname res@pvit-d12-scannet200-wide@iou-tray\endcsname{7.6}
\expandafter\def\csname res@pvit-d12-scannet200-wide@iou-tray-pm\endcsname{7.6}
\expandafter\def\csname res@pvit-d12-scannet200-wide@iou-shower door\endcsname{26.6}
\expandafter\def\csname res@pvit-d12-scannet200-wide@iou-shower door-pm\endcsname{26.6}
\expandafter\def\csname res@pvit-d12-scannet200-wide@iou-pillar\endcsname{6.4}
\expandafter\def\csname res@pvit-d12-scannet200-wide@iou-pillar-pm\endcsname{6.4}
\expandafter\def\csname res@pvit-d12-scannet200-wide@iou-ledge\endcsname{9.3}
\expandafter\def\csname res@pvit-d12-scannet200-wide@iou-ledge-pm\endcsname{9.3}
\expandafter\def\csname res@pvit-d12-scannet200-wide@iou-toaster oven\endcsname{1.0}
\expandafter\def\csname res@pvit-d12-scannet200-wide@iou-toaster oven-pm\endcsname{1.0}
\expandafter\def\csname res@pvit-d12-scannet200-wide@iou-mouse\endcsname{5.6}
\expandafter\def\csname res@pvit-d12-scannet200-wide@iou-mouse-pm\endcsname{5.6}
\expandafter\def\csname res@pvit-d12-scannet200-wide@iou-toilet seat cover dispenser\endcsname{82.4}
\expandafter\def\csname res@pvit-d12-scannet200-wide@iou-toilet seat cover dispenser-pm\endcsname{82.4}
\expandafter\def\csname res@pvit-d12-scannet200-wide@iou-furniture\endcsname{0.0}
\expandafter\def\csname res@pvit-d12-scannet200-wide@iou-furniture-pm\endcsname{0.0}
\expandafter\def\csname res@pvit-d12-scannet200-wide@iou-cart\endcsname{18.1}
\expandafter\def\csname res@pvit-d12-scannet200-wide@iou-cart-pm\endcsname{18.1}
\expandafter\def\csname res@pvit-d12-scannet200-wide@iou-storage container\endcsname{0.0}
\expandafter\def\csname res@pvit-d12-scannet200-wide@iou-storage container-pm\endcsname{0.0}
\expandafter\def\csname res@pvit-d12-scannet200-wide@iou-scale\endcsname{38.1}
\expandafter\def\csname res@pvit-d12-scannet200-wide@iou-scale-pm\endcsname{38.1}
\expandafter\def\csname res@pvit-d12-scannet200-wide@iou-tissue box\endcsname{19.5}
\expandafter\def\csname res@pvit-d12-scannet200-wide@iou-tissue box-pm\endcsname{19.5}
\expandafter\def\csname res@pvit-d12-scannet200-wide@iou-light switch\endcsname{0.0}
\expandafter\def\csname res@pvit-d12-scannet200-wide@iou-light switch-pm\endcsname{0.0}
\expandafter\def\csname res@pvit-d12-scannet200-wide@iou-crate\endcsname{0.0}
\expandafter\def\csname res@pvit-d12-scannet200-wide@iou-crate-pm\endcsname{0.0}
\expandafter\def\csname res@pvit-d12-scannet200-wide@iou-power outlet\endcsname{0.0}
\expandafter\def\csname res@pvit-d12-scannet200-wide@iou-power outlet-pm\endcsname{0.0}
\expandafter\def\csname res@pvit-d12-scannet200-wide@iou-decoration\endcsname{42.1}
\expandafter\def\csname res@pvit-d12-scannet200-wide@iou-decoration-pm\endcsname{42.1}
\expandafter\def\csname res@pvit-d12-scannet200-wide@iou-sign\endcsname{37.3}
\expandafter\def\csname res@pvit-d12-scannet200-wide@iou-sign-pm\endcsname{37.3}
\expandafter\def\csname res@pvit-d12-scannet200-wide@iou-projector\endcsname{0.0}
\expandafter\def\csname res@pvit-d12-scannet200-wide@iou-projector-pm\endcsname{0.0}
\expandafter\def\csname res@pvit-d12-scannet200-wide@iou-closet door\endcsname{8.2}
\expandafter\def\csname res@pvit-d12-scannet200-wide@iou-closet door-pm\endcsname{8.2}
\expandafter\def\csname res@pvit-d12-scannet200-wide@iou-vacuum cleaner\endcsname{84.8}
\expandafter\def\csname res@pvit-d12-scannet200-wide@iou-vacuum cleaner-pm\endcsname{84.8}
\expandafter\def\csname res@pvit-d12-scannet200-wide@iou-candle\endcsname{0.0}
\expandafter\def\csname res@pvit-d12-scannet200-wide@iou-candle-pm\endcsname{0.0}
\expandafter\def\csname res@pvit-d12-scannet200-wide@iou-plunger\endcsname{28.8}
\expandafter\def\csname res@pvit-d12-scannet200-wide@iou-plunger-pm\endcsname{28.8}
\expandafter\def\csname res@pvit-d12-scannet200-wide@iou-stuffed animal\endcsname{0.0}
\expandafter\def\csname res@pvit-d12-scannet200-wide@iou-stuffed animal-pm\endcsname{0.0}
\expandafter\def\csname res@pvit-d12-scannet200-wide@iou-headphones\endcsname{0.0}
\expandafter\def\csname res@pvit-d12-scannet200-wide@iou-headphones-pm\endcsname{0.0}
\expandafter\def\csname res@pvit-d12-scannet200-wide@iou-dish rack\endcsname{91.3}
\expandafter\def\csname res@pvit-d12-scannet200-wide@iou-dish rack-pm\endcsname{91.3}
\expandafter\def\csname res@pvit-d12-scannet200-wide@iou-broom\endcsname{2.1}
\expandafter\def\csname res@pvit-d12-scannet200-wide@iou-broom-pm\endcsname{2.1}
\expandafter\def\csname res@pvit-d12-scannet200-wide@iou-guitar case\endcsname{0.0}
\expandafter\def\csname res@pvit-d12-scannet200-wide@iou-guitar case-pm\endcsname{0.0}
\expandafter\def\csname res@pvit-d12-scannet200-wide@iou-range hood\endcsname{74.1}
\expandafter\def\csname res@pvit-d12-scannet200-wide@iou-range hood-pm\endcsname{74.1}
\expandafter\def\csname res@pvit-d12-scannet200-wide@iou-dustpan\endcsname{1.7}
\expandafter\def\csname res@pvit-d12-scannet200-wide@iou-dustpan-pm\endcsname{1.7}
\expandafter\def\csname res@pvit-d12-scannet200-wide@iou-hair dryer\endcsname{23.8}
\expandafter\def\csname res@pvit-d12-scannet200-wide@iou-hair dryer-pm\endcsname{23.8}
\expandafter\def\csname res@pvit-d12-scannet200-wide@iou-water bottle\endcsname{0.0}
\expandafter\def\csname res@pvit-d12-scannet200-wide@iou-water bottle-pm\endcsname{0.0}
\expandafter\def\csname res@pvit-d12-scannet200-wide@iou-handicap bar\endcsname{6.1}
\expandafter\def\csname res@pvit-d12-scannet200-wide@iou-handicap bar-pm\endcsname{6.1}
\expandafter\def\csname res@pvit-d12-scannet200-wide@iou-purse\endcsname{0.0}
\expandafter\def\csname res@pvit-d12-scannet200-wide@iou-purse-pm\endcsname{0.0}
\expandafter\def\csname res@pvit-d12-scannet200-wide@iou-vent\endcsname{30.5}
\expandafter\def\csname res@pvit-d12-scannet200-wide@iou-vent-pm\endcsname{30.5}
\expandafter\def\csname res@pvit-d12-scannet200-wide@iou-shower floor\endcsname{38.5}
\expandafter\def\csname res@pvit-d12-scannet200-wide@iou-shower floor-pm\endcsname{38.5}
\expandafter\def\csname res@pvit-d12-scannet200-wide@iou-water pitcher\endcsname{0.0}
\expandafter\def\csname res@pvit-d12-scannet200-wide@iou-water pitcher-pm\endcsname{0.0}
\expandafter\def\csname res@pvit-d12-scannet200-wide@iou-mailbox\endcsname{48.6}
\expandafter\def\csname res@pvit-d12-scannet200-wide@iou-mailbox-pm\endcsname{48.6}
\expandafter\def\csname res@pvit-d12-scannet200-wide@iou-bowl\endcsname{7.7}
\expandafter\def\csname res@pvit-d12-scannet200-wide@iou-bowl-pm\endcsname{7.7}
\expandafter\def\csname res@pvit-d12-scannet200-wide@iou-paper bag\endcsname{6.2}
\expandafter\def\csname res@pvit-d12-scannet200-wide@iou-paper bag-pm\endcsname{6.2}
\expandafter\def\csname res@pvit-d12-scannet200-wide@iou-alarm clock\endcsname{0.0}
\expandafter\def\csname res@pvit-d12-scannet200-wide@iou-alarm clock-pm\endcsname{0.0}
\expandafter\def\csname res@pvit-d12-scannet200-wide@iou-music stand\endcsname{0.0}
\expandafter\def\csname res@pvit-d12-scannet200-wide@iou-music stand-pm\endcsname{0.0}
\expandafter\def\csname res@pvit-d12-scannet200-wide@iou-projector screen\endcsname{0.0}
\expandafter\def\csname res@pvit-d12-scannet200-wide@iou-projector screen-pm\endcsname{0.0}
\expandafter\def\csname res@pvit-d12-scannet200-wide@iou-divider\endcsname{67.4}
\expandafter\def\csname res@pvit-d12-scannet200-wide@iou-divider-pm\endcsname{67.4}
\expandafter\def\csname res@pvit-d12-scannet200-wide@iou-laundry detergent\endcsname{0.0}
\expandafter\def\csname res@pvit-d12-scannet200-wide@iou-laundry detergent-pm\endcsname{0.0}
\expandafter\def\csname res@pvit-d12-scannet200-wide@iou-bathroom counter\endcsname{26.9}
\expandafter\def\csname res@pvit-d12-scannet200-wide@iou-bathroom counter-pm\endcsname{26.9}
\expandafter\def\csname res@pvit-d12-scannet200-wide@iou-object\endcsname{7.0}
\expandafter\def\csname res@pvit-d12-scannet200-wide@iou-object-pm\endcsname{7.0}
\expandafter\def\csname res@pvit-d12-scannet200-wide@iou-bathroom vanity\endcsname{49.1}
\expandafter\def\csname res@pvit-d12-scannet200-wide@iou-bathroom vanity-pm\endcsname{49.1}
\expandafter\def\csname res@pvit-d12-scannet200-wide@iou-closet wall\endcsname{10.9}
\expandafter\def\csname res@pvit-d12-scannet200-wide@iou-closet wall-pm\endcsname{10.9}
\expandafter\def\csname res@pvit-d12-scannet200-wide@iou-laundry hamper\endcsname{41.3}
\expandafter\def\csname res@pvit-d12-scannet200-wide@iou-laundry hamper-pm\endcsname{41.3}
\expandafter\def\csname res@pvit-d12-scannet200-wide@iou-bathroom stall door\endcsname{60.1}
\expandafter\def\csname res@pvit-d12-scannet200-wide@iou-bathroom stall door-pm\endcsname{60.1}
\expandafter\def\csname res@pvit-d12-scannet200-wide@iou-ceiling light\endcsname{41.8}
\expandafter\def\csname res@pvit-d12-scannet200-wide@iou-ceiling light-pm\endcsname{41.8}
\expandafter\def\csname res@pvit-d12-scannet200-wide@iou-trash bin\endcsname{31.9}
\expandafter\def\csname res@pvit-d12-scannet200-wide@iou-trash bin-pm\endcsname{31.9}
\expandafter\def\csname res@pvit-d12-scannet200-wide@iou-dumbbell\endcsname{87.4}
\expandafter\def\csname res@pvit-d12-scannet200-wide@iou-dumbbell-pm\endcsname{87.4}
\expandafter\def\csname res@pvit-d12-scannet200-wide@iou-stair rail\endcsname{32.3}
\expandafter\def\csname res@pvit-d12-scannet200-wide@iou-stair rail-pm\endcsname{32.3}
\expandafter\def\csname res@pvit-d12-scannet200-wide@iou-tube\endcsname{35.4}
\expandafter\def\csname res@pvit-d12-scannet200-wide@iou-tube-pm\endcsname{35.4}
\expandafter\def\csname res@pvit-d12-scannet200-wide@iou-bathroom cabinet\endcsname{30.9}
\expandafter\def\csname res@pvit-d12-scannet200-wide@iou-bathroom cabinet-pm\endcsname{30.9}
\expandafter\def\csname res@pvit-d12-scannet200-wide@iou-cd case\endcsname{0.0}
\expandafter\def\csname res@pvit-d12-scannet200-wide@iou-cd case-pm\endcsname{0.0}
\expandafter\def\csname res@pvit-d12-scannet200-wide@iou-closet rod\endcsname{24.4}
\expandafter\def\csname res@pvit-d12-scannet200-wide@iou-closet rod-pm\endcsname{24.4}
\expandafter\def\csname res@pvit-d12-scannet200-wide@iou-coffee kettle\endcsname{39.4}
\expandafter\def\csname res@pvit-d12-scannet200-wide@iou-coffee kettle-pm\endcsname{39.4}
\expandafter\def\csname res@pvit-d12-scannet200-wide@iou-structure\endcsname{0.0}
\expandafter\def\csname res@pvit-d12-scannet200-wide@iou-structure-pm\endcsname{0.0}
\expandafter\def\csname res@pvit-d12-scannet200-wide@iou-shower head\endcsname{34.2}
\expandafter\def\csname res@pvit-d12-scannet200-wide@iou-shower head-pm\endcsname{34.2}
\expandafter\def\csname res@pvit-d12-scannet200-wide@iou-keyboard piano\endcsname{0.0}
\expandafter\def\csname res@pvit-d12-scannet200-wide@iou-keyboard piano-pm\endcsname{0.0}
\expandafter\def\csname res@pvit-d12-scannet200-wide@iou-case of water bottles\endcsname{35.3}
\expandafter\def\csname res@pvit-d12-scannet200-wide@iou-case of water bottles-pm\endcsname{35.3}
\expandafter\def\csname res@pvit-d12-scannet200-wide@iou-coat rack\endcsname{0.0}
\expandafter\def\csname res@pvit-d12-scannet200-wide@iou-coat rack-pm\endcsname{0.0}
\expandafter\def\csname res@pvit-d12-scannet200-wide@iou-storage organizer\endcsname{0.0}
\expandafter\def\csname res@pvit-d12-scannet200-wide@iou-storage organizer-pm\endcsname{0.0}
\expandafter\def\csname res@pvit-d12-scannet200-wide@iou-folded chair\endcsname{0.0}
\expandafter\def\csname res@pvit-d12-scannet200-wide@iou-folded chair-pm\endcsname{0.0}
\expandafter\def\csname res@pvit-d12-scannet200-wide@iou-fire alarm\endcsname{0.0}
\expandafter\def\csname res@pvit-d12-scannet200-wide@iou-fire alarm-pm\endcsname{0.0}
\expandafter\def\csname res@pvit-d12-scannet200-wide@iou-power strip\endcsname{0.0}
\expandafter\def\csname res@pvit-d12-scannet200-wide@iou-power strip-pm\endcsname{0.0}
\expandafter\def\csname res@pvit-d12-scannet200-wide@iou-calendar\endcsname{0.0}
\expandafter\def\csname res@pvit-d12-scannet200-wide@iou-calendar-pm\endcsname{0.0}
\expandafter\def\csname res@pvit-d12-scannet200-wide@iou-poster\endcsname{0.0}
\expandafter\def\csname res@pvit-d12-scannet200-wide@iou-poster-pm\endcsname{0.0}
\expandafter\def\csname res@pvit-d12-scannet200-wide@iou-potted plant\endcsname{0.0}
\expandafter\def\csname res@pvit-d12-scannet200-wide@iou-potted plant-pm\endcsname{0.0}
\expandafter\def\csname res@pvit-d12-scannet200-wide@iou-luggage\endcsname{0.0}
\expandafter\def\csname res@pvit-d12-scannet200-wide@iou-luggage-pm\endcsname{0.0}
\expandafter\def\csname res@pvit-d12-scannet200-wide@iou-mattress\endcsname{3.3}
\expandafter\def\csname res@pvit-d12-scannet200-wide@iou-mattress-pm\endcsname{3.3}
\expandafter\def\csname res@pvit-d12-sd-hard@n\endcsname{1}
\expandafter\def\csname res@pvit-d12-sd-hard@miou\endcsname{77.7}
\expandafter\def\csname res@pvit-d12-sd-hard@miou-pm\endcsname{77.7}
\expandafter\def\csname res@pvit-d12-sd-hard@macc\endcsname{85.6}
\expandafter\def\csname res@pvit-d12-sd-hard@macc-pm\endcsname{85.6}
\expandafter\def\csname res@pvit-d12-sd-hard@allacc\endcsname{91.9}
\expandafter\def\csname res@pvit-d12-sd-hard@allacc-pm\endcsname{91.9}
\expandafter\def\csname res@pvit-d12-sd-hard@iou-wall\endcsname{87.4}
\expandafter\def\csname res@pvit-d12-sd-hard@iou-wall-pm\endcsname{87.4}
\expandafter\def\csname res@pvit-d12-sd-hard@iou-floor\endcsname{95.8}
\expandafter\def\csname res@pvit-d12-sd-hard@iou-floor-pm\endcsname{95.8}
\expandafter\def\csname res@pvit-d12-sd-hard@iou-cabinet\endcsname{72.5}
\expandafter\def\csname res@pvit-d12-sd-hard@iou-cabinet-pm\endcsname{72.5}
\expandafter\def\csname res@pvit-d12-sd-hard@iou-bed\endcsname{83.0}
\expandafter\def\csname res@pvit-d12-sd-hard@iou-bed-pm\endcsname{83.0}
\expandafter\def\csname res@pvit-d12-sd-hard@iou-chair\endcsname{91.5}
\expandafter\def\csname res@pvit-d12-sd-hard@iou-chair-pm\endcsname{91.5}
\expandafter\def\csname res@pvit-d12-sd-hard@iou-sofa\endcsname{85.8}
\expandafter\def\csname res@pvit-d12-sd-hard@iou-sofa-pm\endcsname{85.8}
\expandafter\def\csname res@pvit-d12-sd-hard@iou-table\endcsname{77.1}
\expandafter\def\csname res@pvit-d12-sd-hard@iou-table-pm\endcsname{77.1}
\expandafter\def\csname res@pvit-d12-sd-hard@iou-door\endcsname{71.7}
\expandafter\def\csname res@pvit-d12-sd-hard@iou-door-pm\endcsname{71.7}
\expandafter\def\csname res@pvit-d12-sd-hard@iou-window\endcsname{75.0}
\expandafter\def\csname res@pvit-d12-sd-hard@iou-window-pm\endcsname{75.0}
\expandafter\def\csname res@pvit-d12-sd-hard@iou-bookshelf\endcsname{84.4}
\expandafter\def\csname res@pvit-d12-sd-hard@iou-bookshelf-pm\endcsname{84.4}
\expandafter\def\csname res@pvit-d12-sd-hard@iou-picture\endcsname{39.6}
\expandafter\def\csname res@pvit-d12-sd-hard@iou-picture-pm\endcsname{39.6}
\expandafter\def\csname res@pvit-d12-sd-hard@iou-counter\endcsname{71.2}
\expandafter\def\csname res@pvit-d12-sd-hard@iou-counter-pm\endcsname{71.2}
\expandafter\def\csname res@pvit-d12-sd-hard@iou-desk\endcsname{72.1}
\expandafter\def\csname res@pvit-d12-sd-hard@iou-desk-pm\endcsname{72.1}
\expandafter\def\csname res@pvit-d12-sd-hard@iou-curtain\endcsname{81.2}
\expandafter\def\csname res@pvit-d12-sd-hard@iou-curtain-pm\endcsname{81.2}
\expandafter\def\csname res@pvit-d12-sd-hard@iou-refridgerator\endcsname{74.3}
\expandafter\def\csname res@pvit-d12-sd-hard@iou-refridgerator-pm\endcsname{74.3}
\expandafter\def\csname res@pvit-d12-sd-hard@iou-shower curtain\endcsname{76.2}
\expandafter\def\csname res@pvit-d12-sd-hard@iou-shower curtain-pm\endcsname{76.2}
\expandafter\def\csname res@pvit-d12-sd-hard@iou-toilet\endcsname{95.1}
\expandafter\def\csname res@pvit-d12-sd-hard@iou-toilet-pm\endcsname{95.1}
\expandafter\def\csname res@pvit-d12-sd-hard@iou-sink\endcsname{70.2}
\expandafter\def\csname res@pvit-d12-sd-hard@iou-sink-pm\endcsname{70.2}
\expandafter\def\csname res@pvit-d12-sd-hard@iou-bathtub\endcsname{88.7}
\expandafter\def\csname res@pvit-d12-sd-hard@iou-bathtub-pm\endcsname{88.7}
\expandafter\def\csname res@pvit-d12-sd-hard@iou-otherfurniture\endcsname{61.4}
\expandafter\def\csname res@pvit-d12-sd-hard@iou-otherfurniture-pm\endcsname{61.4}
\expandafter\def\csname res@pvit-d12-sd-hard@miou-delta\endcsname{+0.2}
\expandafter\def\csname res@pvit-d12-sd-hard@macc-delta\endcsname{\ensuremath{-}0.2}
\expandafter\def\csname res@pvit-d12-sd-hard@allacc-delta\endcsname{+0.0}
\expandafter\def\csname res@pvit-d12-sd-hard@iou-wall-delta\endcsname{\ensuremath{-}0.4}
\expandafter\def\csname res@pvit-d12-sd-hard@iou-floor-delta\endcsname{\ensuremath{-}0.1}
\expandafter\def\csname res@pvit-d12-sd-hard@iou-cabinet-delta\endcsname{+0.1}
\expandafter\def\csname res@pvit-d12-sd-hard@iou-bed-delta\endcsname{+0.0}
\expandafter\def\csname res@pvit-d12-sd-hard@iou-chair-delta\endcsname{\ensuremath{-}1.2}
\expandafter\def\csname res@pvit-d12-sd-hard@iou-sofa-delta\endcsname{\ensuremath{-}1.0}
\expandafter\def\csname res@pvit-d12-sd-hard@iou-table-delta\endcsname{+0.0}
\expandafter\def\csname res@pvit-d12-sd-hard@iou-door-delta\endcsname{+2.1}
\expandafter\def\csname res@pvit-d12-sd-hard@iou-window-delta\endcsname{+3.5}
\expandafter\def\csname res@pvit-d12-sd-hard@iou-bookshelf-delta\endcsname{\ensuremath{-}1.1}
\expandafter\def\csname res@pvit-d12-sd-hard@iou-picture-delta\endcsname{\ensuremath{-}0.3}
\expandafter\def\csname res@pvit-d12-sd-hard@iou-counter-delta\endcsname{+0.7}
\expandafter\def\csname res@pvit-d12-sd-hard@iou-desk-delta\endcsname{+0.0}
\expandafter\def\csname res@pvit-d12-sd-hard@iou-curtain-delta\endcsname{+0.5}
\expandafter\def\csname res@pvit-d12-sd-hard@iou-refridgerator-delta\endcsname{+3.4}
\expandafter\def\csname res@pvit-d12-sd-hard@iou-shower curtain-delta\endcsname{+0.8}
\expandafter\def\csname res@pvit-d12-sd-hard@iou-toilet-delta\endcsname{\ensuremath{-}0.9}
\expandafter\def\csname res@pvit-d12-sd-hard@iou-sink-delta\endcsname{\ensuremath{-}1.0}
\expandafter\def\csname res@pvit-d12-sd-hard@iou-bathtub-delta\endcsname{\ensuremath{-}0.1}
\expandafter\def\csname res@pvit-d12-sd-hard@iou-otherfurniture-delta\endcsname{\ensuremath{-}1.1}
\expandafter\def\csname res@pvit-d12-kd-hard@n\endcsname{1}
\expandafter\def\csname res@pvit-d12-kd-hard@miou\endcsname{78.0}
\expandafter\def\csname res@pvit-d12-kd-hard@miou-pm\endcsname{78.0}
\expandafter\def\csname res@pvit-d12-kd-hard@macc\endcsname{85.8}
\expandafter\def\csname res@pvit-d12-kd-hard@macc-pm\endcsname{85.8}
\expandafter\def\csname res@pvit-d12-kd-hard@allacc\endcsname{91.9}
\expandafter\def\csname res@pvit-d12-kd-hard@allacc-pm\endcsname{91.9}
\expandafter\def\csname res@pvit-d12-kd-hard@iou-wall\endcsname{87.4}
\expandafter\def\csname res@pvit-d12-kd-hard@iou-wall-pm\endcsname{87.4}
\expandafter\def\csname res@pvit-d12-kd-hard@iou-floor\endcsname{95.9}
\expandafter\def\csname res@pvit-d12-kd-hard@iou-floor-pm\endcsname{95.9}
\expandafter\def\csname res@pvit-d12-kd-hard@iou-cabinet\endcsname{71.2}
\expandafter\def\csname res@pvit-d12-kd-hard@iou-cabinet-pm\endcsname{71.2}
\expandafter\def\csname res@pvit-d12-kd-hard@iou-bed\endcsname{84.5}
\expandafter\def\csname res@pvit-d12-kd-hard@iou-bed-pm\endcsname{84.5}
\expandafter\def\csname res@pvit-d12-kd-hard@iou-chair\endcsname{92.0}
\expandafter\def\csname res@pvit-d12-kd-hard@iou-chair-pm\endcsname{92.0}
\expandafter\def\csname res@pvit-d12-kd-hard@iou-sofa\endcsname{86.0}
\expandafter\def\csname res@pvit-d12-kd-hard@iou-sofa-pm\endcsname{86.0}
\expandafter\def\csname res@pvit-d12-kd-hard@iou-table\endcsname{78.5}
\expandafter\def\csname res@pvit-d12-kd-hard@iou-table-pm\endcsname{78.5}
\expandafter\def\csname res@pvit-d12-kd-hard@iou-door\endcsname{69.0}
\expandafter\def\csname res@pvit-d12-kd-hard@iou-door-pm\endcsname{69.0}
\expandafter\def\csname res@pvit-d12-kd-hard@iou-window\endcsname{71.4}
\expandafter\def\csname res@pvit-d12-kd-hard@iou-window-pm\endcsname{71.4}
\expandafter\def\csname res@pvit-d12-kd-hard@iou-bookshelf\endcsname{85.7}
\expandafter\def\csname res@pvit-d12-kd-hard@iou-bookshelf-pm\endcsname{85.7}
\expandafter\def\csname res@pvit-d12-kd-hard@iou-picture\endcsname{41.6}
\expandafter\def\csname res@pvit-d12-kd-hard@iou-picture-pm\endcsname{41.6}
\expandafter\def\csname res@pvit-d12-kd-hard@iou-counter\endcsname{72.3}
\expandafter\def\csname res@pvit-d12-kd-hard@iou-counter-pm\endcsname{72.3}
\expandafter\def\csname res@pvit-d12-kd-hard@iou-desk\endcsname{74.7}
\expandafter\def\csname res@pvit-d12-kd-hard@iou-desk-pm\endcsname{74.7}
\expandafter\def\csname res@pvit-d12-kd-hard@iou-curtain\endcsname{81.5}
\expandafter\def\csname res@pvit-d12-kd-hard@iou-curtain-pm\endcsname{81.5}
\expandafter\def\csname res@pvit-d12-kd-hard@iou-refridgerator\endcsname{74.1}
\expandafter\def\csname res@pvit-d12-kd-hard@iou-refridgerator-pm\endcsname{74.1}
\expandafter\def\csname res@pvit-d12-kd-hard@iou-shower curtain\endcsname{72.4}
\expandafter\def\csname res@pvit-d12-kd-hard@iou-shower curtain-pm\endcsname{72.4}
\expandafter\def\csname res@pvit-d12-kd-hard@iou-toilet\endcsname{96.2}
\expandafter\def\csname res@pvit-d12-kd-hard@iou-toilet-pm\endcsname{96.2}
\expandafter\def\csname res@pvit-d12-kd-hard@iou-sink\endcsname{72.2}
\expandafter\def\csname res@pvit-d12-kd-hard@iou-sink-pm\endcsname{72.2}
\expandafter\def\csname res@pvit-d12-kd-hard@iou-bathtub\endcsname{89.1}
\expandafter\def\csname res@pvit-d12-kd-hard@iou-bathtub-pm\endcsname{89.1}
\expandafter\def\csname res@pvit-d12-kd-hard@iou-otherfurniture\endcsname{63.9}
\expandafter\def\csname res@pvit-d12-kd-hard@iou-otherfurniture-pm\endcsname{63.9}
\expandafter\def\csname res@pvit-d12-kd-hard@miou-delta\endcsname{+0.5}
\expandafter\def\csname res@pvit-d12-kd-hard@macc-delta\endcsname{+0.0}
\expandafter\def\csname res@pvit-d12-kd-hard@allacc-delta\endcsname{+0.0}
\expandafter\def\csname res@pvit-d12-kd-hard@iou-wall-delta\endcsname{\ensuremath{-}0.4}
\expandafter\def\csname res@pvit-d12-kd-hard@iou-floor-delta\endcsname{+0.0}
\expandafter\def\csname res@pvit-d12-kd-hard@iou-cabinet-delta\endcsname{\ensuremath{-}1.2}
\expandafter\def\csname res@pvit-d12-kd-hard@iou-bed-delta\endcsname{+1.5}
\expandafter\def\csname res@pvit-d12-kd-hard@iou-chair-delta\endcsname{\ensuremath{-}0.7}
\expandafter\def\csname res@pvit-d12-kd-hard@iou-sofa-delta\endcsname{\ensuremath{-}0.8}
\expandafter\def\csname res@pvit-d12-kd-hard@iou-table-delta\endcsname{+1.4}
\expandafter\def\csname res@pvit-d12-kd-hard@iou-door-delta\endcsname{\ensuremath{-}0.6}
\expandafter\def\csname res@pvit-d12-kd-hard@iou-window-delta\endcsname{\ensuremath{-}0.1}
\expandafter\def\csname res@pvit-d12-kd-hard@iou-bookshelf-delta\endcsname{+0.2}
\expandafter\def\csname res@pvit-d12-kd-hard@iou-picture-delta\endcsname{+1.7}
\expandafter\def\csname res@pvit-d12-kd-hard@iou-counter-delta\endcsname{+1.8}
\expandafter\def\csname res@pvit-d12-kd-hard@iou-desk-delta\endcsname{+2.6}
\expandafter\def\csname res@pvit-d12-kd-hard@iou-curtain-delta\endcsname{+0.8}
\expandafter\def\csname res@pvit-d12-kd-hard@iou-refridgerator-delta\endcsname{+3.2}
\expandafter\def\csname res@pvit-d12-kd-hard@iou-shower curtain-delta\endcsname{\ensuremath{-}3.0}
\expandafter\def\csname res@pvit-d12-kd-hard@iou-toilet-delta\endcsname{+0.2}
\expandafter\def\csname res@pvit-d12-kd-hard@iou-sink-delta\endcsname{+1.0}
\expandafter\def\csname res@pvit-d12-kd-hard@iou-bathtub-delta\endcsname{+0.3}
\expandafter\def\csname res@pvit-d12-kd-hard@iou-otherfurniture-delta\endcsname{+1.4}
\expandafter\def\csname res@pvit-d12-kd-soft@n\endcsname{1}
\expandafter\def\csname res@pvit-d12-kd-soft@miou\endcsname{77.5}
\expandafter\def\csname res@pvit-d12-kd-soft@miou-pm\endcsname{77.5}
\expandafter\def\csname res@pvit-d12-kd-soft@macc\endcsname{85.4}
\expandafter\def\csname res@pvit-d12-kd-soft@macc-pm\endcsname{85.4}
\expandafter\def\csname res@pvit-d12-kd-soft@allacc\endcsname{91.8}
\expandafter\def\csname res@pvit-d12-kd-soft@allacc-pm\endcsname{91.8}
\expandafter\def\csname res@pvit-d12-kd-soft@iou-wall\endcsname{87.4}
\expandafter\def\csname res@pvit-d12-kd-soft@iou-wall-pm\endcsname{87.4}
\expandafter\def\csname res@pvit-d12-kd-soft@iou-floor\endcsname{95.9}
\expandafter\def\csname res@pvit-d12-kd-soft@iou-floor-pm\endcsname{95.9}
\expandafter\def\csname res@pvit-d12-kd-soft@iou-cabinet\endcsname{72.5}
\expandafter\def\csname res@pvit-d12-kd-soft@iou-cabinet-pm\endcsname{72.5}
\expandafter\def\csname res@pvit-d12-kd-soft@iou-bed\endcsname{83.5}
\expandafter\def\csname res@pvit-d12-kd-soft@iou-bed-pm\endcsname{83.5}
\expandafter\def\csname res@pvit-d12-kd-soft@iou-chair\endcsname{91.1}
\expandafter\def\csname res@pvit-d12-kd-soft@iou-chair-pm\endcsname{91.1}
\expandafter\def\csname res@pvit-d12-kd-soft@iou-sofa\endcsname{84.0}
\expandafter\def\csname res@pvit-d12-kd-soft@iou-sofa-pm\endcsname{84.0}
\expandafter\def\csname res@pvit-d12-kd-soft@iou-table\endcsname{78.1}
\expandafter\def\csname res@pvit-d12-kd-soft@iou-table-pm\endcsname{78.1}
\expandafter\def\csname res@pvit-d12-kd-soft@iou-door\endcsname{70.4}
\expandafter\def\csname res@pvit-d12-kd-soft@iou-door-pm\endcsname{70.4}
\expandafter\def\csname res@pvit-d12-kd-soft@iou-window\endcsname{71.5}
\expandafter\def\csname res@pvit-d12-kd-soft@iou-window-pm\endcsname{71.5}
\expandafter\def\csname res@pvit-d12-kd-soft@iou-bookshelf\endcsname{83.5}
\expandafter\def\csname res@pvit-d12-kd-soft@iou-bookshelf-pm\endcsname{83.5}
\expandafter\def\csname res@pvit-d12-kd-soft@iou-picture\endcsname{41.1}
\expandafter\def\csname res@pvit-d12-kd-soft@iou-picture-pm\endcsname{41.1}
\expandafter\def\csname res@pvit-d12-kd-soft@iou-counter\endcsname{72.6}
\expandafter\def\csname res@pvit-d12-kd-soft@iou-counter-pm\endcsname{72.6}
\expandafter\def\csname res@pvit-d12-kd-soft@iou-desk\endcsname{73.8}
\expandafter\def\csname res@pvit-d12-kd-soft@iou-desk-pm\endcsname{73.8}
\expandafter\def\csname res@pvit-d12-kd-soft@iou-curtain\endcsname{80.0}
\expandafter\def\csname res@pvit-d12-kd-soft@iou-curtain-pm\endcsname{80.0}
\expandafter\def\csname res@pvit-d12-kd-soft@iou-refridgerator\endcsname{75.0}
\expandafter\def\csname res@pvit-d12-kd-soft@iou-refridgerator-pm\endcsname{75.0}
\expandafter\def\csname res@pvit-d12-kd-soft@iou-shower curtain\endcsname{69.5}
\expandafter\def\csname res@pvit-d12-kd-soft@iou-shower curtain-pm\endcsname{69.5}
\expandafter\def\csname res@pvit-d12-kd-soft@iou-toilet\endcsname{96.1}
\expandafter\def\csname res@pvit-d12-kd-soft@iou-toilet-pm\endcsname{96.1}
\expandafter\def\csname res@pvit-d12-kd-soft@iou-sink\endcsname{71.6}
\expandafter\def\csname res@pvit-d12-kd-soft@iou-sink-pm\endcsname{71.6}
\expandafter\def\csname res@pvit-d12-kd-soft@iou-bathtub\endcsname{89.4}
\expandafter\def\csname res@pvit-d12-kd-soft@iou-bathtub-pm\endcsname{89.4}
\expandafter\def\csname res@pvit-d12-kd-soft@iou-otherfurniture\endcsname{62.3}
\expandafter\def\csname res@pvit-d12-kd-soft@iou-otherfurniture-pm\endcsname{62.3}
\expandafter\def\csname res@pvit-d12-kd-soft@miou-delta\endcsname{+0.0}
\expandafter\def\csname res@pvit-d12-kd-soft@macc-delta\endcsname{\ensuremath{-}0.4}
\expandafter\def\csname res@pvit-d12-kd-soft@allacc-delta\endcsname{\ensuremath{-}0.1}
\expandafter\def\csname res@pvit-d12-kd-soft@iou-wall-delta\endcsname{\ensuremath{-}0.4}
\expandafter\def\csname res@pvit-d12-kd-soft@iou-floor-delta\endcsname{+0.0}
\expandafter\def\csname res@pvit-d12-kd-soft@iou-cabinet-delta\endcsname{+0.1}
\expandafter\def\csname res@pvit-d12-kd-soft@iou-bed-delta\endcsname{+0.5}
\expandafter\def\csname res@pvit-d12-kd-soft@iou-chair-delta\endcsname{\ensuremath{-}1.6}
\expandafter\def\csname res@pvit-d12-kd-soft@iou-sofa-delta\endcsname{\ensuremath{-}2.8}
\expandafter\def\csname res@pvit-d12-kd-soft@iou-table-delta\endcsname{+1.0}
\expandafter\def\csname res@pvit-d12-kd-soft@iou-door-delta\endcsname{+0.8}
\expandafter\def\csname res@pvit-d12-kd-soft@iou-window-delta\endcsname{+0.0}
\expandafter\def\csname res@pvit-d12-kd-soft@iou-bookshelf-delta\endcsname{\ensuremath{-}2.0}
\expandafter\def\csname res@pvit-d12-kd-soft@iou-picture-delta\endcsname{+1.2}
\expandafter\def\csname res@pvit-d12-kd-soft@iou-counter-delta\endcsname{+2.1}
\expandafter\def\csname res@pvit-d12-kd-soft@iou-desk-delta\endcsname{+1.7}
\expandafter\def\csname res@pvit-d12-kd-soft@iou-curtain-delta\endcsname{\ensuremath{-}0.7}
\expandafter\def\csname res@pvit-d12-kd-soft@iou-refridgerator-delta\endcsname{+4.1}
\expandafter\def\csname res@pvit-d12-kd-soft@iou-shower curtain-delta\endcsname{\ensuremath{-}5.9}
\expandafter\def\csname res@pvit-d12-kd-soft@iou-toilet-delta\endcsname{+0.1}
\expandafter\def\csname res@pvit-d12-kd-soft@iou-sink-delta\endcsname{+0.4}
\expandafter\def\csname res@pvit-d12-kd-soft@iou-bathtub-delta\endcsname{+0.6}
\expandafter\def\csname res@pvit-d12-kd-soft@iou-otherfurniture-delta\endcsname{\ensuremath{-}0.2}
\expandafter\def\csname res@pvit-d12-cnnkd-scannet@n\endcsname{1}
\expandafter\def\csname res@pvit-d12-cnnkd-scannet@miou\endcsname{78.3}
\expandafter\def\csname res@pvit-d12-cnnkd-scannet@miou-pm\endcsname{78.3}
\expandafter\def\csname res@pvit-d12-cnnkd-scannet@macc\endcsname{86.1}
\expandafter\def\csname res@pvit-d12-cnnkd-scannet@macc-pm\endcsname{86.1}
\expandafter\def\csname res@pvit-d12-cnnkd-scannet@allacc\endcsname{92.2}
\expandafter\def\csname res@pvit-d12-cnnkd-scannet@allacc-pm\endcsname{92.2}
\expandafter\def\csname res@pvit-d12-cnnkd-scannet@iou-wall\endcsname{88.1}
\expandafter\def\csname res@pvit-d12-cnnkd-scannet@iou-wall-pm\endcsname{88.1}
\expandafter\def\csname res@pvit-d12-cnnkd-scannet@iou-floor\endcsname{95.9}
\expandafter\def\csname res@pvit-d12-cnnkd-scannet@iou-floor-pm\endcsname{95.9}
\expandafter\def\csname res@pvit-d12-cnnkd-scannet@iou-cabinet\endcsname{72.0}
\expandafter\def\csname res@pvit-d12-cnnkd-scannet@iou-cabinet-pm\endcsname{72.0}
\expandafter\def\csname res@pvit-d12-cnnkd-scannet@iou-bed\endcsname{85.7}
\expandafter\def\csname res@pvit-d12-cnnkd-scannet@iou-bed-pm\endcsname{85.7}
\expandafter\def\csname res@pvit-d12-cnnkd-scannet@iou-chair\endcsname{92.6}
\expandafter\def\csname res@pvit-d12-cnnkd-scannet@iou-chair-pm\endcsname{92.6}
\expandafter\def\csname res@pvit-d12-cnnkd-scannet@iou-sofa\endcsname{86.7}
\expandafter\def\csname res@pvit-d12-cnnkd-scannet@iou-sofa-pm\endcsname{86.7}
\expandafter\def\csname res@pvit-d12-cnnkd-scannet@iou-table\endcsname{79.5}
\expandafter\def\csname res@pvit-d12-cnnkd-scannet@iou-table-pm\endcsname{79.5}
\expandafter\def\csname res@pvit-d12-cnnkd-scannet@iou-door\endcsname{72.4}
\expandafter\def\csname res@pvit-d12-cnnkd-scannet@iou-door-pm\endcsname{72.4}
\expandafter\def\csname res@pvit-d12-cnnkd-scannet@iou-window\endcsname{75.7}
\expandafter\def\csname res@pvit-d12-cnnkd-scannet@iou-window-pm\endcsname{75.7}
\expandafter\def\csname res@pvit-d12-cnnkd-scannet@iou-bookshelf\endcsname{82.9}
\expandafter\def\csname res@pvit-d12-cnnkd-scannet@iou-bookshelf-pm\endcsname{82.9}
\expandafter\def\csname res@pvit-d12-cnnkd-scannet@iou-picture\endcsname{40.8}
\expandafter\def\csname res@pvit-d12-cnnkd-scannet@iou-picture-pm\endcsname{40.8}
\expandafter\def\csname res@pvit-d12-cnnkd-scannet@iou-counter\endcsname{71.7}
\expandafter\def\csname res@pvit-d12-cnnkd-scannet@iou-counter-pm\endcsname{71.7}
\expandafter\def\csname res@pvit-d12-cnnkd-scannet@iou-desk\endcsname{75.6}
\expandafter\def\csname res@pvit-d12-cnnkd-scannet@iou-desk-pm\endcsname{75.6}
\expandafter\def\csname res@pvit-d12-cnnkd-scannet@iou-curtain\endcsname{81.9}
\expandafter\def\csname res@pvit-d12-cnnkd-scannet@iou-curtain-pm\endcsname{81.9}
\expandafter\def\csname res@pvit-d12-cnnkd-scannet@iou-refridgerator\endcsname{72.4}
\expandafter\def\csname res@pvit-d12-cnnkd-scannet@iou-refridgerator-pm\endcsname{72.4}
\expandafter\def\csname res@pvit-d12-cnnkd-scannet@iou-shower curtain\endcsname{70.3}
\expandafter\def\csname res@pvit-d12-cnnkd-scannet@iou-shower curtain-pm\endcsname{70.3}
\expandafter\def\csname res@pvit-d12-cnnkd-scannet@iou-toilet\endcsname{96.2}
\expandafter\def\csname res@pvit-d12-cnnkd-scannet@iou-toilet-pm\endcsname{96.2}
\expandafter\def\csname res@pvit-d12-cnnkd-scannet@iou-sink\endcsname{74.3}
\expandafter\def\csname res@pvit-d12-cnnkd-scannet@iou-sink-pm\endcsname{74.3}
\expandafter\def\csname res@pvit-d12-cnnkd-scannet@iou-bathtub\endcsname{88.9}
\expandafter\def\csname res@pvit-d12-cnnkd-scannet@iou-bathtub-pm\endcsname{88.9}
\expandafter\def\csname res@pvit-d12-cnnkd-scannet@iou-otherfurniture\endcsname{61.6}
\expandafter\def\csname res@pvit-d12-cnnkd-scannet@iou-otherfurniture-pm\endcsname{61.6}
\expandafter\def\csname res@pvit-d12-cnnkd-scannet@miou-delta\endcsname{+0.8}
\expandafter\def\csname res@pvit-d12-cnnkd-scannet@macc-delta\endcsname{+0.3}
\expandafter\def\csname res@pvit-d12-cnnkd-scannet@allacc-delta\endcsname{+0.3}
\expandafter\def\csname res@pvit-d12-cnnkd-scannet@iou-wall-delta\endcsname{+0.3}
\expandafter\def\csname res@pvit-d12-cnnkd-scannet@iou-floor-delta\endcsname{+0.0}
\expandafter\def\csname res@pvit-d12-cnnkd-scannet@iou-cabinet-delta\endcsname{\ensuremath{-}0.4}
\expandafter\def\csname res@pvit-d12-cnnkd-scannet@iou-bed-delta\endcsname{+2.7}
\expandafter\def\csname res@pvit-d12-cnnkd-scannet@iou-chair-delta\endcsname{\ensuremath{-}0.1}
\expandafter\def\csname res@pvit-d12-cnnkd-scannet@iou-sofa-delta\endcsname{\ensuremath{-}0.1}
\expandafter\def\csname res@pvit-d12-cnnkd-scannet@iou-table-delta\endcsname{+2.4}
\expandafter\def\csname res@pvit-d12-cnnkd-scannet@iou-door-delta\endcsname{+2.8}
\expandafter\def\csname res@pvit-d12-cnnkd-scannet@iou-window-delta\endcsname{+4.2}
\expandafter\def\csname res@pvit-d12-cnnkd-scannet@iou-bookshelf-delta\endcsname{\ensuremath{-}2.6}
\expandafter\def\csname res@pvit-d12-cnnkd-scannet@iou-picture-delta\endcsname{+0.9}
\expandafter\def\csname res@pvit-d12-cnnkd-scannet@iou-counter-delta\endcsname{+1.2}
\expandafter\def\csname res@pvit-d12-cnnkd-scannet@iou-desk-delta\endcsname{+3.5}
\expandafter\def\csname res@pvit-d12-cnnkd-scannet@iou-curtain-delta\endcsname{+1.2}
\expandafter\def\csname res@pvit-d12-cnnkd-scannet@iou-refridgerator-delta\endcsname{+1.5}
\expandafter\def\csname res@pvit-d12-cnnkd-scannet@iou-shower curtain-delta\endcsname{\ensuremath{-}5.1}
\expandafter\def\csname res@pvit-d12-cnnkd-scannet@iou-toilet-delta\endcsname{+0.2}
\expandafter\def\csname res@pvit-d12-cnnkd-scannet@iou-sink-delta\endcsname{+3.1}
\expandafter\def\csname res@pvit-d12-cnnkd-scannet@iou-bathtub-delta\endcsname{+0.1}
\expandafter\def\csname res@pvit-d12-cnnkd-scannet@iou-otherfurniture-delta\endcsname{\ensuremath{-}0.9}
\expandafter\def\csname res@pvit-d12-structured3d@n\endcsname{1}
\expandafter\def\csname res@pvit-d12-structured3d@miou\endcsname{83.0}
\expandafter\def\csname res@pvit-d12-structured3d@miou-pm\endcsname{83.0}
\expandafter\def\csname res@pvit-d12-structured3d@macc\endcsname{89.9}
\expandafter\def\csname res@pvit-d12-structured3d@macc-pm\endcsname{89.9}
\expandafter\def\csname res@pvit-d12-structured3d@allacc\endcsname{97.2}
\expandafter\def\csname res@pvit-d12-structured3d@allacc-pm\endcsname{97.2}
\expandafter\def\csname res@pvit-d12-scannet200-cnnkd@n\endcsname{1}
\expandafter\def\csname res@pvit-d12-scannet200-cnnkd@miou\endcsname{36.9}
\expandafter\def\csname res@pvit-d12-scannet200-cnnkd@miou-pm\endcsname{36.9}
\expandafter\def\csname res@pvit-d12-scannet200-cnnkd@macc\endcsname{48.4}
\expandafter\def\csname res@pvit-d12-scannet200-cnnkd@macc-pm\endcsname{48.4}
\expandafter\def\csname res@pvit-d12-scannet200-cnnkd@allacc\endcsname{84.2}
\expandafter\def\csname res@pvit-d12-scannet200-cnnkd@allacc-pm\endcsname{84.2}
\expandafter\def\csname res@pvit-d12-scannet200-cnnkd@iou-wall\endcsname{81.5}
\expandafter\def\csname res@pvit-d12-scannet200-cnnkd@iou-wall-pm\endcsname{81.5}
\expandafter\def\csname res@pvit-d12-scannet200-cnnkd@iou-chair\endcsname{75.3}
\expandafter\def\csname res@pvit-d12-scannet200-cnnkd@iou-chair-pm\endcsname{75.3}
\expandafter\def\csname res@pvit-d12-scannet200-cnnkd@iou-floor\endcsname{95.2}
\expandafter\def\csname res@pvit-d12-scannet200-cnnkd@iou-floor-pm\endcsname{95.2}
\expandafter\def\csname res@pvit-d12-scannet200-cnnkd@iou-table\endcsname{70.7}
\expandafter\def\csname res@pvit-d12-scannet200-cnnkd@iou-table-pm\endcsname{70.7}
\expandafter\def\csname res@pvit-d12-scannet200-cnnkd@iou-door\endcsname{70.2}
\expandafter\def\csname res@pvit-d12-scannet200-cnnkd@iou-door-pm\endcsname{70.2}
\expandafter\def\csname res@pvit-d12-scannet200-cnnkd@iou-couch\endcsname{81.5}
\expandafter\def\csname res@pvit-d12-scannet200-cnnkd@iou-couch-pm\endcsname{81.5}
\expandafter\def\csname res@pvit-d12-scannet200-cnnkd@iou-cabinet\endcsname{42.7}
\expandafter\def\csname res@pvit-d12-scannet200-cnnkd@iou-cabinet-pm\endcsname{42.7}
\expandafter\def\csname res@pvit-d12-scannet200-cnnkd@iou-shelf\endcsname{48.8}
\expandafter\def\csname res@pvit-d12-scannet200-cnnkd@iou-shelf-pm\endcsname{48.8}
\expandafter\def\csname res@pvit-d12-scannet200-cnnkd@iou-desk\endcsname{68.7}
\expandafter\def\csname res@pvit-d12-scannet200-cnnkd@iou-desk-pm\endcsname{68.7}
\expandafter\def\csname res@pvit-d12-scannet200-cnnkd@iou-office chair\endcsname{20.0}
\expandafter\def\csname res@pvit-d12-scannet200-cnnkd@iou-office chair-pm\endcsname{20.0}
\expandafter\def\csname res@pvit-d12-scannet200-cnnkd@iou-bed\endcsname{82.2}
\expandafter\def\csname res@pvit-d12-scannet200-cnnkd@iou-bed-pm\endcsname{82.2}
\expandafter\def\csname res@pvit-d12-scannet200-cnnkd@iou-pillow\endcsname{63.1}
\expandafter\def\csname res@pvit-d12-scannet200-cnnkd@iou-pillow-pm\endcsname{63.1}
\expandafter\def\csname res@pvit-d12-scannet200-cnnkd@iou-sink\endcsname{71.5}
\expandafter\def\csname res@pvit-d12-scannet200-cnnkd@iou-sink-pm\endcsname{71.5}
\expandafter\def\csname res@pvit-d12-scannet200-cnnkd@iou-picture\endcsname{35.6}
\expandafter\def\csname res@pvit-d12-scannet200-cnnkd@iou-picture-pm\endcsname{35.6}
\expandafter\def\csname res@pvit-d12-scannet200-cnnkd@iou-window\endcsname{72.0}
\expandafter\def\csname res@pvit-d12-scannet200-cnnkd@iou-window-pm\endcsname{72.0}
\expandafter\def\csname res@pvit-d12-scannet200-cnnkd@iou-toilet\endcsname{94.7}
\expandafter\def\csname res@pvit-d12-scannet200-cnnkd@iou-toilet-pm\endcsname{94.7}
\expandafter\def\csname res@pvit-d12-scannet200-cnnkd@iou-bookshelf\endcsname{52.3}
\expandafter\def\csname res@pvit-d12-scannet200-cnnkd@iou-bookshelf-pm\endcsname{52.3}
\expandafter\def\csname res@pvit-d12-scannet200-cnnkd@iou-monitor\endcsname{90.6}
\expandafter\def\csname res@pvit-d12-scannet200-cnnkd@iou-monitor-pm\endcsname{90.6}
\expandafter\def\csname res@pvit-d12-scannet200-cnnkd@iou-curtain\endcsname{75.3}
\expandafter\def\csname res@pvit-d12-scannet200-cnnkd@iou-curtain-pm\endcsname{75.3}
\expandafter\def\csname res@pvit-d12-scannet200-cnnkd@iou-book\endcsname{44.3}
\expandafter\def\csname res@pvit-d12-scannet200-cnnkd@iou-book-pm\endcsname{44.3}
\expandafter\def\csname res@pvit-d12-scannet200-cnnkd@iou-armchair\endcsname{48.9}
\expandafter\def\csname res@pvit-d12-scannet200-cnnkd@iou-armchair-pm\endcsname{48.9}
\expandafter\def\csname res@pvit-d12-scannet200-cnnkd@iou-coffee table\endcsname{55.5}
\expandafter\def\csname res@pvit-d12-scannet200-cnnkd@iou-coffee table-pm\endcsname{55.5}
\expandafter\def\csname res@pvit-d12-scannet200-cnnkd@iou-box\endcsname{40.8}
\expandafter\def\csname res@pvit-d12-scannet200-cnnkd@iou-box-pm\endcsname{40.8}
\expandafter\def\csname res@pvit-d12-scannet200-cnnkd@iou-refrigerator\endcsname{75.1}
\expandafter\def\csname res@pvit-d12-scannet200-cnnkd@iou-refrigerator-pm\endcsname{75.1}
\expandafter\def\csname res@pvit-d12-scannet200-cnnkd@iou-lamp\endcsname{74.0}
\expandafter\def\csname res@pvit-d12-scannet200-cnnkd@iou-lamp-pm\endcsname{74.0}
\expandafter\def\csname res@pvit-d12-scannet200-cnnkd@iou-kitchen cabinet\endcsname{73.7}
\expandafter\def\csname res@pvit-d12-scannet200-cnnkd@iou-kitchen cabinet-pm\endcsname{73.7}
\expandafter\def\csname res@pvit-d12-scannet200-cnnkd@iou-towel\endcsname{61.7}
\expandafter\def\csname res@pvit-d12-scannet200-cnnkd@iou-towel-pm\endcsname{61.7}
\expandafter\def\csname res@pvit-d12-scannet200-cnnkd@iou-clothes\endcsname{23.2}
\expandafter\def\csname res@pvit-d12-scannet200-cnnkd@iou-clothes-pm\endcsname{23.2}
\expandafter\def\csname res@pvit-d12-scannet200-cnnkd@iou-tv\endcsname{92.1}
\expandafter\def\csname res@pvit-d12-scannet200-cnnkd@iou-tv-pm\endcsname{92.1}
\expandafter\def\csname res@pvit-d12-scannet200-cnnkd@iou-nightstand\endcsname{68.5}
\expandafter\def\csname res@pvit-d12-scannet200-cnnkd@iou-nightstand-pm\endcsname{68.5}
\expandafter\def\csname res@pvit-d12-scannet200-cnnkd@iou-counter\endcsname{36.9}
\expandafter\def\csname res@pvit-d12-scannet200-cnnkd@iou-counter-pm\endcsname{36.9}
\expandafter\def\csname res@pvit-d12-scannet200-cnnkd@iou-dresser\endcsname{53.5}
\expandafter\def\csname res@pvit-d12-scannet200-cnnkd@iou-dresser-pm\endcsname{53.5}
\expandafter\def\csname res@pvit-d12-scannet200-cnnkd@iou-stool\endcsname{61.1}
\expandafter\def\csname res@pvit-d12-scannet200-cnnkd@iou-stool-pm\endcsname{61.1}
\expandafter\def\csname res@pvit-d12-scannet200-cnnkd@iou-cushion\endcsname{0.0}
\expandafter\def\csname res@pvit-d12-scannet200-cnnkd@iou-cushion-pm\endcsname{0.0}
\expandafter\def\csname res@pvit-d12-scannet200-cnnkd@iou-plant\endcsname{75.5}
\expandafter\def\csname res@pvit-d12-scannet200-cnnkd@iou-plant-pm\endcsname{75.5}
\expandafter\def\csname res@pvit-d12-scannet200-cnnkd@iou-ceiling\endcsname{93.1}
\expandafter\def\csname res@pvit-d12-scannet200-cnnkd@iou-ceiling-pm\endcsname{93.1}
\expandafter\def\csname res@pvit-d12-scannet200-cnnkd@iou-bathtub\endcsname{82.3}
\expandafter\def\csname res@pvit-d12-scannet200-cnnkd@iou-bathtub-pm\endcsname{82.3}
\expandafter\def\csname res@pvit-d12-scannet200-cnnkd@iou-end table\endcsname{47.0}
\expandafter\def\csname res@pvit-d12-scannet200-cnnkd@iou-end table-pm\endcsname{47.0}
\expandafter\def\csname res@pvit-d12-scannet200-cnnkd@iou-dining table\endcsname{8.6}
\expandafter\def\csname res@pvit-d12-scannet200-cnnkd@iou-dining table-pm\endcsname{8.6}
\expandafter\def\csname res@pvit-d12-scannet200-cnnkd@iou-keyboard\endcsname{39.9}
\expandafter\def\csname res@pvit-d12-scannet200-cnnkd@iou-keyboard-pm\endcsname{39.9}
\expandafter\def\csname res@pvit-d12-scannet200-cnnkd@iou-bag\endcsname{24.3}
\expandafter\def\csname res@pvit-d12-scannet200-cnnkd@iou-bag-pm\endcsname{24.3}
\expandafter\def\csname res@pvit-d12-scannet200-cnnkd@iou-backpack\endcsname{65.7}
\expandafter\def\csname res@pvit-d12-scannet200-cnnkd@iou-backpack-pm\endcsname{65.7}
\expandafter\def\csname res@pvit-d12-scannet200-cnnkd@iou-toilet paper\endcsname{48.1}
\expandafter\def\csname res@pvit-d12-scannet200-cnnkd@iou-toilet paper-pm\endcsname{48.1}
\expandafter\def\csname res@pvit-d12-scannet200-cnnkd@iou-printer\endcsname{62.0}
\expandafter\def\csname res@pvit-d12-scannet200-cnnkd@iou-printer-pm\endcsname{62.0}
\expandafter\def\csname res@pvit-d12-scannet200-cnnkd@iou-tv stand\endcsname{48.8}
\expandafter\def\csname res@pvit-d12-scannet200-cnnkd@iou-tv stand-pm\endcsname{48.8}
\expandafter\def\csname res@pvit-d12-scannet200-cnnkd@iou-whiteboard\endcsname{54.8}
\expandafter\def\csname res@pvit-d12-scannet200-cnnkd@iou-whiteboard-pm\endcsname{54.8}
\expandafter\def\csname res@pvit-d12-scannet200-cnnkd@iou-blanket\endcsname{20.1}
\expandafter\def\csname res@pvit-d12-scannet200-cnnkd@iou-blanket-pm\endcsname{20.1}
\expandafter\def\csname res@pvit-d12-scannet200-cnnkd@iou-shower curtain\endcsname{70.5}
\expandafter\def\csname res@pvit-d12-scannet200-cnnkd@iou-shower curtain-pm\endcsname{70.5}
\expandafter\def\csname res@pvit-d12-scannet200-cnnkd@iou-trash can\endcsname{67.4}
\expandafter\def\csname res@pvit-d12-scannet200-cnnkd@iou-trash can-pm\endcsname{67.4}
\expandafter\def\csname res@pvit-d12-scannet200-cnnkd@iou-closet\endcsname{11.4}
\expandafter\def\csname res@pvit-d12-scannet200-cnnkd@iou-closet-pm\endcsname{11.4}
\expandafter\def\csname res@pvit-d12-scannet200-cnnkd@iou-stairs\endcsname{47.9}
\expandafter\def\csname res@pvit-d12-scannet200-cnnkd@iou-stairs-pm\endcsname{47.9}
\expandafter\def\csname res@pvit-d12-scannet200-cnnkd@iou-microwave\endcsname{58.4}
\expandafter\def\csname res@pvit-d12-scannet200-cnnkd@iou-microwave-pm\endcsname{58.4}
\expandafter\def\csname res@pvit-d12-scannet200-cnnkd@iou-stove\endcsname{67.5}
\expandafter\def\csname res@pvit-d12-scannet200-cnnkd@iou-stove-pm\endcsname{67.5}
\expandafter\def\csname res@pvit-d12-scannet200-cnnkd@iou-shoe\endcsname{54.8}
\expandafter\def\csname res@pvit-d12-scannet200-cnnkd@iou-shoe-pm\endcsname{54.8}
\expandafter\def\csname res@pvit-d12-scannet200-cnnkd@iou-computer tower\endcsname{64.1}
\expandafter\def\csname res@pvit-d12-scannet200-cnnkd@iou-computer tower-pm\endcsname{64.1}
\expandafter\def\csname res@pvit-d12-scannet200-cnnkd@iou-bottle\endcsname{18.7}
\expandafter\def\csname res@pvit-d12-scannet200-cnnkd@iou-bottle-pm\endcsname{18.7}
\expandafter\def\csname res@pvit-d12-scannet200-cnnkd@iou-bin\endcsname{3.5}
\expandafter\def\csname res@pvit-d12-scannet200-cnnkd@iou-bin-pm\endcsname{3.5}
\expandafter\def\csname res@pvit-d12-scannet200-cnnkd@iou-ottoman\endcsname{44.1}
\expandafter\def\csname res@pvit-d12-scannet200-cnnkd@iou-ottoman-pm\endcsname{44.1}
\expandafter\def\csname res@pvit-d12-scannet200-cnnkd@iou-bench\endcsname{48.0}
\expandafter\def\csname res@pvit-d12-scannet200-cnnkd@iou-bench-pm\endcsname{48.0}
\expandafter\def\csname res@pvit-d12-scannet200-cnnkd@iou-board\endcsname{20.6}
\expandafter\def\csname res@pvit-d12-scannet200-cnnkd@iou-board-pm\endcsname{20.6}
\expandafter\def\csname res@pvit-d12-scannet200-cnnkd@iou-washing machine\endcsname{84.0}
\expandafter\def\csname res@pvit-d12-scannet200-cnnkd@iou-washing machine-pm\endcsname{84.0}
\expandafter\def\csname res@pvit-d12-scannet200-cnnkd@iou-mirror\endcsname{50.1}
\expandafter\def\csname res@pvit-d12-scannet200-cnnkd@iou-mirror-pm\endcsname{50.1}
\expandafter\def\csname res@pvit-d12-scannet200-cnnkd@iou-copier\endcsname{88.5}
\expandafter\def\csname res@pvit-d12-scannet200-cnnkd@iou-copier-pm\endcsname{88.5}
\expandafter\def\csname res@pvit-d12-scannet200-cnnkd@iou-basket\endcsname{0.0}
\expandafter\def\csname res@pvit-d12-scannet200-cnnkd@iou-basket-pm\endcsname{0.0}
\expandafter\def\csname res@pvit-d12-scannet200-cnnkd@iou-sofa chair\endcsname{16.7}
\expandafter\def\csname res@pvit-d12-scannet200-cnnkd@iou-sofa chair-pm\endcsname{16.7}
\expandafter\def\csname res@pvit-d12-scannet200-cnnkd@iou-file cabinet\endcsname{61.4}
\expandafter\def\csname res@pvit-d12-scannet200-cnnkd@iou-file cabinet-pm\endcsname{61.4}
\expandafter\def\csname res@pvit-d12-scannet200-cnnkd@iou-fan\endcsname{56.4}
\expandafter\def\csname res@pvit-d12-scannet200-cnnkd@iou-fan-pm\endcsname{56.4}
\expandafter\def\csname res@pvit-d12-scannet200-cnnkd@iou-laptop\endcsname{55.0}
\expandafter\def\csname res@pvit-d12-scannet200-cnnkd@iou-laptop-pm\endcsname{55.0}
\expandafter\def\csname res@pvit-d12-scannet200-cnnkd@iou-shower\endcsname{32.7}
\expandafter\def\csname res@pvit-d12-scannet200-cnnkd@iou-shower-pm\endcsname{32.7}
\expandafter\def\csname res@pvit-d12-scannet200-cnnkd@iou-paper\endcsname{5.9}
\expandafter\def\csname res@pvit-d12-scannet200-cnnkd@iou-paper-pm\endcsname{5.9}
\expandafter\def\csname res@pvit-d12-scannet200-cnnkd@iou-person\endcsname{35.9}
\expandafter\def\csname res@pvit-d12-scannet200-cnnkd@iou-person-pm\endcsname{35.9}
\expandafter\def\csname res@pvit-d12-scannet200-cnnkd@iou-paper towel dispenser\endcsname{59.7}
\expandafter\def\csname res@pvit-d12-scannet200-cnnkd@iou-paper towel dispenser-pm\endcsname{59.7}
\expandafter\def\csname res@pvit-d12-scannet200-cnnkd@iou-oven\endcsname{16.2}
\expandafter\def\csname res@pvit-d12-scannet200-cnnkd@iou-oven-pm\endcsname{16.2}
\expandafter\def\csname res@pvit-d12-scannet200-cnnkd@iou-blinds\endcsname{0.0}
\expandafter\def\csname res@pvit-d12-scannet200-cnnkd@iou-blinds-pm\endcsname{0.0}
\expandafter\def\csname res@pvit-d12-scannet200-cnnkd@iou-rack\endcsname{0.0}
\expandafter\def\csname res@pvit-d12-scannet200-cnnkd@iou-rack-pm\endcsname{0.0}
\expandafter\def\csname res@pvit-d12-scannet200-cnnkd@iou-plate\endcsname{0.0}
\expandafter\def\csname res@pvit-d12-scannet200-cnnkd@iou-plate-pm\endcsname{0.0}
\expandafter\def\csname res@pvit-d12-scannet200-cnnkd@iou-blackboard\endcsname{85.6}
\expandafter\def\csname res@pvit-d12-scannet200-cnnkd@iou-blackboard-pm\endcsname{85.6}
\expandafter\def\csname res@pvit-d12-scannet200-cnnkd@iou-piano\endcsname{64.1}
\expandafter\def\csname res@pvit-d12-scannet200-cnnkd@iou-piano-pm\endcsname{64.1}
\expandafter\def\csname res@pvit-d12-scannet200-cnnkd@iou-suitcase\endcsname{74.3}
\expandafter\def\csname res@pvit-d12-scannet200-cnnkd@iou-suitcase-pm\endcsname{74.3}
\expandafter\def\csname res@pvit-d12-scannet200-cnnkd@iou-rail\endcsname{4.3}
\expandafter\def\csname res@pvit-d12-scannet200-cnnkd@iou-rail-pm\endcsname{4.3}
\expandafter\def\csname res@pvit-d12-scannet200-cnnkd@iou-radiator\endcsname{69.2}
\expandafter\def\csname res@pvit-d12-scannet200-cnnkd@iou-radiator-pm\endcsname{69.2}
\expandafter\def\csname res@pvit-d12-scannet200-cnnkd@iou-recycling bin\endcsname{51.7}
\expandafter\def\csname res@pvit-d12-scannet200-cnnkd@iou-recycling bin-pm\endcsname{51.7}
\expandafter\def\csname res@pvit-d12-scannet200-cnnkd@iou-container\endcsname{12.3}
\expandafter\def\csname res@pvit-d12-scannet200-cnnkd@iou-container-pm\endcsname{12.3}
\expandafter\def\csname res@pvit-d12-scannet200-cnnkd@iou-wardrobe\endcsname{29.6}
\expandafter\def\csname res@pvit-d12-scannet200-cnnkd@iou-wardrobe-pm\endcsname{29.6}
\expandafter\def\csname res@pvit-d12-scannet200-cnnkd@iou-soap dispenser\endcsname{76.4}
\expandafter\def\csname res@pvit-d12-scannet200-cnnkd@iou-soap dispenser-pm\endcsname{76.4}
\expandafter\def\csname res@pvit-d12-scannet200-cnnkd@iou-telephone\endcsname{39.6}
\expandafter\def\csname res@pvit-d12-scannet200-cnnkd@iou-telephone-pm\endcsname{39.6}
\expandafter\def\csname res@pvit-d12-scannet200-cnnkd@iou-bucket\endcsname{35.8}
\expandafter\def\csname res@pvit-d12-scannet200-cnnkd@iou-bucket-pm\endcsname{35.8}
\expandafter\def\csname res@pvit-d12-scannet200-cnnkd@iou-clock\endcsname{21.8}
\expandafter\def\csname res@pvit-d12-scannet200-cnnkd@iou-clock-pm\endcsname{21.8}
\expandafter\def\csname res@pvit-d12-scannet200-cnnkd@iou-stand\endcsname{0.0}
\expandafter\def\csname res@pvit-d12-scannet200-cnnkd@iou-stand-pm\endcsname{0.0}
\expandafter\def\csname res@pvit-d12-scannet200-cnnkd@iou-light\endcsname{5.1}
\expandafter\def\csname res@pvit-d12-scannet200-cnnkd@iou-light-pm\endcsname{5.1}
\expandafter\def\csname res@pvit-d12-scannet200-cnnkd@iou-laundry basket\endcsname{33.8}
\expandafter\def\csname res@pvit-d12-scannet200-cnnkd@iou-laundry basket-pm\endcsname{33.8}
\expandafter\def\csname res@pvit-d12-scannet200-cnnkd@iou-pipe\endcsname{2.7}
\expandafter\def\csname res@pvit-d12-scannet200-cnnkd@iou-pipe-pm\endcsname{2.7}
\expandafter\def\csname res@pvit-d12-scannet200-cnnkd@iou-clothes dryer\endcsname{67.8}
\expandafter\def\csname res@pvit-d12-scannet200-cnnkd@iou-clothes dryer-pm\endcsname{67.8}
\expandafter\def\csname res@pvit-d12-scannet200-cnnkd@iou-guitar\endcsname{61.2}
\expandafter\def\csname res@pvit-d12-scannet200-cnnkd@iou-guitar-pm\endcsname{61.2}
\expandafter\def\csname res@pvit-d12-scannet200-cnnkd@iou-toilet paper holder\endcsname{0.0}
\expandafter\def\csname res@pvit-d12-scannet200-cnnkd@iou-toilet paper holder-pm\endcsname{0.0}
\expandafter\def\csname res@pvit-d12-scannet200-cnnkd@iou-seat\endcsname{0.0}
\expandafter\def\csname res@pvit-d12-scannet200-cnnkd@iou-seat-pm\endcsname{0.0}
\expandafter\def\csname res@pvit-d12-scannet200-cnnkd@iou-speaker\endcsname{0.0}
\expandafter\def\csname res@pvit-d12-scannet200-cnnkd@iou-speaker-pm\endcsname{0.0}
\expandafter\def\csname res@pvit-d12-scannet200-cnnkd@iou-column\endcsname{0.0}
\expandafter\def\csname res@pvit-d12-scannet200-cnnkd@iou-column-pm\endcsname{0.0}
\expandafter\def\csname res@pvit-d12-scannet200-cnnkd@iou-bicycle\endcsname{0.0}
\expandafter\def\csname res@pvit-d12-scannet200-cnnkd@iou-bicycle-pm\endcsname{0.0}
\expandafter\def\csname res@pvit-d12-scannet200-cnnkd@iou-ladder\endcsname{70.4}
\expandafter\def\csname res@pvit-d12-scannet200-cnnkd@iou-ladder-pm\endcsname{70.4}
\expandafter\def\csname res@pvit-d12-scannet200-cnnkd@iou-bathroom stall\endcsname{58.4}
\expandafter\def\csname res@pvit-d12-scannet200-cnnkd@iou-bathroom stall-pm\endcsname{58.4}
\expandafter\def\csname res@pvit-d12-scannet200-cnnkd@iou-shower wall\endcsname{49.8}
\expandafter\def\csname res@pvit-d12-scannet200-cnnkd@iou-shower wall-pm\endcsname{49.8}
\expandafter\def\csname res@pvit-d12-scannet200-cnnkd@iou-cup\endcsname{24.1}
\expandafter\def\csname res@pvit-d12-scannet200-cnnkd@iou-cup-pm\endcsname{24.1}
\expandafter\def\csname res@pvit-d12-scannet200-cnnkd@iou-jacket\endcsname{18.0}
\expandafter\def\csname res@pvit-d12-scannet200-cnnkd@iou-jacket-pm\endcsname{18.0}
\expandafter\def\csname res@pvit-d12-scannet200-cnnkd@iou-storage bin\endcsname{0.1}
\expandafter\def\csname res@pvit-d12-scannet200-cnnkd@iou-storage bin-pm\endcsname{0.1}
\expandafter\def\csname res@pvit-d12-scannet200-cnnkd@iou-coffee maker\endcsname{86.1}
\expandafter\def\csname res@pvit-d12-scannet200-cnnkd@iou-coffee maker-pm\endcsname{86.1}
\expandafter\def\csname res@pvit-d12-scannet200-cnnkd@iou-dishwasher\endcsname{47.5}
\expandafter\def\csname res@pvit-d12-scannet200-cnnkd@iou-dishwasher-pm\endcsname{47.5}
\expandafter\def\csname res@pvit-d12-scannet200-cnnkd@iou-paper towel roll\endcsname{33.4}
\expandafter\def\csname res@pvit-d12-scannet200-cnnkd@iou-paper towel roll-pm\endcsname{33.4}
\expandafter\def\csname res@pvit-d12-scannet200-cnnkd@iou-machine\endcsname{0.0}
\expandafter\def\csname res@pvit-d12-scannet200-cnnkd@iou-machine-pm\endcsname{0.0}
\expandafter\def\csname res@pvit-d12-scannet200-cnnkd@iou-mat\endcsname{37.5}
\expandafter\def\csname res@pvit-d12-scannet200-cnnkd@iou-mat-pm\endcsname{37.5}
\expandafter\def\csname res@pvit-d12-scannet200-cnnkd@iou-windowsill\endcsname{12.8}
\expandafter\def\csname res@pvit-d12-scannet200-cnnkd@iou-windowsill-pm\endcsname{12.8}
\expandafter\def\csname res@pvit-d12-scannet200-cnnkd@iou-bar\endcsname{14.6}
\expandafter\def\csname res@pvit-d12-scannet200-cnnkd@iou-bar-pm\endcsname{14.6}
\expandafter\def\csname res@pvit-d12-scannet200-cnnkd@iou-toaster\endcsname{86.3}
\expandafter\def\csname res@pvit-d12-scannet200-cnnkd@iou-toaster-pm\endcsname{86.3}
\expandafter\def\csname res@pvit-d12-scannet200-cnnkd@iou-bulletin board\endcsname{37.1}
\expandafter\def\csname res@pvit-d12-scannet200-cnnkd@iou-bulletin board-pm\endcsname{37.1}
\expandafter\def\csname res@pvit-d12-scannet200-cnnkd@iou-ironing board\endcsname{34.5}
\expandafter\def\csname res@pvit-d12-scannet200-cnnkd@iou-ironing board-pm\endcsname{34.5}
\expandafter\def\csname res@pvit-d12-scannet200-cnnkd@iou-fireplace\endcsname{68.9}
\expandafter\def\csname res@pvit-d12-scannet200-cnnkd@iou-fireplace-pm\endcsname{68.9}
\expandafter\def\csname res@pvit-d12-scannet200-cnnkd@iou-soap dish\endcsname{39.9}
\expandafter\def\csname res@pvit-d12-scannet200-cnnkd@iou-soap dish-pm\endcsname{39.9}
\expandafter\def\csname res@pvit-d12-scannet200-cnnkd@iou-kitchen counter\endcsname{60.1}
\expandafter\def\csname res@pvit-d12-scannet200-cnnkd@iou-kitchen counter-pm\endcsname{60.1}
\expandafter\def\csname res@pvit-d12-scannet200-cnnkd@iou-doorframe\endcsname{47.2}
\expandafter\def\csname res@pvit-d12-scannet200-cnnkd@iou-doorframe-pm\endcsname{47.2}
\expandafter\def\csname res@pvit-d12-scannet200-cnnkd@iou-toilet paper dispenser\endcsname{55.1}
\expandafter\def\csname res@pvit-d12-scannet200-cnnkd@iou-toilet paper dispenser-pm\endcsname{55.1}
\expandafter\def\csname res@pvit-d12-scannet200-cnnkd@iou-mini fridge\endcsname{67.7}
\expandafter\def\csname res@pvit-d12-scannet200-cnnkd@iou-mini fridge-pm\endcsname{67.7}
\expandafter\def\csname res@pvit-d12-scannet200-cnnkd@iou-fire extinguisher\endcsname{21.9}
\expandafter\def\csname res@pvit-d12-scannet200-cnnkd@iou-fire extinguisher-pm\endcsname{21.9}
\expandafter\def\csname res@pvit-d12-scannet200-cnnkd@iou-ball\endcsname{77.2}
\expandafter\def\csname res@pvit-d12-scannet200-cnnkd@iou-ball-pm\endcsname{77.2}
\expandafter\def\csname res@pvit-d12-scannet200-cnnkd@iou-hat\endcsname{0.0}
\expandafter\def\csname res@pvit-d12-scannet200-cnnkd@iou-hat-pm\endcsname{0.0}
\expandafter\def\csname res@pvit-d12-scannet200-cnnkd@iou-shower curtain rod\endcsname{41.1}
\expandafter\def\csname res@pvit-d12-scannet200-cnnkd@iou-shower curtain rod-pm\endcsname{41.1}
\expandafter\def\csname res@pvit-d12-scannet200-cnnkd@iou-water cooler\endcsname{92.4}
\expandafter\def\csname res@pvit-d12-scannet200-cnnkd@iou-water cooler-pm\endcsname{92.4}
\expandafter\def\csname res@pvit-d12-scannet200-cnnkd@iou-paper cutter\endcsname{22.2}
\expandafter\def\csname res@pvit-d12-scannet200-cnnkd@iou-paper cutter-pm\endcsname{22.2}
\expandafter\def\csname res@pvit-d12-scannet200-cnnkd@iou-tray\endcsname{23.0}
\expandafter\def\csname res@pvit-d12-scannet200-cnnkd@iou-tray-pm\endcsname{23.0}
\expandafter\def\csname res@pvit-d12-scannet200-cnnkd@iou-shower door\endcsname{23.6}
\expandafter\def\csname res@pvit-d12-scannet200-cnnkd@iou-shower door-pm\endcsname{23.6}
\expandafter\def\csname res@pvit-d12-scannet200-cnnkd@iou-pillar\endcsname{7.2}
\expandafter\def\csname res@pvit-d12-scannet200-cnnkd@iou-pillar-pm\endcsname{7.2}
\expandafter\def\csname res@pvit-d12-scannet200-cnnkd@iou-ledge\endcsname{20.7}
\expandafter\def\csname res@pvit-d12-scannet200-cnnkd@iou-ledge-pm\endcsname{20.7}
\expandafter\def\csname res@pvit-d12-scannet200-cnnkd@iou-toaster oven\endcsname{3.0}
\expandafter\def\csname res@pvit-d12-scannet200-cnnkd@iou-toaster oven-pm\endcsname{3.0}
\expandafter\def\csname res@pvit-d12-scannet200-cnnkd@iou-mouse\endcsname{6.1}
\expandafter\def\csname res@pvit-d12-scannet200-cnnkd@iou-mouse-pm\endcsname{6.1}
\expandafter\def\csname res@pvit-d12-scannet200-cnnkd@iou-toilet seat cover dispenser\endcsname{83.3}
\expandafter\def\csname res@pvit-d12-scannet200-cnnkd@iou-toilet seat cover dispenser-pm\endcsname{83.3}
\expandafter\def\csname res@pvit-d12-scannet200-cnnkd@iou-furniture\endcsname{0.0}
\expandafter\def\csname res@pvit-d12-scannet200-cnnkd@iou-furniture-pm\endcsname{0.0}
\expandafter\def\csname res@pvit-d12-scannet200-cnnkd@iou-cart\endcsname{37.8}
\expandafter\def\csname res@pvit-d12-scannet200-cnnkd@iou-cart-pm\endcsname{37.8}
\expandafter\def\csname res@pvit-d12-scannet200-cnnkd@iou-storage container\endcsname{0.0}
\expandafter\def\csname res@pvit-d12-scannet200-cnnkd@iou-storage container-pm\endcsname{0.0}
\expandafter\def\csname res@pvit-d12-scannet200-cnnkd@iou-scale\endcsname{34.7}
\expandafter\def\csname res@pvit-d12-scannet200-cnnkd@iou-scale-pm\endcsname{34.7}
\expandafter\def\csname res@pvit-d12-scannet200-cnnkd@iou-tissue box\endcsname{16.2}
\expandafter\def\csname res@pvit-d12-scannet200-cnnkd@iou-tissue box-pm\endcsname{16.2}
\expandafter\def\csname res@pvit-d12-scannet200-cnnkd@iou-light switch\endcsname{0.0}
\expandafter\def\csname res@pvit-d12-scannet200-cnnkd@iou-light switch-pm\endcsname{0.0}
\expandafter\def\csname res@pvit-d12-scannet200-cnnkd@iou-crate\endcsname{0.0}
\expandafter\def\csname res@pvit-d12-scannet200-cnnkd@iou-crate-pm\endcsname{0.0}
\expandafter\def\csname res@pvit-d12-scannet200-cnnkd@iou-power outlet\endcsname{0.0}
\expandafter\def\csname res@pvit-d12-scannet200-cnnkd@iou-power outlet-pm\endcsname{0.0}
\expandafter\def\csname res@pvit-d12-scannet200-cnnkd@iou-decoration\endcsname{45.3}
\expandafter\def\csname res@pvit-d12-scannet200-cnnkd@iou-decoration-pm\endcsname{45.3}
\expandafter\def\csname res@pvit-d12-scannet200-cnnkd@iou-sign\endcsname{35.7}
\expandafter\def\csname res@pvit-d12-scannet200-cnnkd@iou-sign-pm\endcsname{35.7}
\expandafter\def\csname res@pvit-d12-scannet200-cnnkd@iou-projector\endcsname{6.7}
\expandafter\def\csname res@pvit-d12-scannet200-cnnkd@iou-projector-pm\endcsname{6.7}
\expandafter\def\csname res@pvit-d12-scannet200-cnnkd@iou-closet door\endcsname{2.5}
\expandafter\def\csname res@pvit-d12-scannet200-cnnkd@iou-closet door-pm\endcsname{2.5}
\expandafter\def\csname res@pvit-d12-scannet200-cnnkd@iou-vacuum cleaner\endcsname{87.7}
\expandafter\def\csname res@pvit-d12-scannet200-cnnkd@iou-vacuum cleaner-pm\endcsname{87.7}
\expandafter\def\csname res@pvit-d12-scannet200-cnnkd@iou-candle\endcsname{0.0}
\expandafter\def\csname res@pvit-d12-scannet200-cnnkd@iou-candle-pm\endcsname{0.0}
\expandafter\def\csname res@pvit-d12-scannet200-cnnkd@iou-plunger\endcsname{35.9}
\expandafter\def\csname res@pvit-d12-scannet200-cnnkd@iou-plunger-pm\endcsname{35.9}
\expandafter\def\csname res@pvit-d12-scannet200-cnnkd@iou-stuffed animal\endcsname{0.0}
\expandafter\def\csname res@pvit-d12-scannet200-cnnkd@iou-stuffed animal-pm\endcsname{0.0}
\expandafter\def\csname res@pvit-d12-scannet200-cnnkd@iou-headphones\endcsname{0.0}
\expandafter\def\csname res@pvit-d12-scannet200-cnnkd@iou-headphones-pm\endcsname{0.0}
\expandafter\def\csname res@pvit-d12-scannet200-cnnkd@iou-dish rack\endcsname{65.9}
\expandafter\def\csname res@pvit-d12-scannet200-cnnkd@iou-dish rack-pm\endcsname{65.9}
\expandafter\def\csname res@pvit-d12-scannet200-cnnkd@iou-broom\endcsname{7.9}
\expandafter\def\csname res@pvit-d12-scannet200-cnnkd@iou-broom-pm\endcsname{7.9}
\expandafter\def\csname res@pvit-d12-scannet200-cnnkd@iou-guitar case\endcsname{0.0}
\expandafter\def\csname res@pvit-d12-scannet200-cnnkd@iou-guitar case-pm\endcsname{0.0}
\expandafter\def\csname res@pvit-d12-scannet200-cnnkd@iou-range hood\endcsname{78.6}
\expandafter\def\csname res@pvit-d12-scannet200-cnnkd@iou-range hood-pm\endcsname{78.6}
\expandafter\def\csname res@pvit-d12-scannet200-cnnkd@iou-dustpan\endcsname{1.2}
\expandafter\def\csname res@pvit-d12-scannet200-cnnkd@iou-dustpan-pm\endcsname{1.2}
\expandafter\def\csname res@pvit-d12-scannet200-cnnkd@iou-hair dryer\endcsname{7.3}
\expandafter\def\csname res@pvit-d12-scannet200-cnnkd@iou-hair dryer-pm\endcsname{7.3}
\expandafter\def\csname res@pvit-d12-scannet200-cnnkd@iou-water bottle\endcsname{0.0}
\expandafter\def\csname res@pvit-d12-scannet200-cnnkd@iou-water bottle-pm\endcsname{0.0}
\expandafter\def\csname res@pvit-d12-scannet200-cnnkd@iou-handicap bar\endcsname{5.5}
\expandafter\def\csname res@pvit-d12-scannet200-cnnkd@iou-handicap bar-pm\endcsname{5.5}
\expandafter\def\csname res@pvit-d12-scannet200-cnnkd@iou-purse\endcsname{0.0}
\expandafter\def\csname res@pvit-d12-scannet200-cnnkd@iou-purse-pm\endcsname{0.0}
\expandafter\def\csname res@pvit-d12-scannet200-cnnkd@iou-vent\endcsname{0.0}
\expandafter\def\csname res@pvit-d12-scannet200-cnnkd@iou-vent-pm\endcsname{0.0}
\expandafter\def\csname res@pvit-d12-scannet200-cnnkd@iou-shower floor\endcsname{61.0}
\expandafter\def\csname res@pvit-d12-scannet200-cnnkd@iou-shower floor-pm\endcsname{61.0}
\expandafter\def\csname res@pvit-d12-scannet200-cnnkd@iou-water pitcher\endcsname{0.0}
\expandafter\def\csname res@pvit-d12-scannet200-cnnkd@iou-water pitcher-pm\endcsname{0.0}
\expandafter\def\csname res@pvit-d12-scannet200-cnnkd@iou-mailbox\endcsname{55.1}
\expandafter\def\csname res@pvit-d12-scannet200-cnnkd@iou-mailbox-pm\endcsname{55.1}
\expandafter\def\csname res@pvit-d12-scannet200-cnnkd@iou-bowl\endcsname{30.6}
\expandafter\def\csname res@pvit-d12-scannet200-cnnkd@iou-bowl-pm\endcsname{30.6}
\expandafter\def\csname res@pvit-d12-scannet200-cnnkd@iou-paper bag\endcsname{11.9}
\expandafter\def\csname res@pvit-d12-scannet200-cnnkd@iou-paper bag-pm\endcsname{11.9}
\expandafter\def\csname res@pvit-d12-scannet200-cnnkd@iou-alarm clock\endcsname{0.0}
\expandafter\def\csname res@pvit-d12-scannet200-cnnkd@iou-alarm clock-pm\endcsname{0.0}
\expandafter\def\csname res@pvit-d12-scannet200-cnnkd@iou-music stand\endcsname{0.0}
\expandafter\def\csname res@pvit-d12-scannet200-cnnkd@iou-music stand-pm\endcsname{0.0}
\expandafter\def\csname res@pvit-d12-scannet200-cnnkd@iou-projector screen\endcsname{1.6}
\expandafter\def\csname res@pvit-d12-scannet200-cnnkd@iou-projector screen-pm\endcsname{1.6}
\expandafter\def\csname res@pvit-d12-scannet200-cnnkd@iou-divider\endcsname{64.4}
\expandafter\def\csname res@pvit-d12-scannet200-cnnkd@iou-divider-pm\endcsname{64.4}
\expandafter\def\csname res@pvit-d12-scannet200-cnnkd@iou-laundry detergent\endcsname{0.0}
\expandafter\def\csname res@pvit-d12-scannet200-cnnkd@iou-laundry detergent-pm\endcsname{0.0}
\expandafter\def\csname res@pvit-d12-scannet200-cnnkd@iou-bathroom counter\endcsname{30.9}
\expandafter\def\csname res@pvit-d12-scannet200-cnnkd@iou-bathroom counter-pm\endcsname{30.9}
\expandafter\def\csname res@pvit-d12-scannet200-cnnkd@iou-object\endcsname{8.9}
\expandafter\def\csname res@pvit-d12-scannet200-cnnkd@iou-object-pm\endcsname{8.9}
\expandafter\def\csname res@pvit-d12-scannet200-cnnkd@iou-bathroom vanity\endcsname{54.1}
\expandafter\def\csname res@pvit-d12-scannet200-cnnkd@iou-bathroom vanity-pm\endcsname{54.1}
\expandafter\def\csname res@pvit-d12-scannet200-cnnkd@iou-closet wall\endcsname{7.5}
\expandafter\def\csname res@pvit-d12-scannet200-cnnkd@iou-closet wall-pm\endcsname{7.5}
\expandafter\def\csname res@pvit-d12-scannet200-cnnkd@iou-laundry hamper\endcsname{25.8}
\expandafter\def\csname res@pvit-d12-scannet200-cnnkd@iou-laundry hamper-pm\endcsname{25.8}
\expandafter\def\csname res@pvit-d12-scannet200-cnnkd@iou-bathroom stall door\endcsname{61.1}
\expandafter\def\csname res@pvit-d12-scannet200-cnnkd@iou-bathroom stall door-pm\endcsname{61.1}
\expandafter\def\csname res@pvit-d12-scannet200-cnnkd@iou-ceiling light\endcsname{63.0}
\expandafter\def\csname res@pvit-d12-scannet200-cnnkd@iou-ceiling light-pm\endcsname{63.0}
\expandafter\def\csname res@pvit-d12-scannet200-cnnkd@iou-trash bin\endcsname{55.6}
\expandafter\def\csname res@pvit-d12-scannet200-cnnkd@iou-trash bin-pm\endcsname{55.6}
\expandafter\def\csname res@pvit-d12-scannet200-cnnkd@iou-dumbbell\endcsname{59.2}
\expandafter\def\csname res@pvit-d12-scannet200-cnnkd@iou-dumbbell-pm\endcsname{59.2}
\expandafter\def\csname res@pvit-d12-scannet200-cnnkd@iou-stair rail\endcsname{25.5}
\expandafter\def\csname res@pvit-d12-scannet200-cnnkd@iou-stair rail-pm\endcsname{25.5}
\expandafter\def\csname res@pvit-d12-scannet200-cnnkd@iou-tube\endcsname{34.8}
\expandafter\def\csname res@pvit-d12-scannet200-cnnkd@iou-tube-pm\endcsname{34.8}
\expandafter\def\csname res@pvit-d12-scannet200-cnnkd@iou-bathroom cabinet\endcsname{54.5}
\expandafter\def\csname res@pvit-d12-scannet200-cnnkd@iou-bathroom cabinet-pm\endcsname{54.5}
\expandafter\def\csname res@pvit-d12-scannet200-cnnkd@iou-cd case\endcsname{0.0}
\expandafter\def\csname res@pvit-d12-scannet200-cnnkd@iou-cd case-pm\endcsname{0.0}
\expandafter\def\csname res@pvit-d12-scannet200-cnnkd@iou-closet rod\endcsname{29.0}
\expandafter\def\csname res@pvit-d12-scannet200-cnnkd@iou-closet rod-pm\endcsname{29.0}
\expandafter\def\csname res@pvit-d12-scannet200-cnnkd@iou-coffee kettle\endcsname{36.1}
\expandafter\def\csname res@pvit-d12-scannet200-cnnkd@iou-coffee kettle-pm\endcsname{36.1}
\expandafter\def\csname res@pvit-d12-scannet200-cnnkd@iou-structure\endcsname{0.0}
\expandafter\def\csname res@pvit-d12-scannet200-cnnkd@iou-structure-pm\endcsname{0.0}
\expandafter\def\csname res@pvit-d12-scannet200-cnnkd@iou-shower head\endcsname{29.5}
\expandafter\def\csname res@pvit-d12-scannet200-cnnkd@iou-shower head-pm\endcsname{29.5}
\expandafter\def\csname res@pvit-d12-scannet200-cnnkd@iou-keyboard piano\endcsname{0.0}
\expandafter\def\csname res@pvit-d12-scannet200-cnnkd@iou-keyboard piano-pm\endcsname{0.0}
\expandafter\def\csname res@pvit-d12-scannet200-cnnkd@iou-case of water bottles\endcsname{4.8}
\expandafter\def\csname res@pvit-d12-scannet200-cnnkd@iou-case of water bottles-pm\endcsname{4.8}
\expandafter\def\csname res@pvit-d12-scannet200-cnnkd@iou-coat rack\endcsname{24.3}
\expandafter\def\csname res@pvit-d12-scannet200-cnnkd@iou-coat rack-pm\endcsname{24.3}
\expandafter\def\csname res@pvit-d12-scannet200-cnnkd@iou-storage organizer\endcsname{0.0}
\expandafter\def\csname res@pvit-d12-scannet200-cnnkd@iou-storage organizer-pm\endcsname{0.0}
\expandafter\def\csname res@pvit-d12-scannet200-cnnkd@iou-folded chair\endcsname{36.4}
\expandafter\def\csname res@pvit-d12-scannet200-cnnkd@iou-folded chair-pm\endcsname{36.4}
\expandafter\def\csname res@pvit-d12-scannet200-cnnkd@iou-fire alarm\endcsname{7.8}
\expandafter\def\csname res@pvit-d12-scannet200-cnnkd@iou-fire alarm-pm\endcsname{7.8}
\expandafter\def\csname res@pvit-d12-scannet200-cnnkd@iou-power strip\endcsname{0.0}
\expandafter\def\csname res@pvit-d12-scannet200-cnnkd@iou-power strip-pm\endcsname{0.0}
\expandafter\def\csname res@pvit-d12-scannet200-cnnkd@iou-calendar\endcsname{0.0}
\expandafter\def\csname res@pvit-d12-scannet200-cnnkd@iou-calendar-pm\endcsname{0.0}
\expandafter\def\csname res@pvit-d12-scannet200-cnnkd@iou-poster\endcsname{0.0}
\expandafter\def\csname res@pvit-d12-scannet200-cnnkd@iou-poster-pm\endcsname{0.0}
\expandafter\def\csname res@pvit-d12-scannet200-cnnkd@iou-potted plant\endcsname{0.0}
\expandafter\def\csname res@pvit-d12-scannet200-cnnkd@iou-potted plant-pm\endcsname{0.0}
\expandafter\def\csname res@pvit-d12-scannet200-cnnkd@iou-luggage\endcsname{0.0}
\expandafter\def\csname res@pvit-d12-scannet200-cnnkd@iou-luggage-pm\endcsname{0.0}
\expandafter\def\csname res@pvit-d12-scannet200-cnnkd@iou-mattress\endcsname{15.7}
\expandafter\def\csname res@pvit-d12-scannet200-cnnkd@iou-mattress-pm\endcsname{15.7}
\expandafter\def\csname res@pvit-d12-scannet200-cnnkd@miou-delta\endcsname{+0.8}
\expandafter\def\csname res@pvit-d12-scannet200-cnnkd@macc-delta\endcsname{+1.9}
\expandafter\def\csname res@pvit-d12-scannet200-cnnkd@allacc-delta\endcsname{+0.5}
\expandafter\def\csname res@pvit-d12-scannet200-cnnkd@iou-wall-delta\endcsname{\ensuremath{-}0.2}
\expandafter\def\csname res@pvit-d12-scannet200-cnnkd@iou-chair-delta\endcsname{\ensuremath{-}0.5}
\expandafter\def\csname res@pvit-d12-scannet200-cnnkd@iou-floor-delta\endcsname{+0.0}
\expandafter\def\csname res@pvit-d12-scannet200-cnnkd@iou-table-delta\endcsname{+0.8}
\expandafter\def\csname res@pvit-d12-scannet200-cnnkd@iou-door-delta\endcsname{+3.5}
\expandafter\def\csname res@pvit-d12-scannet200-cnnkd@iou-couch-delta\endcsname{\ensuremath{-}0.3}
\expandafter\def\csname res@pvit-d12-scannet200-cnnkd@iou-cabinet-delta\endcsname{+0.3}
\expandafter\def\csname res@pvit-d12-scannet200-cnnkd@iou-shelf-delta\endcsname{+6.0}
\expandafter\def\csname res@pvit-d12-scannet200-cnnkd@iou-desk-delta\endcsname{+1.8}
\expandafter\def\csname res@pvit-d12-scannet200-cnnkd@iou-office chair-delta\endcsname{\ensuremath{-}10.7}
\expandafter\def\csname res@pvit-d12-scannet200-cnnkd@iou-bed-delta\endcsname{+0.4}
\expandafter\def\csname res@pvit-d12-scannet200-cnnkd@iou-pillow-delta\endcsname{+3.8}
\expandafter\def\csname res@pvit-d12-scannet200-cnnkd@iou-sink-delta\endcsname{+1.0}
\expandafter\def\csname res@pvit-d12-scannet200-cnnkd@iou-picture-delta\endcsname{+2.1}
\expandafter\def\csname res@pvit-d12-scannet200-cnnkd@iou-window-delta\endcsname{+1.0}
\expandafter\def\csname res@pvit-d12-scannet200-cnnkd@iou-toilet-delta\endcsname{+0.7}
\expandafter\def\csname res@pvit-d12-scannet200-cnnkd@iou-bookshelf-delta\endcsname{+4.1}
\expandafter\def\csname res@pvit-d12-scannet200-cnnkd@iou-monitor-delta\endcsname{+0.3}
\expandafter\def\csname res@pvit-d12-scannet200-cnnkd@iou-curtain-delta\endcsname{+1.8}
\expandafter\def\csname res@pvit-d12-scannet200-cnnkd@iou-book-delta\endcsname{+1.5}
\expandafter\def\csname res@pvit-d12-scannet200-cnnkd@iou-armchair-delta\endcsname{\ensuremath{-}6.2}
\expandafter\def\csname res@pvit-d12-scannet200-cnnkd@iou-coffee table-delta\endcsname{\ensuremath{-}0.3}
\expandafter\def\csname res@pvit-d12-scannet200-cnnkd@iou-box-delta\endcsname{\ensuremath{-}1.6}
\expandafter\def\csname res@pvit-d12-scannet200-cnnkd@iou-refrigerator-delta\endcsname{\ensuremath{-}5.3}
\expandafter\def\csname res@pvit-d12-scannet200-cnnkd@iou-lamp-delta\endcsname{+1.1}
\expandafter\def\csname res@pvit-d12-scannet200-cnnkd@iou-kitchen cabinet-delta\endcsname{+3.8}
\expandafter\def\csname res@pvit-d12-scannet200-cnnkd@iou-towel-delta\endcsname{+0.3}
\expandafter\def\csname res@pvit-d12-scannet200-cnnkd@iou-clothes-delta\endcsname{+0.6}
\expandafter\def\csname res@pvit-d12-scannet200-cnnkd@iou-tv-delta\endcsname{\ensuremath{-}0.7}
\expandafter\def\csname res@pvit-d12-scannet200-cnnkd@iou-nightstand-delta\endcsname{\ensuremath{-}1.5}
\expandafter\def\csname res@pvit-d12-scannet200-cnnkd@iou-counter-delta\endcsname{\ensuremath{-}1.2}
\expandafter\def\csname res@pvit-d12-scannet200-cnnkd@iou-dresser-delta\endcsname{\ensuremath{-}0.7}
\expandafter\def\csname res@pvit-d12-scannet200-cnnkd@iou-stool-delta\endcsname{\ensuremath{-}10.4}
\expandafter\def\csname res@pvit-d12-scannet200-cnnkd@iou-cushion-delta\endcsname{+0.0}
\expandafter\def\csname res@pvit-d12-scannet200-cnnkd@iou-plant-delta\endcsname{+3.0}
\expandafter\def\csname res@pvit-d12-scannet200-cnnkd@iou-ceiling-delta\endcsname{+0.5}
\expandafter\def\csname res@pvit-d12-scannet200-cnnkd@iou-bathtub-delta\endcsname{\ensuremath{-}0.5}
\expandafter\def\csname res@pvit-d12-scannet200-cnnkd@iou-end table-delta\endcsname{+12.5}
\expandafter\def\csname res@pvit-d12-scannet200-cnnkd@iou-dining table-delta\endcsname{\ensuremath{-}0.1}
\expandafter\def\csname res@pvit-d12-scannet200-cnnkd@iou-keyboard-delta\endcsname{\ensuremath{-}1.8}
\expandafter\def\csname res@pvit-d12-scannet200-cnnkd@iou-bag-delta\endcsname{+1.9}
\expandafter\def\csname res@pvit-d12-scannet200-cnnkd@iou-backpack-delta\endcsname{+3.5}
\expandafter\def\csname res@pvit-d12-scannet200-cnnkd@iou-toilet paper-delta\endcsname{\ensuremath{-}2.7}
\expandafter\def\csname res@pvit-d12-scannet200-cnnkd@iou-printer-delta\endcsname{+8.5}
\expandafter\def\csname res@pvit-d12-scannet200-cnnkd@iou-tv stand-delta\endcsname{\ensuremath{-}11.2}
\expandafter\def\csname res@pvit-d12-scannet200-cnnkd@iou-whiteboard-delta\endcsname{+0.6}
\expandafter\def\csname res@pvit-d12-scannet200-cnnkd@iou-blanket-delta\endcsname{+12.4}
\expandafter\def\csname res@pvit-d12-scannet200-cnnkd@iou-shower curtain-delta\endcsname{+5.6}
\expandafter\def\csname res@pvit-d12-scannet200-cnnkd@iou-trash can-delta\endcsname{+1.6}
\expandafter\def\csname res@pvit-d12-scannet200-cnnkd@iou-closet-delta\endcsname{+0.7}
\expandafter\def\csname res@pvit-d12-scannet200-cnnkd@iou-stairs-delta\endcsname{\ensuremath{-}9.1}
\expandafter\def\csname res@pvit-d12-scannet200-cnnkd@iou-microwave-delta\endcsname{+2.8}
\expandafter\def\csname res@pvit-d12-scannet200-cnnkd@iou-stove-delta\endcsname{+0.9}
\expandafter\def\csname res@pvit-d12-scannet200-cnnkd@iou-shoe-delta\endcsname{\ensuremath{-}1.1}
\expandafter\def\csname res@pvit-d12-scannet200-cnnkd@iou-computer tower-delta\endcsname{+4.8}
\expandafter\def\csname res@pvit-d12-scannet200-cnnkd@iou-bottle-delta\endcsname{+2.4}
\expandafter\def\csname res@pvit-d12-scannet200-cnnkd@iou-bin-delta\endcsname{+3.5}
\expandafter\def\csname res@pvit-d12-scannet200-cnnkd@iou-ottoman-delta\endcsname{+4.0}
\expandafter\def\csname res@pvit-d12-scannet200-cnnkd@iou-bench-delta\endcsname{+16.8}
\expandafter\def\csname res@pvit-d12-scannet200-cnnkd@iou-board-delta\endcsname{+0.6}
\expandafter\def\csname res@pvit-d12-scannet200-cnnkd@iou-washing machine-delta\endcsname{\ensuremath{-}2.7}
\expandafter\def\csname res@pvit-d12-scannet200-cnnkd@iou-mirror-delta\endcsname{\ensuremath{-}3.2}
\expandafter\def\csname res@pvit-d12-scannet200-cnnkd@iou-copier-delta\endcsname{\ensuremath{-}2.2}
\expandafter\def\csname res@pvit-d12-scannet200-cnnkd@iou-basket-delta\endcsname{+0.0}
\expandafter\def\csname res@pvit-d12-scannet200-cnnkd@iou-sofa chair-delta\endcsname{\ensuremath{-}2.8}
\expandafter\def\csname res@pvit-d12-scannet200-cnnkd@iou-file cabinet-delta\endcsname{\ensuremath{-}4.0}
\expandafter\def\csname res@pvit-d12-scannet200-cnnkd@iou-fan-delta\endcsname{\ensuremath{-}6.0}
\expandafter\def\csname res@pvit-d12-scannet200-cnnkd@iou-laptop-delta\endcsname{+4.0}
\expandafter\def\csname res@pvit-d12-scannet200-cnnkd@iou-shower-delta\endcsname{+5.9}
\expandafter\def\csname res@pvit-d12-scannet200-cnnkd@iou-paper-delta\endcsname{+2.0}
\expandafter\def\csname res@pvit-d12-scannet200-cnnkd@iou-person-delta\endcsname{+13.0}
\expandafter\def\csname res@pvit-d12-scannet200-cnnkd@iou-paper towel dispenser-delta\endcsname{+1.7}
\expandafter\def\csname res@pvit-d12-scannet200-cnnkd@iou-oven-delta\endcsname{+7.0}
\expandafter\def\csname res@pvit-d12-scannet200-cnnkd@iou-blinds-delta\endcsname{+0.0}
\expandafter\def\csname res@pvit-d12-scannet200-cnnkd@iou-rack-delta\endcsname{+0.0}
\expandafter\def\csname res@pvit-d12-scannet200-cnnkd@iou-plate-delta\endcsname{+0.0}
\expandafter\def\csname res@pvit-d12-scannet200-cnnkd@iou-blackboard-delta\endcsname{\ensuremath{-}4.0}
\expandafter\def\csname res@pvit-d12-scannet200-cnnkd@iou-piano-delta\endcsname{\ensuremath{-}25.8}
\expandafter\def\csname res@pvit-d12-scannet200-cnnkd@iou-suitcase-delta\endcsname{+14.2}
\expandafter\def\csname res@pvit-d12-scannet200-cnnkd@iou-rail-delta\endcsname{+0.7}
\expandafter\def\csname res@pvit-d12-scannet200-cnnkd@iou-radiator-delta\endcsname{+2.3}
\expandafter\def\csname res@pvit-d12-scannet200-cnnkd@iou-recycling bin-delta\endcsname{\ensuremath{-}3.1}
\expandafter\def\csname res@pvit-d12-scannet200-cnnkd@iou-container-delta\endcsname{+9.3}
\expandafter\def\csname res@pvit-d12-scannet200-cnnkd@iou-wardrobe-delta\endcsname{\ensuremath{-}14.2}
\expandafter\def\csname res@pvit-d12-scannet200-cnnkd@iou-soap dispenser-delta\endcsname{+5.1}
\expandafter\def\csname res@pvit-d12-scannet200-cnnkd@iou-telephone-delta\endcsname{+2.8}
\expandafter\def\csname res@pvit-d12-scannet200-cnnkd@iou-bucket-delta\endcsname{+2.7}
\expandafter\def\csname res@pvit-d12-scannet200-cnnkd@iou-clock-delta\endcsname{\ensuremath{-}2.8}
\expandafter\def\csname res@pvit-d12-scannet200-cnnkd@iou-stand-delta\endcsname{+0.0}
\expandafter\def\csname res@pvit-d12-scannet200-cnnkd@iou-light-delta\endcsname{\ensuremath{-}2.4}
\expandafter\def\csname res@pvit-d12-scannet200-cnnkd@iou-laundry basket-delta\endcsname{+24.8}
\expandafter\def\csname res@pvit-d12-scannet200-cnnkd@iou-pipe-delta\endcsname{\ensuremath{-}5.4}
\expandafter\def\csname res@pvit-d12-scannet200-cnnkd@iou-clothes dryer-delta\endcsname{\ensuremath{-}14.5}
\expandafter\def\csname res@pvit-d12-scannet200-cnnkd@iou-guitar-delta\endcsname{\ensuremath{-}22.2}
\expandafter\def\csname res@pvit-d12-scannet200-cnnkd@iou-toilet paper holder-delta\endcsname{+0.0}
\expandafter\def\csname res@pvit-d12-scannet200-cnnkd@iou-seat-delta\endcsname{+0.0}
\expandafter\def\csname res@pvit-d12-scannet200-cnnkd@iou-speaker-delta\endcsname{+0.0}
\expandafter\def\csname res@pvit-d12-scannet200-cnnkd@iou-column-delta\endcsname{+0.0}
\expandafter\def\csname res@pvit-d12-scannet200-cnnkd@iou-bicycle-delta\endcsname{+0.0}
\expandafter\def\csname res@pvit-d12-scannet200-cnnkd@iou-ladder-delta\endcsname{+20.2}
\expandafter\def\csname res@pvit-d12-scannet200-cnnkd@iou-bathroom stall-delta\endcsname{\ensuremath{-}2.8}
\expandafter\def\csname res@pvit-d12-scannet200-cnnkd@iou-shower wall-delta\endcsname{+1.9}
\expandafter\def\csname res@pvit-d12-scannet200-cnnkd@iou-cup-delta\endcsname{\ensuremath{-}1.0}
\expandafter\def\csname res@pvit-d12-scannet200-cnnkd@iou-jacket-delta\endcsname{\ensuremath{-}1.4}
\expandafter\def\csname res@pvit-d12-scannet200-cnnkd@iou-storage bin-delta\endcsname{\ensuremath{-}3.2}
\expandafter\def\csname res@pvit-d12-scannet200-cnnkd@iou-coffee maker-delta\endcsname{\ensuremath{-}5.3}
\expandafter\def\csname res@pvit-d12-scannet200-cnnkd@iou-dishwasher-delta\endcsname{+0.1}
\expandafter\def\csname res@pvit-d12-scannet200-cnnkd@iou-paper towel roll-delta\endcsname{\ensuremath{-}2.9}
\expandafter\def\csname res@pvit-d12-scannet200-cnnkd@iou-machine-delta\endcsname{+0.0}
\expandafter\def\csname res@pvit-d12-scannet200-cnnkd@iou-mat-delta\endcsname{+0.2}
\expandafter\def\csname res@pvit-d12-scannet200-cnnkd@iou-windowsill-delta\endcsname{\ensuremath{-}6.1}
\expandafter\def\csname res@pvit-d12-scannet200-cnnkd@iou-bar-delta\endcsname{\ensuremath{-}0.5}
\expandafter\def\csname res@pvit-d12-scannet200-cnnkd@iou-toaster-delta\endcsname{+22.4}
\expandafter\def\csname res@pvit-d12-scannet200-cnnkd@iou-bulletin board-delta\endcsname{\ensuremath{-}10.8}
\expandafter\def\csname res@pvit-d12-scannet200-cnnkd@iou-ironing board-delta\endcsname{\ensuremath{-}12.1}
\expandafter\def\csname res@pvit-d12-scannet200-cnnkd@iou-fireplace-delta\endcsname{+3.1}
\expandafter\def\csname res@pvit-d12-scannet200-cnnkd@iou-soap dish-delta\endcsname{+7.1}
\expandafter\def\csname res@pvit-d12-scannet200-cnnkd@iou-kitchen counter-delta\endcsname{+3.7}
\expandafter\def\csname res@pvit-d12-scannet200-cnnkd@iou-doorframe-delta\endcsname{+1.9}
\expandafter\def\csname res@pvit-d12-scannet200-cnnkd@iou-toilet paper dispenser-delta\endcsname{\ensuremath{-}2.1}
\expandafter\def\csname res@pvit-d12-scannet200-cnnkd@iou-mini fridge-delta\endcsname{+11.5}
\expandafter\def\csname res@pvit-d12-scannet200-cnnkd@iou-fire extinguisher-delta\endcsname{\ensuremath{-}10.7}
\expandafter\def\csname res@pvit-d12-scannet200-cnnkd@iou-ball-delta\endcsname{\ensuremath{-}2.5}
\expandafter\def\csname res@pvit-d12-scannet200-cnnkd@iou-hat-delta\endcsname{+0.0}
\expandafter\def\csname res@pvit-d12-scannet200-cnnkd@iou-shower curtain rod-delta\endcsname{\ensuremath{-}8.3}
\expandafter\def\csname res@pvit-d12-scannet200-cnnkd@iou-water cooler-delta\endcsname{+4.3}
\expandafter\def\csname res@pvit-d12-scannet200-cnnkd@iou-paper cutter-delta\endcsname{\ensuremath{-}0.2}
\expandafter\def\csname res@pvit-d12-scannet200-cnnkd@iou-tray-delta\endcsname{+15.4}
\expandafter\def\csname res@pvit-d12-scannet200-cnnkd@iou-shower door-delta\endcsname{\ensuremath{-}3.0}
\expandafter\def\csname res@pvit-d12-scannet200-cnnkd@iou-pillar-delta\endcsname{+0.8}
\expandafter\def\csname res@pvit-d12-scannet200-cnnkd@iou-ledge-delta\endcsname{+11.4}
\expandafter\def\csname res@pvit-d12-scannet200-cnnkd@iou-toaster oven-delta\endcsname{+2.0}
\expandafter\def\csname res@pvit-d12-scannet200-cnnkd@iou-mouse-delta\endcsname{+0.5}
\expandafter\def\csname res@pvit-d12-scannet200-cnnkd@iou-toilet seat cover dispenser-delta\endcsname{+0.9}
\expandafter\def\csname res@pvit-d12-scannet200-cnnkd@iou-furniture-delta\endcsname{+0.0}
\expandafter\def\csname res@pvit-d12-scannet200-cnnkd@iou-cart-delta\endcsname{+19.7}
\expandafter\def\csname res@pvit-d12-scannet200-cnnkd@iou-storage container-delta\endcsname{+0.0}
\expandafter\def\csname res@pvit-d12-scannet200-cnnkd@iou-scale-delta\endcsname{\ensuremath{-}3.4}
\expandafter\def\csname res@pvit-d12-scannet200-cnnkd@iou-tissue box-delta\endcsname{\ensuremath{-}3.3}
\expandafter\def\csname res@pvit-d12-scannet200-cnnkd@iou-light switch-delta\endcsname{+0.0}
\expandafter\def\csname res@pvit-d12-scannet200-cnnkd@iou-crate-delta\endcsname{+0.0}
\expandafter\def\csname res@pvit-d12-scannet200-cnnkd@iou-power outlet-delta\endcsname{+0.0}
\expandafter\def\csname res@pvit-d12-scannet200-cnnkd@iou-decoration-delta\endcsname{+3.2}
\expandafter\def\csname res@pvit-d12-scannet200-cnnkd@iou-sign-delta\endcsname{\ensuremath{-}1.6}
\expandafter\def\csname res@pvit-d12-scannet200-cnnkd@iou-projector-delta\endcsname{+6.7}
\expandafter\def\csname res@pvit-d12-scannet200-cnnkd@iou-closet door-delta\endcsname{\ensuremath{-}5.7}
\expandafter\def\csname res@pvit-d12-scannet200-cnnkd@iou-vacuum cleaner-delta\endcsname{+2.9}
\expandafter\def\csname res@pvit-d12-scannet200-cnnkd@iou-candle-delta\endcsname{+0.0}
\expandafter\def\csname res@pvit-d12-scannet200-cnnkd@iou-plunger-delta\endcsname{+7.1}
\expandafter\def\csname res@pvit-d12-scannet200-cnnkd@iou-stuffed animal-delta\endcsname{+0.0}
\expandafter\def\csname res@pvit-d12-scannet200-cnnkd@iou-headphones-delta\endcsname{+0.0}
\expandafter\def\csname res@pvit-d12-scannet200-cnnkd@iou-dish rack-delta\endcsname{\ensuremath{-}25.4}
\expandafter\def\csname res@pvit-d12-scannet200-cnnkd@iou-broom-delta\endcsname{+5.8}
\expandafter\def\csname res@pvit-d12-scannet200-cnnkd@iou-guitar case-delta\endcsname{+0.0}
\expandafter\def\csname res@pvit-d12-scannet200-cnnkd@iou-range hood-delta\endcsname{+4.5}
\expandafter\def\csname res@pvit-d12-scannet200-cnnkd@iou-dustpan-delta\endcsname{\ensuremath{-}0.5}
\expandafter\def\csname res@pvit-d12-scannet200-cnnkd@iou-hair dryer-delta\endcsname{\ensuremath{-}16.5}
\expandafter\def\csname res@pvit-d12-scannet200-cnnkd@iou-water bottle-delta\endcsname{+0.0}
\expandafter\def\csname res@pvit-d12-scannet200-cnnkd@iou-handicap bar-delta\endcsname{\ensuremath{-}0.6}
\expandafter\def\csname res@pvit-d12-scannet200-cnnkd@iou-purse-delta\endcsname{+0.0}
\expandafter\def\csname res@pvit-d12-scannet200-cnnkd@iou-vent-delta\endcsname{\ensuremath{-}30.5}
\expandafter\def\csname res@pvit-d12-scannet200-cnnkd@iou-shower floor-delta\endcsname{+22.5}
\expandafter\def\csname res@pvit-d12-scannet200-cnnkd@iou-water pitcher-delta\endcsname{+0.0}
\expandafter\def\csname res@pvit-d12-scannet200-cnnkd@iou-mailbox-delta\endcsname{+6.5}
\expandafter\def\csname res@pvit-d12-scannet200-cnnkd@iou-bowl-delta\endcsname{+22.9}
\expandafter\def\csname res@pvit-d12-scannet200-cnnkd@iou-paper bag-delta\endcsname{+5.7}
\expandafter\def\csname res@pvit-d12-scannet200-cnnkd@iou-alarm clock-delta\endcsname{+0.0}
\expandafter\def\csname res@pvit-d12-scannet200-cnnkd@iou-music stand-delta\endcsname{+0.0}
\expandafter\def\csname res@pvit-d12-scannet200-cnnkd@iou-projector screen-delta\endcsname{+1.6}
\expandafter\def\csname res@pvit-d12-scannet200-cnnkd@iou-divider-delta\endcsname{\ensuremath{-}3.0}
\expandafter\def\csname res@pvit-d12-scannet200-cnnkd@iou-laundry detergent-delta\endcsname{+0.0}
\expandafter\def\csname res@pvit-d12-scannet200-cnnkd@iou-bathroom counter-delta\endcsname{+4.0}
\expandafter\def\csname res@pvit-d12-scannet200-cnnkd@iou-object-delta\endcsname{+1.9}
\expandafter\def\csname res@pvit-d12-scannet200-cnnkd@iou-bathroom vanity-delta\endcsname{+5.0}
\expandafter\def\csname res@pvit-d12-scannet200-cnnkd@iou-closet wall-delta\endcsname{\ensuremath{-}3.4}
\expandafter\def\csname res@pvit-d12-scannet200-cnnkd@iou-laundry hamper-delta\endcsname{\ensuremath{-}15.5}
\expandafter\def\csname res@pvit-d12-scannet200-cnnkd@iou-bathroom stall door-delta\endcsname{+1.0}
\expandafter\def\csname res@pvit-d12-scannet200-cnnkd@iou-ceiling light-delta\endcsname{+21.2}
\expandafter\def\csname res@pvit-d12-scannet200-cnnkd@iou-trash bin-delta\endcsname{+23.7}
\expandafter\def\csname res@pvit-d12-scannet200-cnnkd@iou-dumbbell-delta\endcsname{\ensuremath{-}28.2}
\expandafter\def\csname res@pvit-d12-scannet200-cnnkd@iou-stair rail-delta\endcsname{\ensuremath{-}6.8}
\expandafter\def\csname res@pvit-d12-scannet200-cnnkd@iou-tube-delta\endcsname{\ensuremath{-}0.6}
\expandafter\def\csname res@pvit-d12-scannet200-cnnkd@iou-bathroom cabinet-delta\endcsname{+23.6}
\expandafter\def\csname res@pvit-d12-scannet200-cnnkd@iou-cd case-delta\endcsname{+0.0}
\expandafter\def\csname res@pvit-d12-scannet200-cnnkd@iou-closet rod-delta\endcsname{+4.6}
\expandafter\def\csname res@pvit-d12-scannet200-cnnkd@iou-coffee kettle-delta\endcsname{\ensuremath{-}3.3}
\expandafter\def\csname res@pvit-d12-scannet200-cnnkd@iou-structure-delta\endcsname{+0.0}
\expandafter\def\csname res@pvit-d12-scannet200-cnnkd@iou-shower head-delta\endcsname{\ensuremath{-}4.7}
\expandafter\def\csname res@pvit-d12-scannet200-cnnkd@iou-keyboard piano-delta\endcsname{+0.0}
\expandafter\def\csname res@pvit-d12-scannet200-cnnkd@iou-case of water bottles-delta\endcsname{\ensuremath{-}30.5}
\expandafter\def\csname res@pvit-d12-scannet200-cnnkd@iou-coat rack-delta\endcsname{+24.3}
\expandafter\def\csname res@pvit-d12-scannet200-cnnkd@iou-storage organizer-delta\endcsname{+0.0}
\expandafter\def\csname res@pvit-d12-scannet200-cnnkd@iou-folded chair-delta\endcsname{+36.4}
\expandafter\def\csname res@pvit-d12-scannet200-cnnkd@iou-fire alarm-delta\endcsname{+7.8}
\expandafter\def\csname res@pvit-d12-scannet200-cnnkd@iou-power strip-delta\endcsname{+0.0}
\expandafter\def\csname res@pvit-d12-scannet200-cnnkd@iou-calendar-delta\endcsname{+0.0}
\expandafter\def\csname res@pvit-d12-scannet200-cnnkd@iou-poster-delta\endcsname{+0.0}
\expandafter\def\csname res@pvit-d12-scannet200-cnnkd@iou-potted plant-delta\endcsname{+0.0}
\expandafter\def\csname res@pvit-d12-scannet200-cnnkd@iou-luggage-delta\endcsname{+0.0}
\expandafter\def\csname res@pvit-d12-scannet200-cnnkd@iou-mattress-delta\endcsname{+12.4}
\expandafter\def\csname res@pvit-d12-nuscenes-cnnkd@n\endcsname{1}
\expandafter\def\csname res@pvit-d12-nuscenes-cnnkd@miou\endcsname{82.0}
\expandafter\def\csname res@pvit-d12-nuscenes-cnnkd@miou-pm\endcsname{82.0}
\expandafter\def\csname res@pvit-d12-nuscenes-cnnkd@macc\endcsname{88.8}
\expandafter\def\csname res@pvit-d12-nuscenes-cnnkd@macc-pm\endcsname{88.8}
\expandafter\def\csname res@pvit-d12-nuscenes-cnnkd@allacc\endcsname{95.0}
\expandafter\def\csname res@pvit-d12-nuscenes-cnnkd@allacc-pm\endcsname{95.0}
\expandafter\def\csname res@pvit-d12-nuscenes-cnnkd@iou-barrier\endcsname{81.1}
\expandafter\def\csname res@pvit-d12-nuscenes-cnnkd@iou-barrier-pm\endcsname{81.1}
\expandafter\def\csname res@pvit-d12-nuscenes-cnnkd@iou-bicycle\endcsname{55.1}
\expandafter\def\csname res@pvit-d12-nuscenes-cnnkd@iou-bicycle-pm\endcsname{55.1}
\expandafter\def\csname res@pvit-d12-nuscenes-cnnkd@iou-bus\endcsname{97.4}
\expandafter\def\csname res@pvit-d12-nuscenes-cnnkd@iou-bus-pm\endcsname{97.4}
\expandafter\def\csname res@pvit-d12-nuscenes-cnnkd@iou-car\endcsname{94.0}
\expandafter\def\csname res@pvit-d12-nuscenes-cnnkd@iou-car-pm\endcsname{94.0}
\expandafter\def\csname res@pvit-d12-nuscenes-cnnkd@iou-construction_vehicle\endcsname{63.7}
\expandafter\def\csname res@pvit-d12-nuscenes-cnnkd@iou-construction_vehicle-pm\endcsname{63.7}
\expandafter\def\csname res@pvit-d12-nuscenes-cnnkd@iou-motorcycle\endcsname{90.5}
\expandafter\def\csname res@pvit-d12-nuscenes-cnnkd@iou-motorcycle-pm\endcsname{90.5}
\expandafter\def\csname res@pvit-d12-nuscenes-cnnkd@iou-pedestrian\endcsname{84.9}
\expandafter\def\csname res@pvit-d12-nuscenes-cnnkd@iou-pedestrian-pm\endcsname{84.9}
\expandafter\def\csname res@pvit-d12-nuscenes-cnnkd@iou-traffic_cone\endcsname{71.5}
\expandafter\def\csname res@pvit-d12-nuscenes-cnnkd@iou-traffic_cone-pm\endcsname{71.5}
\expandafter\def\csname res@pvit-d12-nuscenes-cnnkd@iou-trailer\endcsname{73.9}
\expandafter\def\csname res@pvit-d12-nuscenes-cnnkd@iou-trailer-pm\endcsname{73.9}
\expandafter\def\csname res@pvit-d12-nuscenes-cnnkd@iou-truck\endcsname{87.8}
\expandafter\def\csname res@pvit-d12-nuscenes-cnnkd@iou-truck-pm\endcsname{87.8}
\expandafter\def\csname res@pvit-d12-nuscenes-cnnkd@iou-driveable_surface\endcsname{97.2}
\expandafter\def\csname res@pvit-d12-nuscenes-cnnkd@iou-driveable_surface-pm\endcsname{97.2}
\expandafter\def\csname res@pvit-d12-nuscenes-cnnkd@iou-other_flat\endcsname{77.0}
\expandafter\def\csname res@pvit-d12-nuscenes-cnnkd@iou-other_flat-pm\endcsname{77.0}
\expandafter\def\csname res@pvit-d12-nuscenes-cnnkd@iou-sidewalk\endcsname{78.0}
\expandafter\def\csname res@pvit-d12-nuscenes-cnnkd@iou-sidewalk-pm\endcsname{78.0}
\expandafter\def\csname res@pvit-d12-nuscenes-cnnkd@iou-terrain\endcsname{77.3}
\expandafter\def\csname res@pvit-d12-nuscenes-cnnkd@iou-terrain-pm\endcsname{77.3}
\expandafter\def\csname res@pvit-d12-nuscenes-cnnkd@iou-manmade\endcsname{91.8}
\expandafter\def\csname res@pvit-d12-nuscenes-cnnkd@iou-manmade-pm\endcsname{91.8}
\expandafter\def\csname res@pvit-d12-nuscenes-cnnkd@iou-vegetation\endcsname{90.3}
\expandafter\def\csname res@pvit-d12-nuscenes-cnnkd@iou-vegetation-pm\endcsname{90.3}
\expandafter\def\csname res@pvit-d12-nuscenes-cnnkd@miou-delta\endcsname{+0.4}
\expandafter\def\csname res@pvit-d12-nuscenes-cnnkd@macc-delta\endcsname{+0.3}
\expandafter\def\csname res@pvit-d12-nuscenes-cnnkd@allacc-delta\endcsname{+0.1}
\expandafter\def\csname res@pvit-d12-nuscenes-cnnkd@iou-barrier-delta\endcsname{+0.1}
\expandafter\def\csname res@pvit-d12-nuscenes-cnnkd@iou-bicycle-delta\endcsname{+2.7}
\expandafter\def\csname res@pvit-d12-nuscenes-cnnkd@iou-bus-delta\endcsname{+0.2}
\expandafter\def\csname res@pvit-d12-nuscenes-cnnkd@iou-car-delta\endcsname{+0.2}
\expandafter\def\csname res@pvit-d12-nuscenes-cnnkd@iou-construction_vehicle-delta\endcsname{\ensuremath{-}4.6}
\expandafter\def\csname res@pvit-d12-nuscenes-cnnkd@iou-motorcycle-delta\endcsname{+2.6}
\expandafter\def\csname res@pvit-d12-nuscenes-cnnkd@iou-pedestrian-delta\endcsname{+0.2}
\expandafter\def\csname res@pvit-d12-nuscenes-cnnkd@iou-traffic_cone-delta\endcsname{\ensuremath{-}0.2}
\expandafter\def\csname res@pvit-d12-nuscenes-cnnkd@iou-trailer-delta\endcsname{+2.5}
\expandafter\def\csname res@pvit-d12-nuscenes-cnnkd@iou-truck-delta\endcsname{\ensuremath{-}0.1}
\expandafter\def\csname res@pvit-d12-nuscenes-cnnkd@iou-driveable_surface-delta\endcsname{+0.0}
\expandafter\def\csname res@pvit-d12-nuscenes-cnnkd@iou-other_flat-delta\endcsname{+1.2}
\expandafter\def\csname res@pvit-d12-nuscenes-cnnkd@iou-sidewalk-delta\endcsname{+0.3}
\expandafter\def\csname res@pvit-d12-nuscenes-cnnkd@iou-terrain-delta\endcsname{+0.6}
\expandafter\def\csname res@pvit-d12-nuscenes-cnnkd@iou-manmade-delta\endcsname{+0.0}
\expandafter\def\csname res@pvit-d12-nuscenes-cnnkd@iou-vegetation-delta\endcsname{+0.0}
\expandafter\def\csname res@pvit-d12-semantickitti-voltaug@n\endcsname{1}
\expandafter\def\csname res@pvit-d12-semantickitti-voltaug@miou\endcsname{70.6}
\expandafter\def\csname res@pvit-d12-semantickitti-voltaug@miou-pm\endcsname{70.6}
\expandafter\def\csname res@pvit-d12-semantickitti-voltaug@macc\endcsname{77.8}
\expandafter\def\csname res@pvit-d12-semantickitti-voltaug@macc-pm\endcsname{77.8}
\expandafter\def\csname res@pvit-d12-semantickitti-voltaug@allacc\endcsname{92.9}
\expandafter\def\csname res@pvit-d12-semantickitti-voltaug@allacc-pm\endcsname{92.9}
\expandafter\def\csname res@pvit-d12-semantickitti-voltaug@iou-car\endcsname{97.2}
\expandafter\def\csname res@pvit-d12-semantickitti-voltaug@iou-car-pm\endcsname{97.2}
\expandafter\def\csname res@pvit-d12-semantickitti-voltaug@iou-bicycle\endcsname{58.8}
\expandafter\def\csname res@pvit-d12-semantickitti-voltaug@iou-bicycle-pm\endcsname{58.8}
\expandafter\def\csname res@pvit-d12-semantickitti-voltaug@iou-motorcycle\endcsname{82.7}
\expandafter\def\csname res@pvit-d12-semantickitti-voltaug@iou-motorcycle-pm\endcsname{82.7}
\expandafter\def\csname res@pvit-d12-semantickitti-voltaug@iou-truck\endcsname{88.0}
\expandafter\def\csname res@pvit-d12-semantickitti-voltaug@iou-truck-pm\endcsname{88.0}
\expandafter\def\csname res@pvit-d12-semantickitti-voltaug@iou-other-vehicle\endcsname{73.2}
\expandafter\def\csname res@pvit-d12-semantickitti-voltaug@iou-other-vehicle-pm\endcsname{73.2}
\expandafter\def\csname res@pvit-d12-semantickitti-voltaug@iou-person\endcsname{82.4}
\expandafter\def\csname res@pvit-d12-semantickitti-voltaug@iou-person-pm\endcsname{82.4}
\expandafter\def\csname res@pvit-d12-semantickitti-voltaug@iou-bicyclist\endcsname{91.6}
\expandafter\def\csname res@pvit-d12-semantickitti-voltaug@iou-bicyclist-pm\endcsname{91.6}
\expandafter\def\csname res@pvit-d12-semantickitti-voltaug@iou-motorcyclist\endcsname{9.4}
\expandafter\def\csname res@pvit-d12-semantickitti-voltaug@iou-motorcyclist-pm\endcsname{9.4}
\expandafter\def\csname res@pvit-d12-semantickitti-voltaug@iou-road\endcsname{96.0}
\expandafter\def\csname res@pvit-d12-semantickitti-voltaug@iou-road-pm\endcsname{96.0}
\expandafter\def\csname res@pvit-d12-semantickitti-voltaug@iou-parking\endcsname{56.6}
\expandafter\def\csname res@pvit-d12-semantickitti-voltaug@iou-parking-pm\endcsname{56.6}
\expandafter\def\csname res@pvit-d12-semantickitti-voltaug@iou-sidewalk\endcsname{84.6}
\expandafter\def\csname res@pvit-d12-semantickitti-voltaug@iou-sidewalk-pm\endcsname{84.6}
\expandafter\def\csname res@pvit-d12-semantickitti-voltaug@iou-other-ground\endcsname{1.1}
\expandafter\def\csname res@pvit-d12-semantickitti-voltaug@iou-other-ground-pm\endcsname{1.1}
\expandafter\def\csname res@pvit-d12-semantickitti-voltaug@iou-building\endcsname{92.2}
\expandafter\def\csname res@pvit-d12-semantickitti-voltaug@iou-building-pm\endcsname{92.2}
\expandafter\def\csname res@pvit-d12-semantickitti-voltaug@iou-fence\endcsname{67.2}
\expandafter\def\csname res@pvit-d12-semantickitti-voltaug@iou-fence-pm\endcsname{67.2}
\expandafter\def\csname res@pvit-d12-semantickitti-voltaug@iou-vegetation\endcsname{88.2}
\expandafter\def\csname res@pvit-d12-semantickitti-voltaug@iou-vegetation-pm\endcsname{88.2}
\expandafter\def\csname res@pvit-d12-semantickitti-voltaug@iou-trunk\endcsname{75.2}
\expandafter\def\csname res@pvit-d12-semantickitti-voltaug@iou-trunk-pm\endcsname{75.2}
\expandafter\def\csname res@pvit-d12-semantickitti-voltaug@iou-terrain\endcsname{74.0}
\expandafter\def\csname res@pvit-d12-semantickitti-voltaug@iou-terrain-pm\endcsname{74.0}
\expandafter\def\csname res@pvit-d12-semantickitti-voltaug@iou-pole\endcsname{68.4}
\expandafter\def\csname res@pvit-d12-semantickitti-voltaug@iou-pole-pm\endcsname{68.4}
\expandafter\def\csname res@pvit-d12-semantickitti-voltaug@iou-traffic-sign\endcsname{54.1}
\expandafter\def\csname res@pvit-d12-semantickitti-voltaug@iou-traffic-sign-pm\endcsname{54.1}
\expandafter\def\csname res@pvit-d12-semantickitti-cnnkd@n\endcsname{1}
\expandafter\def\csname res@pvit-d12-semantickitti-cnnkd@miou\endcsname{71.4}
\expandafter\def\csname res@pvit-d12-semantickitti-cnnkd@miou-pm\endcsname{71.4}
\expandafter\def\csname res@pvit-d12-semantickitti-cnnkd@macc\endcsname{80.2}
\expandafter\def\csname res@pvit-d12-semantickitti-cnnkd@macc-pm\endcsname{80.2}
\expandafter\def\csname res@pvit-d12-semantickitti-cnnkd@allacc\endcsname{93.6}
\expandafter\def\csname res@pvit-d12-semantickitti-cnnkd@allacc-pm\endcsname{93.6}
\expandafter\def\csname res@pvit-d12-semantickitti-cnnkd@iou-car\endcsname{95.7}
\expandafter\def\csname res@pvit-d12-semantickitti-cnnkd@iou-car-pm\endcsname{95.7}
\expandafter\def\csname res@pvit-d12-semantickitti-cnnkd@iou-bicycle\endcsname{53.1}
\expandafter\def\csname res@pvit-d12-semantickitti-cnnkd@iou-bicycle-pm\endcsname{53.1}
\expandafter\def\csname res@pvit-d12-semantickitti-cnnkd@iou-motorcycle\endcsname{78.3}
\expandafter\def\csname res@pvit-d12-semantickitti-cnnkd@iou-motorcycle-pm\endcsname{78.3}
\expandafter\def\csname res@pvit-d12-semantickitti-cnnkd@iou-truck\endcsname{88.8}
\expandafter\def\csname res@pvit-d12-semantickitti-cnnkd@iou-truck-pm\endcsname{88.8}
\expandafter\def\csname res@pvit-d12-semantickitti-cnnkd@iou-other-vehicle\endcsname{58.7}
\expandafter\def\csname res@pvit-d12-semantickitti-cnnkd@iou-other-vehicle-pm\endcsname{58.7}
\expandafter\def\csname res@pvit-d12-semantickitti-cnnkd@iou-person\endcsname{76.0}
\expandafter\def\csname res@pvit-d12-semantickitti-cnnkd@iou-person-pm\endcsname{76.0}
\expandafter\def\csname res@pvit-d12-semantickitti-cnnkd@iou-bicyclist\endcsname{88.0}
\expandafter\def\csname res@pvit-d12-semantickitti-cnnkd@iou-bicyclist-pm\endcsname{88.0}
\expandafter\def\csname res@pvit-d12-semantickitti-cnnkd@iou-motorcyclist\endcsname{48.3}
\expandafter\def\csname res@pvit-d12-semantickitti-cnnkd@iou-motorcyclist-pm\endcsname{48.3}
\expandafter\def\csname res@pvit-d12-semantickitti-cnnkd@iou-road\endcsname{95.5}
\expandafter\def\csname res@pvit-d12-semantickitti-cnnkd@iou-road-pm\endcsname{95.5}
\expandafter\def\csname res@pvit-d12-semantickitti-cnnkd@iou-parking\endcsname{57.6}
\expandafter\def\csname res@pvit-d12-semantickitti-cnnkd@iou-parking-pm\endcsname{57.6}
\expandafter\def\csname res@pvit-d12-semantickitti-cnnkd@iou-sidewalk\endcsname{83.6}
\expandafter\def\csname res@pvit-d12-semantickitti-cnnkd@iou-sidewalk-pm\endcsname{83.6}
\expandafter\def\csname res@pvit-d12-semantickitti-cnnkd@iou-other-ground\endcsname{4.2}
\expandafter\def\csname res@pvit-d12-semantickitti-cnnkd@iou-other-ground-pm\endcsname{4.2}
\expandafter\def\csname res@pvit-d12-semantickitti-cnnkd@iou-building\endcsname{92.1}
\expandafter\def\csname res@pvit-d12-semantickitti-cnnkd@iou-building-pm\endcsname{92.1}
\expandafter\def\csname res@pvit-d12-semantickitti-cnnkd@iou-fence\endcsname{68.5}
\expandafter\def\csname res@pvit-d12-semantickitti-cnnkd@iou-fence-pm\endcsname{68.5}
\expandafter\def\csname res@pvit-d12-semantickitti-cnnkd@iou-vegetation\endcsname{90.7}
\expandafter\def\csname res@pvit-d12-semantickitti-cnnkd@iou-vegetation-pm\endcsname{90.7}
\expandafter\def\csname res@pvit-d12-semantickitti-cnnkd@iou-trunk\endcsname{74.5}
\expandafter\def\csname res@pvit-d12-semantickitti-cnnkd@iou-trunk-pm\endcsname{74.5}
\expandafter\def\csname res@pvit-d12-semantickitti-cnnkd@iou-terrain\endcsname{80.7}
\expandafter\def\csname res@pvit-d12-semantickitti-cnnkd@iou-terrain-pm\endcsname{80.7}
\expandafter\def\csname res@pvit-d12-semantickitti-cnnkd@iou-pole\endcsname{67.3}
\expandafter\def\csname res@pvit-d12-semantickitti-cnnkd@iou-pole-pm\endcsname{67.3}
\expandafter\def\csname res@pvit-d12-semantickitti-cnnkd@iou-traffic-sign\endcsname{54.0}
\expandafter\def\csname res@pvit-d12-semantickitti-cnnkd@iou-traffic-sign-pm\endcsname{54.0}
\expandafter\def\csname res@rt-litept@params\endcsname{12.7M}
\expandafter\def\csname res@rt-litept@infer-ms\endcsname{31}
\expandafter\def\csname res@rt-litept@train-ms\endcsname{75}
\expandafter\def\csname res@rt-litept@infer-mem\endcsname{0.3}
\expandafter\def\csname res@rt-litept@train-mem\endcsname{1.1}
\expandafter\def\csname res@rt-litept@points\endcsname{96k}
\expandafter\def\csname res@rt-litept@infer-ms-scene\endcsname{31}
\expandafter\def\csname res@rt-litept@train-ms-scene\endcsname{75}
\expandafter\def\csname res@rt-litept_b8@params\endcsname{12.7M}
\expandafter\def\csname res@rt-litept_b8@infer-ms\endcsname{135}
\expandafter\def\csname res@rt-litept_b8@train-ms\endcsname{338}
\expandafter\def\csname res@rt-litept_b8@infer-mem\endcsname{1.8}
\expandafter\def\csname res@rt-litept_b8@train-mem\endcsname{7.0}
\expandafter\def\csname res@rt-litept_b8@points\endcsname{869k}
\expandafter\def\csname res@rt-litept_b8@infer-ms-scene\endcsname{17}
\expandafter\def\csname res@rt-litept_b8@train-ms-scene\endcsname{42}
\expandafter\def\csname res@rt-phasorpt-novo@params\endcsname{22.6M}
\expandafter\def\csname res@rt-phasorpt-novo@infer-ms\endcsname{48}
\expandafter\def\csname res@rt-phasorpt-novo@train-ms\endcsname{126}
\expandafter\def\csname res@rt-phasorpt-novo@infer-mem\endcsname{0.3}
\expandafter\def\csname res@rt-phasorpt-novo@train-mem\endcsname{3.6}
\expandafter\def\csname res@rt-phasorpt-novo@points\endcsname{96k}
\expandafter\def\csname res@rt-phasorpt-novo@infer-ms-scene\endcsname{48}
\expandafter\def\csname res@rt-phasorpt-novo@train-ms-scene\endcsname{126}
\expandafter\def\csname res@rt-phasorpt-novo_b8@params\endcsname{22.6M}
\expandafter\def\csname res@rt-phasorpt-novo_b8@infer-ms\endcsname{199}
\expandafter\def\csname res@rt-phasorpt-novo_b8@train-ms\endcsname{563}
\expandafter\def\csname res@rt-phasorpt-novo_b8@infer-mem\endcsname{1.9}
\expandafter\def\csname res@rt-phasorpt-novo_b8@train-mem\endcsname{22.0}
\expandafter\def\csname res@rt-phasorpt-novo_b8@points\endcsname{869k}
\expandafter\def\csname res@rt-phasorpt-novo_b8@infer-ms-scene\endcsname{25}
\expandafter\def\csname res@rt-phasorpt-novo_b8@train-ms-scene\endcsname{70}
\expandafter\def\csname res@rt-phasorpt-voxtok@params\endcsname{22.5M}
\expandafter\def\csname res@rt-phasorpt-voxtok@infer-ms\endcsname{54}
\expandafter\def\csname res@rt-phasorpt-voxtok@train-ms\endcsname{138}
\expandafter\def\csname res@rt-phasorpt-voxtok@infer-mem\endcsname{0.3}
\expandafter\def\csname res@rt-phasorpt-voxtok@train-mem\endcsname{3.4}
\expandafter\def\csname res@rt-phasorpt-voxtok@points\endcsname{96k}
\expandafter\def\csname res@rt-phasorpt-voxtok@infer-ms-scene\endcsname{54}
\expandafter\def\csname res@rt-phasorpt-voxtok@train-ms-scene\endcsname{138}
\expandafter\def\csname res@rt-phasorpt-voxtok_b8@params\endcsname{22.5M}
\expandafter\def\csname res@rt-phasorpt-voxtok_b8@infer-ms\endcsname{204}
\expandafter\def\csname res@rt-phasorpt-voxtok_b8@train-ms\endcsname{562}
\expandafter\def\csname res@rt-phasorpt-voxtok_b8@infer-mem\endcsname{1.9}
\expandafter\def\csname res@rt-phasorpt-voxtok_b8@train-mem\endcsname{20.8}
\expandafter\def\csname res@rt-phasorpt-voxtok_b8@points\endcsname{869k}
\expandafter\def\csname res@rt-phasorpt-voxtok_b8@infer-ms-scene\endcsname{25}
\expandafter\def\csname res@rt-phasorpt-voxtok_b8@train-ms-scene\endcsname{70}
\expandafter\def\csname res@rt-phasorpt@params\endcsname{22.6M}
\expandafter\def\csname res@rt-phasorpt@infer-ms\endcsname{55}
\expandafter\def\csname res@rt-phasorpt@train-ms\endcsname{142}
\expandafter\def\csname res@rt-phasorpt@infer-mem\endcsname{0.3}
\expandafter\def\csname res@rt-phasorpt@train-mem\endcsname{3.7}
\expandafter\def\csname res@rt-phasorpt@points\endcsname{96k}
\expandafter\def\csname res@rt-phasorpt@infer-ms-scene\endcsname{55}
\expandafter\def\csname res@rt-phasorpt@train-ms-scene\endcsname{142}
\expandafter\def\csname res@rt-phasorpt_b8@params\endcsname{22.6M}
\expandafter\def\csname res@rt-phasorpt_b8@infer-ms\endcsname{219}
\expandafter\def\csname res@rt-phasorpt_b8@train-ms\endcsname{611}
\expandafter\def\csname res@rt-phasorpt_b8@infer-mem\endcsname{1.9}
\expandafter\def\csname res@rt-phasorpt_b8@train-mem\endcsname{22.6}
\expandafter\def\csname res@rt-phasorpt_b8@points\endcsname{869k}
\expandafter\def\csname res@rt-phasorpt_b8@infer-ms-scene\endcsname{27}
\expandafter\def\csname res@rt-phasorpt_b8@train-ms-scene\endcsname{76}
\expandafter\def\csname res@rt-ptv3@params\endcsname{46.2M}
\expandafter\def\csname res@rt-ptv3@infer-ms\endcsname{78}
\expandafter\def\csname res@rt-ptv3@train-ms\endcsname{204}
\expandafter\def\csname res@rt-ptv3@infer-mem\endcsname{0.5}
\expandafter\def\csname res@rt-ptv3@train-mem\endcsname{3.4}
\expandafter\def\csname res@rt-ptv3@points\endcsname{96k}
\expandafter\def\csname res@rt-ptv3@infer-ms-scene\endcsname{78}
\expandafter\def\csname res@rt-ptv3@train-ms-scene\endcsname{204}
\expandafter\def\csname res@rt-ptv3_b8@params\endcsname{46.2M}
\expandafter\def\csname res@rt-ptv3_b8@infer-ms\endcsname{403}
\expandafter\def\csname res@rt-ptv3_b8@train-ms\endcsname{981}
\expandafter\def\csname res@rt-ptv3_b8@infer-mem\endcsname{2.8}
\expandafter\def\csname res@rt-ptv3_b8@train-mem\endcsname{20.7}
\expandafter\def\csname res@rt-ptv3_b8@points\endcsname{869k}
\expandafter\def\csname res@rt-ptv3_b8@infer-ms-scene\endcsname{50}
\expandafter\def\csname res@rt-ptv3_b8@train-ms-scene\endcsname{123}
\expandafter\def\csname res@rt-volt-vo@params\endcsname{23.7M}
\expandafter\def\csname res@rt-volt-vo@infer-ms\endcsname{42}
\expandafter\def\csname res@rt-volt-vo@train-ms\endcsname{184}
\expandafter\def\csname res@rt-volt-vo@infer-mem\endcsname{0.3}
\expandafter\def\csname res@rt-volt-vo@train-mem\endcsname{3.0}
\expandafter\def\csname res@rt-volt-vo@points\endcsname{96k}
\expandafter\def\csname res@rt-volt-vo@infer-ms-scene\endcsname{42}
\expandafter\def\csname res@rt-volt-vo@train-ms-scene\endcsname{184}
\expandafter\def\csname res@rt-volt-vo_b8@params\endcsname{23.7M}
\expandafter\def\csname res@rt-volt-vo_b8@infer-ms\endcsname{306}
\expandafter\def\csname res@rt-volt-vo_b8@train-ms\endcsname{1123}
\expandafter\def\csname res@rt-volt-vo_b8@infer-mem\endcsname{1.5}
\expandafter\def\csname res@rt-volt-vo_b8@train-mem\endcsname{18.3}
\expandafter\def\csname res@rt-volt-vo_b8@points\endcsname{869k}
\expandafter\def\csname res@rt-volt-vo_b8@infer-ms-scene\endcsname{38}
\expandafter\def\csname res@rt-volt-vo_b8@train-ms-scene\endcsname{140}
\expandafter\def\csname res@rt-volt@params\endcsname{23.7M}
\expandafter\def\csname res@rt-volt@infer-ms\endcsname{37}
\expandafter\def\csname res@rt-volt@train-ms\endcsname{104}
\expandafter\def\csname res@rt-volt@infer-mem\endcsname{0.3}
\expandafter\def\csname res@rt-volt@train-mem\endcsname{2.5}
\expandafter\def\csname res@rt-volt@points\endcsname{96k}
\expandafter\def\csname res@rt-volt@infer-ms-scene\endcsname{37}
\expandafter\def\csname res@rt-volt@train-ms-scene\endcsname{104}
\expandafter\def\csname res@rt-volt_b8@params\endcsname{23.7M}
\expandafter\def\csname res@rt-volt_b8@infer-ms\endcsname{262}
\expandafter\def\csname res@rt-volt_b8@train-ms\endcsname{736}
\expandafter\def\csname res@rt-volt_b8@infer-mem\endcsname{1.5}
\expandafter\def\csname res@rt-volt_b8@train-mem\endcsname{15.3}
\expandafter\def\csname res@rt-volt_b8@points\endcsname{869k}
\expandafter\def\csname res@rt-volt_b8@infer-ms-scene\endcsname{33}
\expandafter\def\csname res@rt-volt_b8@train-ms-scene\endcsname{92}
\expandafter\def\csname res@ips@repeats\endcsname{3}
\expandafter\def\csname res@ips@maxstd\endcsname{0.1}
\expandafter\def\csname res@ips-512@miou\endcsname{74.8}
\expandafter\def\csname res@ips-512@macc\endcsname{83.6}
\expandafter\def\csname res@ips-512@allacc\endcsname{90.8}
\expandafter\def\csname res@ips-512@miou-delta\endcsname{\ensuremath{-}1.4}
\expandafter\def\csname res@ips-512@macc-delta\endcsname{\ensuremath{-}1.1}
\expandafter\def\csname res@ips-512@allacc-delta\endcsname{\ensuremath{-}0.5}
\expandafter\def\csname res@ips-1024@miou\endcsname{76.2}
\expandafter\def\csname res@ips-1024@macc\endcsname{84.7}
\expandafter\def\csname res@ips-1024@allacc\endcsname{91.3}
\expandafter\def\csname res@ips-1024@miou-delta\endcsname{+0.0}
\expandafter\def\csname res@ips-1024@macc-delta\endcsname{+0.0}
\expandafter\def\csname res@ips-1024@allacc-delta\endcsname{+0.0}
\expandafter\def\csname res@ips-2048@miou\endcsname{77.0}
\expandafter\def\csname res@ips-2048@macc\endcsname{85.3}
\expandafter\def\csname res@ips-2048@allacc\endcsname{91.6}
\expandafter\def\csname res@ips-2048@miou-delta\endcsname{+0.8}
\expandafter\def\csname res@ips-2048@macc-delta\endcsname{+0.6}
\expandafter\def\csname res@ips-2048@allacc-delta\endcsname{+0.3}
\expandafter\def\csname res@ips-4096@miou\endcsname{76.9}
\expandafter\def\csname res@ips-4096@macc\endcsname{85.1}
\expandafter\def\csname res@ips-4096@allacc\endcsname{91.5}
\expandafter\def\csname res@ips-4096@miou-delta\endcsname{+0.7}
\expandafter\def\csname res@ips-4096@macc-delta\endcsname{+0.4}
\expandafter\def\csname res@ips-4096@allacc-delta\endcsname{+0.2}
\expandafter\def\csname res@gs@repeats\endcsname{3}
\expandafter\def\csname res@gs-on@miou\endcsname{76.1}
\expandafter\def\csname res@gs-on@macc\endcsname{84.8}
\expandafter\def\csname res@gs-on@allacc\endcsname{91.3}
\expandafter\def\csname res@gs-on@miou-std\endcsname{0.1}
\expandafter\def\csname res@gs-on@miou-delta\endcsname{+0.0}
\expandafter\def\csname res@gs-on@macc-delta\endcsname{+0.0}
\expandafter\def\csname res@gs-on@allacc-delta\endcsname{+0.0}
\expandafter\def\csname res@gs-off@miou\endcsname{75.6}
\expandafter\def\csname res@gs-off@macc\endcsname{84.6}
\expandafter\def\csname res@gs-off@allacc\endcsname{91.0}
\expandafter\def\csname res@gs-off@miou-std\endcsname{0.1}
\expandafter\def\csname res@gs-off@miou-delta\endcsname{\ensuremath{-}0.5}
\expandafter\def\csname res@gs-off@macc-delta\endcsname{\ensuremath{-}0.2}
\expandafter\def\csname res@gs-off@allacc-delta\endcsname{\ensuremath{-}0.3}

%% file: sec/0_abstract.tex
\begin{abstract}
Most state-of-the-art 3D point cloud backbones are hybrids that combine attention with sparse convolutions.
While the convolutions supply local geometry, they add architectural complexity, hinder deployment, and tie the network to a discrete voxel grid.
Convolution-free point transformers avoid these costs, yet they still trail the hybrids when trained on a single dataset with matched supervision.
Dropping the convolutions frees the network from the voxel grid, but we hypothesize that current convolution-free backbones do not fully exploit the continuous point geometry.
Pooling the points onto voxels discards their sub-voxel positions, and in attention, positions only influence where a query attends, while the values it aggregates carry no information about where they come from.
We address both with operators built on phasors of the continuous point coordinates.
\textbf{Phasor pooling} embeds each token from the exact positions of its points, and \textbf{phasor attention} makes the values geometry-aware while staying compatible with efficient attention kernels.
Building on both, \textbf{PhasorPT} is a convolution-free point transformer that operates on continuous coordinates throughout.
It matches or exceeds the hybrid PTv3 on multiple indoor and outdoor datasets with half its parameters and without a single convolution.
With additional supervision from a convolutional teacher trained on the same data, it surpasses all baselines on ScanNet, ScanNet200, and SemanticKITTI.
\end{abstract}

%% file: sec/1_intro.tex
\section{Introduction}
\label{sec:intro}

\begin{figure}[t]
  \centering
  \includegraphics{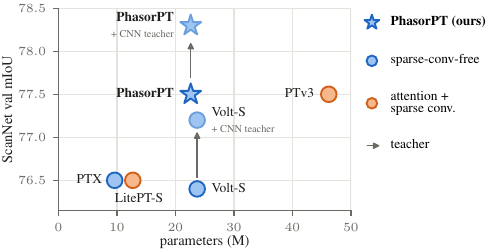}
  \caption{\textbf{ScanNet val mIoU over parameters}, trained from scratch on a single dataset. Arrows: the gain from distillation from Volt's sparse-convolutional U-Net trained on the same data (\cref{sec:exp_selfdistill}). Without a teacher, PhasorPT matches PTv3 at half its parameters and without sparse convolutions.}
  \label{fig:teaser}
\end{figure}

In its pure form, attention~\cite{vaswani2017attention} learns to exchange information between tokens, but it is inherently permutation equivariant and thus blind to the spatial relations between them.
Most transformer architectures therefore encode the positions of their tokens explicitly.
In 3D point clouds, where arguably most of the structure lies in the relative spatial offsets between neighboring points, this encoding is particularly important.
Early point transformers encoded these offsets through learned functions for every pair of points, added to both the attention weights and the values~\cite{zhao2021pt,wu2022ptv2,lai2022stratified}; this is effective, but computationally expensive and incompatible with efficient attention kernels.

Point Transformer V3 (PTv3)~\cite{wu2024ptv3} scaled point transformers by serializing the points along space-filling curves and attending within fixed-size patches.
To use kernels such as FlashAttention~\cite{dao2023flashattention2}, it replaced the per-pair encodings with a sparse convolution in front of every attention block (xCPE), which encodes positions implicitly through the local geometry.
This convolution is key to its accuracy: removing it without replacement degrades PTv3 substantially~\cite{reichardt2026ptx,yue2026litept}, which makes it effectively a hybrid of convolution and attention.

Sparse convolutions come at a cost, however.
Dense voxel grids scale cubically with the resolution, so sparse convolutions rely on custom kernels and data structures~\cite{choy2019minkowski,spconv2022}, which complicate the architecture and its deployment.
Recent backbones therefore increasingly adopt 3D rotary position embeddings (RoPE)~\cite{su2024roformer}, an efficient, parameter-free relative encoding that is compatible with efficient attention kernels.
Yet RoPE has not replaced the convolution so far:
LitePT~\cite{yue2026litept} keeps convolutional stages at high resolution, Utonia~\cite{zhang2026utonia} and SpaCeFormer~\cite{choy2026spaceformer} add RoPE to the attention but keep the sparse convolutions of PTv3, and PointTransformerX (PTX)~\cite{reichardt2026ptx} replaces xCPE with RoPE without fully recovering its accuracy.
Volume Transformer (Volt)~\cite{yilmaz2026volt}, a plain Vision Transformer (ViT)~\cite{dosovitskiy2021vit} with 3D RoPE, comes closest: with distillation from a convolutional teacher trained on the same data, it matches hybrid backbones on most benchmarks.

Without the teacher, however, Volt still trails the hybrids when trained on a single dataset with matched supervision.
We hypothesize that current convolution-free architectures do not yet make full use of the continuous point geometry that dropping the convolutions makes available, and identify two places where it is lost.
First, the tokens.
A sparse convolution indexes its weights by the discrete slot of each neighbor and therefore needs the points pooled onto a grid; without convolutions, this constraint disappears.
Volt nevertheless embeds its tokens on a discrete voxel grid: it maps each voxel slot of a patch through its own weight with a strided convolution and positions the tokens at integer patch coordinates, so that the exact layout of the points inside each cell is lost.
Second, the values.
Relative position encodings such as RoPE act only on the attention scores: they decide where a query attends, but the values it aggregates carry no information about where they come from (\cref{sec:method_blind}).

We address both with operators built on \emph{phasors} $e^{\mathrm{i}\boldsymbol{\omega}^\top\mathbf{p}}$: unit complex numbers whose phase is the continuous point position $\mathbf{p}$ projected onto a frequency vector $\boldsymbol{\omega}$, \ie, a plane wave evaluated at the point.
RoPE applies the same phasors to the queries and keys; we apply them to the features before these are aggregated.

\textbf{Phasor pooling} keeps the voxel grid only to group the points into tokens efficiently, and embeds each token from the phasors of the exact offsets of its points, a continuous kernel instead of one weight per voxel slot.
Each token is positioned at the centroid of its points, so that the attention scores also see continuous positions.
\textbf{Phasor attention} makes the attention values geometry-aware: it additionally rotates the values by their positions, so that each value enters the output relative to the query, while the attention kernel stays unchanged; a comparable operation has been proposed for language and multi-view transformers~\cite{harik2022roper,garciacastellanos2026rove,miyato2024gta,li2025prope}.
Building on both operators, we introduce \textbf{PhasorPT}, a convolution-free point transformer that operates on continuous coordinates throughout.
It matches or exceeds PTv3 on ScanNet, ScanNet200, Structured3D, and nuScenes with half its parameters and, without any teacher, matches or exceeds Volt distilled from its convolutional teacher on ScanNet, ScanNet200, and nuScenes.
Distilled from the same teacher with Volt's loss, it surpasses the distilled Volt on every dataset we evaluate with the teacher, and all baselines on ScanNet, ScanNet200, and SemanticKITTI.

In summary, our contributions are:
\begin{itemize}
    \item \textbf{Phasor pooling}, which embeds each token from the exact positions of its points instead of a discrete voxel grid.
    \item \textbf{Phasor attention}, which makes the attention values geometry-aware without per-pair computation and remains compatible with efficient attention kernels.
    \item \textbf{PhasorPT}, a convolution-free point transformer that matches or exceeds PTv3 with half its parameters.%
\end{itemize}

%% file: sec/2_related.tex
\section{Related Work}
\label{sec:related}

\paragraph{Point cloud backbones.}
Point-based networks operate directly on the raw points, aggregating local neighborhoods with shared MLPs and symmetric pooling~\cite{qi2017pointnet,qi2017pointnetpp}, graph convolutions~\cite{wang2019dgcnn}, or continuous convolutions with kernel points or learned weight functions of the relative position~\cite{thomas2019kpconv,wu2019pointconv}.
Their per-neighborhood computation limits their scalability to large scenes.
Sparse convolutional networks~\cite{choy2019minkowski,tang2020spvnas,peng2024oacnns} voxelize the point cloud and restrict the convolutions to the occupied voxels, which requires dedicated sparse kernels and data structures.
Point transformers apply attention within local neighborhoods~\cite{zhao2021pt,wu2022ptv2} or windows~\cite{lai2022stratified,yang2023swin3d,wang2023dsvt}, and scale to large scenes by sorting the points into a locality-preserving 1D sequence and attending within fixed-size groups of it~\cite{wang2023octformer,liu2023flatformer,wu2024ptv3}, which keeps the cost of attention linear in the number of points.
PTv3~\cite{wu2024ptv3} became the backbone of large-scale pretraining~\cite{wu2025sonata,zhang2026utonia}, while keeping sparse convolutions for the positional encoding.
Recent works reduce this reliance.
LitePT~\cite{yue2026litept} uses convolutional stages only at the high-resolution levels and RoPE-based attention at the coarse levels.
SpaCeFormer~\cite{choy2026spaceformer} builds on PTv3 and replaces the serialized attention at the high-resolution levels with spatial window attention with RoPE, while keeping sparse convolutions in every block.
PTX~\cite{reichardt2026ptx} removes all operations that are not native to PyTorch from PTv3 for portability, replacing PTv3's convolutional position embeddings (xCPE) with axial 3D RoPE in all blocks, and recovers part of the lost accuracy with learned per-head RoPE frames, larger patches at inference and a modified FFN.
Volt~\cite{yilmaz2026volt} revisits a vanilla ViT on volumetric patches with 3D RoPE, and closes the gap to convolutional backbones through a data-efficient training recipe and distillation from a convolutional teacher.
For automotive lidar, VaViT~\cite{puy2026vavit} applies a plain ViT to bird's-eye-view pillar tokens.
Like PTX, Volt, and VaViT, PhasorPT uses no sparse convolutions; instead, it encodes the local geometry directly with phasors of the continuous point offsets: in its tokenizer, where learned waves modulate the point features before pooling, and in the value path of the attention.

\paragraph{Positional encoding in transformers.}
Absolute encodings add a function of the position to the input tokens~\cite{vaswani2017attention}.
Relative encodings add learned biases for the relative positions to the attention logits~\cite{liu2021swin}, learned functions of the offset of every pair of tokens to the logits and the values~\cite{zhao2021pt,wu2022ptv2}, or learned tables indexed by the quantized relative position, which interact with the queries and keys and are added to the values~\cite{lai2022stratified}.
As these per-pair computations scale poorly and do not fit efficient attention kernels, recent architectures resort to two alternatives that do.
Conditional positional encodings generate the encoding from the local neighborhood with a convolution~\cite{chu2023cpe}, as in hybrid architectures that add convolutions to vision transformers~\cite{wu2021cvt}; PTv3 and SpaCeFormer place such a sparse convolution in front of every attention block.
RoPE~\cite{su2024roformer} rotates the queries and keys by their positions, so that their dot product only depends on the relative position, without parameters or per-pair computation.
It has been extended to 2D images, with per-axis or learned mixed frequencies~\cite{heo2024rope}, to learned rotations of multi-dimensional positions~\cite{ostmeier2025liere,schenck2025string}, and, with per-axis frequencies, to the 3D backbones above.
All of them encode the geometry only in the attention weights.

\paragraph{Geometry in the value path.}
The early point transformers added a function of the offset to the attention values as well~\cite{zhao2021pt,wu2022ptv2}, but PTv3~\cite{wu2024ptv3} dropped these per-pair encodings in favor of efficient attention kernels.
RoPER~\cite{harik2022roper} was, to our knowledge, the first to make the values position-aware by rotating them: each value is rotated by its key position and the attention output back by the query position, so that the values enter relative to the query while the computation remains compatible with efficient attention kernels.
Geometric Transform Attention (GTA)~\cite{miyato2024gta} independently derived a comparable structure for multi-view transformers, transforming features between the camera frames of different views in novel view synthesis; PRoPE~\cite{li2025prope} and RayRoPE~\cite{wu2026rayrope} extend it to projective camera encodings.
RoVE~\cite{garciacastellanos2026rove}, noting that the construction has been discovered independently in several communities, interprets it as attentive convolution and finds consistent gains in language models.
These works operate on 1D sequences or a discrete set of camera frames.
We apply the mechanism to continuous, sparse, and irregularly sampled 3D positions, show that it encodes the local geometry otherwise supplied by convolutions without any per-pair computation, and derive from it a pooling operator that tokenizes the raw points.

%% file: sec/3_method.tex
\section{Phasors for Continuous Geometry}
\label{sec:method}

Without the fixed voxel grid imposed by sparse convolutions, the points can keep their continuous coordinates throughout the network.
Geometry can still be lost in two places, however: when the points are pooled into tokens on a voxel grid, and in the values of RoPE attention.
We address both with one encoding, a phasor of the continuous offset multiplied onto a feature before it is aggregated: phasor pooling tokenizes the raw points (\cref{sec:method_pool}), and phasor attention makes the attention values geometry-aware (\cref{sec:method_phasor}).
\Cref{sec:method_analysis} shows that both aggregate a characteristic function of the offsets.

\subsection{Phasor Pooling}
\label{sec:method_pool}

\paragraph{Tokenization on a grid.}
Volt~\cite{yilmaz2026volt} adopts the patch embedding of ViT~\cite{dosovitskiy2021vit}:
the points are first voxelized on a fixed grid, a strided convolution with kernel size $s$ equal to its stride embeds each patch of $s^3$ voxels into one token, and a transposed convolution maps the tokens back. 
Both kernels are piecewise constant over the voxel slots of a patch: all points inside a slot share one weight, their exact positions are lost, and the number of weights grows with $s^3$.
PTv3~\cite{wu2024ptv3} instead max-pools the features of the points in each cell, which discards their positions entirely and leaves the encoding of their spatial arrangement to the sparse convolution in front of every block.

\begin{figure}[t]
  \centering
  \includegraphics[width=\linewidth]{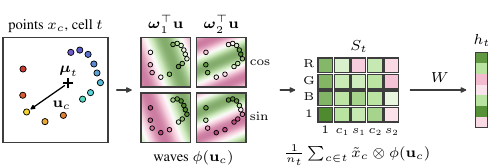}
  \caption{\textbf{Phasor pooling} (2D, one cell, two waves). Every point samples the waves $\phi$ at its offset $\mathbf{u}_c$ to the token centroid $\boldsymbol{\mu}_t$; the mean of the outer products with its features gives $S_t$ (rows: the features $\tilde{x}_c$, here the RGB color and the constant; columns: the constant and the cosine and sine of each wave), which a linear map projects to the token $h_t$ (\cref{eq:pw_pool}). Unlike the one weight per voxel slot of a strided convolution, the resulting kernel is a smooth function of the exact offset.}
  \label{fig:planewave}
\end{figure}

\paragraph{Phasor pooling.}
We keep the one-step, ViT-style tokenization, but replace the per-slot weights with phasors of the exact point offsets (\cref{fig:planewave}).
Each token $t$ collects the $n_t$ points of its grid cell and is positioned at their centroid; let $\mathbf{u}_c$ be the offset of point $c$ from it, in cell units.
We describe the offset by $M$ learned waves $\boldsymbol{\omega}_m \in \mathbb{R}^3$ and pool, per token, the outer products of each point's features $\tilde{x}_c = [x_c, 1]$ with them:
\begin{equation}
\begin{aligned}
    \phi(\mathbf{u}) &= \big[1, \cos \boldsymbol{\omega}_m^\top \mathbf{u}, \sin \boldsymbol{\omega}_m^\top \mathbf{u}\big]_{m=1}^M, \\
    S_t &= \frac{1}{n_t} \sum\nolimits_{c \in t} \tilde{x}_c \otimes \phi(\mathbf{u}_c), \\
    h_t &= W \big[\mathrm{vec}(S_t),\, \log n_t\big].
\end{aligned}
    \label{eq:pw_pool}
\end{equation}
Each phase $\boldsymbol{\omega}_m^\top \mathbf{u}$ is the offset projected onto the direction of wave $m$, scaled by its frequency $\lVert\boldsymbol{\omega}_m\rVert$.
Since $W$ is linear, $h_t$ is a strided continuous convolution~\cite{wang2018pcc,wu2019pointconv} of the point features whose kernel is a Fourier series in the offset, with one weight matrix per wave instead of one per voxel slot, as in Fourier neural operators~\cite{li2021fno}.
The constant input channel pools the waves alone and thus describes the shape of the cell's point set (its empirical characteristic function~\cite{feuerverger1977ecf}), and $\log n_t$ restores the density that the mean discards.
Each cosine and sine pair is the phasor $e^{\mathrm{i}\boldsymbol{\omega}_m^\top \mathbf{u}_c}$, the same rotation that RoPE applies to a position (\cref{sec:method_rope}), here applied to the exact offset of a point inside its cell; we therefore call it phasor pooling.
As in Fourier feature embeddings~\cite{rahimi2007rff,tancik2020fourier,mildenhall2020nerf} and the fixed trigonometric encodings of neighbor offsets in Point-NN~\cite{zhang2023pointnn}, the waves lift the low-dimensional offsets to sinusoids of several frequencies; here both their lengths and directions are learned (initialization in \cref{sec:suppl_arch}).
A residual FFN on the pooled tokens then reads out the within-cell layout non-linearly, before the first attention block.

\paragraph{Phasor unpooling.}
The inverse maps each token to a field over its cell that every point reads at its own offset,
\begin{equation}
    y_c = W_b\, \bar{h}_t + W_r\, (A\, \bar{h}_t)^\top \psi(\mathbf{u}_c),
    \label{eq:pw_unpool}
\end{equation}
with the normalized token $\bar{h}_t$, a second set of $M'$ learned waves $\psi$, and a linear map $A$ from the token to a $2M' \times r$ matrix, whose $r$ columns are the wave coefficients of $r$ scalar fields over the cell.
The first term is a plain broadcast, the second mixes the $r$ fields into the output with $W_r$, so that the variation inside a cell spans at most $r$ output directions (rank $r$); it replaces the per-slot output weights of a transposed convolution.
The rank is independent of the number of waves and bounds the size of $A$ (\cref{sec:suppl_arch}); with $W_r$ initialized to zero, unpooling starts as the broadcast.
\Cref{tab:components} compares both directions with Volt's strided and transposed convolutions under the same backbone.

\subsection{Phasor Attention}
\label{sec:method_phasor}

\paragraph{Rotary position embeddings.}
\label{sec:method_rope}
On its own, standard scaled dot-product self-attention~\cite{vaswani2017attention} is permutation equivariant: it treats its tokens as an unordered set, so an explicit positional encoding is needed to make it position-aware.
1D RoPE~\cite{su2024roformer} encodes the relative position between a query $q_n$ at position $n$ and a key $k_m$ at position $m$ by splitting each token's feature into 2D slices and rotating them based on the token's position in the sequence, encoding each slice with a different frequency.
Following \cite{su2024roformer}, we write the rotation of a $d$-dimensional feature at position $n$ as the block-diagonal matrix
\begin{equation}
    \mathbf{R}(n) = \mathrm{diag}\big(\mathbf{R}_{\theta_1 n}, \dots, \mathbf{R}_{\theta_{d/2} n}\big),
    \label{eq:rope}
\end{equation}
where $\mathbf{R}_\alpha \in SO(2)$ is the $2 \times 2$ rotation by the angle $\alpha$, with frequencies $\theta_\ell = b^{-2(\ell-1)/d}$ and base $b$.
As rotations in the same plane compose by adding their angles, $\mathbf{R}(n)\mathbf{R}(m) = \mathbf{R}(n + m)$ and $\mathbf{R}(n)^\top = \mathbf{R}(-n)$, the dot product of the rotated query and key depends only on their relative position:
\begin{equation}
    \big(\mathbf{R}(n)\, q_n\big)^\top \mathbf{R}(m)\, k_m = q_n^\top\, \mathbf{R}(m - n)\, k_m.
    \label{eq:rope_dot}
\end{equation}

\begin{figure}[t]
  \centering
  \includegraphics[width=\linewidth]{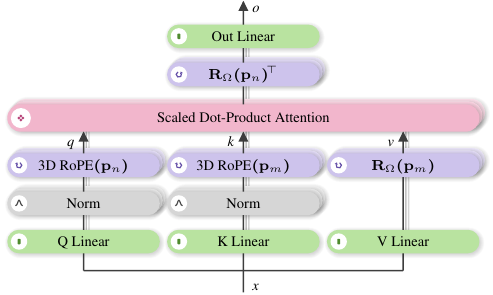}
  \caption{\textbf{Multi-head phasor attention.}
  Queries and keys are normalized (QK-Norm) and rotated by 3D RoPE as in standard RoPE attention.
  In addition, the values are rotated by their positions $\mathbf{p}_m$ before, and the per-head attention outputs back by the query positions $\mathbf{p}_n$ after the scaled dot-product attention, so that each value enters the output relative to the query (\cref{eq:vo_attn}).
  The value path rotates by $\mathbf{R}_\Omega$, with its own learned 3D frequency per feature pair (as in RoPE-Mixed~\cite{heo2024rope}) instead of the fixed axial frequencies.}
  \label{fig:vo_attn}
\end{figure}

Axial 3D RoPE extends the same principle to 3D by dividing the slices into three groups, one per coordinate axis.
Each slice $\ell$ is assigned an axis $a(\ell) \in \{x, y, z\}$, and $\mathbf{R}(\mathbf{p})$ rotates it by the angle $\theta_\ell\, p_{a(\ell)}$ of the token position $\mathbf{p} \in \mathbb{R}^3$.
The same identities hold, so that the attention score depends on $\mathbf{p}_m - \mathbf{p}_n$ only.
We denote this rotation as 3D RoPE$(\mathbf{p})$ in \cref{fig:vo_attn}.

\paragraph{Geometry-blind values.}
\label{sec:method_blind}
Analyzing the attention of a trained 3D RoPE transformer distinguishes geometric attention heads, whose scores are largely explained by the relative position of query and key, from heads that score by content (\cref{sec:suppl_heads}).
A geometric head concentrates its attention on certain offsets, yet its output does not reveal where the mass actually landed.
In sparse 3D data, a given offset may contain no token; the softmax then forces the attention mass onto other tokens, possibly from entirely different regions.
As the value path encodes no geometry, the output carries no sign of whether the mass landed at the intended offset or elsewhere, and the following blocks have to tell such a miss from a hit by the content alone.
We argue that this limitation is especially detrimental for point clouds, where much of the meaning lies in exactly this shape of the neighborhood.

\paragraph{Value rotation.}
We make the value path geometry-aware by rotating each value $v_m$ by its key position $\mathbf{p}_m$ before the attention (\cref{fig:vo_attn}).
To keep the absolute key positions out of the output, we rotate the per-head attention output back by the query position $\mathbf{p}_n$.
For the values, we do not tie each slice to a fixed axis, but give every slice $\ell$ of every head its own learned 3D frequency $\boldsymbol{\omega}_\ell \in \mathbb{R}^3$ as in RoPE-Mixed~\cite{heo2024rope}, and collect them in $\Omega = \{\boldsymbol{\omega}_\ell\}_\ell$.
We denote by $\mathbf{R}_\Omega(\mathbf{p})$ the block-diagonal rotation that rotates each slice $\ell$ by the angle $\boldsymbol{\omega}_\ell^\top \mathbf{p}$.
As each angle $\boldsymbol{\omega}_\ell^\top \mathbf{p}$ is linear in $\mathbf{p}$, the rotations compose, and the output becomes
\begin{equation}
\begin{aligned}
    o_n
    &= \mathbf{R}_\Omega(\mathbf{p}_n)^\top \sum\nolimits_m a_{nm}\, \mathbf{R}_\Omega(\mathbf{p}_m)\, v_m \\
    &= \sum\nolimits_m a_{nm}\, \mathbf{R}_\Omega(\mathbf{p}_m - \mathbf{p}_n)\, v_m, \\
    a_{nm} &= \mathrm{softmax}_m\big(q_n^\top \mathbf{R}(\mathbf{p}_m - \mathbf{p}_n)\, k_m / \sqrt{d}\big),
\end{aligned}
    \label{eq:vo_attn}
\end{equation}
where $a_{nm}$ are the standard RoPE attention weights (\cref{eq:rope_dot}).
Each value thus enters the output expressed relative to the query position, while both rotations are applied outside of the attention itself:
the sum over $m$ is a standard scaled-dot-product attention on the rotated $q$, $k$, $v$, so efficient kernels can be used unchanged.
We initialize $\boldsymbol{\omega}_\ell = 0$, so that every block starts as standard RoPE attention and each slice learns whether, in which direction, and at which frequency to encode the geometry in its values.
Each rotated slice costs three parameters for $\boldsymbol{\omega}_\ell$; which portion of the value channels is rotated can be chosen freely (\cref{sec:suppl_arch}).
Rotating the values and the output back follows the structure of GTA~\cite{miyato2024gta} and RoVE~\cite{garciacastellanos2026rove} (\cref{sec:related}); we apply it to 3D positions, with a learned frequency per slice.

\subsection{Analysis}
\label{sec:method_analysis}

\paragraph{Attention as a characteristic function.}
Writing slice $\ell$ of a feature as a complex number, $\tilde{v}_{m,\ell} = v_{m,2\ell-1} + \mathrm{i}\, v_{m,2\ell}$, turns the rotation $\mathbf{R}_\alpha$ into a multiplication with the phasor $e^{\mathrm{i}\alpha}$, and slice $\ell$ of the output into a weighted sum of phasors,
\begin{equation}
    \tilde{o}_{n,\ell} = \sum\nolimits_m a_{nm}\, \tilde{v}_{m,\ell}\, e^{\mathrm{i} \boldsymbol{\omega}_\ell^\top \boldsymbol{\Delta}_{nm}},
    \label{eq:phasor}
\end{equation}
with the offset $\boldsymbol{\Delta}_{nm} = \mathbf{p}_m - \mathbf{p}_n$.
For a content-independent value, $\tilde{v}_{m,\ell} = \bar{v}$, this is $\bar{v}\, \varphi_n(\boldsymbol{\omega}_\ell)$, with the characteristic function of the offsets under the attention distribution
\begin{equation}
    \varphi_n(\boldsymbol{\omega}) = \sum\nolimits_m a_{nm}\, e^{\mathrm{i} \boldsymbol{\omega}^\top \boldsymbol{\Delta}_{nm}},
    \label{eq:charfn}
\end{equation}
evaluated at the frequency of the slice.
With the attention-weighted mean $\boldsymbol{\mu}_n$ and covariance $\mathbf{C}_n$ of the offsets, it is approximately (\cref{sec:suppl_charfn})
\begin{equation}
    \varphi_n(\boldsymbol{\omega}) \approx e^{\mathrm{i} \boldsymbol{\omega}^\top \boldsymbol{\mu}_n} \big(1 - \tfrac{1}{2}\, \boldsymbol{\omega}^\top \mathbf{C}_n \boldsymbol{\omega}\big),
    \label{eq:charfn_approx}
\end{equation}
for low frequencies, $\boldsymbol{\omega}^\top \mathbf{C}_n \boldsymbol{\omega} \ll 1$, \ie, while the spread of the attention mass along $\boldsymbol{\omega}$ is small compared to the wavelength.
The phase of the output thus measures where the attention mass landed along $\boldsymbol{\omega}_\ell$, and its magnitude how spread out it is; together, the slices sample $\varphi_n$ at their frequencies.
This addresses the sparse neighborhoods of \cref{sec:method_blind}: if the softmax spreads mass to distant tokens because the queried direction is empty, the output reveals this, for similar values, by a shifted phase and a reduced magnitude, instead of a geometry-free average of the retrieved values.
Phasor pooling computes the same quantity with uniform weights over the points of a cell: its constant input channel is the empirical characteristic function of their offsets (\cref{eq:pw_pool}).

\paragraph{Relation to convolution.}
With the value and output projections $W_v$, $W_o$, the output of a head is
\begin{equation}
    W_o\, o_n = \sum\nolimits_m a_{nm}\, W_o \mathbf{R}_\Omega(\mathbf{p}_m - \mathbf{p}_n)\, W_v\, x_m,
    \label{eq:conv}
\end{equation}
where each neighbor contributes a linear map of its feature that depends continuously on its offset.
With uniform weights, this is a continuous convolution~\cite{thomas2019kpconv,wu2019pointconv} with a learned trigonometric kernel, which RoVE calls attentive convolution~\cite{garciacastellanos2026rove}; the attention weights make it content adaptive.
Phasor attention can thus learn an offset-dependent aggregation of the same kind as the sparse convolution that supplies the local geometry in front of the attention blocks of previous architectures~\cite{wu2024ptv3}, without evaluating a kernel per pair and without being bound to the voxel grid.

%% file: sec/4_phasorpt.tex
\begin{figure*}[t]
  \centering
  \includegraphics[width=\textwidth]{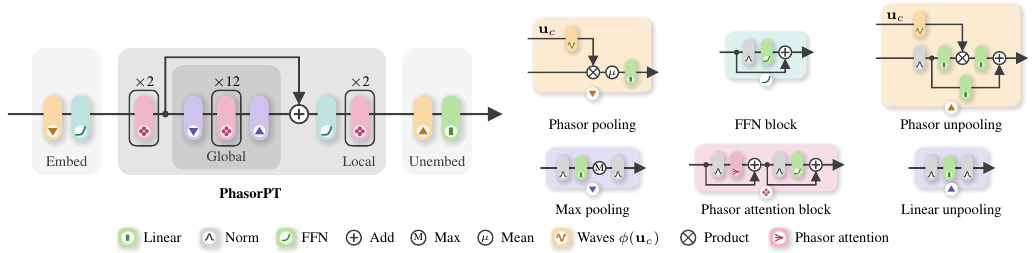}
  \caption{\textbf{PhasorPT overview.}
  Left: phasor pooling (\cref{sec:method_pool}) and an FFN block embed the points directly into the tokens of the local stream (4\,cm indoors). The global branch, a residual block of the local stream, max-pools them to a coarser grid (16\,cm), runs the main serialized phasor attention blocks there, and broadcasts the result back; an FFN block and further local blocks merge it into the local stream. Phasor unpooling and a linear head unembed the tokens back to the points.
  Right: the pooling and unpooling operations ($\mathbf{u}_c$: offset of point $c$ from the centroid of its token) and the pre-norm phasor attention and FFN blocks.}
  \label{fig:overview}
\end{figure*}

\section{PhasorPT}
\label{sec:phasorpt}

Volt~\cite{yilmaz2026volt} has recently shown that a plain ViT~\cite{dosovitskiy2021vit} is a strong 3D backbone by itself.
It samples the points onto a fine voxel grid, embeds non-overlapping patches of voxels into tokens with a strided convolution, and runs global attention over all of its tokens.
Trained jointly on several datasets, it surpasses PTv3~\cite{wu2024ptv3}, while in a supervision-matched, single-dataset setting it needs distillation from a sparse-convolutional network trained on the same data to match it.
Its global attention scales quadratically with the number of tokens, and a single resolution has to capture both the local geometry and the scene context.
PTv3 instead follows a multi-stage U-Net that runs most of its compute on a strongly downsampled point cloud, and serializes the tokens along space-filling curves into fixed-size patches, which keeps the cost of attention linear in the number of tokens.
For its local geometry, it relies on a sparse convolution in every block (xCPE), which requires pooling the points onto a fixed grid at every level.

PhasorPT (\cref{fig:overview}) keeps close to the architectural simplicity of Volt, while retaining the serialized attention and the downsampled main stage of PTv3.
It consists of a fine local stream and a coarse global branch (4 and 16\,cm indoors; \cref{sec:suppl_impl}).
Phasor pooling (\cref{sec:method_pool}) tokenizes the raw points directly into the local stream, and two serialized phasor attention blocks accumulate the geometry of each token's neighborhood. 
To capture long-range dependencies, the global branch pools the local tokens to the coarse grid and runs most of the compute there.
The output of the branch is broadcast back and added to the local stream, and further local blocks merge the global context with the local stream's features before phasor unpooling maps the tokens back to per-point features.
PhasorPT operates on the continuous point coordinates throughout: each coarser token is positioned at the mean of its children instead of a grid cell, and attention, pooling, and unpooling see their exact offsets.

\paragraph{Serialized phasor attention.}
All blocks consist of pre-norm phasor attention (\cref{sec:method_phasor}) and a GELU~\cite{hendrycks2016gelu} FFN with an expansion ratio of 4.
As in PTv3~\cite{wu2024ptv3}, attention runs within fixed-size patches of the serialized Hilbert order, with a different axis permutation per block, so that consecutive blocks group different neighborhoods (\cref{sec:suppl_impl}).
Queries and keys are normalized (QK-Norm~\cite{henry2020qknorm,dehghani2023vit22b}), as in Volt~\cite{yilmaz2026volt}.
As 3D RoPE probes only the three coordinate axes, each head learns a rotation of the positions, following PTX~\cite{reichardt2026ptx}, so that the heads sample different spatial directions (\cref{sec:suppl_impl}).

\paragraph{Global branch.}
The global branch max-pools the child features of each coarse cell, inspired by the grid pooling of PTv3~\cite{wu2024ptv3}.
Its output is broadcast back to the children and added to the local stream, which still holds the intra-cell structure that the max discards; an FFN block follows to merge the two feature streams.

%% file: sec/5_experiments.tex
\begin{table*}[t]
  \centering
  \small
  \newcommand{\sdc}[2]{#2}
  \begin{tabular}{lcrccc!{\hspace{0.4em}}cc}
    \toprule
    & & & \multicolumn{3}{c}{Indoor} & \multicolumn{2}{c}{Outdoor} \\
    \cmidrule(lr){4-6} \cmidrule(lr){7-8}
    Method & Conv-free & Params & ScanNet & ScanNet200 & \sdc{S3DIS A5}{Structured3D} & nuScenes & SemanticKITTI \\
    \midrule
    MinkUNet~\cite{choy2019minkowski} & -- & 37.9M & 72.2 & 25.0 & \sdc{65.4}{76.4} & 73.3 & 63.8 \\
    OA-CNNs~\cite{peng2024oacnns} & -- & 51.5M & 76.1 & 32.3 & \sdc{71.1}{--} & 78.9 & 70.6 \\
    PTv2~\cite{wu2022ptv2} & -- & 12.8M & 75.4 & 30.2 & \sdc{\second{71.6}}{79.0} & 80.2 & 70.3 \\
    PTv3~\cite{wu2024ptv3} & -- & 46.2M & \second{77.5} & 35.2 & \sdc{\best{73.4}}{82.4} & 80.4 & \second{70.8} \\
    LitePT-S~\cite{yue2026litept} & -- & 12.7M & 76.5 & -- & \sdc{--}{\best{83.6}} & \best{82.2} & -- \\
    PTX~\cite{reichardt2026ptx} & $\checkmark$ & 9.6M & 76.5 & -- & -- & -- & -- \\
    VaViT-B~\cite{puy2026vavit} & $\checkmark$ & 85.8M & -- & -- & -- & 81.6 & 68.7 \\
    Volt-S~\cite{yilmaz2026volt} & $\checkmark$ & 23.7M & \result{volt-s-official-scannet}{miou} & \result{volt-s-official-scannet200}{miou} & \sdc{--}{--} & \result{volt-s-official-nuscenes}{miou} & \result{volt-s-official-semantickitti}{miou} \\
    \quad \textit{+ CNN teacher} & & & 77.2 & \second{36.2} & -- & 81.0 & 70.5 \\
    \midrule
    \ours PhasorPT (ours) & $\checkmark$ & 22.6M & \second{\result{pvit-d12-scannet}{miou}} & \result{pvit-d12-scannet200-wide}{miou}\rlap{$^\dagger$} & \sdc{\result{pvit-d12-s3dis}{miou}}{\second{\result{pvit-d12-structured3d}{miou}}\rlap{$^\ddagger$}} & \multicolumn{1}{>{\columncolor{gray!15}[\dimexpr\tabcolsep+0.4em\relax][\tabcolsep]}c}{\result{pvit-d12-nuscenes}{miou}} & \result{pvit-d12-semantickitti-voltaug}{miou} \\
    \ours \quad \textit{+ CNN teacher} & & & \best{\result{pvit-d12-cnnkd-scannet}{miou}} & \best{\result{pvit-d12-scannet200-cnnkd}{miou}}\rlap{$^\dagger$} & -- & \multicolumn{1}{>{\columncolor{gray!15}[\dimexpr\tabcolsep+0.4em\relax][\tabcolsep]}c}{\second{\result{pvit-d12-nuscenes-cnnkd}{miou}}} & \best{\result{pvit-d12-semantickitti-cnnkd}{miou}} \\
    \bottomrule
  \end{tabular}
  \caption{\textbf{Semantic segmentation} (val mIoU, TTA, single-dataset training). Conv-free: no (sparse) convolution in the positional encoding, pooling, or embedding. \textit{CNN teacher}: additional supervision from a sparse-convolutional network trained on the same data, used only during training (Volt-S: as reported by Volt). \sdc{}{Structured3D: baselines as reported by LitePT~\cite{yue2026litept}. }SemanticKITTI: sequence 08; PhasorPT with Volt's augmentations~\cite{yilmaz2026volt}. $^\dagger$Wider per-point output for the 200 classes (22.7M parameters; \cref{sec:suppl_arch}). $^\ddagger$Shorter, staged schedule (\cref{sec:suppl_s3d}). Best in \textbf{bold}, second best \underline{underlined}.}
  \label{tab:main}
\end{table*}

\section{Experiments}
\label{sec:experiments}

\subsection{Setup}
\label{sec:exp_setup}
\paragraph{Datasets.}
We evaluate semantic segmentation indoors on ScanNet~\cite{dai2017scannet}, ScanNet200~\cite{rozenberszki2022scannet200}, and Structured3D~\cite{zheng2020structured3d}, and outdoors on nuScenes~\cite{caesar2020nuscenes} and SemanticKITTI~\cite{behley2019semantickitti}, and report the mIoU on the validation splits.
All models train from scratch on a single dataset; ablations use ScanNet.

\paragraph{Training.}
We follow the protocol of Volt~\cite{yilmaz2026volt}, which adopts the augmentations of PTv3~\cite{wu2024ptv3}, including Mix3D~\cite{nekrasov2021mix3d}, and evaluates the EMA of the weights.
We train with AdamW~\cite{loshchilov2019adamw}, a one-cycle cosine schedule~\cite{smith2019superconvergence}, and cross-entropy plus the Lovász loss~\cite{berman2018lovasz}.
All our models train on a single GPU, for 320 epochs on ScanNet (Volt and PTv3: 800); \cref{tab:suppl_train} lists all settings.
Following Volt's single-dataset setting, we also report PhasorPT distilled from Volt's sparse-convolutional teacher (\cref{sec:exp_selfdistill}); the ablations train without a teacher.

\paragraph{Evaluation.}
Following common practice, we use test-time augmentation (TTA): the predictions are averaged over rotated and flipped copies of each full-resolution scene.
Following PTX~\cite{reichardt2026ptx}, we enlarge the serialized patches at inference, to twice their training size (\cref{tab:suppl_patch}).

\paragraph{Models and baselines.}
We use PhasorPT (22.6M; 22.7M on ScanNet200) for all experiments; every ablation changes one component of it and trains without a teacher (configurations in \cref{sec:suppl_impl}).
Baseline numbers are taken from the published papers (sources in \cref{tab:main}); Volt-S without a teacher from its released runs.
We call a model convolution-free if neither its positional encoding, nor its pooling, nor its input embedding uses a sparse convolution over a neighborhood (\cref{sec:intro}).
Volt's patch embedding is a strided convolution with a kernel size equal to its stride, \ie, a linear layer applied to every patch independently, so we count Volt as convolution-free.

\paragraph{Runtime.}
All models are timed on the same ScanNet val scenes on one NVIDIA A40, with mixed precision after warm-up and each with its own optimized kernels: inference is a single forward pass over a whole scene, at batch 1 and, per scene, at batch 8; training is one forward and backward pass of the loss on one scene, together with its peak memory.

\subsection{Main Results}
\label{sec:exp_main}

Without a teacher, PhasorPT matches or exceeds PTv3 on ScanNet, ScanNet200, Structured3D, and nuScenes at half its parameters and without sparse convolutions (\cref{tab:main}).
With Volt's convolutional teacher (\cref{sec:exp_selfdistill}), it surpasses all baselines on ScanNet, ScanNet200, and SemanticKITTI and is second only to LitePT-S on nuScenes.

\subsection{Where the Geometry Enters}
\label{sec:exp_components}

\begin{table}[t]
  \centering
  \small
  \setlength{\tabcolsep}{4.5pt}
  \newcommand{\dv}[2][\relax]{#1{\result{#2}{miou}} & \,{\scriptsize(\result{#2}{miou-dvox})}}
  \newcommand{\gc}[2]{\multicolumn{1}{>{\columncolor{gray!15}}#1}{#2}}
  \newcommand{\gv}[2]{\multicolumn{1}{>{\columncolor{gray!15}[\tabcolsep][0pt]}r@{}}{#1{\result{#2}{miou}}}
    & \multicolumn{1}{@{}>{\columncolor{gray!15}[0pt][\tabcolsep]}l}{\,{\scriptsize(\result{#2}{miou-dvox})}}}
  \begin{tabular}{lll*{2}{r@{}l}}
    \toprule
    Tokenizer & Positions & Attention & \multicolumn{2}{c}{full} & \multicolumn{2}{c}{geom.} \\
    \midrule
    voxel & grid & RoPE & \result{pvit-d12-voxtok-rope-grid-scannet}{miou} & & \result{pvit-d12-voxtok-rope-grid-geom}{miou} & \\
    voxel & grid & PA & \dv{pvit-d12-voxtok-vo-grid-scannet} & \dv[\second]{pvit-d12-voxtok-vo-grid-geom} \\
    voxel & centroid & PA & \dv{pvit-d12-voxtok-scannet} & \dv{pvit-d12-voxtok-geom} \\
    phasor & centroid & RoPE & \dv[\best]{pvit-d12-rope-scannet} & \dv[\second]{pvit-d12-geom-rope} \\
    \midrule
    \gc{l}{phasor} & \gc{l}{centroid} & \gc{l}{PA} & \gv{\second}{pvit-d12-scannet} & \gv{\best}{pvit-d12-geom} \\
    \bottomrule
  \end{tabular}
  \caption{\textbf{Where the geometry enters PhasorPT} (ScanNet val mIoU, TTA, no teacher). Tokenizer: Volt's voxel patch embedding and unpooling, or phasor pooling and unpooling. Token positions for RoPE and phasor attention (PA): cell centers or point centroids. Input: full (with color and normals) or geometry only (color and normals removed in training and test). In parentheses: change from the first row with the same input. \graybg{Gray}: PhasorPT. Best per column in \textbf{bold}, second best \underline{underlined}.}
  \label{tab:components}
\end{table}

We ablate where continuous geometry enters PhasorPT (\cref{tab:components}).
The baseline keeps the grid in both places of \cref{sec:method} (voxel tokenizer, tokens at the cell centers, RoPE attention); the following rows add phasor attention, centroid positions, and phasor pooling.
Without color and normals, the input carries no appearance and the geometry has to come from the point positions alone, which separates the variants more clearly.
With color and normals, the baseline falls behind all other variants, while these lie within the noise of a single seed of each other.
We hypothesize that color and normals already let the model infer much of the local geometry around each token, which limits the further gains from continuous geometry.
Without them, the differences grow: phasor attention and phasor pooling each improve the baseline by about the same amount, and together they improve it further, which suggests that the two supply largely complementary geometry.
Centroid positions alone do not explain the gain of phasor pooling: with the voxel tokenizer, they lower the geometry-only result.

\subsection{Phasor Attention in Volt}
\label{sec:exp_dropin}

\begin{table}[t]
  \centering
  \small
  \setlength{\tabcolsep}{4pt}
  \newcommand{\dl}[2][\relax]{#1{\result{#2}{miou}} & \,{\scriptsize(\result{#2}{miou-delta})}}
  \newcommand{\nd}[1]{\result{#1}{miou} &}
  \begin{tabular}{l*{4}{r@{}l}}
    \toprule
    & \multicolumn{4}{c}{ScanNet} & \multicolumn{4}{c}{nuScenes} \\
    \cmidrule(lr){2-5} \cmidrule(lr){6-9}
    Teacher & \multicolumn{2}{c}{RoPE} & \multicolumn{2}{c}{+ PA} & \multicolumn{2}{c}{RoPE} & \multicolumn{2}{c}{+ PA} \\
    \midrule
    -- & \nd{volt-s-official-scannet} & \dl{volt-s-vo}
      & \nd{volt-s-official-nuscenes} & \dl{volt-s-vo-nuscenes} \\
    CNN & \dl{volt-s-distill} & \dl[\best]{volt-s-distill-vo}
      & \dl{volt-s-distill-nuscenes} & \dl[\best]{volt-s-distill-vo-nuscenes} \\
    \bottomrule
  \end{tabular}
  \caption{\textbf{Phasor attention in Volt-S} (val mIoU, TTA), Volt's single-dataset recipe. RoPE: Volt's released runs. PA: phasor attention. CNN: Volt's convolutional teacher. In parentheses: change from RoPE without a teacher. Best in \textbf{bold}.}
  \label{tab:volt}
\end{table}

To test phasor attention outside PhasorPT, we add it to Volt-S~\cite{yilmaz2026volt}.
Volt-S keeps its voxel tokenizer and its tokens at integer patch coordinates, so this isolates the effect of geometry-aware values from that of continuous positions (see the grid rows of \cref{tab:components}).
We keep Volt-S's single-dataset training configurations unchanged and only add the value rotation of \cref{sec:method_phasor} to every attention block, on Volt's axial frequencies with a learned scale per head and axis, initialized to zero, so every run starts out exactly as the Volt-S baseline.
On ScanNet, phasor attention improves Volt-S both with and without distillation from the convolutional teacher (\cref{tab:volt}).
On nuScenes, it is slightly below RoPE without the teacher and improves it with the teacher.

\subsection{Distillation}
\label{sec:exp_selfdistill}

Volt reaches its single-dataset results by distilling from a sparse-convolutional U-Net \textit{trained on the same data}~\cite{yilmaz2026volt}, which transfers the inductive biases of the convolutional teacher to the transformer~\cite{abnar2020transferring,chen2022dearkd,zhao2023cskd}; on ScanNet, the teacher also labels the $\approx25\%$ of the training points without a ground-truth label.
Even without a teacher, PhasorPT matches or exceeds Volt-S with its convolutional teacher on ScanNet, ScanNet200, and nuScenes (\cref{tab:main}).
With the same teacher and loss, PhasorPT surpasses the distilled Volt-S on every dataset we evaluate with the teacher, and all baselines on ScanNet, ScanNet200, and SemanticKITTI.

\subsection{Runtime and Memory}
\label{sec:exp_runtime}

\begin{table}[t]
  \centering
  \small
  \setlength{\tabcolsep}{3pt}
  \newcommand{\rt}[2]{\result{rt-#1}{#2}}
  \newcolumntype{C}[1]{>{\centering\arraybackslash}p{#1}}
  \begin{tabular}{lcc C{3.0em}C{3.0em} C{1.9em}C{1.9em}}
    \toprule
    & Conv- & & \multicolumn{2}{c}{Inference (ms)} & \multicolumn{2}{c}{Training} \\
    \cmidrule(lr){4-5} \cmidrule(lr){6-7}
    Method & free & Params & batch 1 & batch 8 & ms & GB \\
    \midrule
    PTv3~\cite{wu2024ptv3} & -- & \rt{ptv3}{params} & \rt{ptv3}{infer-ms} & \rt{ptv3_b8}{infer-ms-scene} & \rt{ptv3}{train-ms} & \rt{ptv3}{train-mem} \\
    LitePT-S~\cite{yue2026litept} & -- & \rt{litept}{params} & \best{\rt{litept}{infer-ms}} & \best{\rt{litept_b8}{infer-ms-scene}} & \best{\rt{litept}{train-ms}} & \best{\rt{litept}{train-mem}} \\
    Volt-S~\cite{yilmaz2026volt} & $\checkmark$ & \rt{volt}{params} & \second{\rt{volt}{infer-ms}} & \rt{volt_b8}{infer-ms-scene} & \second{\rt{volt}{train-ms}} & \second{\rt{volt}{train-mem}} \\
    \quad + PA & $\checkmark$ & \rt{volt-vo}{params} & \rt{volt-vo}{infer-ms} & \rt{volt-vo_b8}{infer-ms-scene} & \rt{volt-vo}{train-ms} & \rt{volt-vo}{train-mem} \\
    \midrule
    \ours PhasorPT (ours) & $\checkmark$ & \rt{phasorpt}{params} & \rt{phasorpt}{infer-ms} & \rt{phasorpt_b8}{infer-ms-scene} & \rt{phasorpt}{train-ms} & \rt{phasorpt}{train-mem} \\
    \quad RoPE only & $\checkmark$ & \rt{phasorpt-novo}{params} & \rt{phasorpt-novo}{infer-ms} & \second{\rt{phasorpt-novo_b8}{infer-ms-scene}} & \rt{phasorpt-novo}{train-ms} & \rt{phasorpt-novo}{train-mem} \\
    \quad voxel tok. & $\checkmark$ & \rt{phasorpt-voxtok}{params} & \rt{phasorpt-voxtok}{infer-ms} & \second{\rt{phasorpt-voxtok_b8}{infer-ms-scene}} & \rt{phasorpt-voxtok}{train-ms} & \rt{phasorpt-voxtok}{train-mem} \\
    \bottomrule
  \end{tabular}
  \caption{\textbf{Efficiency} (ScanNet val; protocol in \cref{sec:exp_setup}). PTv3 and LitePT-S are hybrids of sparse convolutions and attention; Conv-free as in \cref{tab:main}. PA: phasor attention. RoPE only: without the value rotation. Voxel tok.: Volt's patch embedding and unpooling instead of phasor pooling and unpooling. PTX~\cite{reichardt2026ptx} is not included, as its code is not public. Best in \textbf{bold}, second best \underline{underlined}.}
  \label{tab:runtime}
\end{table}

Phasor pooling and unpooling are a single pass over the points, $\mathcal{O}(N C M)$ for $N$ points, $C$ channels and $M$ waves, without neighbor search or per-slot weights; the $[N, C, 2M+1]$ outer products can be fused into the reduction and need not be stored.
Pooling the points directly into tokens also skips the finest stage of the PTv3 hierarchy, where LitePT~\cite{yue2026litept} found attention to be the most expensive and to contribute the least to accuracy.

At the scale of a ScanNet scene, LitePT-S, which uses convolutions instead of attention in its early stages, is the fastest and leanest model (\cref{tab:runtime}).
Among the convolution-free models, Volt-S is the fastest at batch 1: the pooling, value rotation, and two streams of PhasorPT launch many more and smaller kernels (\cref{sec:suppl_runtime}).
At batch 8, this overhead is amortized, and PhasorPT becomes faster per scene than both Volt-S and PTv3.
On larger scenes, merged from several ScanNet scenes, the latency of Volt-S grows superlinearly with the number of points, while PhasorPT and PTv3 grow about linearly (\cref{fig:scaling}).
At every scene size, the value rotation adds only a moderate fraction to the inference time.

%% file: sec/6_conclusion.tex
\section{Conclusion}
\label{sec:conclusion}

Dropping sparse convolutions removes the need for a voxel grid, but convolution-free point transformers do not yet make full use of the continuous point geometry: their tokenizers discard where the points lie inside each cell, and the value path of their attention is unaware of geometry.
We introduced two operators that take the geometry directly from the continuous point coordinates.
Phasor pooling embeds each token from the exact positions of its points and keeps the grid only to group them.
Phasor attention makes the attention values geometry-aware by rotating them with the same phasors that RoPE applies to queries and keys, at no per-pair cost while remaining compatible with efficient attention kernels.
Their gains are largest and largely complementary when the model has to rely on the geometry alone, without color and normals, which indicates that they improve how the model uses the point geometry.
Building on both, PhasorPT operates on continuous coordinates throughout and matches or exceeds PTv3 on four datasets with half its parameters, without a single convolution; distilled from a convolutional teacher as in Volt, it surpasses all baselines on ScanNet, ScanNet200, and SemanticKITTI.
As a drop-in, phasor attention also improves Volt itself in three of four tested settings, which suggests that geometry-aware values help beyond our architecture.

\paragraph{Limitations.}
We train on single datasets only, without pretraining, multi-dataset training, or a scaling study, and we evaluate only semantic segmentation.
Our best results still rely on a convolutional teacher during training, which reintroduces sparse convolutions and a second model to train.
Compared to Volt, PhasorPT gives up part of its simplicity: its two streams and serialized patches reintroduce locality biases that Volt's global attention removes, and the pooling and value rotation add complexity and a runtime overhead at small batch sizes.

%% file: sec/7_statements.tex
\section*{LLM Usage Statement}
We used an LLM-based coding assistant (Claude Code) for parts of the implementation: parts of the model code, tools for training, evaluation, benchmarking, and analysis, experiment configurations, plotting scripts, and the script that parses the training logs into the result tables.
We also used it for writing: drafting and revising text, presenting the derivations and checking consistency.
The authors designed the method and the experiments, reviewed and tested all code, verified all text, and take full responsibility for it.

\iftoggle{cvprfinal}{%
\section*{Acknowledgments}
The project on which this publication is based was funded by the Federal Ministry of Research, Technology and Space under the funding code ``TwinSim4Brownfield'' 16IS24052B.
Parts of this work were supported by the KI-Servicezentrum Berlin-Brandenburg at the Hasso Plattner Institute, funded by the Federal Ministry of Research, Technology and Space under the funding code 16IS22092.
The responsibility for the content of this publication lies with the authors.
}{}

%% file: sec/X_suppl.tex
\clearpage
\ifdefined\arxivversion\else\setcounter{page}{1}\fi  %
\maketitlesupplementary
\appendix

\section{Geometric Heads in RoPE Attention}
\label{sec:suppl_heads}
\Cref{sec:method_blind} rests on two observations about trained 3D RoPE transformers: some heads attend by relative position, and on sparse data the offset such a head prefers is often empty, so that its attention mass lands elsewhere.
We test both in Volt-S with RoPE, trained on ScanNet (\cref{sec:exp_dropin}), whose tokens lie at the integer coordinates $\mathbf{p} \in \mathbb{Z}^3$ of its 10\,cm patches.

\begin{figure}[t]
  \centering
  \includegraphics[width=\linewidth]{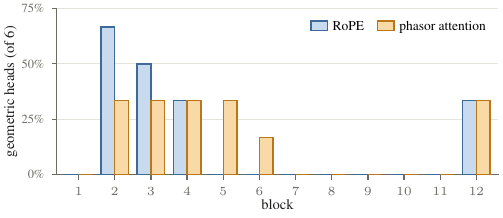}
  \caption{\textbf{The geometric heads of Volt-S concentrate in the early blocks and the last one.} ScanNet val, with RoPE attention and with phasor attention: share per block of the heads whose attention logits a function of the offset alone predicts on held-out scenes ($R^2 > 0.5$).}
  \label{fig:geom_heads}
\end{figure}

\paragraph{Geometric heads.}
For one head, let $s_{ij}$ be the attention logit of query token $i$ for key token $j$, after RoPE.
For 1000 sampled queries in each of 12 validation scenes, we take the keys within $\pm 4$ patches ($\pm 40$\,cm) of the query and fit
\begin{equation}
  s_{ij} \approx b_{\boldsymbol{\delta}} + c_i, \quad \boldsymbol{\delta} = \mathbf{p}_j - \mathbf{p}_i \in \{-4, \dots, 4\}^3,
\end{equation}
with a free value $b_{\boldsymbol{\delta}}$ per offset, shared by all queries and scenes, and a constant $c_i$ per query, which absorbs everything that is equal for all keys of a query and thus does not change the softmax; empty offsets drop out of the fit.
The score of a head is the $R^2$ of this fit on a held-out scene, with $b$ fitted on the other scenes, \ie the share of the variation of a query's logits over its keys that the offset alone explains.
We call a head geometric if its score exceeds $0.5$.
By this criterion, 11 of the 72 heads of Volt-S are geometric (\cref{tab:suppl_heads}); they concentrate in the early blocks and the last one (\cref{fig:geom_heads}).
Their queries and keys are dominated by a component shared by all tokens ($74\%$ and $80\%$ of their energy), so their logits come from the RoPE rotation rather than from content that happens to correlate with the offset.

\paragraph{Empty offsets.}
We take the non-self offset with the largest $b_{\boldsymbol{\delta}}$ as the preferred offset of a geometric head.
It holds a token for only about half of the queries.
For the others, the softmax distributes all of the mass over the other offsets, mostly within $\pm 40$\,cm.
With RoPE, the values carry no position, so the output of the head is the same kind of content average in both cases; whether the mass landed at the preferred offset or elsewhere is not visible in it, which \cref{sec:suppl_vo_phase} tests directly.

\paragraph{First block.}
The first block of Volt-S has no geometric head, and its attention is nearly switched off: its update to the residual stream is far smaller than that of its FFN, and replacing its attention output by a constant vector (its mean over the evaluation scenes) costs only $0.4$ mIoU ($73.5 \to 73.1$, 24 val scenes, no TTA), while removing it costs $3.9$ ($69.6$).
The block thus acts mainly as a bias and an FFN on the patch embedding, which is why PhasorPT places an FFN block directly after the tokenizer, before the first attention (\cref{fig:overview}).

\begin{table}[t]
  \centering
  \small
  \setlength{\tabcolsep}{4pt}
  \begin{tabular}{lcc}
    \toprule
    & RoPE & phasor attn. \\
    \midrule
    heads, $R^2 > 0.3 \,/\, 0.5 \,/\, 0.7$ (of 72) & 21/11/5 & 15/11/2 \\
    \midrule
    \multicolumn{3}{l}{\emph{Probe of the mean offset ($\pm 40$\,cm), geometric heads}} \\
    from the block input & 0.52 & 0.48 \\
    from the head output & 0.33 & 0.65 \\
    \bottomrule
  \end{tabular}
  \caption{\textbf{Geometric heads of Volt-S} (ScanNet val, 12 scenes). Top: number of heads whose logits the offset predicts on held-out scenes above an $R^2$ threshold. Bottom: test $R^2$ of probes (\cref{sec:suppl_vo_phase}) to the attention-weighted mean offset of the keys within $\pm 40$\,cm, from the input of the block or from the output of the head.}
  \label{tab:suppl_heads}
\end{table}

\section{Phasor Attention in Trained Models}
\label{sec:suppl_vo_phase}
We test whether phasor attention makes the head outputs reveal where the attention landed (\cref{sec:method_analysis}), in Volt-S trained on ScanNet with phasor attention, whose value-rotation scales start at zero (\cref{sec:exp_dropin}), on the same scenes as \cref{sec:suppl_heads}.

\paragraph{Learned rotation.}
Training moves the scales clearly away from zero.
After the first block, the values of the attended keys are rotated relative to the query by $10$--$20^\circ$ on average, and by about $50^\circ$ in the highest-frequency channel pairs.
The attention-weighted sum of these phasors keeps a magnitude of $0.86$--$0.94$ on average (about $0.6$ in the highest-frequency pairs), so the rotation mostly operates in the regime of \cref{eq:charfn_approx}: it shifts the phase of the output while passing most of the content.
The learned scale of a head is rank-correlated with its geometric score (Spearman $\rho = 0.59$), but hardly with its semantic lift ($\rho = 0.16$), the share of its attention mass on keys of the query's label minus the share expected from the distances of these keys alone: the model rotates the values where the attention is geometric, not where it is semantic.

\begin{figure}[t]
  \centering
  \includegraphics[width=\linewidth]{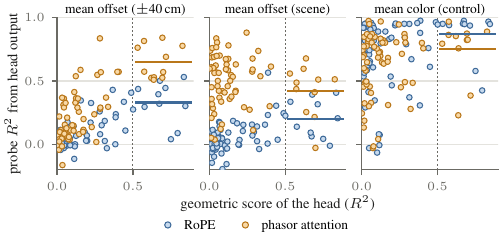}
  \caption{\textbf{Phasor attention makes the head output reveal where the attention landed.} Volt-S on ScanNet val, with RoPE and with phasor attention. Each point is one of the 72 heads, placed at its geometric score (\cref{sec:suppl_heads}; dotted: threshold $0.5$). Test $R^2$ of probes from the head output to the attention-weighted mean offset of the keys within $\pm 40$\,cm (left) and over the scene (middle), and to their mean color (right, control), averaged over the three coordinates or channels; lines: mean over the geometric heads. Not shown: two non-geometric heads below $-0.2$ (middle, phasor attention).}
  \label{fig:probe}
\end{figure}

\paragraph{Where the attention landed.}
We train probes to reconstruct, from the output of a head, the attention-weighted mean offset of its keys, within $\pm 40$\,cm and over the whole scene, and, as a content control, the attention-weighted mean color of its keys (\cref{fig:probe}).
With RoPE, the output reveals little of either offset, even in the geometric heads; with phasor attention, the mean offset within $\pm 40$\,cm becomes decodable about twice as well in the geometric heads ($R^2$ $0.33 \to 0.65$), and the mean offset over the scene across all heads ($0.07 \to 0.44$).
The color stays decodable in both models ($0.87$ and $0.75$ in the geometric heads), so the rotation adds the geometry at a small cost in content.
The head output is not the only source of this information: the features of a query already partly encode its occupied neighborhood, and the input of the block alone predicts the local mean offset of a geometric head with $R^2 \approx 0.5$ in both models (\cref{tab:suppl_heads}).
With RoPE, the output of a geometric head thus tells less about where its attention landed than its query already knew ($0.33$), with phasor attention more ($0.65$).
The probes are linear (ridge regression) or a small MLP, whichever is better on two held-out training scenes, trained on 8 scenes and tested on 4 others, with 500 queries each; their test $R^2$ varies by about $\pm 0.1$ at this sample size.

\section{Phasor Attention as a Characteristic Function}
\label{sec:suppl_charfn}
With weights $a_{nm} \geq 0$ that sum to one, $\varphi_n$ (\cref{eq:charfn}) is the characteristic function of the offsets under the attention distribution of query $n$, \ie its Fourier transform, and \cref{eq:charfn_approx} is its second-order expansion around the mean.
We derive it to make explicit what a single slice of the output reports about this distribution, and under which condition.

\paragraph{Projection.}
A slice with frequency $\boldsymbol{\omega}$ sees the offsets only through their projections $x_m = \boldsymbol{\omega}^\top \boldsymbol{\Delta}_{nm}$, so that $\varphi_n(\boldsymbol{\omega}) = \sum_m a_{nm}\, e^{\mathrm{i} x_m}$ depends only on the distribution of the offsets along $\boldsymbol{\omega}$, scaled by $\|\boldsymbol{\omega}\|$.
As the projection is linear, the attention-weighted mean and variance of the $x_m$ follow from the mean $\boldsymbol{\mu}_n$ and covariance $\mathbf{C}_n$ of the offsets:
\begin{equation}
\begin{aligned}
  \bar{x} &= \sum\nolimits_m a_{nm}\, \boldsymbol{\omega}^\top \boldsymbol{\Delta}_{nm} = \boldsymbol{\omega}^\top \boldsymbol{\mu}_n, \\
  \sigma^2 &= \sum\nolimits_m a_{nm} \big(\boldsymbol{\omega}^\top (\boldsymbol{\Delta}_{nm} - \boldsymbol{\mu}_n)\big)^2 = \boldsymbol{\omega}^\top \mathbf{C}_n \boldsymbol{\omega}.
\end{aligned}
\end{equation}

\paragraph{Expansion.}
We pull the phase of the mean out of the sum, which leaves the deviations $\epsilon_m = x_m - \bar{x}$, and expand the remaining phasor, $e^{\mathrm{i}\epsilon} = 1 + \mathrm{i}\epsilon - \tfrac{1}{2}\epsilon^2 - \tfrac{\mathrm{i}}{6}\epsilon^3 + \dots$:
\begin{equation}
\begin{aligned}
  \varphi_n(\boldsymbol{\omega})
  &= e^{\mathrm{i}\bar{x}} \sum\nolimits_m a_{nm}\, e^{\mathrm{i}\epsilon_m} \\
  &\approx e^{\mathrm{i}\bar{x}} \sum\nolimits_m a_{nm} \big(1 + \mathrm{i}\epsilon_m - \tfrac{1}{2}\epsilon_m^2\big) \\
  &= e^{\mathrm{i}\bar{x}} \big(1 - \tfrac{1}{2}\sigma^2\big).
\end{aligned}
\end{equation}
The weights sum to one and the deviations from the mean average to zero, so the first-order term vanishes, while the second-order term gives the variance $\sigma^2$; this is \cref{eq:charfn_approx}.

\paragraph{Reading the output.}
For a content-independent value, the slice is $\bar{v}\, \varphi_n(\boldsymbol{\omega})$ (\cref{sec:method_analysis}).
Its phase, $\boldsymbol{\omega}^\top \boldsymbol{\mu}_n$, is a linear measurement of where the attention mass landed, along $\boldsymbol{\omega}$, and its magnitude, $1 - \tfrac{1}{2}\boldsymbol{\omega}^\top \mathbf{C}_n \boldsymbol{\omega}$, decreases with how far the mass spreads along $\boldsymbol{\omega}$.
Slices with at least three linearly independent frequencies thus determine the full mean offset $\boldsymbol{\mu}_n$ in 3D, which the probes of \cref{sec:suppl_vo_phase} recover from the head output.
The first dropped term, $-\tfrac{\mathrm{i}}{6} \sum_m a_{nm} \epsilon_m^3$, is the skewness of the projected offsets; it shifts the phase away from the mean when the mass is spread asymmetrically along $\boldsymbol{\omega}$.

\paragraph{Validity.}
The terms of third and higher order are small while $\sigma \ll 1$, \ie while the spread of the attention mass along $\boldsymbol{\omega}$ is small compared to the wavelength $2\pi / \|\boldsymbol{\omega}\|$.
For larger spreads, $|\varphi_n(\boldsymbol{\omega})| \leq 1$ still holds, with equality only if the projected offsets agree up to multiples of $2\pi$, \ie up to multiples of the wavelength along $\boldsymbol{\omega}$.
A single high frequency thus cannot tell such offsets apart, but the slices together, with frequencies spanning several orders of magnitude, resolve the attention distribution at coarse and fine scales.

\section{Distillation}
\label{sec:suppl_selfdistill}
Volt's sparse-convolutional teacher, trained on the same data, improves both Volt-S and PhasorPT (\cref{tab:volt,tab:main}).
This gain is commonly attributed to the inductive biases of the convolutions~\cite{abnar2020transferring,chen2022dearkd,zhao2023cskd}, but distillation also helps with teachers of the student's own architecture~\cite{furlanello2018born} or weaker ones, which Yuan \etal attribute to the regularization by the soft targets~\cite{yuan2020revisiting}.

\begin{figure}[t]
  \centering
  \includegraphics[width=\linewidth]{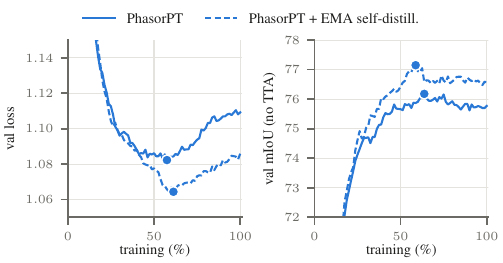}
  \caption{\textbf{EMA self-distillation keeps the validation loss of PhasorPT lower and its mIoU higher from about the middle of training on.} ScanNet val, per evaluation epoch over the fraction of training (320 epochs; no TTA; dots: minimum loss, maximum mIoU). The validation loss is shown without self-distillation loss, so the curves are comparable.}
  \label{fig:curves}
\end{figure}

\paragraph{Results.}
We distill PhasorPT from teachers of its own architecture: a frozen PhasorPT trained without a teacher, as in born-again networks~\cite{furlanello2018born}, and the EMA of the student's own weights, which needs no separately trained network (\cref{tab:suppl_teachers}).
On ScanNet, the EMA teacher with soft targets comes close to the convolutional teacher, and on nuScenes it is on par with it; on ScanNet200 and for PhasorPT with RoPE only, it does not help (\cref{tab:suppl_sd_datasets}).
During training, the EMA teacher keeps the validation loss of PhasorPT lower and its mIoU higher from about the middle of training on, although the loss still rises late in training (\cref{fig:curves}), similar to the effect of the convolutional teacher on Volt-S shown by Volt~\cite{yilmaz2026volt}.
With single seeds, differences of a few tenths are within the noise, so these results suggest, but do not establish, that part of the gain of a teacher comes from regularization.
As the EMA teacher does not help on ScanNet200, the main results use the convolutional teacher, as Volt does.

\paragraph{Loss.}
As in Volt~\cite{yilmaz2026volt}, a second linear head learns from the teacher on all points, including the $25\%$ of the ScanNet training points without a benchmark label, with the loss
\begin{equation}
  \mathcal{L} = (1 - w)\,\mathcal{L}_\text{gt} + w\,\mathcal{L}_\text{distill}, \qquad w = \tfrac{1}{2},
  \label{eq:distill_loss}
\end{equation}
where $\mathcal{L}_\text{gt}$ is the loss of the ground-truth head (\cref{tab:suppl_train}) and $\mathcal{L}_\text{distill}$ that of the second head: the same loss to the teacher's argmax (hard targets) or the cross-entropy to its softmax (soft targets).
Only the ground-truth head is used at test time.
The EMA teacher (decay $0.9995$) labels each batch in evaluation mode, and its weight $w$ ramps from $0$ at $10\%$ to $\tfrac{1}{2}$ at $20\%$ of the updates.

\begin{table}[t]
  \centering
  \small
  \begin{tabular}{llc}
    \toprule
    Teacher & Targets & mIoU \\
    \midrule
    -- & -- & \result{pvit-d12-scannet}{miou} \\
    Volt's sparse-conv.\ U-Net & hard & \best{\result{pvit-d12-cnnkd-scannet}{miou}} \\
    EMA of the weights & hard & \result{pvit-d12-sd-hard}{miou} \\
    EMA of the weights & soft & \second{\result{pvit-d12-sd-scannet}{miou}} \\
    trained PhasorPT, frozen & hard & \result{pvit-d12-kd-hard}{miou} \\
    trained PhasorPT, frozen & soft & \result{pvit-d12-kd-soft}{miou} \\
    \bottomrule
  \end{tabular}
  \caption{\textbf{Teachers for PhasorPT} (ScanNet val mIoU, TTA, single seed). Every teacher is distilled into a second head with weight $\tfrac{1}{2}$, as in Volt~\cite{yilmaz2026volt}. Frozen: the PhasorPT of the first row. Best in \textbf{bold}, second best \underline{underlined}.}
  \label{tab:suppl_teachers}
\end{table}

\begin{table}[t]
  \centering
  \small
  \newcommand{\dl}[2][\relax]{#1{\result{#2}{miou}} & \,{\scriptsize(\result{#2}{miou-delta})}}
  \setlength{\tabcolsep}{4pt}
  \begin{tabular}{lcr@{}lr@{}l}
    \toprule
    & & \multicolumn{4}{c}{Teacher} \\
    \cmidrule(lr){3-6}
    Dataset & -- & \multicolumn{2}{c}{EMA} & \multicolumn{2}{c}{CNN} \\
    \midrule
    ScanNet & \result{pvit-d12-scannet}{miou} & \dl{pvit-d12-sd-scannet} & \dl[\textbf]{pvit-d12-cnnkd-scannet} \\
    ScanNet, RoPE only & \textbf{\result{pvit-d12-rope-scannet}{miou}} & \dl{pvit-d12-rope-sd-scannet} & \multicolumn{2}{c}{--} \\
    ScanNet200 & \result{pvit-d12-scannet200-wide}{miou} & \dl{pvit-d12-scannet200-sd-wide} & \dl[\textbf]{pvit-d12-scannet200-cnnkd} \\
    nuScenes & \result{pvit-d12-nuscenes}{miou} & \dl[\textbf]{pvit-d12-nuscenes-sd} & \dl{pvit-d12-nuscenes-cnnkd} \\
    \bottomrule
  \end{tabular}
  \caption{\textbf{The EMA teacher comes close to the CNN teacher on ScanNet and matches it on nuScenes}, but does not help on ScanNet200 or with RoPE only. Val mIoU, TTA, single seed. EMA: EMA self-distillation with soft targets. CNN: Volt's convolutional teacher, as in \cref{tab:main}. RoPE only: PhasorPT without the value rotation. ScanNet200: 128-channel point output, as in \cref{tab:main}. In parentheses: change from the same model without a teacher. Best per row in \textbf{bold}.}
  \label{tab:suppl_sd_datasets}
\end{table}

\section{Runtime Breakdown}
\label{sec:suppl_runtime}
\paragraph{Kernel dispatch.}
At batch 1, the additional operations of PhasorPT, such as the value rotation, pooling and unpooling, and the second, global stream, launch many more and smaller kernels than Volt-S.
Their launch overhead leaves the NVIDIA A40 idle for about half of the forward pass, although the kernels themselves take less time than those of Volt-S, as the serialized attention is cheaper.
At batch 8, the launches are shared by eight scenes, and PhasorPT becomes faster per scene than Volt-S and PTv3 (\cref{tab:runtime}).

\begin{figure}[t]
  \centering
  \includegraphics[width=\linewidth]{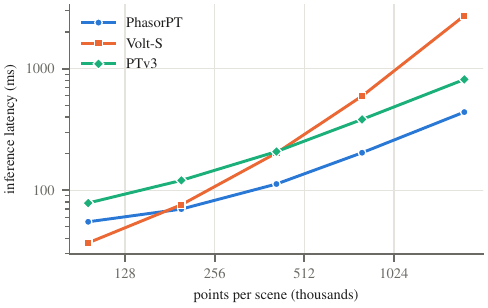}
  \caption{\textbf{PhasorPT grows about linearly with the scene size and is faster than PTv3 at every size.} Volt-S attends globally over all tokens and grows superlinearly; PhasorPT and PTv3 attend in serialized patches. Inference latency for 1 to 16 ScanNet val scenes merged into one virtual scene (side by side on a grid in the $xy$ plane; batch 1, mixed precision, A40; mean over 3 merged scenes per size).}
  \label{fig:scaling}
\end{figure}

\paragraph{Scaling with scene size.}
To measure larger scenes than ScanNet provides, we merge 1 to 16 ScanNet val scenes into one virtual scene, placed side by side in the $xy$ plane, and run it at batch 1 (\cref{fig:scaling}).
The latency of Volt-S, with global attention, grows superlinearly with the number of points, while PhasorPT and PTv3 grow about linearly; PhasorPT overtakes Volt-S from two merged scenes on and is faster than PTv3 at every size.

\section{Additional Ablations}
\label{sec:suppl_ablation}

\paragraph{Inference patch size.}
Following PTX~\cite{reichardt2026ptx}, which enlarges them fourfold, we enlarge the serialized patches at inference (\cref{tab:suppl_patch}).
With phasor attention in our architecture, the returns diminish beyond twice the training size, which all other results use.

\begin{table}[t]
  \centering
  \small
  \newcommand{\ip}[2]{\result{ips-#1}{#2} & \,{\scriptsize(\result{ips-#1}{#2-delta})}}
  \newcommand{\ipb}[3]{#3{\result{ips-#1}{#2}} & \,{\scriptsize(\result{ips-#1}{#2-delta})}}
  \setlength{\tabcolsep}{4pt}
  \begin{tabular}{rc*{3}{r@{}l}}
    \toprule
    Patch & Scale & \multicolumn{2}{c}{mIoU} & \multicolumn{2}{c}{mAcc} & \multicolumn{2}{c}{allAcc} \\
    \midrule
    512 & $\times 0.5$ & \ip{512}{miou} & \ip{512}{macc} & \ip{512}{allacc} \\
    1024 & $\times 1$ & \result{ips-1024}{miou} & & \result{ips-1024}{macc} & & \result{ips-1024}{allacc} & \\
    \rowcolor{gray!15}[0pt][\dimexpr2\tabcolsep\relax]\multicolumn{1}{>{\columncolor{gray!15}[\tabcolsep][\dimexpr2\tabcolsep\relax]}r}{2048} & $\times 2$
      & \ipb{2048}{miou}{\best} & \ipb{2048}{macc}{\best} & \best{\result{ips-2048}{allacc}}
      & \multicolumn{1}{@{}>{\columncolor{gray!15}[0pt][\tabcolsep]}l}{\,{\scriptsize(\result{ips-2048}{allacc-delta})}} \\
    4096 & $\times 4$ & \ipb{4096}{miou}{\second} & \ipb{4096}{macc}{\second} & \ipb{4096}{allacc}{\second} \\
    \bottomrule
  \end{tabular}
  \caption{\textbf{Larger inference patches help up to twice the training size}, which all other results use; beyond it the returns diminish. ScanNet val mIoU, no TTA. One PhasorPT checkpoint (with EMA self-distillation, trained with patches of 1024 tokens), tested with other patch sizes in both streams; scale: relative to the training size. Mean of \result{ips}{repeats} passes, which differ in the random point kept per voxel (std at most \result{ips}{maxstd}); every pass keeps the same points for all patch sizes. In parentheses: change from the training patch size. \graybg{Gray}: the setting of all other results.}
  \label{tab:suppl_patch}
\end{table}

\paragraph{Design choices.}
\Cref{tab:suppl_design} varies single options of PhasorPT.
Removing the norms around the global branch and the U-Net-style fusion cost the most; the other changes stay within the run-to-run noise of a single seed.

\begin{table}[t]
  \centering
  \small
  \newcommand{\dl}[1]{\result{#1}{miou}\,{\scriptsize(\result{#1}{miou-delta})}}
  \begin{tabular}{lc}
    \toprule
    Variant & mIoU \\
    \midrule
    \ours PhasorPT & \result{pvit-d12-scannet}{miou} \\
    \quad main-stage patches of 4096 tokens & \dl{pvit-d12-global} \\
    \quad global branch with 6 heads of 64, as Volt-S & \dl{pvit-d12-hd64} \\
    \quad no QK-norm & \dl{pvit-d12-noqknorm} \\
    \quad no norms around the global branch$^\dagger$ & \dl{pvit-d12-nobranchnorm} \\
    \quad U-Net-style fusion$^\ddagger$ & \dl{pvit-d12-unet} \\
    \quad per-scene drop path & \dl{pvit-d12-dpscene} \\
    \bottomrule
  \end{tabular}
  \caption{\textbf{Removing the norms around the global branch and the U-Net-style fusion cost the most}; the other options stay within the noise of a single seed. ScanNet val mIoU, TTA, no teacher. Each indented row changes one option of PhasorPT; in parentheses: change from PhasorPT. \graybg{Gray}: PhasorPT. $^\dagger$Max pooling without the norm on its input, and unpooling as a plain broadcast of $W x$ instead of norm $\to$ linear $\to$ norm. $^\ddagger$The global branch is fused into the 4\,cm tokens as in the PTv3 decoder (linear $\to$ norm on both inputs) instead of the residual update.}
  \label{tab:suppl_design}
\end{table}

\section{Implementation Details}
\label{sec:suppl_impl}
\subsection{Serialization}
\label{sec:suppl_serialization}
\begin{figure}[t]
  \centering
  \includegraphics[width=\linewidth]{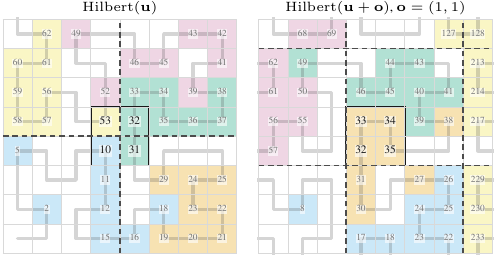}
  \caption{\textbf{Grid offset of the serialization} in 2D, on a sparse set of occupied cells. Numbers: Hilbert code of each occupied cell, gray line: the curve; colors: patches of 8 consecutive occupied cells in the sorted order.
  Left: the curve fills aligned $4 \times 4$ blocks (dashed) one after another, so the four center cells, mutual neighbors across a high bit flip, get the codes 10, 31, 32, 53 and fall into three different patches.
  Right: shifting the grid by $\mathbf{o} = (1, 1)$ moves the block boundaries, and the same cells get the consecutive codes 32--35 and share a patch; the patches elsewhere are cut differently, but not less compact.}
  \label{fig:grid_offset}
\end{figure}

We serialize in Hilbert order, with one axis permutation per block and patches of 1024 tokens in both the local stream and the global branch, enlarged twofold at inference (\cref{tab:suppl_patch}).
In addition to the axis permutation, the $i$-th order shifts the grid by $(i, i, i)$ cells before encoding.
Space-filling curves visit aligned blocks of $2^n$ cells one after another, so patches split the grid along the same block boundaries in every block, and neighboring cells across a boundary at a high bit flip lie far apart in the order (\cref{fig:grid_offset}).
An axis permutation alone does not move these boundaries, since an aligned block stays aligned under any permutation of the axes; the offset moves them between blocks.
Removing the offset at inference from a model trained with it lowers the mIoU (\cref{tab:suppl_gridshift}), so the trained model relies on the varying boundaries.%

\subsection{Architecture}
\label{sec:suppl_arch}
\Cref{tab:suppl_model} lists all model settings per dataset.

\paragraph{Streams.}
The local stream has 128 channels (4 heads of 32), with two blocks before and two after the global branch.
The global branch has 384 channels (12 heads of 32) and 12 blocks, the depth and width of Volt-S~\cite{yilmaz2026volt}.
Its max pooling acts on the pre-normalized local stream; its broadcast update is normalized, with the norm gain initialized to $0.1$, and followed by an FFN block.
The global branch thus acts as one residual block on the local stream, normalized at its input and its output as in the sandwich norm~\cite{ding2021cogview}; without these norms, PhasorPT performs worse (\cref{tab:suppl_design}).
All norms are RMSNorm~\cite{zhang2019rmsnorm}; only the max pooling into the global branch is followed by a LayerNorm, as in the grid pooling of PTv3~\cite{wu2024ptv3}, which also removes the mean that the max adds to the pooled features.
Its update is the same for all tokens of a global cell; future work could look into better ways to merge the global context with the fine spatial detail of the local stream, or integrate phasor pooling and attention into a Volt-style model with a single stream of tokens, as done here for phasor attention (\cref{tab:volt}).

\paragraph{Hierarchy.}
All levels (the points and the 4 and 16\,cm cells) are built from a single sort of the Morton codes of the points (\cref{alg:morton}).
Dropping the $3s$ lowest bits of a Morton code gives the code of the cell $2^s$ times coarser, $\textsc{Morton}(\mathbf{u} \gg s) = \textsc{Morton}(\mathbf{u}) \gg 3s$, and keeps sorted codes sorted, so every pooling step only shifts the codes and merges runs of equal codes in the sorted sequence, without sorting again, and the children of each parent are contiguous.

\begin{algorithm}[t]
  \caption{Level hierarchy from a single sort.}
  \label{alg:morton}
  \begin{algorithmic}[1]
    \Require grid coordinates $\mathbf{u} \in \mathbb{N}^{N \times 3}$, strides $2^{s_1}, \dots, 2^{s_L}$
    \State $z \gets \textsc{Morton}(\mathbf{u})$ \Comment{scene index in the high bits}
    \State $z \gets \textsc{Sort}(z)$ \Comment{the only sort}
    \For{$l = 1, \dots, L$}
      \State $z \gets z \gg 3 s_l$ \Comment{still sorted}
      \State $z, \mathit{inv}_l, n_l \gets \textsc{UniqueConsecutive}(z)$
    \EndFor
    \State \Return parent indices $\mathit{inv}_l$, child counts $n_l$
  \end{algorithmic}
\end{algorithm}

\paragraph{RoPE frames.}
Each head rotates the positions by a learned rotation $\mathbf{F}$, parametrized by its axis-angle vector $\mathbf{r} \in \mathbb{R}^3$ (rotation by $\lVert\mathbf{r}\rVert$ about $\mathbf{r}/\lVert\mathbf{r}\rVert$, computed with Rodrigues' formula), initialized to $\mathbf{r} = \mathbf{0}$ (identity frame) and excluded from weight decay; PTX instead orthonormalizes learned vectors per head (Gram-Schmidt).
With fixed frames, every head encodes offsets along the same three coordinate axes; the learned frame lets each head choose its own axes, as the learned rotations of LieRE~\cite{ostmeier2025liere} and STRING~\cite{schenck2025string} and the learned frames of PTX~\cite{reichardt2026ptx}.
The per-head, per-axis position scale is initialized log-uniformly in $[0.8, 1.2]$ over the heads, so that the heads start at slightly different frequencies instead of identical ones.

\paragraph{Phasor attention.}
Queries and keys use axial RoPE: 15 of the 16 channel pairs of each 32-dimensional head are split into 5 bands per axis, at the frequencies $\theta_k = b^{-k/5}$, $k = 0, \dots, 4$, with the geometric falloff of RoPE~\cite{su2024roformer} and base $b$, in units of the cell size of the stream (\cref{tab:suppl_model}).
Since 32 is not divisible by 6, the last pair stays unrotated, as in partial RoPE~\cite{black2022gptneox}.
The values are rotated in the same 15 pairs, but not along the axes: each pair $\ell$ has its own learned 3D frequency, as the mixed frequencies of RoPE-Mixed~\cite{heo2024rope}, so the value path adds 45 parameters per head.
It is parametrized as $\boldsymbol{\omega}_\ell = \theta_\ell\, \mathbf{w}_\ell$, with the RoPE frequency $\theta_\ell$ of the pair, so that all $\mathbf{w}_\ell \in \mathbb{R}^3$ share one scale under the optimizer and the weight decay; $\mathbf{w}_\ell$ is initialized to zero.

\paragraph{Phasor pooling and unpooling.}
Pooling into the 4\,cm tokens uses $M = 32$ waves.
Each wave vector is parametrized by its frequency $\|\boldsymbol{\omega}_m\| = c \cdot \mathrm{sigmoid}(l_m)$ and its direction $\mathbf{d}_m / \|\mathbf{d}_m\|$, both learned without weight decay; the sigmoid acts on the log-scale, so low and high frequencies move alike under Adam.
The frequencies are initialized log-spaced from $\pi/2$ per cell (a quarter period across the cell) to $\pi$ per point spacing, the directions on a Fibonacci lattice of the upper hemisphere~\cite{gonzalez2010fibonacci} ($\pm\boldsymbol{\omega}$ give the same pair), with the magnitudes assigned to the directions in a golden-ratio (low-discrepancy) permutation, so that the waves of every frequency range point in all directions.
The waves thus start as a near-uniform sampling of the frequencies over all directions and scales of the cell, instead of a random one with gaps and clusters.
The bound $c$ is $1.2\times$ the Nyquist frequency of the input grid, $\pi$ per point spacing (2\,cm indoors, 5\,cm outdoors).
Above the Nyquist frequency, a wave advances its phase by more than $\pi$ from one grid point to the next, so the grid cannot resolve it: its value at a point then mostly reflects where in its voxel grid sampling kept the point, which is random, rather than the geometry of the cell.
Unpooling uses $M' = 16$ waves, initialized in the same way, and rank $r = 16$; $W_r$ is initialized to zero.

\paragraph{ScanNet200.}
In the base model, the per-point output has 64 channels, which feed the linear head, and a second one under distillation.
With the 200 classes of ScanNet200, this base model falls short of PTv3; widening the output to 128 channels with rank $r = 32$, which adds 0.1M parameters, closes the gap (\cref{tab:suppl_s200,tab:main}).

\begin{table}[t]
  \centering
  \small
  \begin{tabular}{lc}
    \toprule
    Point output & mIoU \\
    \midrule
    64 channels, $r = 16$ & \result{pvit-d12-scannet200}{miou} \\
    \ours 128 channels, $r = 32$ & \best{\result{pvit-d12-scannet200-wide}{miou}} \\
    \bottomrule
  \end{tabular}
  \caption{\textbf{A 128-channel per-point output closes the gap to PTv3 on ScanNet200} at 0.1M additional parameters. Val mIoU, TTA, single seed. \graybg{Gray}: as in \cref{tab:main}. Best in \textbf{bold}.}
  \label{tab:suppl_s200}
\end{table}

\subsection{Per-Point Drop Path}
\label{sec:suppl_droppath}
Stochastic depth~\cite{huang2016stochasticdepth} drops the residual branch of a block for a whole mini-batch, and in common implementations for a whole sample.
Point transformers built on Pointcept~\cite{pointcept2023} concatenate the scenes of a batch into one flat tensor of shape $N \times C$, with a separate buffer of the scene offsets, and their drop path, such as that of PTv3~\cite{wu2024ptv3}, draws the mask over the first dimension, \ie per point instead of per scene.
The resulting regularization is a per-token residual dropout rather than a random reduction of the depth: every scene passes through all blocks, while each block updates a random subset of its points.
We keep this per-point drop path for consistency with PTv3, LitePT~\cite{yue2026litept}, and Volt~\cite{yilmaz2026volt}, which all use it, with a rate increasing linearly over the block sequence (local, global, refinement) up to $0.3$ ($0.2$ on SemanticKITTI, \cref{tab:suppl_train}).
Dropping blocks per scene instead performs on par (\cref{tab:suppl_design}).

\subsection{Training Settings}
\label{sec:suppl_train}
\Cref{tab:suppl_train} lists the training settings on all datasets.
Every model trains on a single GPU with 48\,GB, an NVIDIA A40 or L40S.
Without the validation during training, a run takes about 5.5\,h on ScanNet, 8--9\,h on ScanNet200, 12--14\,h on nuScenes, and 7.5\,h on SemanticKITTI; training with the convolutional teacher takes about as long.
The staged schedule on Structured3D takes about 34\,h in total (\cref{sec:suppl_s3d}).

\begin{table}[t]
  \centering
  \footnotesize
  \setlength{\tabcolsep}{3pt}
  \begin{tabular}{@{}llll@{}}
    \toprule
    Config & Indoor & nuScenes & SemanticKITTI \\
    \midrule
    optimizer & \multicolumn{3}{l}{AdamW, $\beta = (0.9, 0.99)$} \\
    learning rate & $10^{-3}$ & $10^{-3}$ & $2 \cdot 10^{-3}$ \\
    weight decay & $0.05$ & $0.005$ & $0.02$ \\
    schedule & \multicolumn{3}{l}{one-cycle, cosine; warm-up $1\%$} \\
    batch size & \multicolumn{3}{l}{12, one GPU} \\
    epochs & 320$^*$ & 50 & 20 \\
    precision & \multicolumn{3}{l}{mixed (AMP)} \\
    gradient clipping & \multicolumn{3}{l}{5} \\
    loss & \multicolumn{3}{l}{CE (label smoothing $0.1$) + Lovász} \\
    EMA decay & \multicolumn{3}{l}{$0.9995$} \\
    drop path & $0.3$ & $0.3$ & $0.2$ \\
    \midrule
    grid sampling & 2\,cm ($50\%$)$^\ddagger$ & 5\,cm & 5\,cm \\
    crop (points) & sphere, 102{,}400 & -- & sphere, 120{,}000 \\
    augmentations & as PTv3 & as PTv3 & as Volt$^\dagger$ \\
    Mix3D & $p = 0.8$ & $p = 0.8$ & $p = 0.85$ \\
    test & \multicolumn{3}{l}{TTA (rotations, flips), patches $\times 2$} \\
    \bottomrule
  \end{tabular}
  \caption{\textbf{Training settings of PhasorPT.} Indoor: ScanNet, ScanNet200, Structured3D. No weight decay on the waves, position scales, and frames. $^\ddagger$On $50\%$ of the samples; the others enter at full resolution. $^*$Structured3D: \cref{sec:suppl_s3d}. $^\dagger$Instance cut-mix, 50\,m range, random crop to $60$--$100\%$ of the points~\cite{yilmaz2026volt}.}
  \label{tab:suppl_train}
\end{table}

\begin{table}[t]
  \centering
  \footnotesize
  \setlength{\tabcolsep}{4pt}
  \begin{tabular}{@{}lll@{}}
    \toprule
    & Indoor & Outdoor \\
    \midrule
    input features & color, normal & strength, log range, height$^*$ \\
    input grid & 2\,cm & 5\,cm \\
    tokens & 4\,cm & 10\,cm \\
    global branch & 16\,cm & 80\,cm \\
    \midrule
    local stream & \multicolumn{2}{l}{128 channels, 4 heads, 2 + 2 blocks} \\
    global branch & \multicolumn{2}{l}{384 channels, 12 heads, 12 blocks} \\
    FFN & \multicolumn{2}{l}{GELU, ratio 4} \\
    patches & \multicolumn{2}{l}{1024 tokens (training), 2048 (inference)} \\
    \midrule
    RoPE base, local & 100 & 350 \\
    RoPE base, global & 100 & 100 \\
    \midrule
    phasor pooling & \multicolumn{2}{l}{$M = 32$ waves} \\
    phasor unpooling & \multicolumn{2}{l}{$M' = 16$ waves, $r = 16$, 64 channels$^\dagger$} \\
    parameters & \multicolumn{2}{l}{22.6M$^\dagger$} \\
    \bottomrule
  \end{tabular}
  \caption{\textbf{Model settings of PhasorPT.} Indoor: ScanNet, ScanNet200, Structured3D; outdoor: nuScenes, SemanticKITTI. Tokens and global branch: cell size. RoPE: positions in units of the cell size of the stream (times 1.2), five bands per axis at $b^{-k/5}$, $k = 0, \dots, 4$, for base $b$; the per-head frames and scales and the waves are learned, with the initialization in \cref{sec:suppl_arch}. $^*$SemanticKITTI: $xyz$, strength. $^\dagger$ScanNet200: $r = 32$, 128 channels, 22.7M.}
  \label{tab:suppl_model}
\end{table}

\begin{table}[t]
  \centering
  \small
  \setlength{\tabcolsep}{4pt}
  \begin{tabular}{lr@{}lr@{}lr@{}l}
    \toprule
    Offset at inference & \multicolumn{2}{c}{mIoU} & \multicolumn{2}{c}{mAcc} & \multicolumn{2}{c}{allAcc} \\
    \midrule
    \rowcolor{gray!15}[0pt][\dimexpr2\tabcolsep\relax]\multicolumn{1}{>{\columncolor{gray!15}[\tabcolsep][\dimexpr2\tabcolsep\relax]}l}{with offset (as trained)}
      & \best{\result{gs-on}{miou}} & & \best{\result{gs-on}{macc}} & & \best{\result{gs-on}{allacc}}
      & \multicolumn{1}{@{}>{\columncolor{gray!15}[0pt][\tabcolsep]}l}{} \\
    without offset & \result{gs-off}{miou} & \,{\scriptsize(\result{gs-off}{miou-delta})} & \result{gs-off}{macc} & \,{\scriptsize(\result{gs-off}{macc-delta})} & \result{gs-off}{allacc} & \,{\scriptsize(\result{gs-off}{allacc-delta})} \\
    \bottomrule
  \end{tabular}
  \caption{\textbf{Removing the grid offset at inference lowers the mIoU}: the trained model relies on the varying patch boundaries. ScanNet val, no TTA. One PhasorPT checkpoint (without a teacher), trained with the per-block grid offset, tested with and without it; Hilbert order and patches of 2048 tokens in both cases. Mean of \result{gs}{repeats} passes (std at most \result{gs-off}{miou-std}). In parentheses: change from the offset. \graybg{Gray}: the setting of all other results.}
  \label{tab:suppl_gridshift}
\end{table}

\subsection{Structured3D Schedule}
\label{sec:suppl_s3d}
Structured3D is by far the largest of our indoor datasets, with about 18k training rooms against 1.2k scenes in ScanNet.
Due to limited compute, PhasorPT trained on it with a shorter schedule than on the other datasets: three consecutive runs, each starting a new one-cycle schedule of 40 epochs from the weights of the previous one, of which the second was stopped after 15 epochs, followed by 8 epochs without Mix3D at a tenth of the learning rate.
In total, these are 103 epochs, about half of the 200 epochs of LitePT~\cite{yue2026litept}; even so, PhasorPT matches PTv3 without a teacher (\cref{tab:main}).

\section{Qualitative Examples}
\label{sec:suppl_examples}
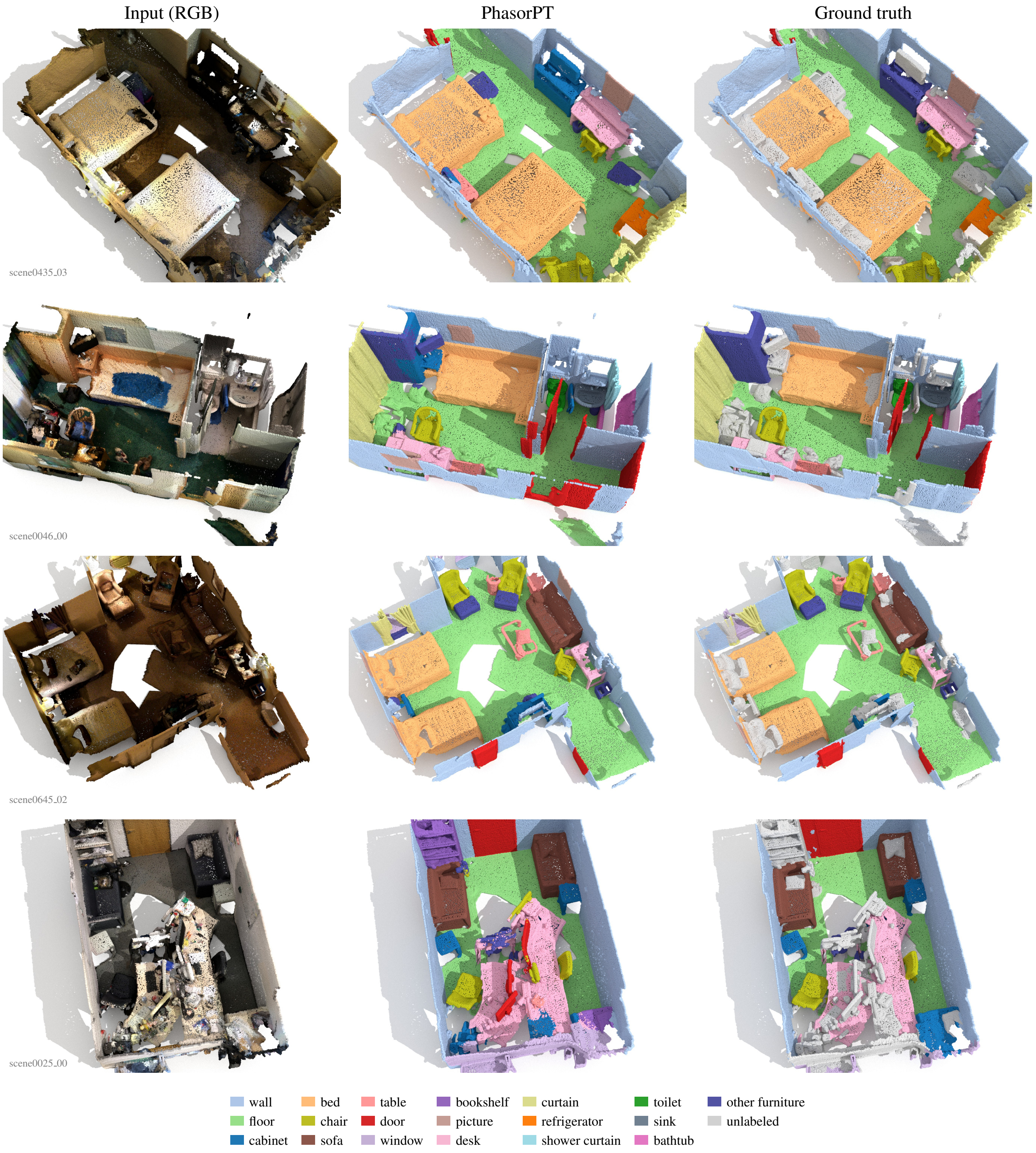
\begin{figure*}[p]
  \centering
  \input{figures/examples_scannet.tex}
  \caption{\textbf{ScanNet validation scenes.} Input colors, prediction of PhasorPT with the CNN teacher, and ground truth; unlabeled points in gray.}
  \label{fig:suppl_ex_scannet}
\end{figure*}

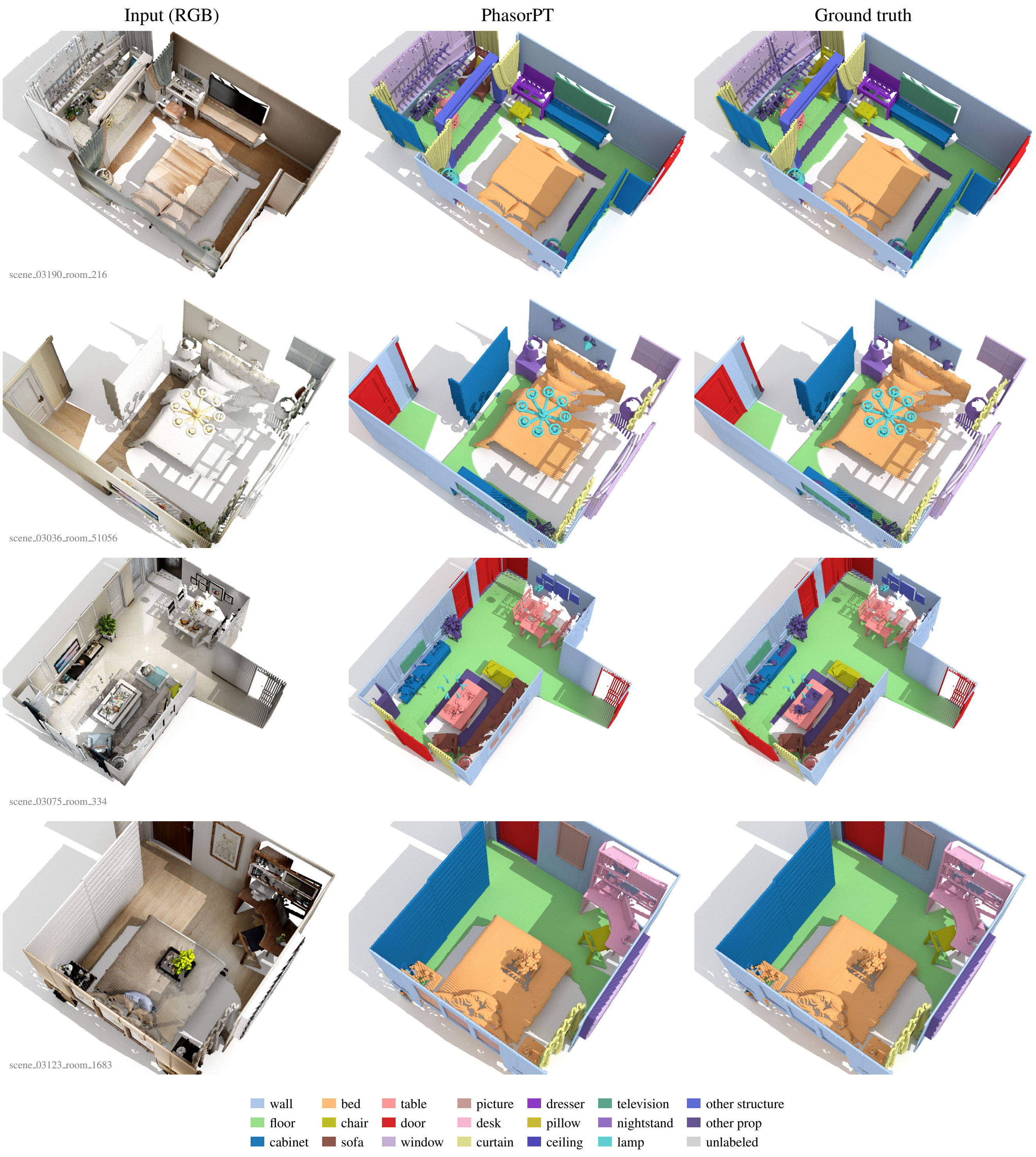
\begin{figure*}[p]
  \centering
  \input{figures/examples_structured3d.tex}
  \caption{\textbf{Structured3D validation rooms.} Input colors, prediction of PhasorPT without a teacher, and ground truth; unlabeled points in gray.}
  \label{fig:suppl_ex_s3d}
\end{figure*}

\begin{figure*}[p]
  \centering
  \input{figures/examples_kitti.tex}
  \caption{\textbf{SemanticKITTI validation scans} (sequence 08, top view of $50 \times 37.5$\,m around the sensor). Input remission, prediction of PhasorPT with the CNN teacher, and ground truth; unlabeled points in gray.}
  \label{fig:suppl_ex_kitti}
\end{figure*}
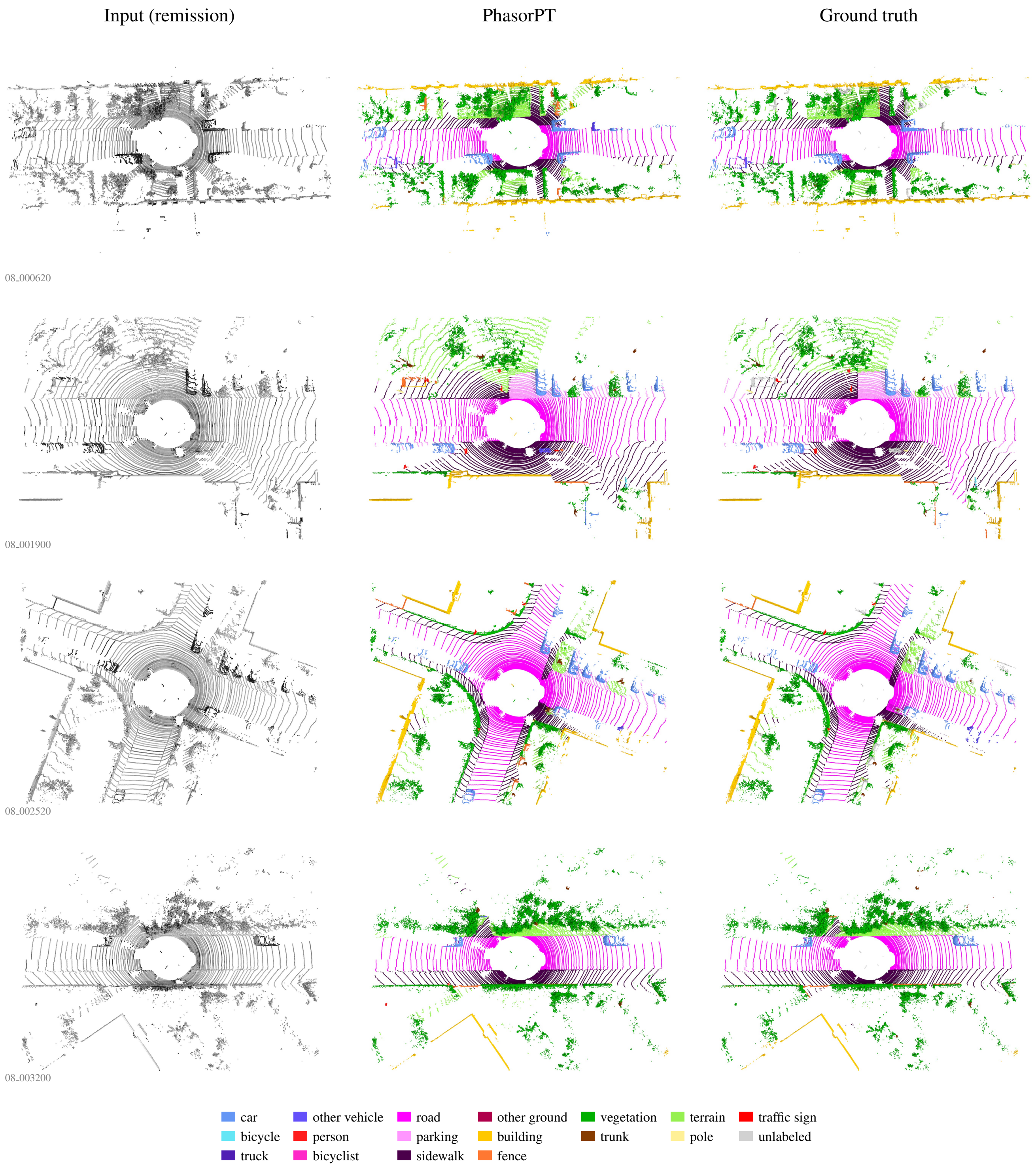

\Cref{fig:suppl_ex_scannet,fig:suppl_ex_s3d,fig:suppl_ex_kitti} show PhasorPT predictions on validation scenes of ScanNet, Structured3D, and SemanticKITTI.
The scenes are representative rather than the best: among the validation scenes with at least the median number of points and of classes, we show, among those whose per-scene mIoU is closest to the median, rooms that are easy to read in a single view.
The predictions on ScanNet and SemanticKITTI are from PhasorPT with the CNN teacher, those on Structured3D, which has no teacher, from PhasorPT without one (\cref{tab:main}).

%% file: figures/examples_scannet.tex
\ifdefined\exw\else\newlength\exw\newlength\exgap\fi
\setlength\exgap{4pt}\setlength\exw{\dimexpr(\linewidth-2\exgap)/3-0.5pt\relax}
\makebox[\exw]{\small Input (RGB)}\hspace{\exgap}\makebox[\exw]{\small PhasorPT}\hspace{\exgap}\makebox[\exw]{\small Ground truth}\\[2pt]
\includegraphics[width=\exw]{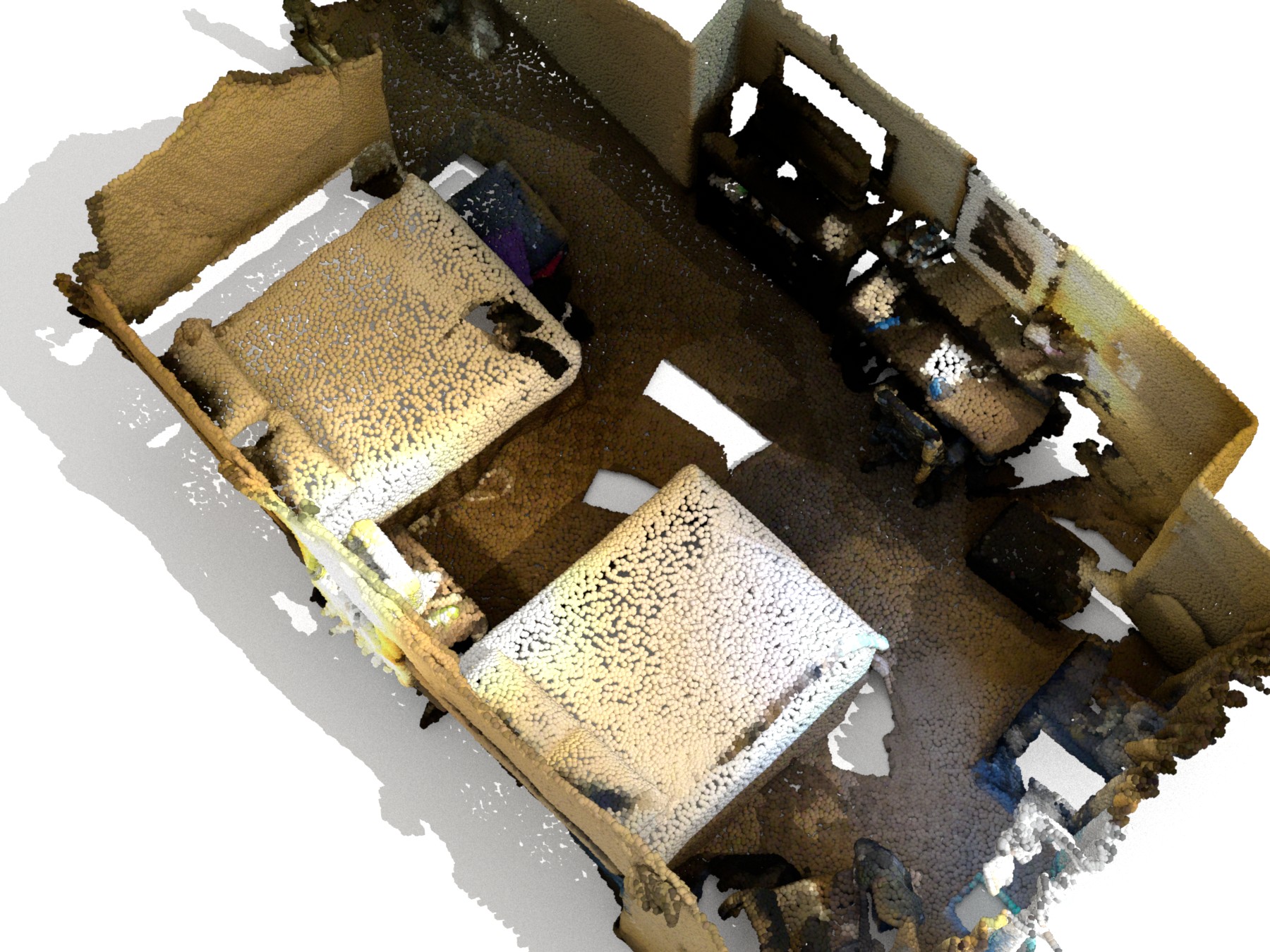}\llap{\makebox[\exw][l]{\hspace{3pt}\raisebox{3pt}{\tiny\color{gray}scene0435\_03}}}\hspace{\exgap}\includegraphics[width=\exw]{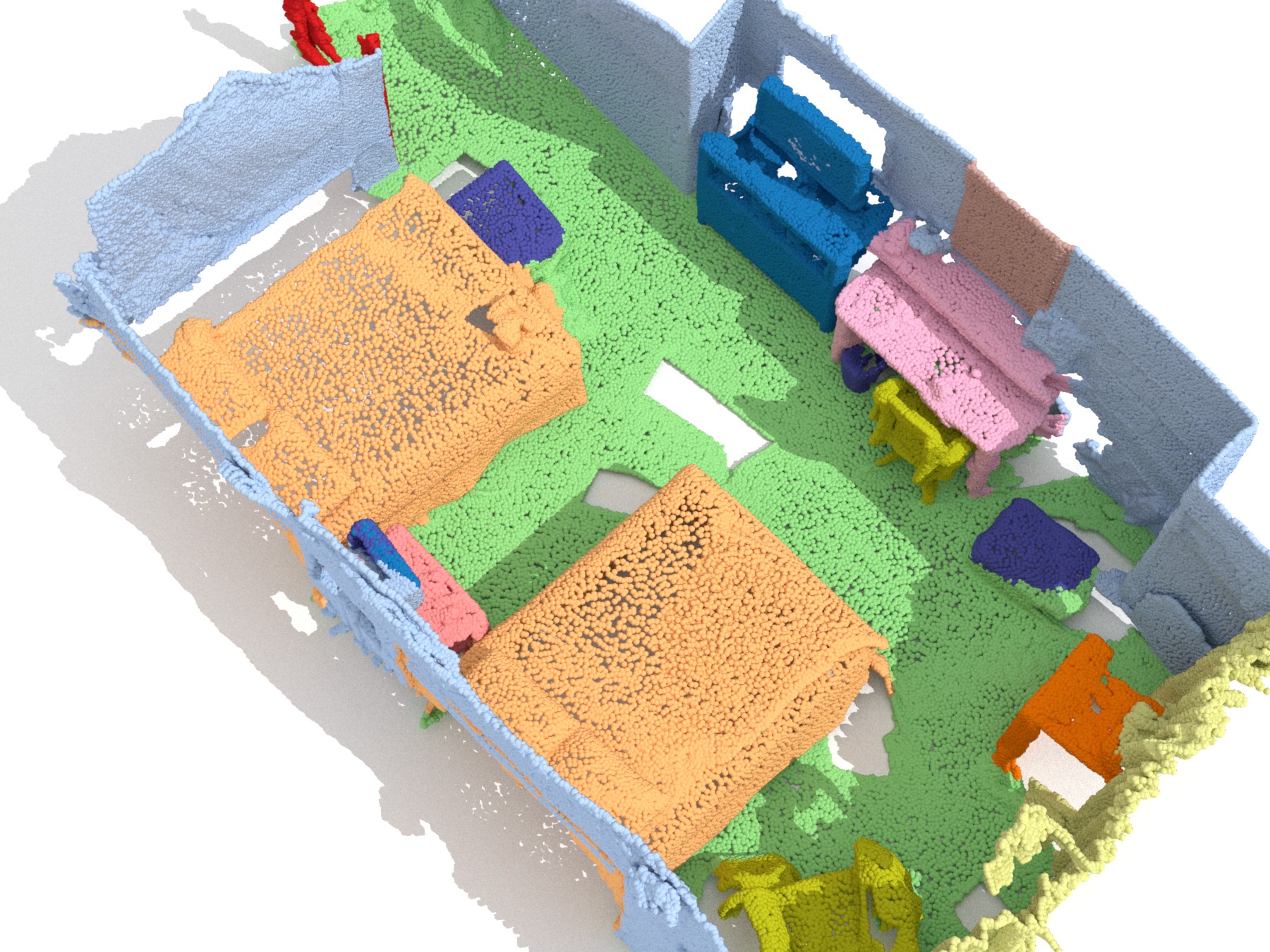}\hspace{\exgap}\includegraphics[width=\exw]{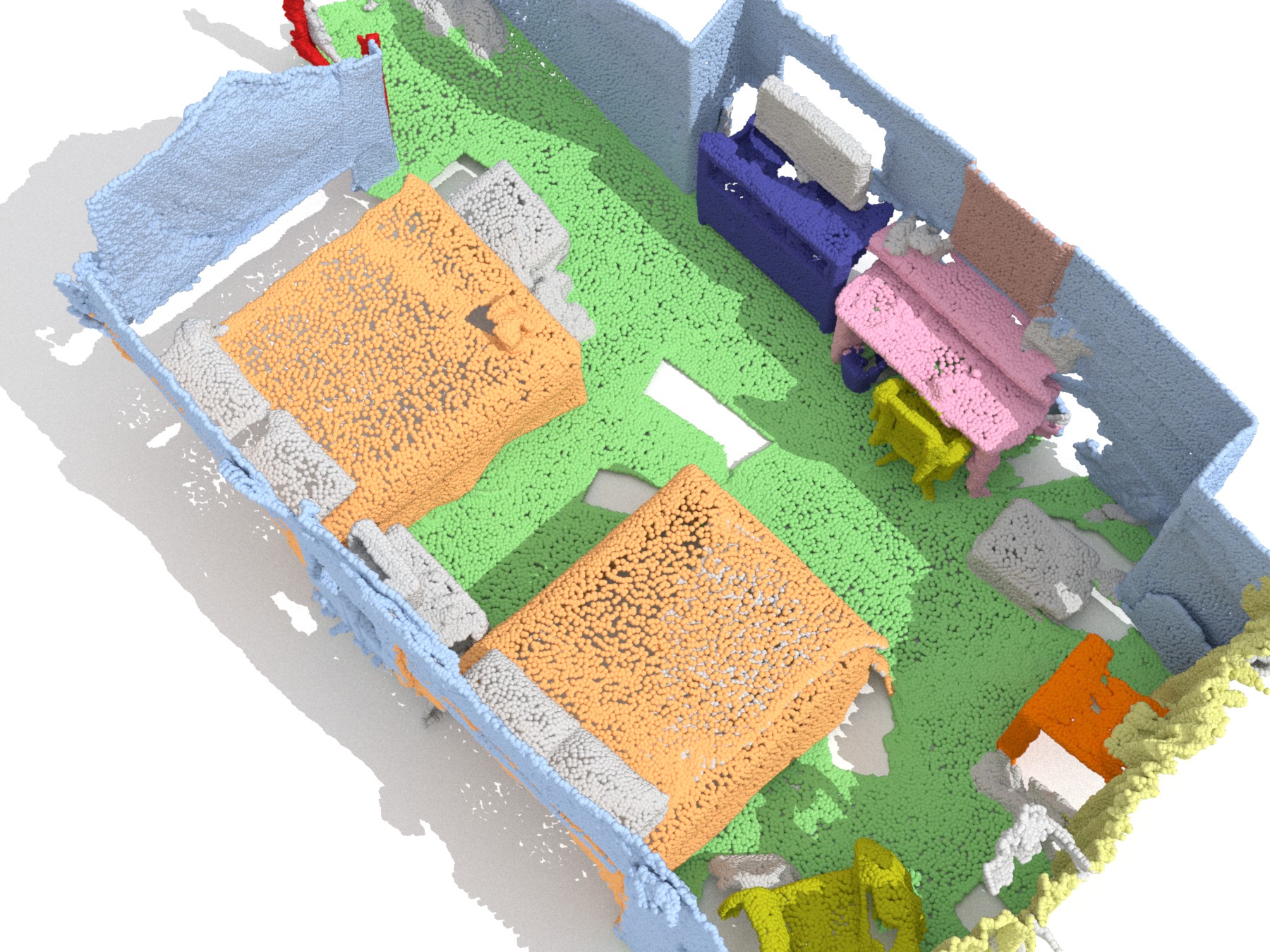}\\[\exgap]
\includegraphics[width=\exw]{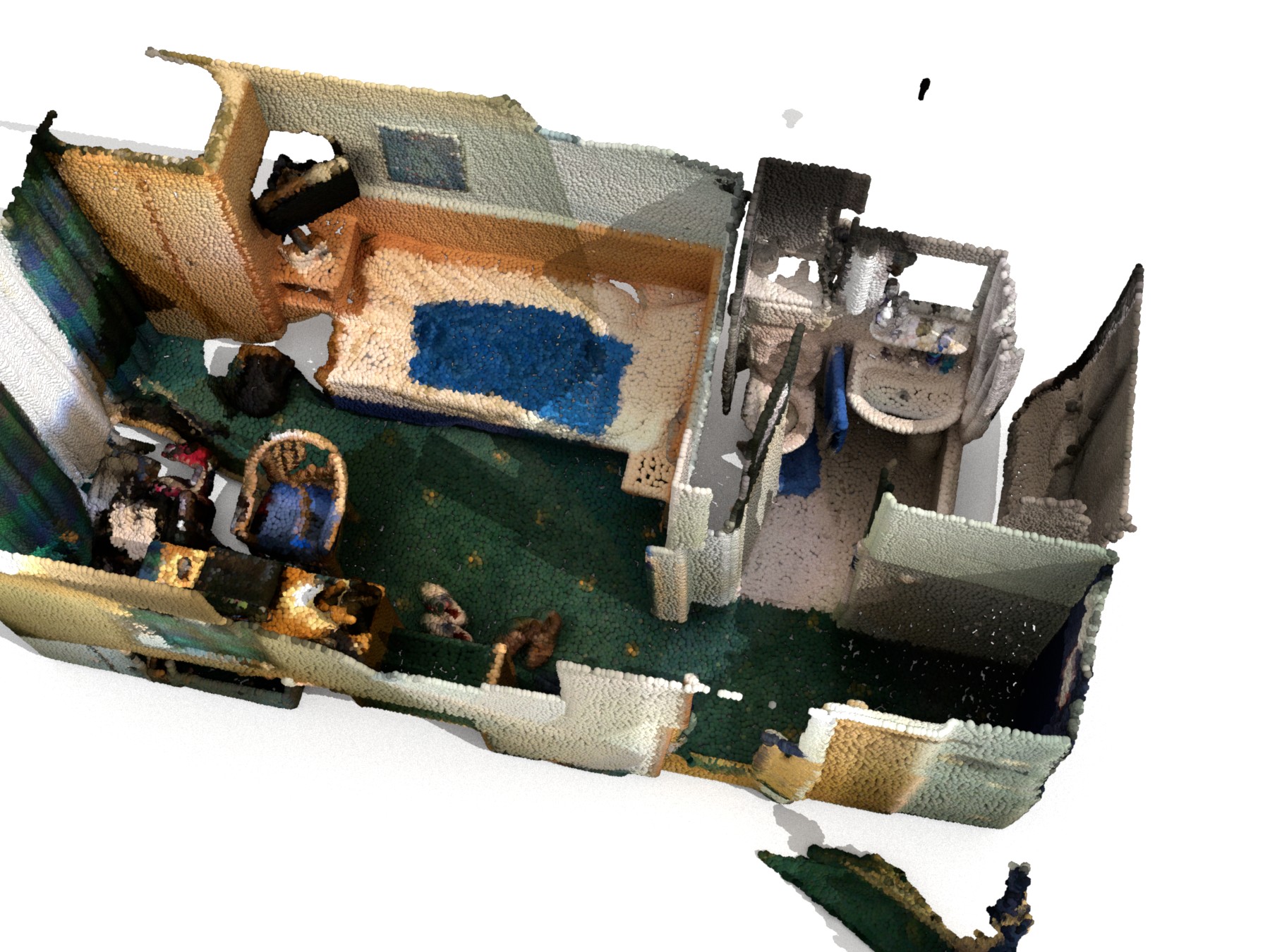}\llap{\makebox[\exw][l]{\hspace{3pt}\raisebox{3pt}{\tiny\color{gray}scene0046\_00}}}\hspace{\exgap}\includegraphics[width=\exw]{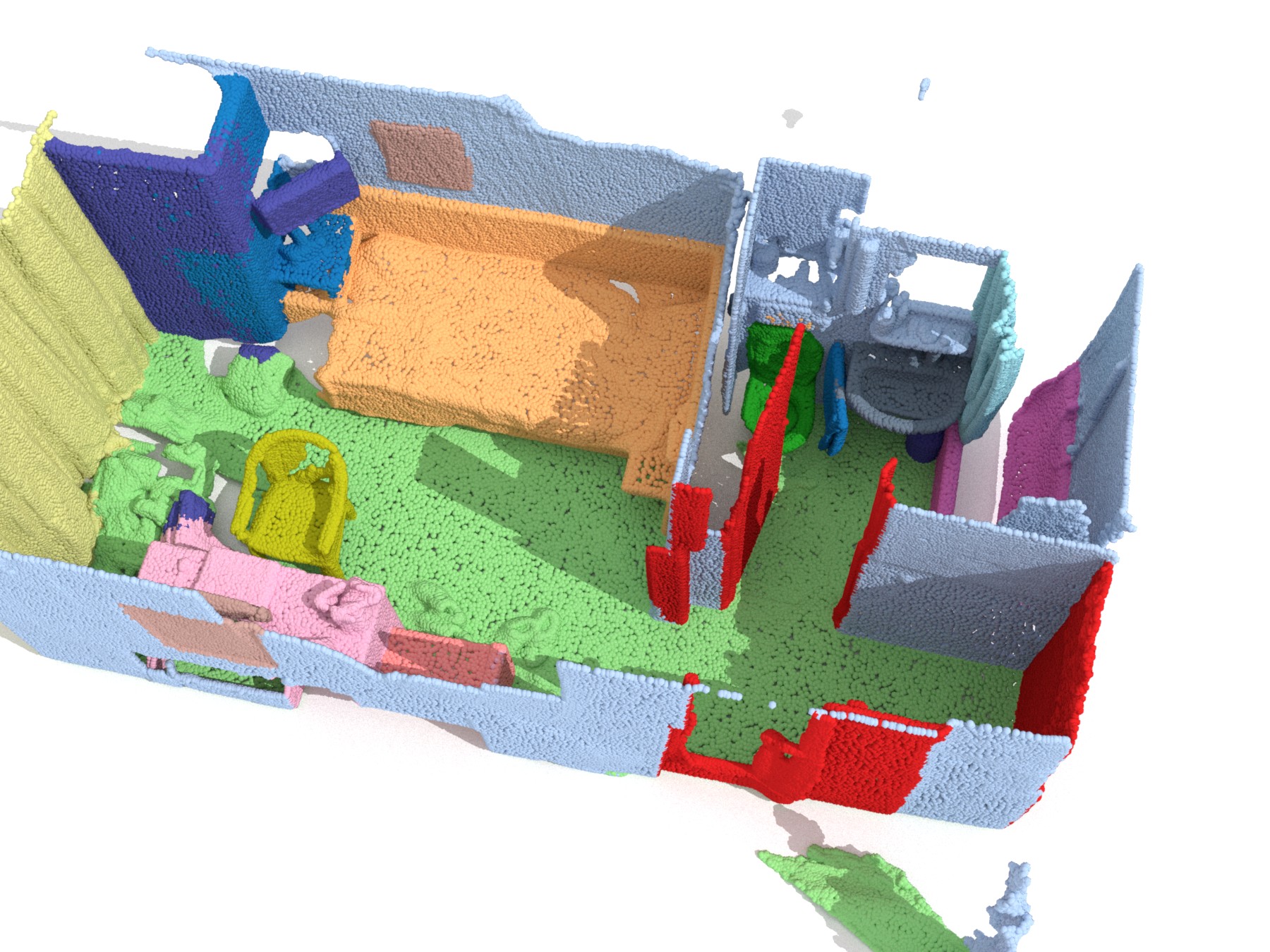}\hspace{\exgap}\includegraphics[width=\exw]{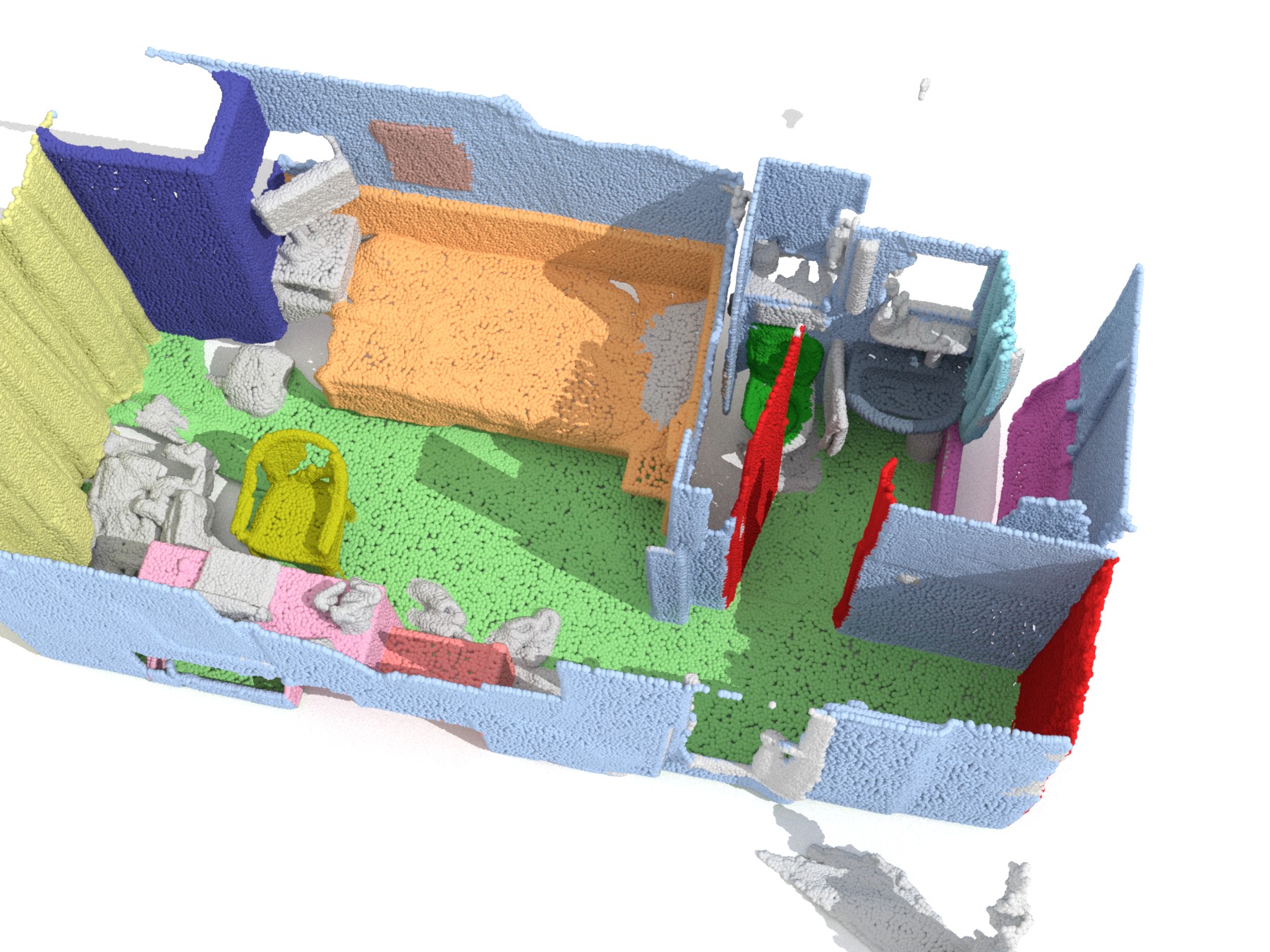}\\[\exgap]
\includegraphics[width=\exw]{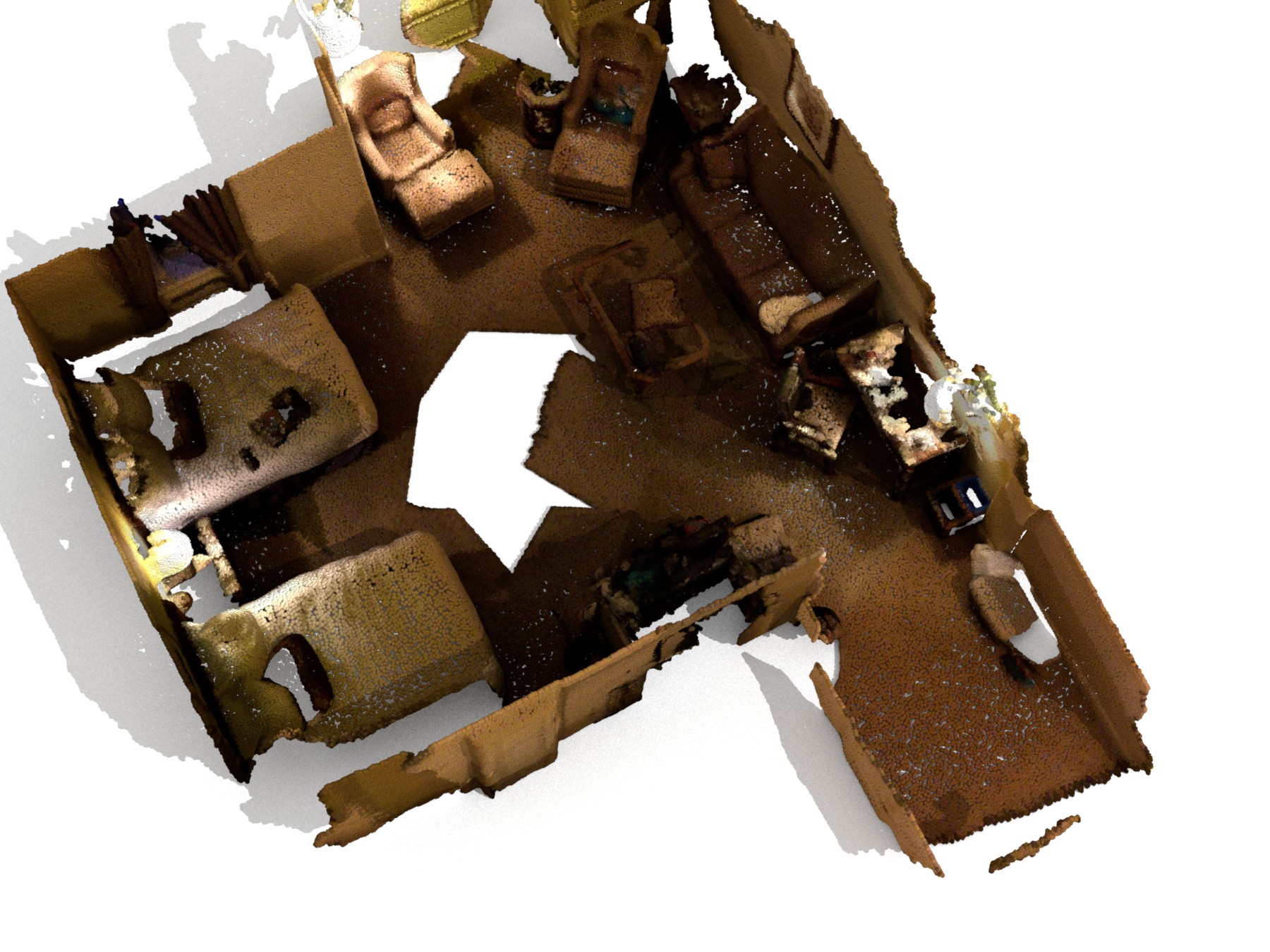}\llap{\makebox[\exw][l]{\hspace{3pt}\raisebox{3pt}{\tiny\color{gray}scene0645\_02}}}\hspace{\exgap}\includegraphics[width=\exw]{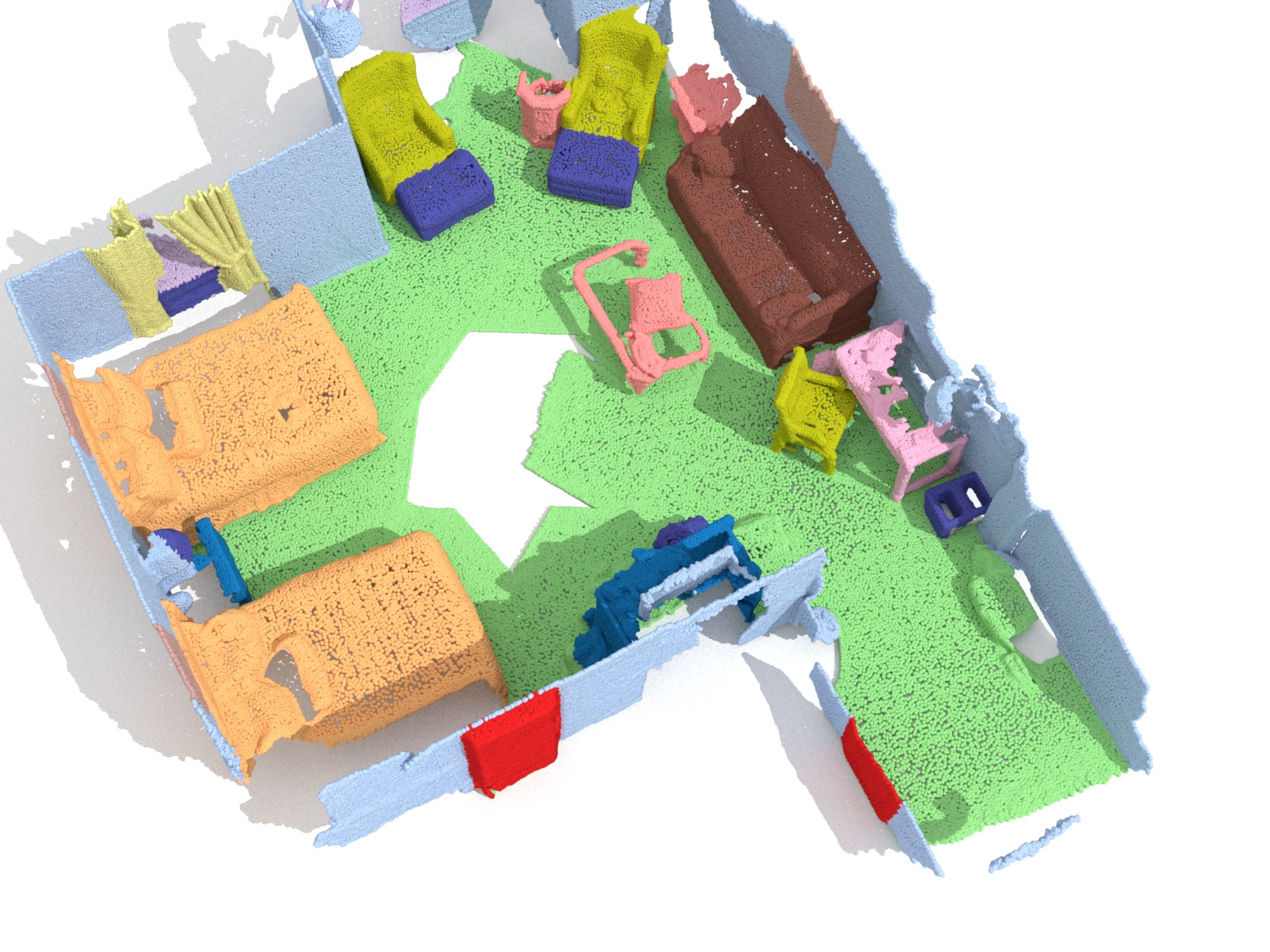}\hspace{\exgap}\includegraphics[width=\exw]{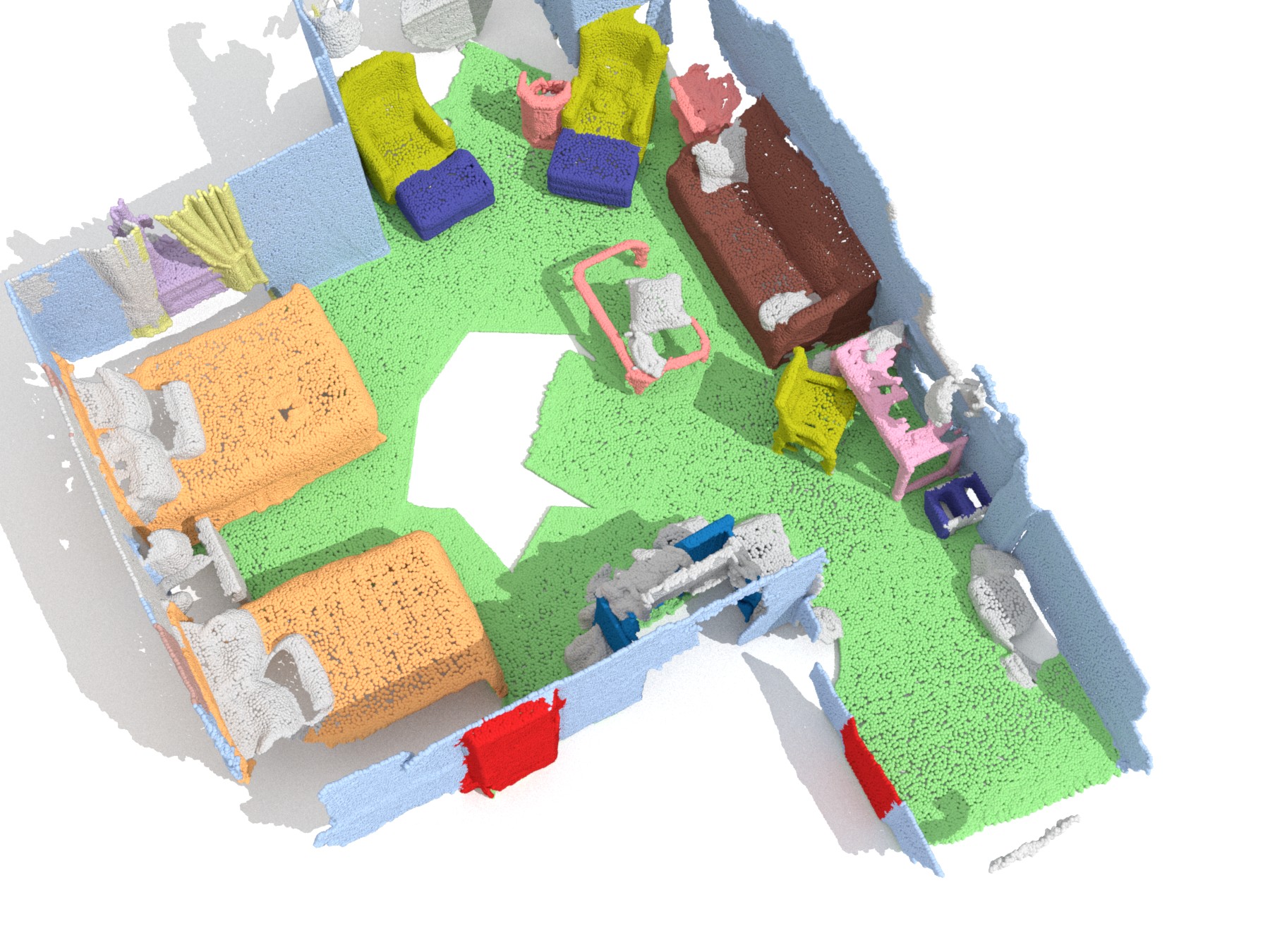}\\[\exgap]
\includegraphics[width=\exw]{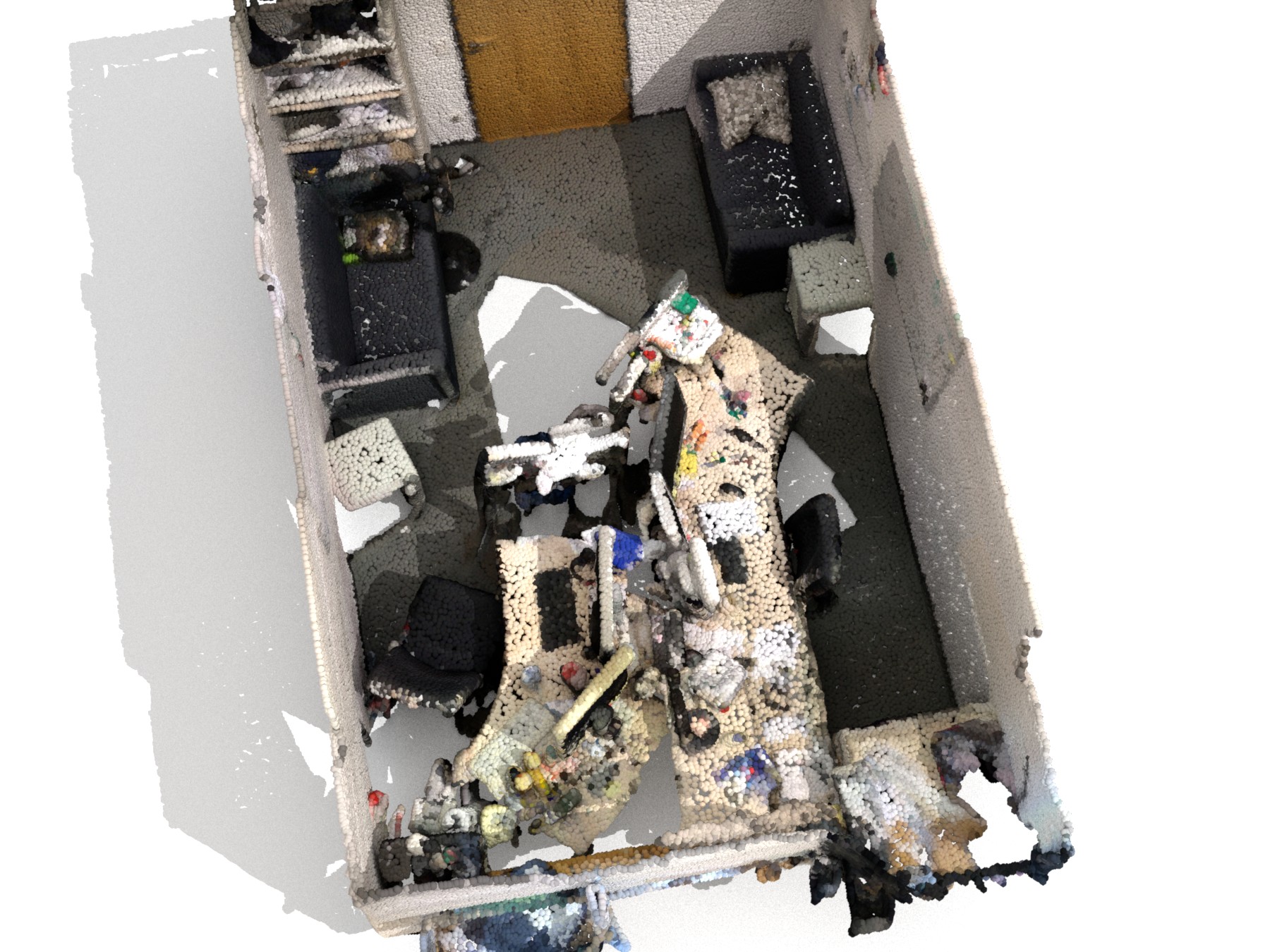}\llap{\makebox[\exw][l]{\hspace{3pt}\raisebox{3pt}{\tiny\color{gray}scene0025\_00}}}\hspace{\exgap}\includegraphics[width=\exw]{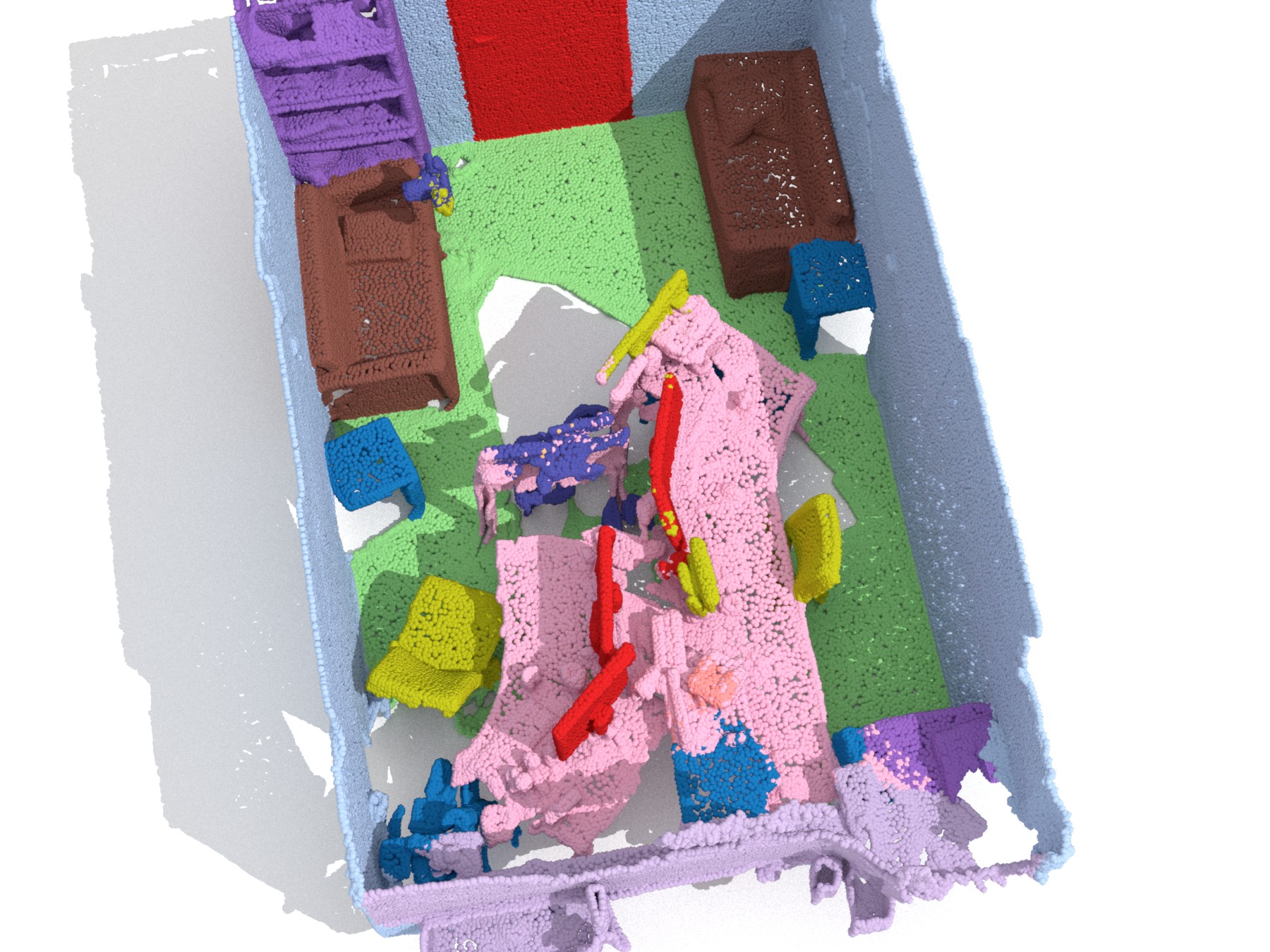}\hspace{\exgap}\includegraphics[width=\exw]{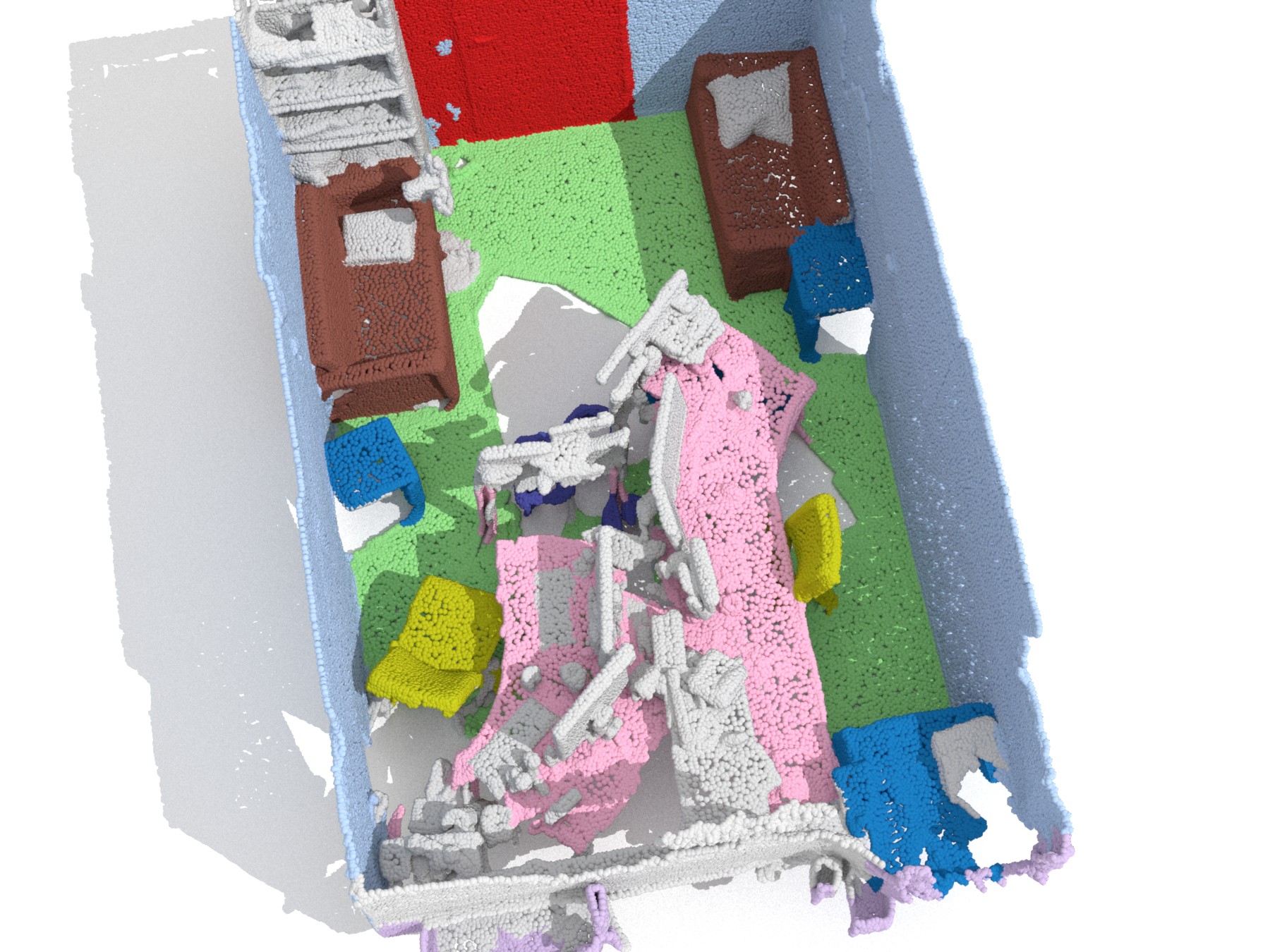}\\[4pt]
\includegraphics[width=\linewidth]{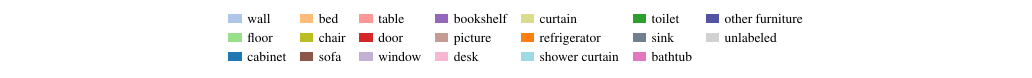}

%% file: figures/examples_structured3d.tex
\ifdefined\exw\else\newlength\exw\newlength\exgap\fi
\setlength\exgap{4pt}\setlength\exw{\dimexpr(\linewidth-2\exgap)/3-0.5pt\relax}
\makebox[\exw]{\small Input (RGB)}\hspace{\exgap}\makebox[\exw]{\small PhasorPT}\hspace{\exgap}\makebox[\exw]{\small Ground truth}\\[2pt]
\includegraphics[width=\exw]{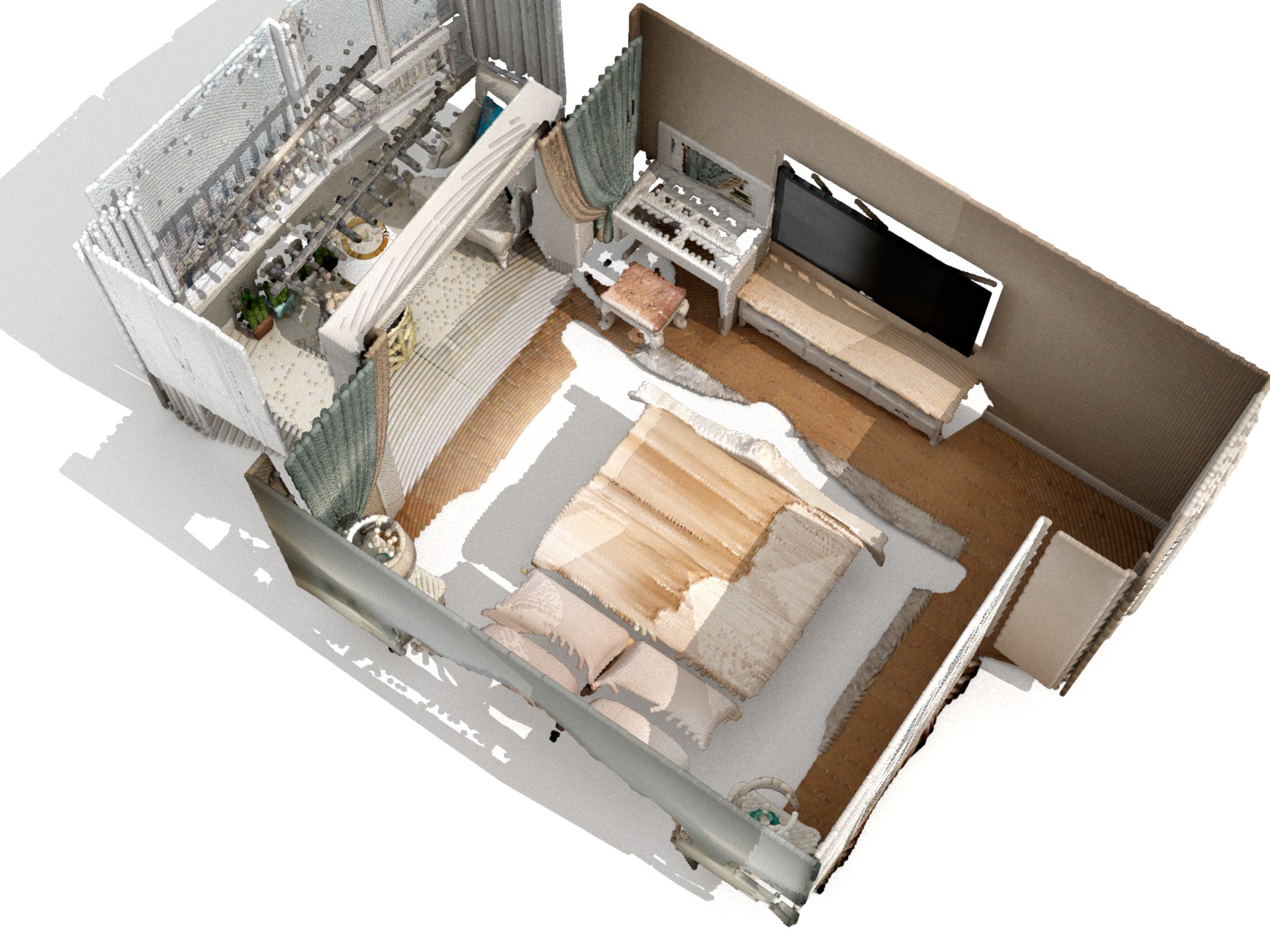}\llap{\makebox[\exw][l]{\hspace{3pt}\raisebox{3pt}{\tiny\color{gray}scene\_03190\_room\_216}}}\hspace{\exgap}\includegraphics[width=\exw]{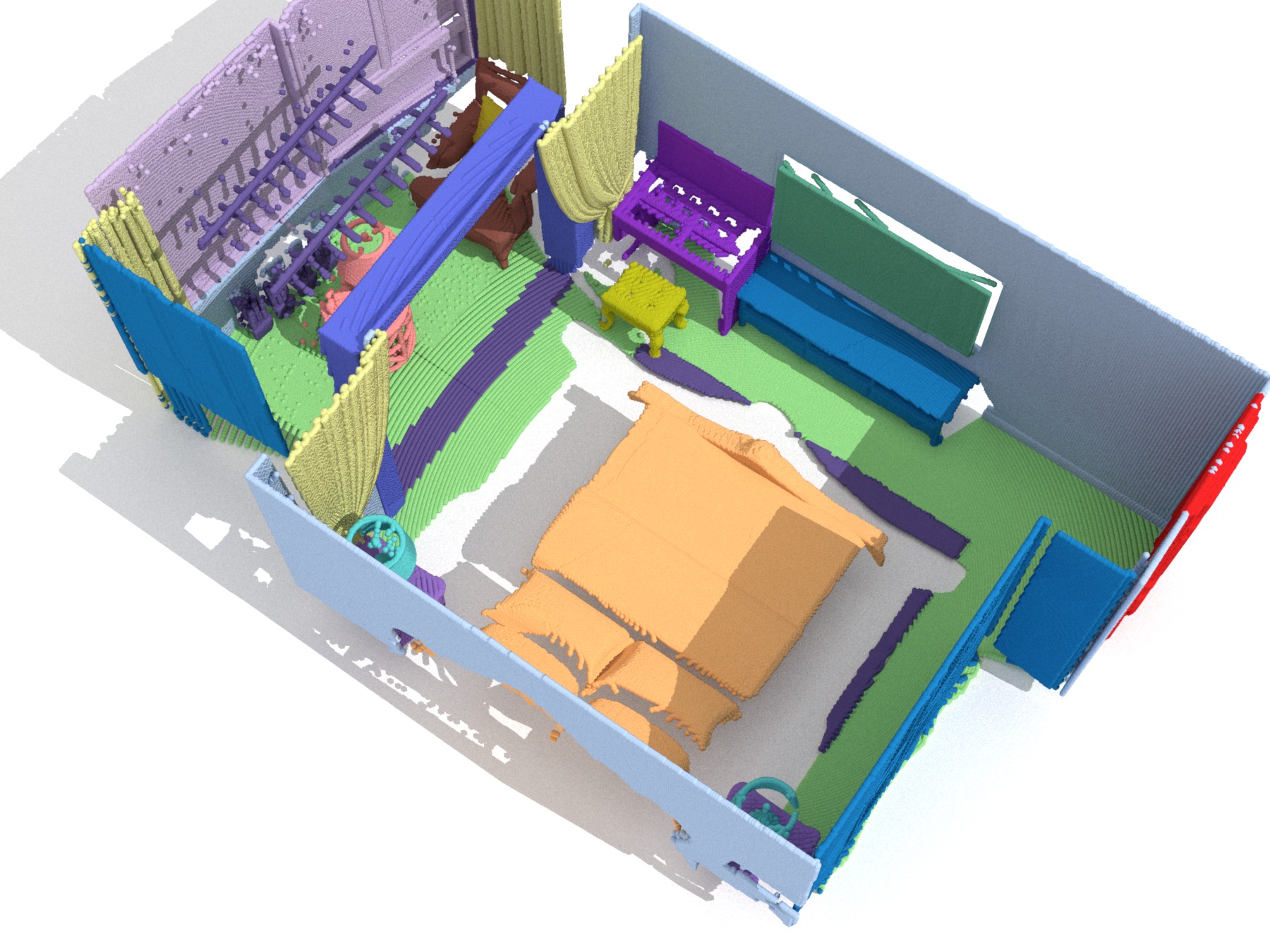}\hspace{\exgap}\includegraphics[width=\exw]{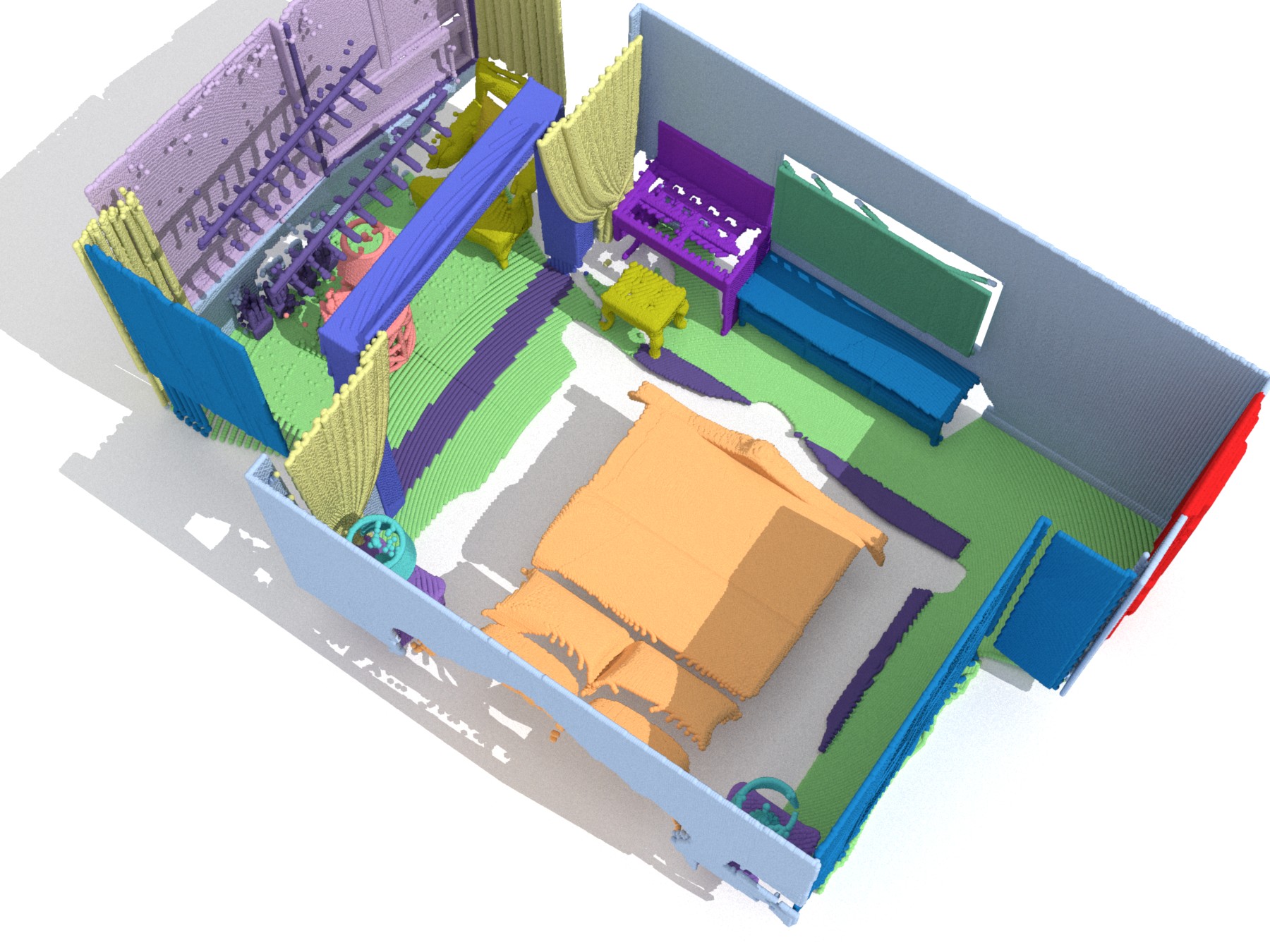}\\[\exgap]
\includegraphics[width=\exw]{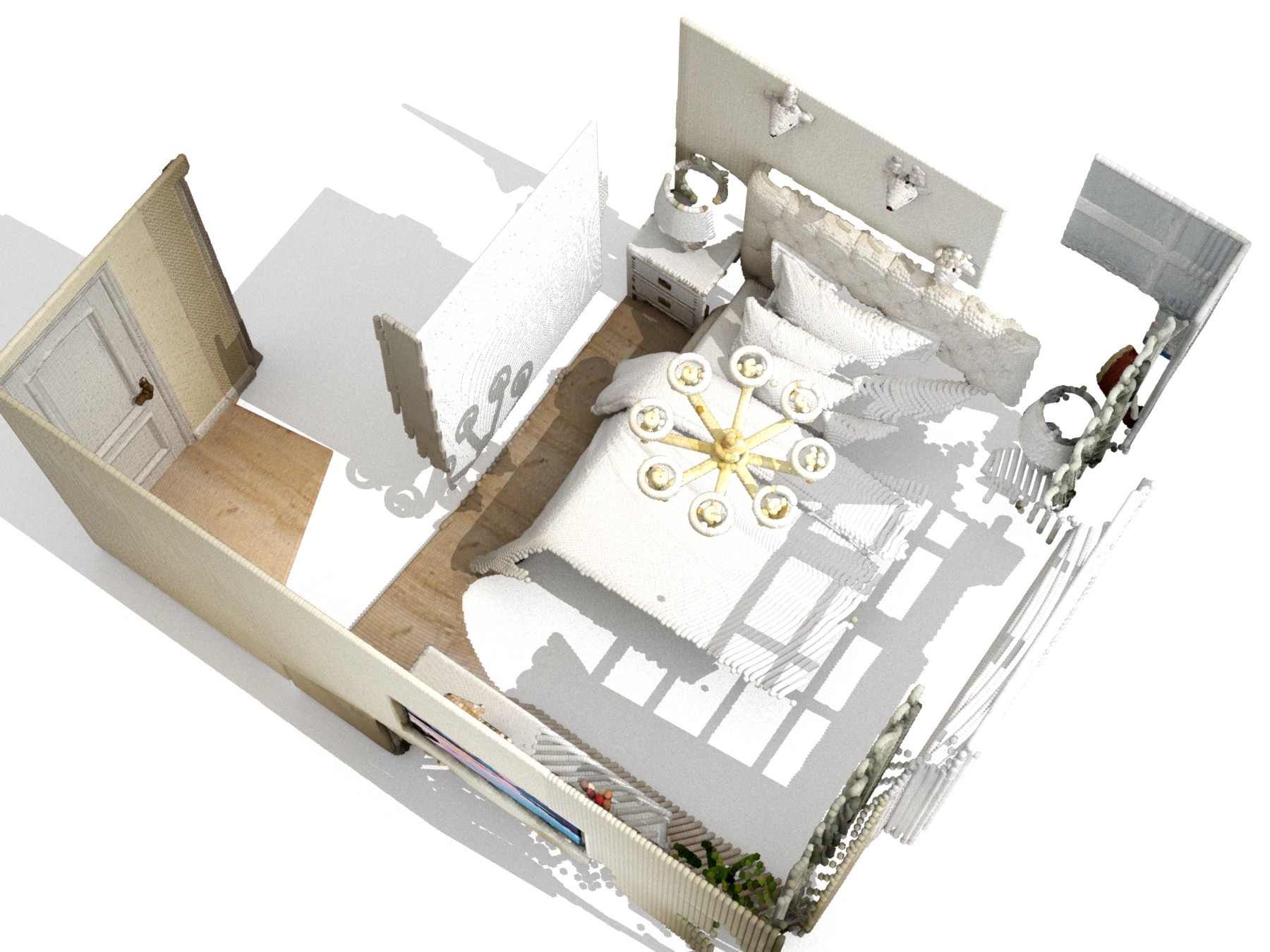}\llap{\makebox[\exw][l]{\hspace{3pt}\raisebox{3pt}{\tiny\color{gray}scene\_03036\_room\_51056}}}\hspace{\exgap}\includegraphics[width=\exw]{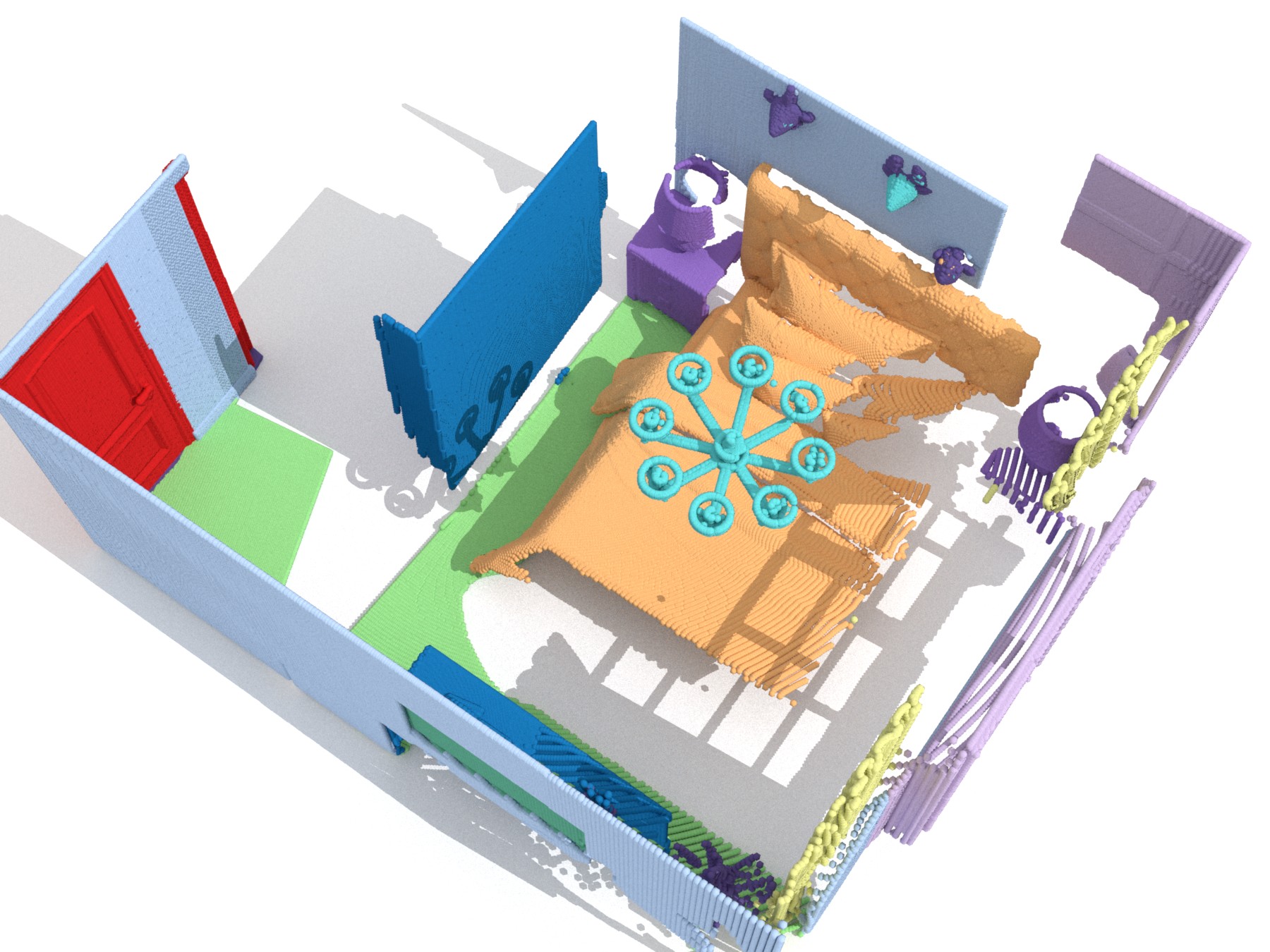}\hspace{\exgap}\includegraphics[width=\exw]{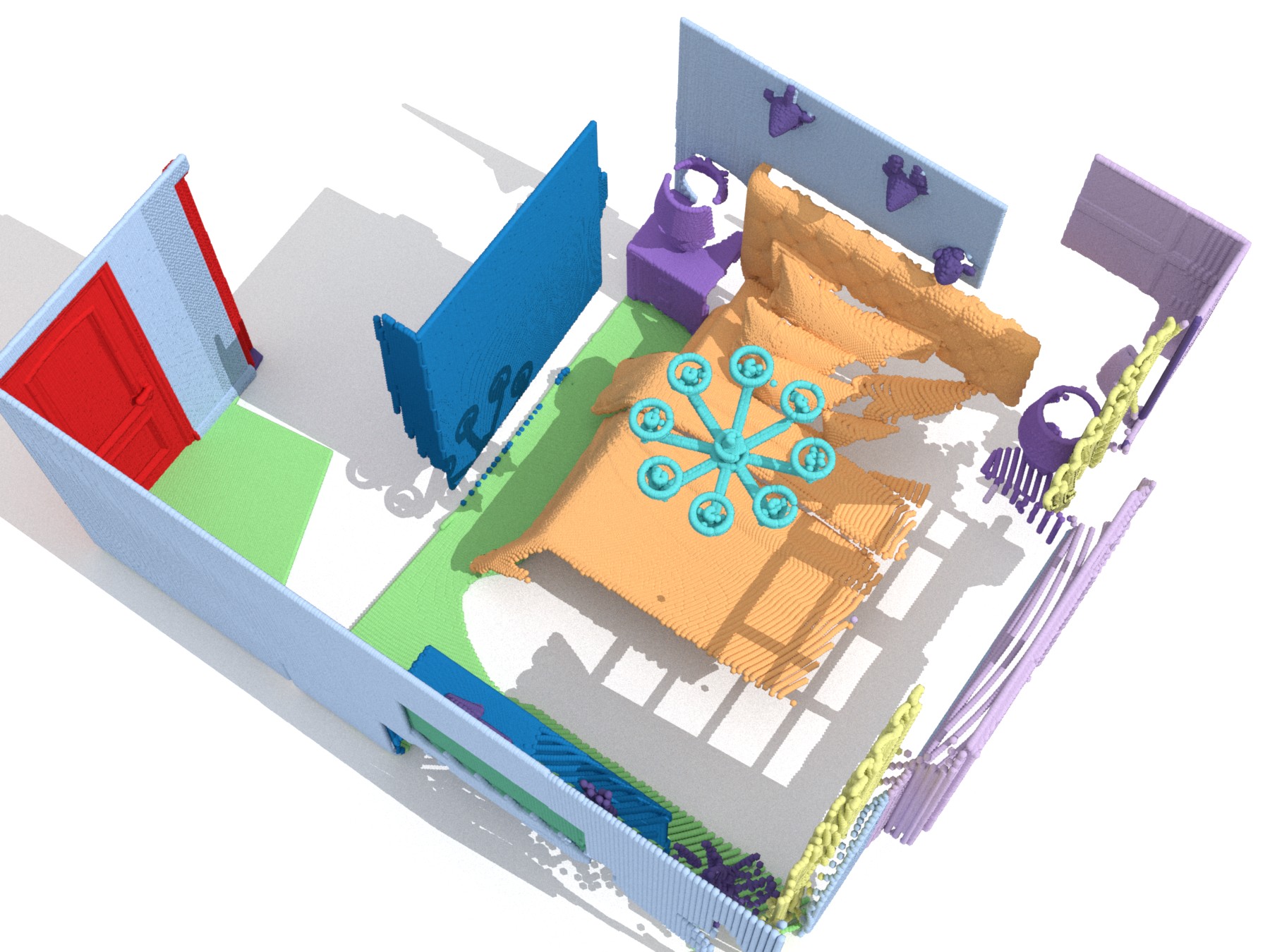}\\[\exgap]
\includegraphics[width=\exw]{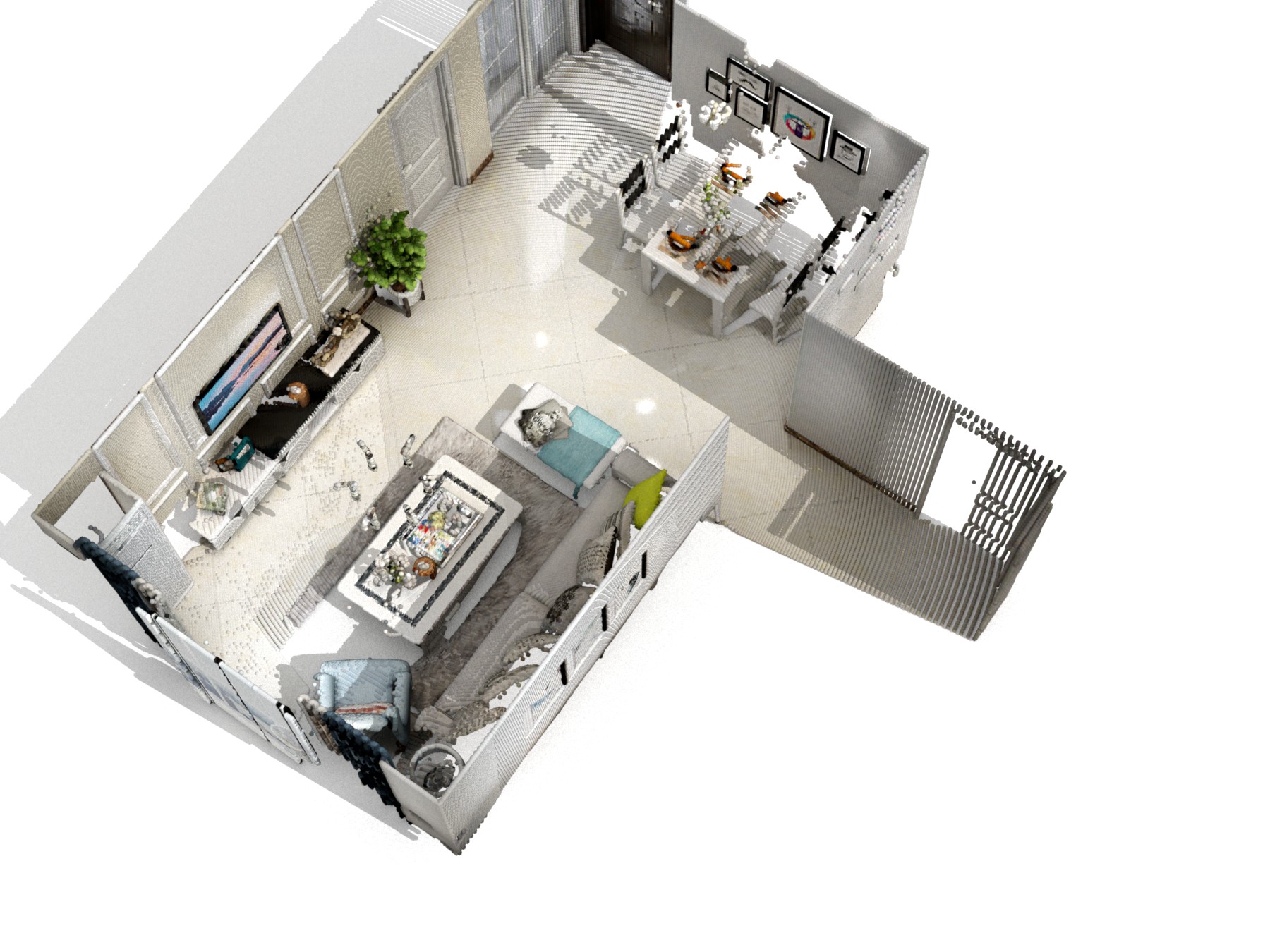}\llap{\makebox[\exw][l]{\hspace{3pt}\raisebox{3pt}{\tiny\color{gray}scene\_03075\_room\_334}}}\hspace{\exgap}\includegraphics[width=\exw]{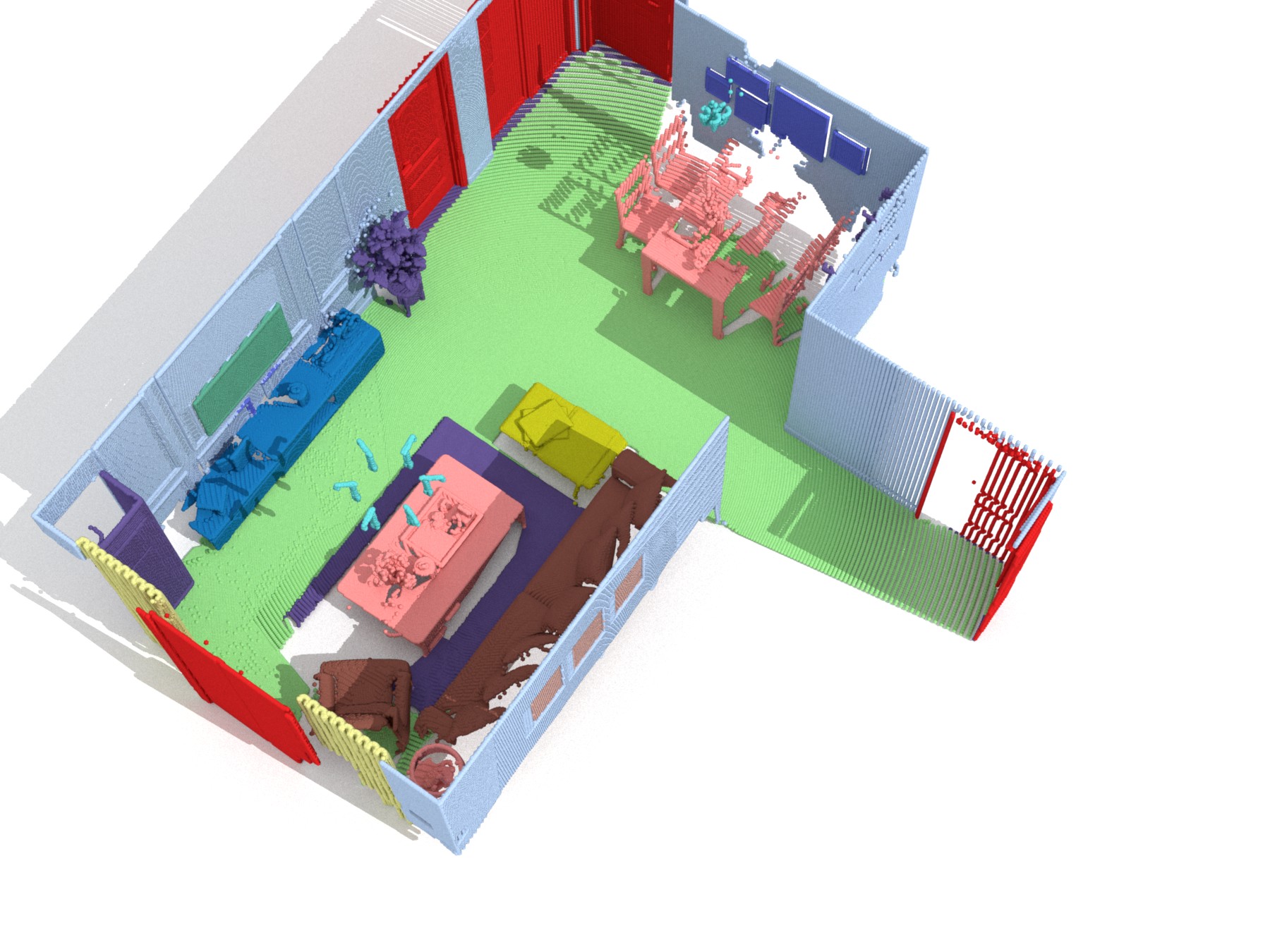}\hspace{\exgap}\includegraphics[width=\exw]{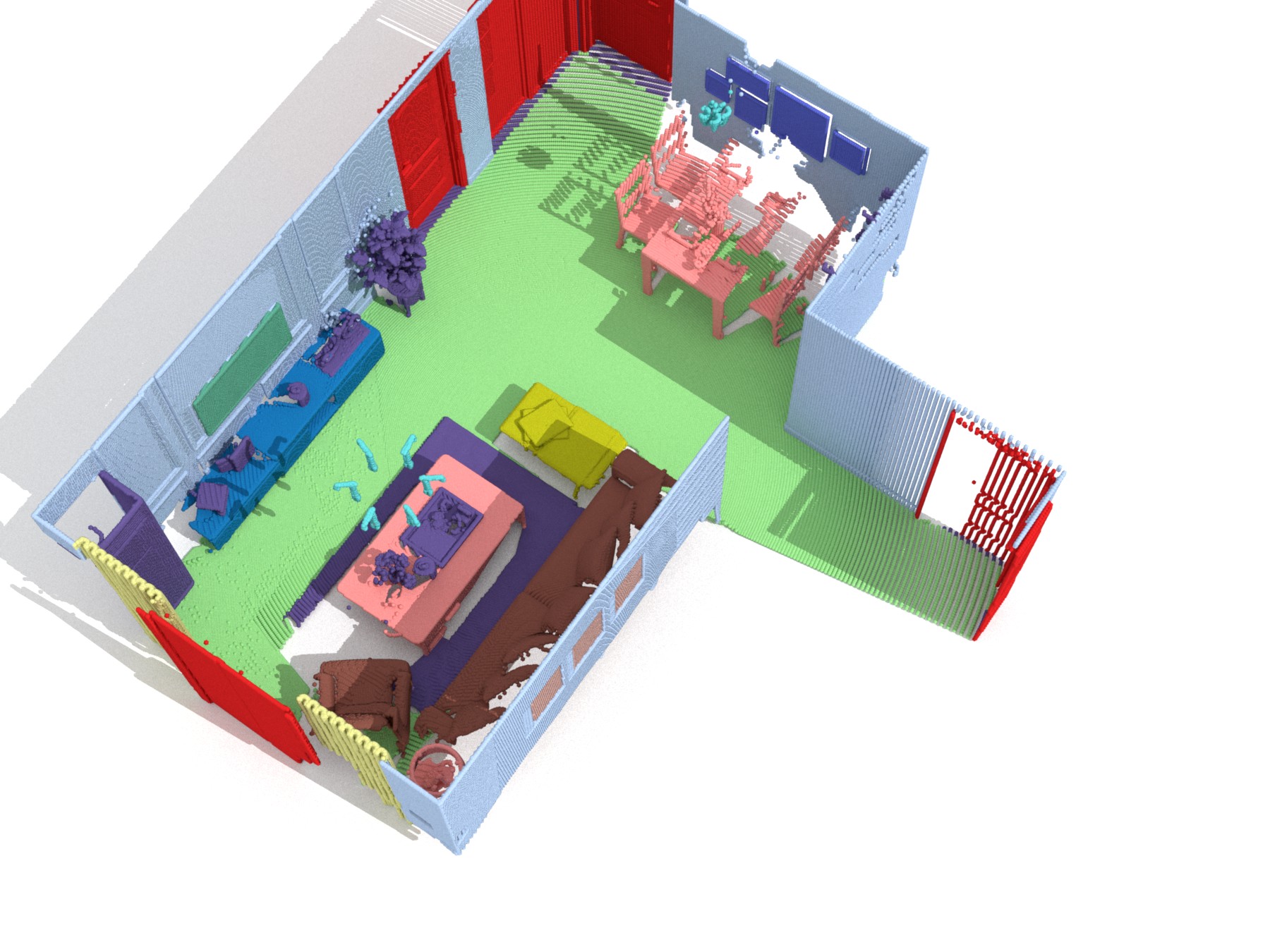}\\[\exgap]
\includegraphics[width=\exw]{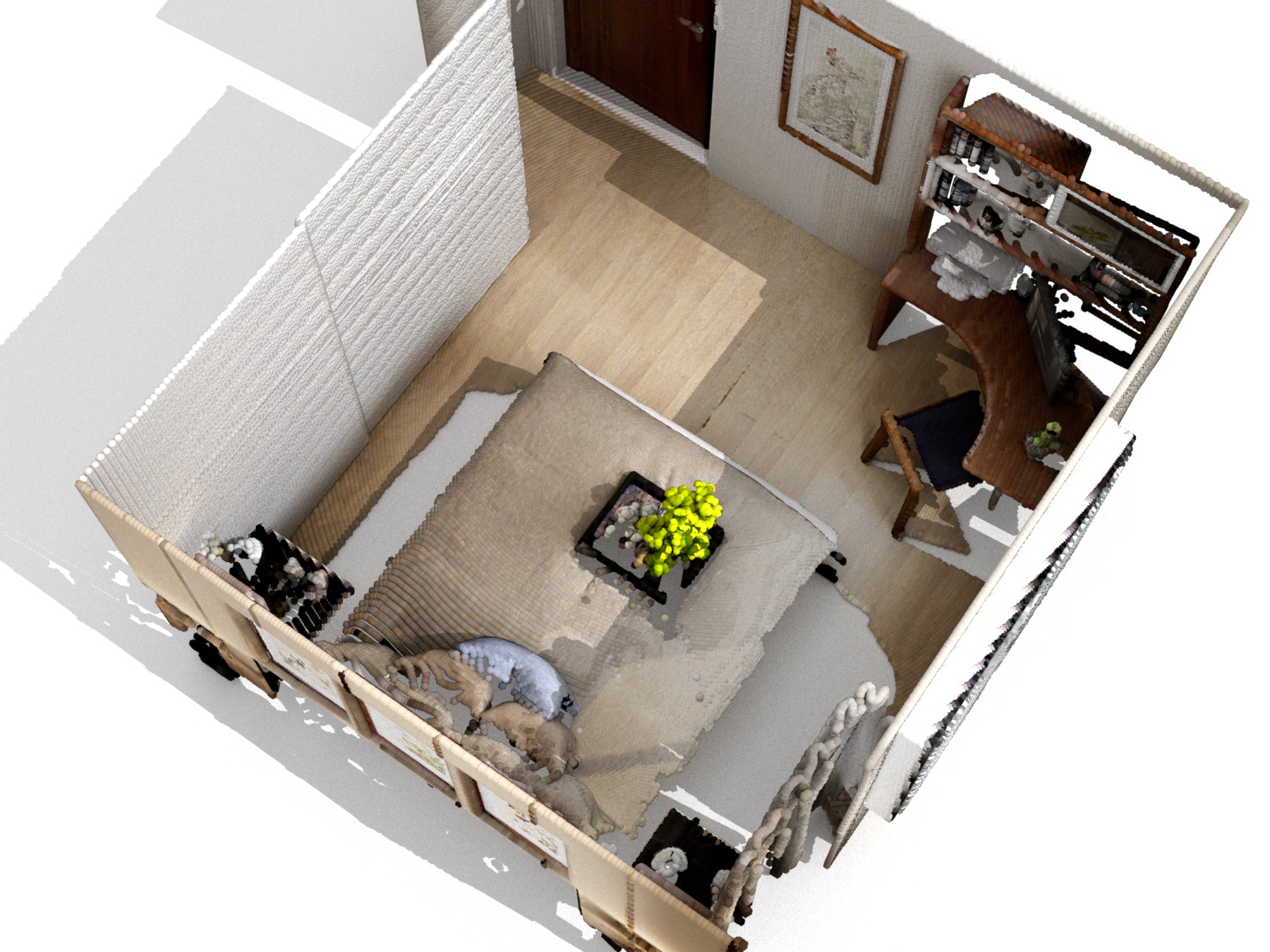}\llap{\makebox[\exw][l]{\hspace{3pt}\raisebox{3pt}{\tiny\color{gray}scene\_03123\_room\_1683}}}\hspace{\exgap}\includegraphics[width=\exw]{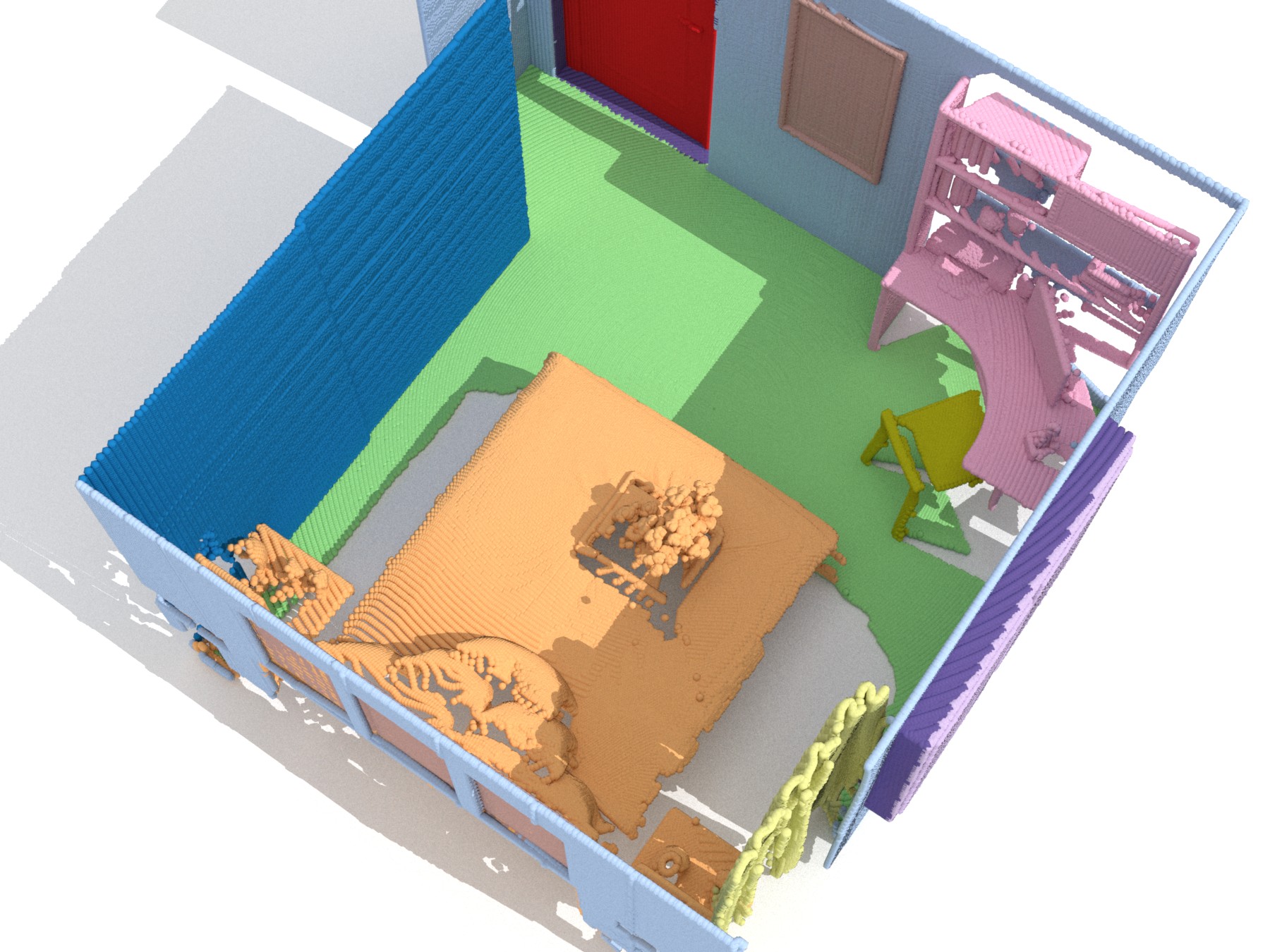}\hspace{\exgap}\includegraphics[width=\exw]{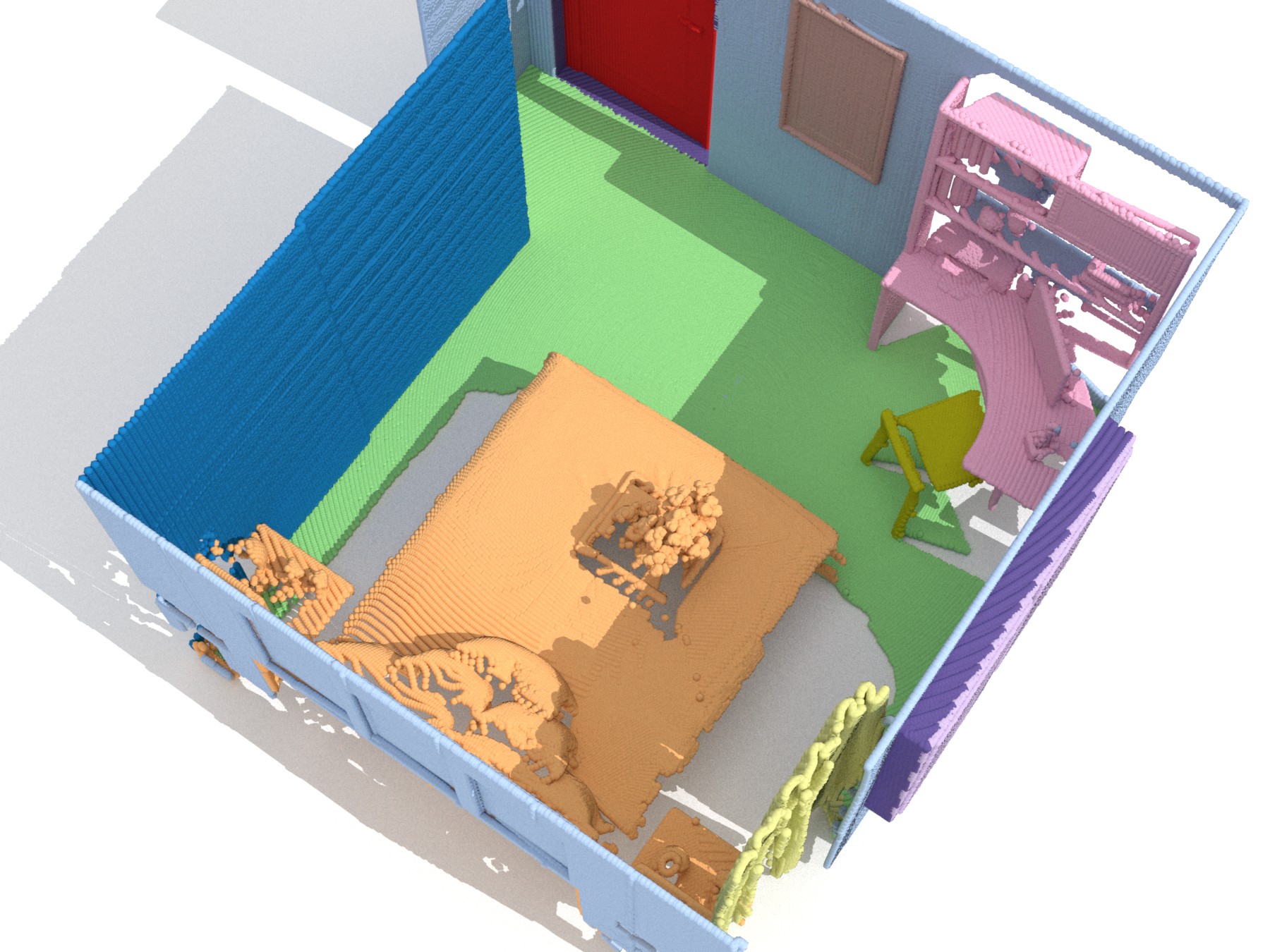}\\[4pt]
\includegraphics[width=\linewidth]{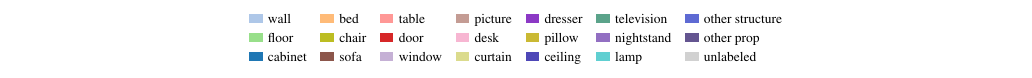}

%% file: figures/examples_kitti.tex
\ifdefined\exw\else\newlength\exw\newlength\exgap\fi
\setlength\exgap{4pt}\setlength\exw{\dimexpr(\linewidth-2\exgap)/3-0.5pt\relax}
\makebox[\exw]{\small Input (remission)}\hspace{\exgap}\makebox[\exw]{\small PhasorPT}\hspace{\exgap}\makebox[\exw]{\small Ground truth}\\[2pt]
\includegraphics[width=\exw]{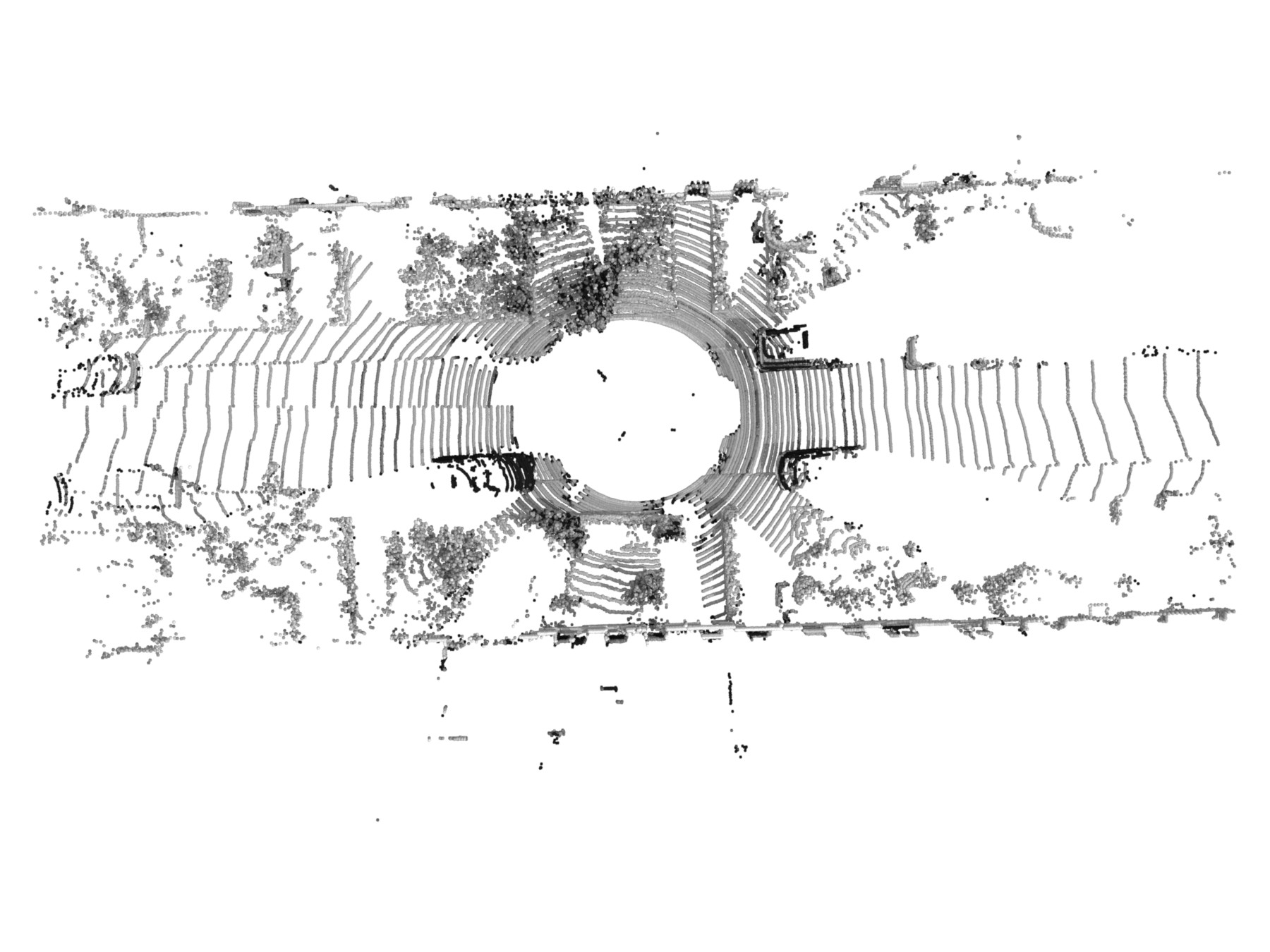}\llap{\makebox[\exw][l]{\hspace{3pt}\raisebox{3pt}{\tiny\color{gray}08\_000620}}}\hspace{\exgap}\includegraphics[width=\exw]{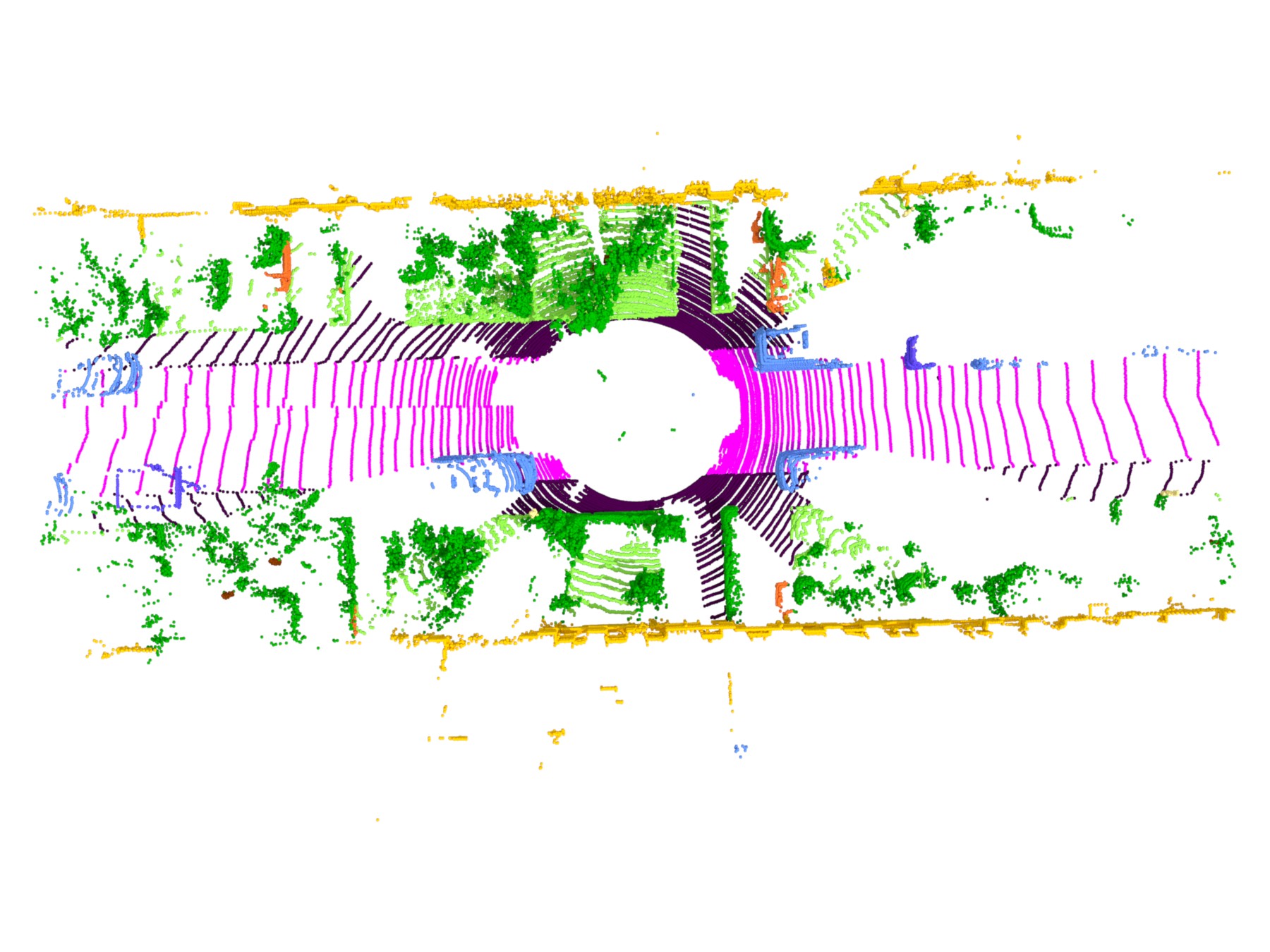}\hspace{\exgap}\includegraphics[width=\exw]{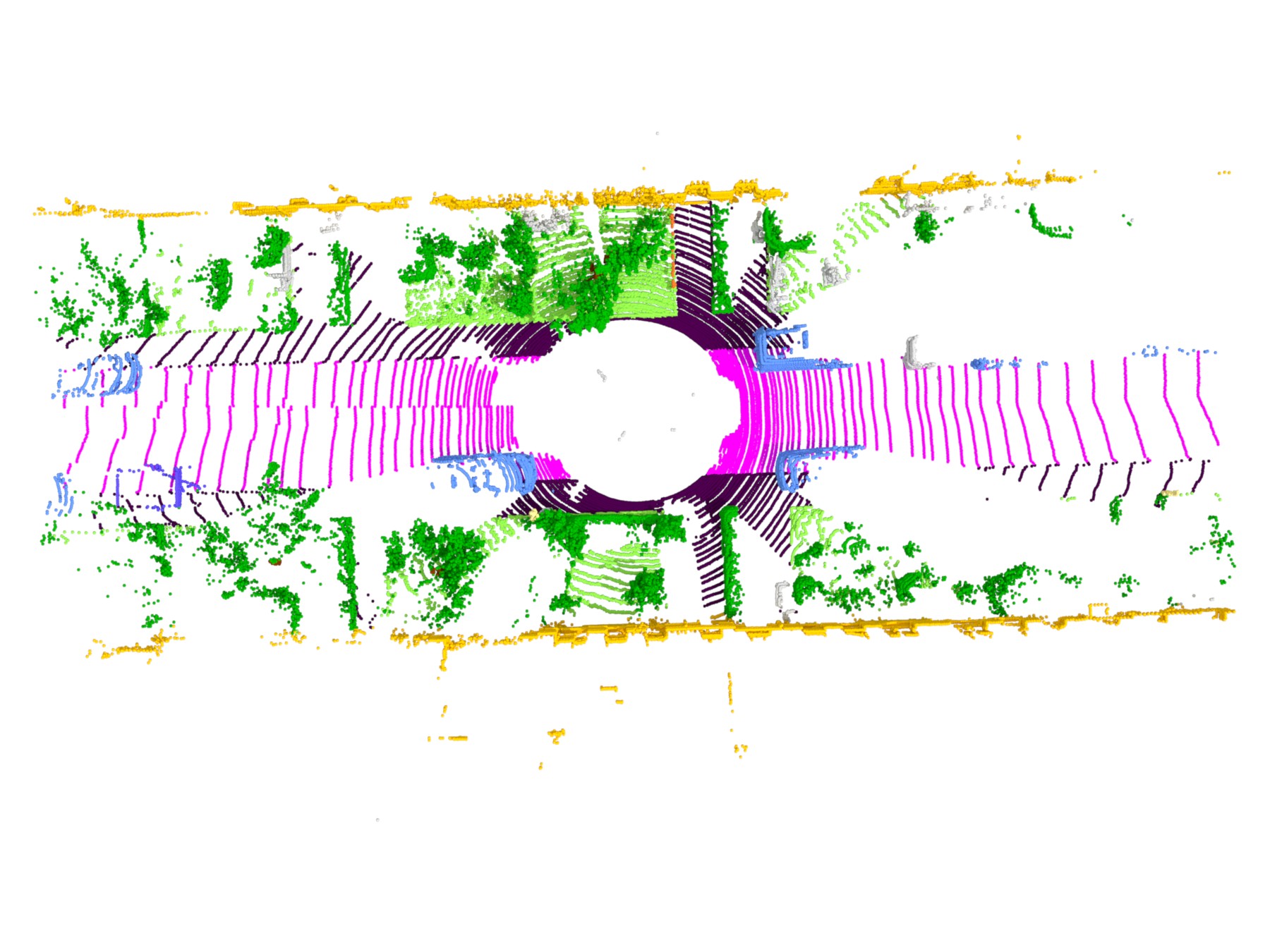}\\[\exgap]
\includegraphics[width=\exw]{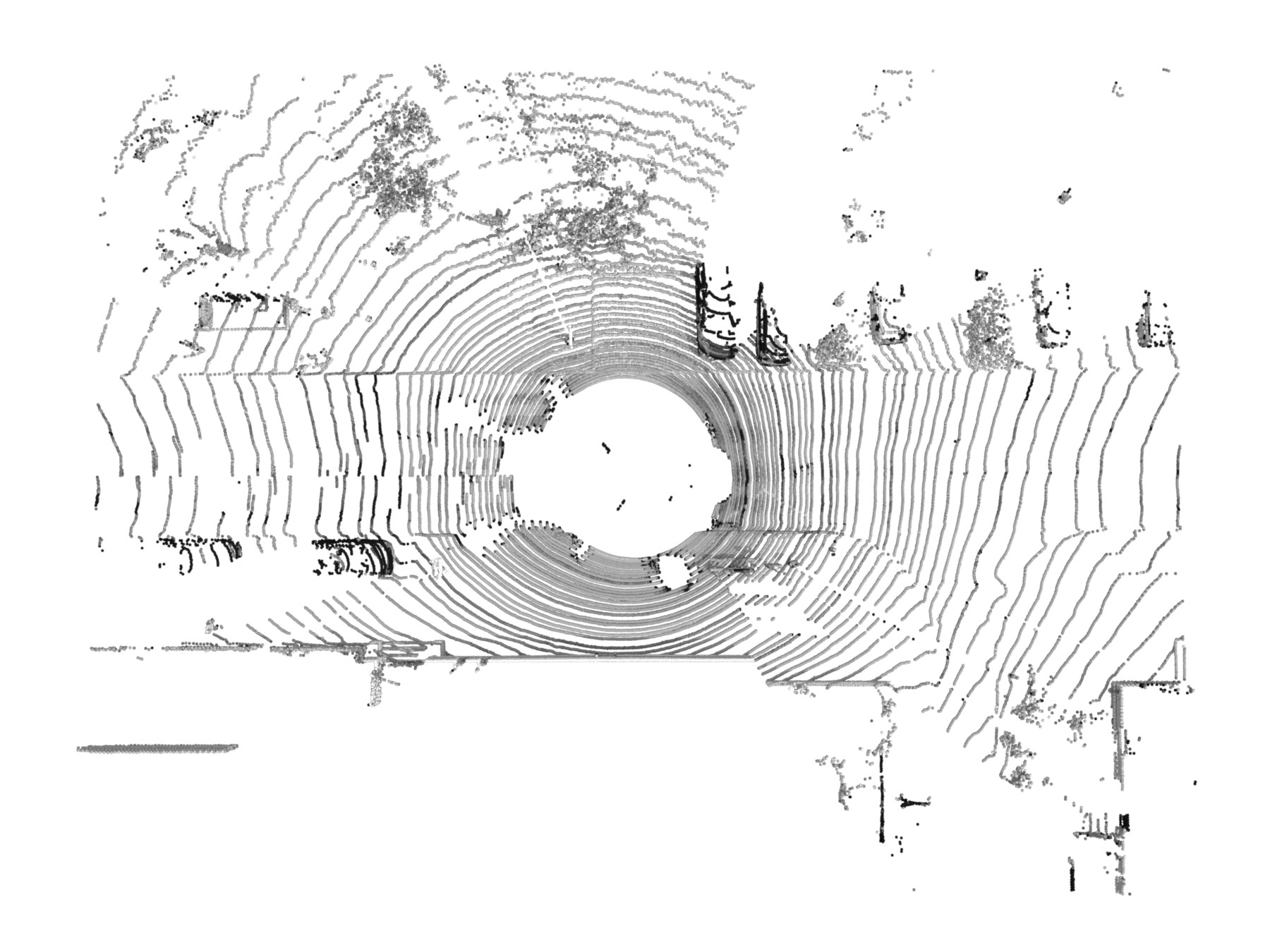}\llap{\makebox[\exw][l]{\hspace{3pt}\raisebox{3pt}{\tiny\color{gray}08\_001900}}}\hspace{\exgap}\includegraphics[width=\exw]{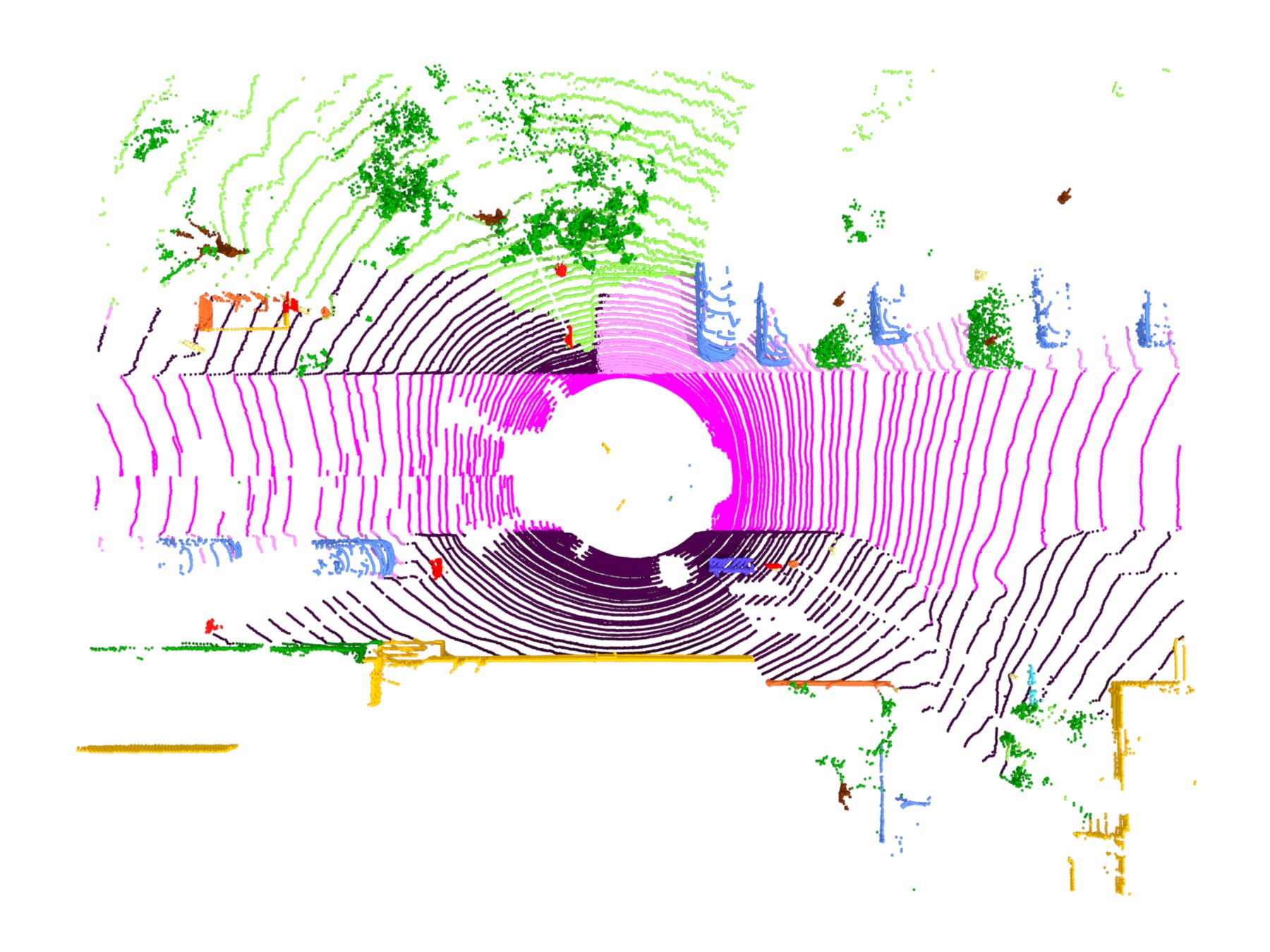}\hspace{\exgap}\includegraphics[width=\exw]{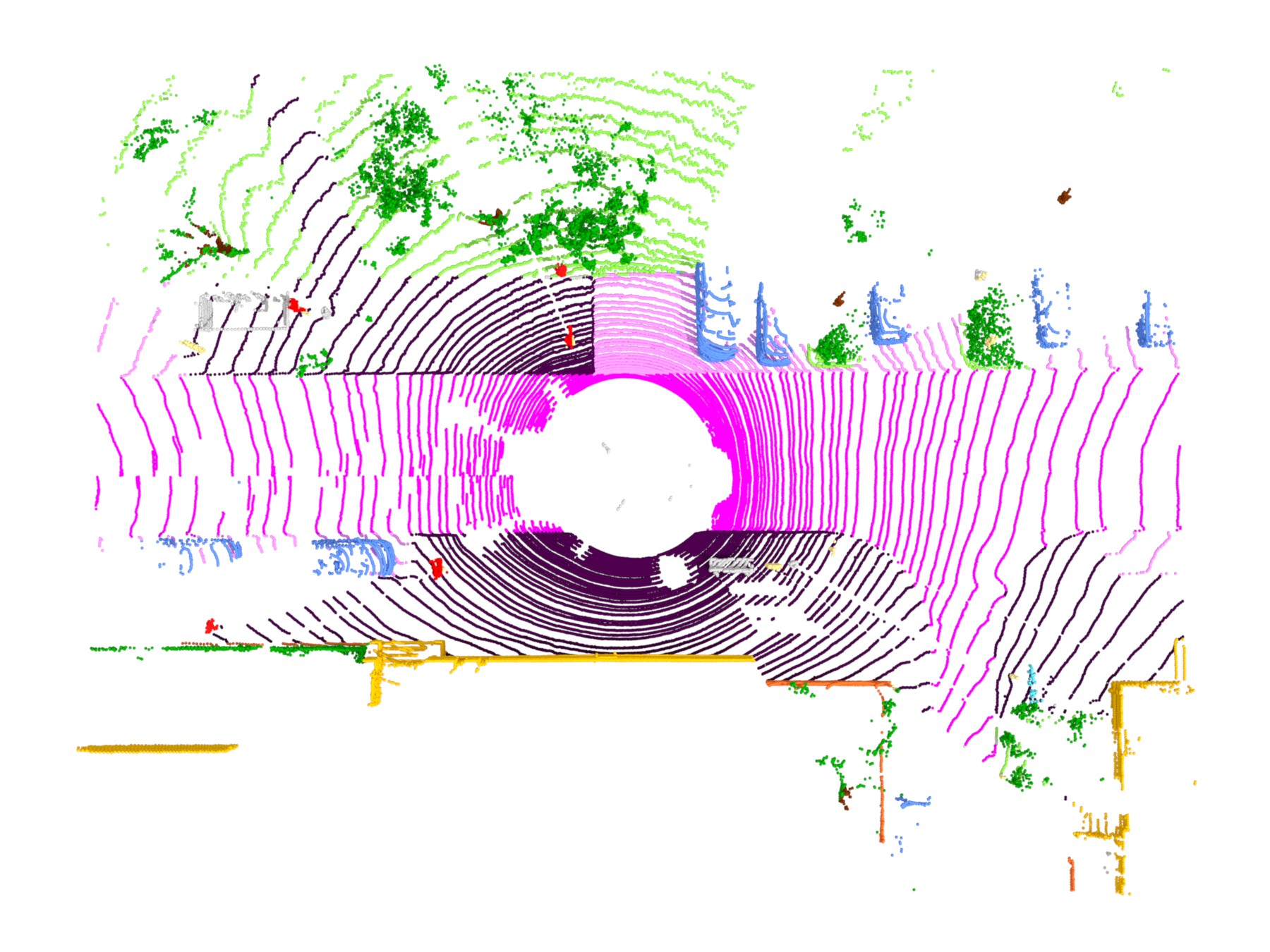}\\[\exgap]
\includegraphics[width=\exw]{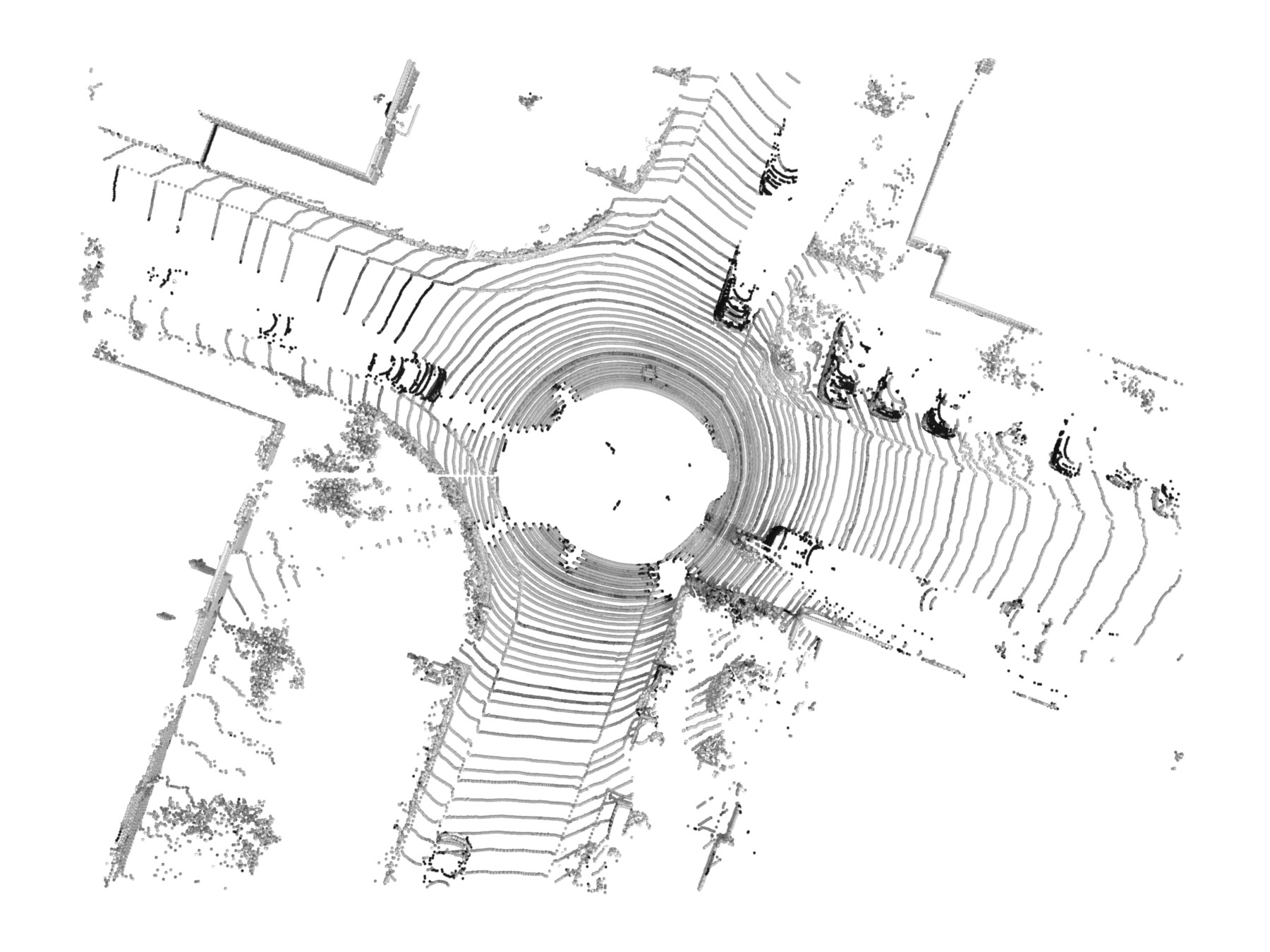}\llap{\makebox[\exw][l]{\hspace{3pt}\raisebox{3pt}{\tiny\color{gray}08\_002520}}}\hspace{\exgap}\includegraphics[width=\exw]{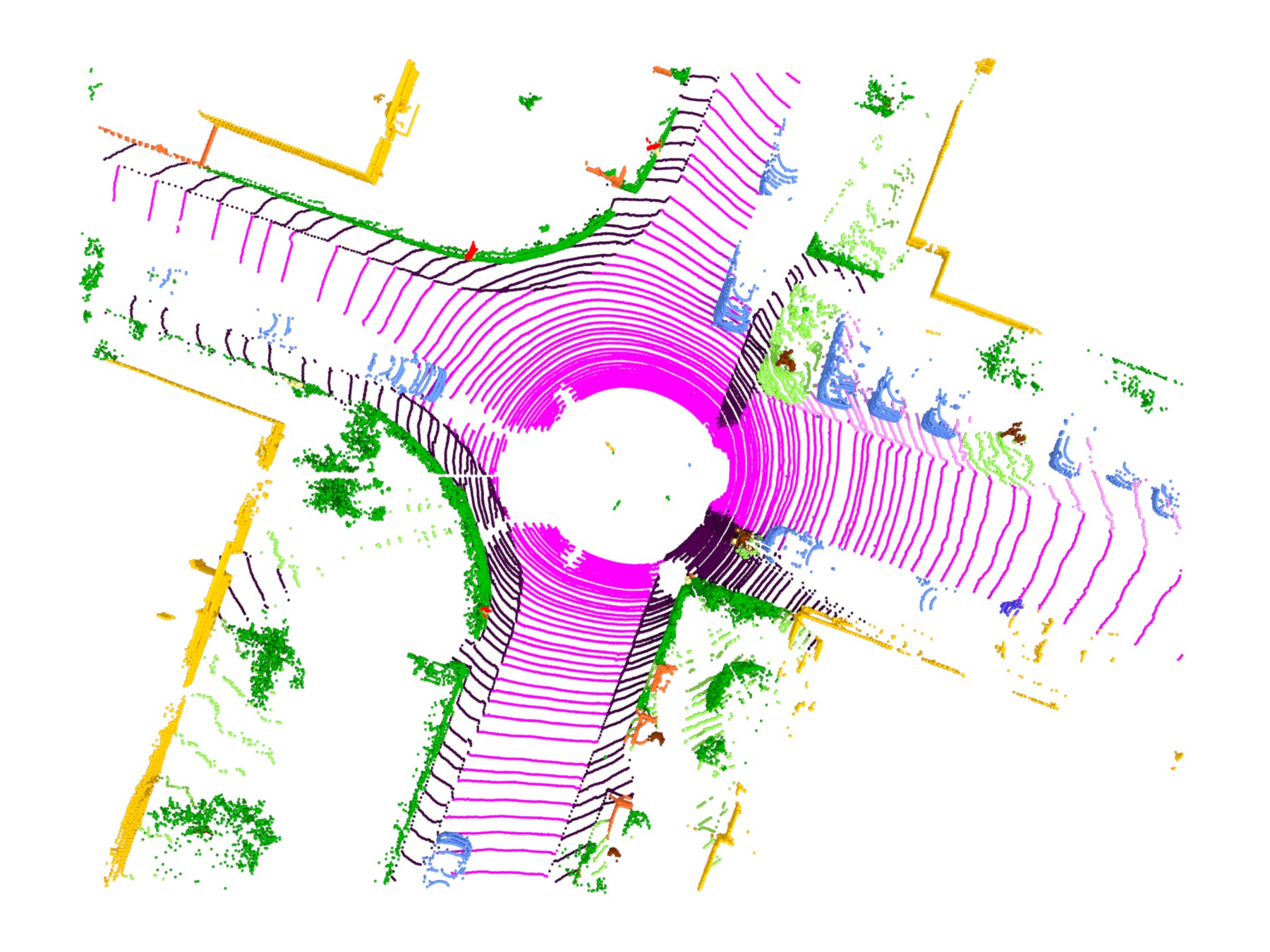}\hspace{\exgap}\includegraphics[width=\exw]{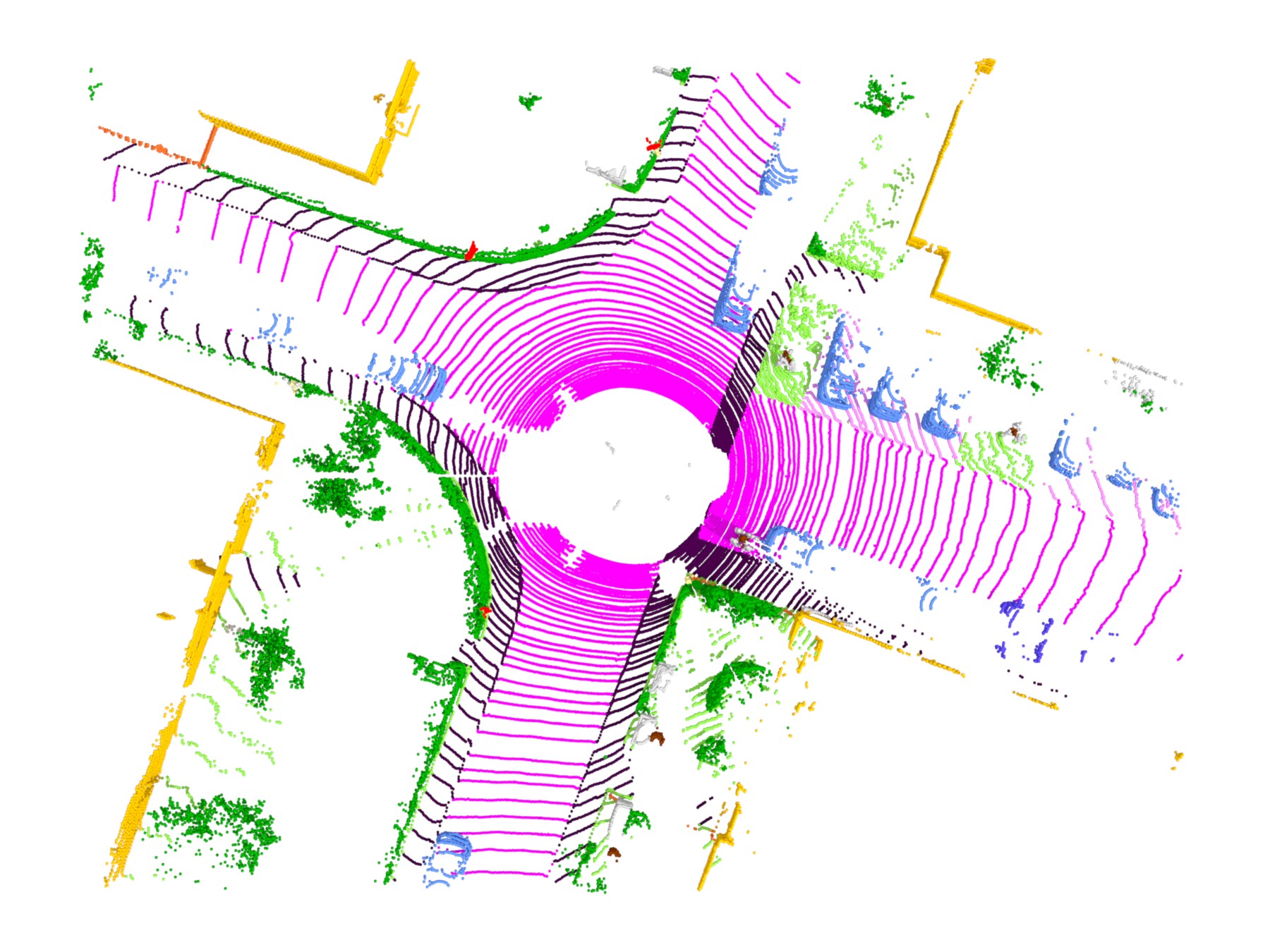}\\[\exgap]
\includegraphics[width=\exw]{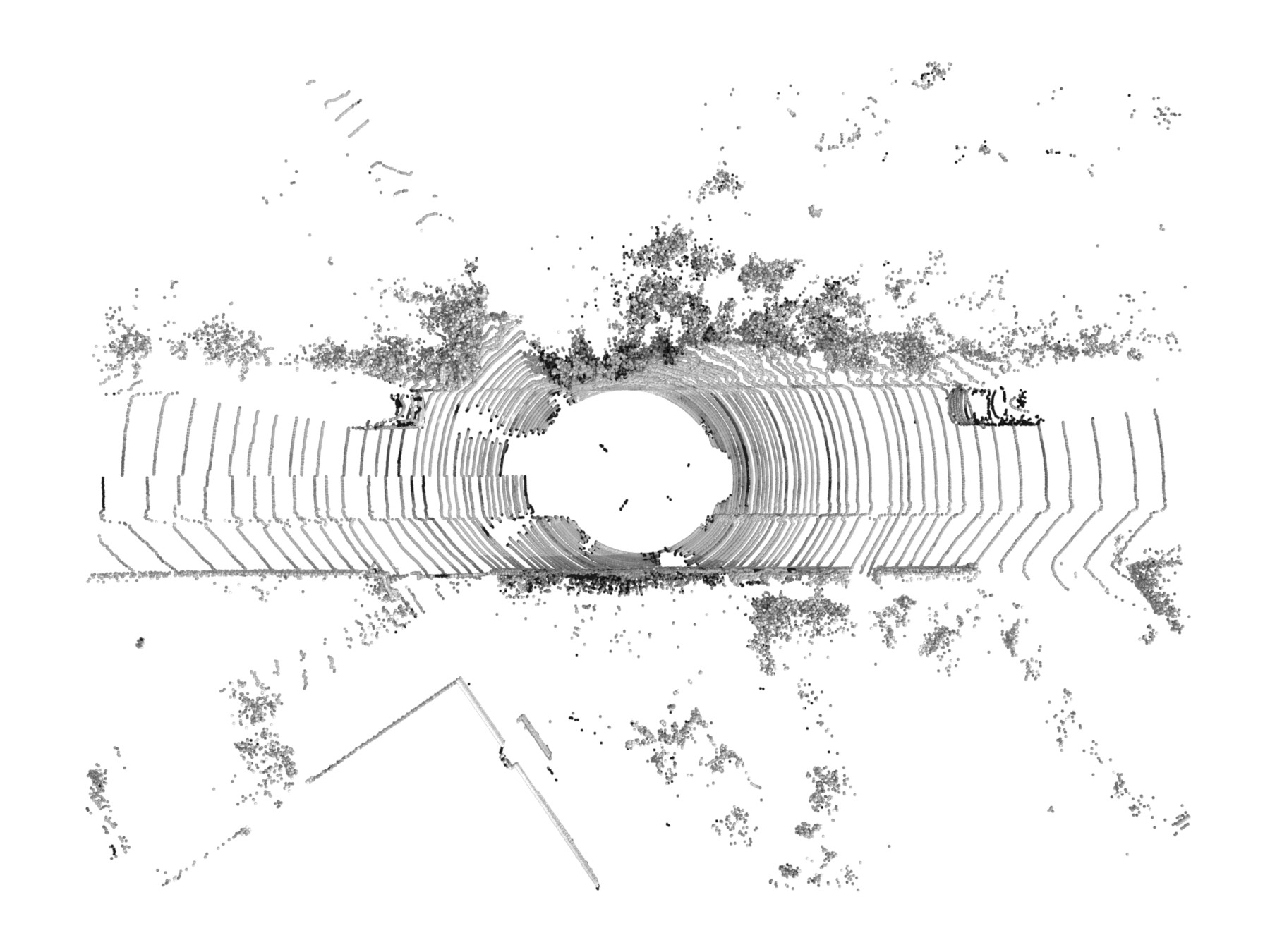}\llap{\makebox[\exw][l]{\hspace{3pt}\raisebox{3pt}{\tiny\color{gray}08\_003200}}}\hspace{\exgap}\includegraphics[width=\exw]{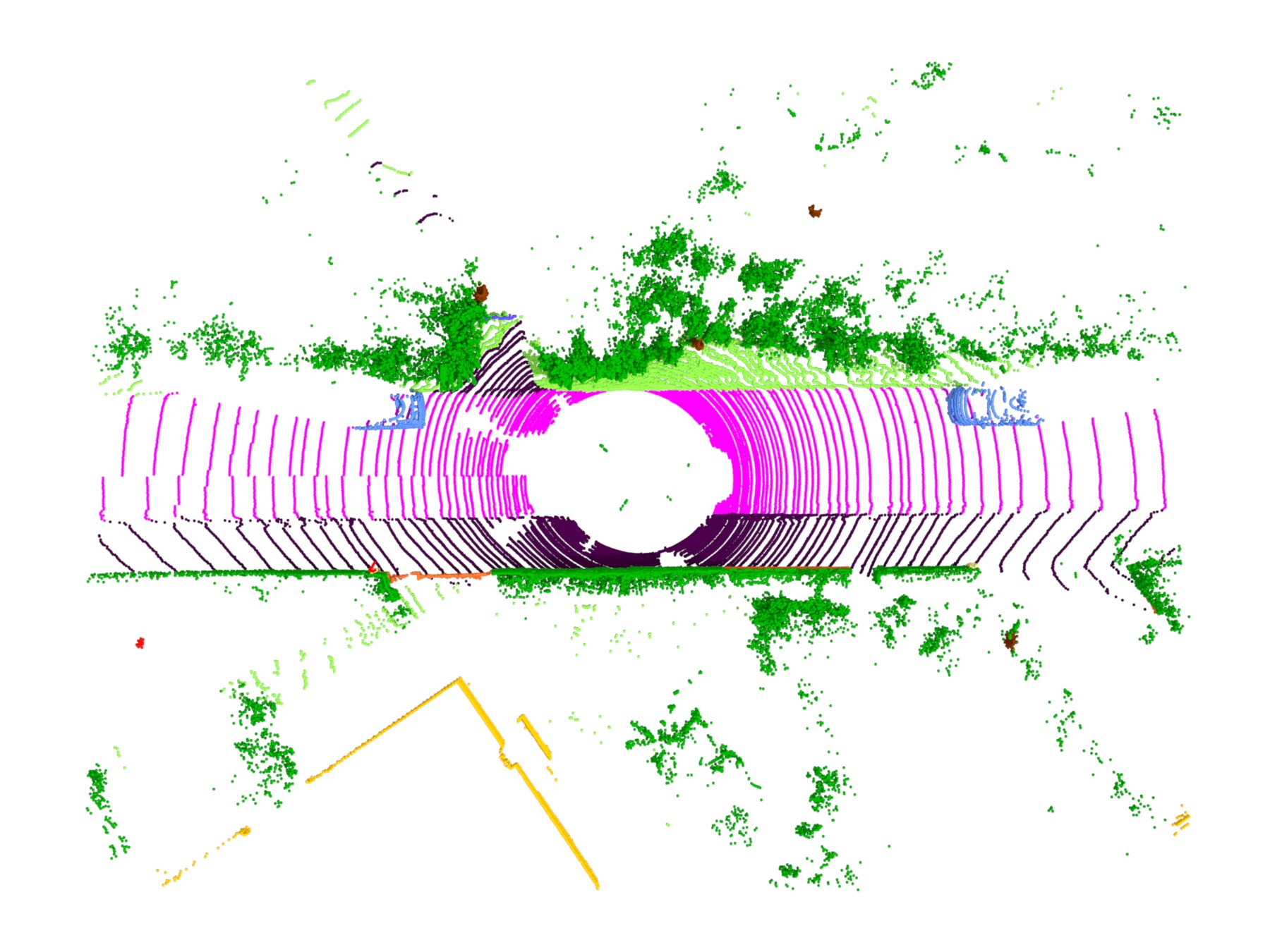}\hspace{\exgap}\includegraphics[width=\exw]{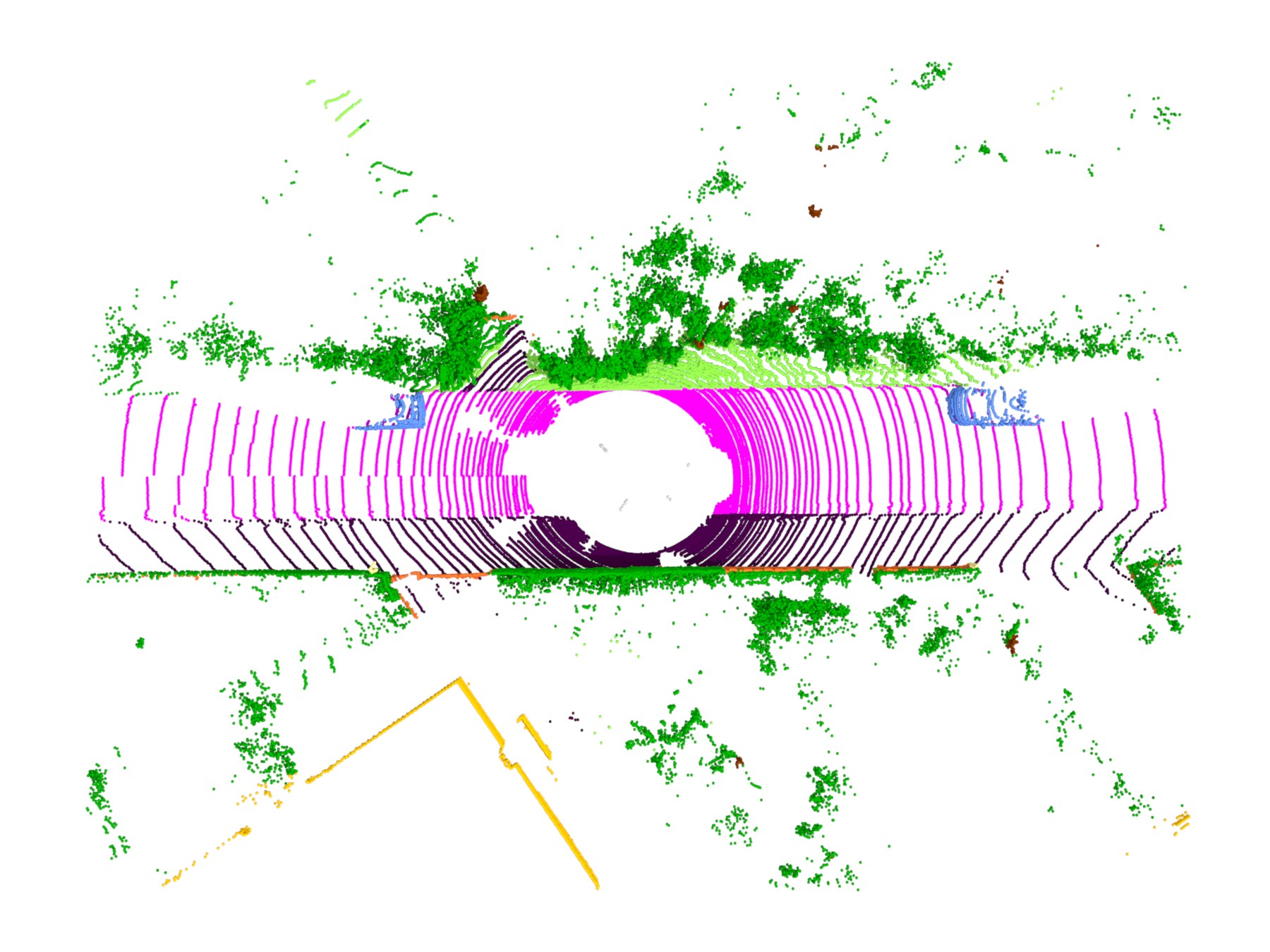}\\[4pt]
\includegraphics[width=\linewidth]{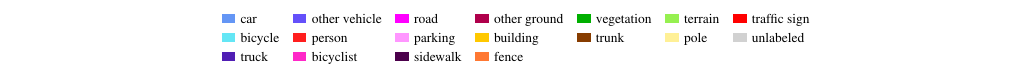}